\documentclass{article}

\usepackage[main, final]{neurips_2026}
\usepackage[utf8]{inputenc} % allow utf-8 input
\usepackage[T1]{fontenc}    % use 8-bit T1 fonts
\usepackage{hyperref}       % hyperlinks
\usepackage{url}            % simple URL typesetting
\usepackage{booktabs}       % professional-quality tables
\usepackage{amsfonts}       % blackboard math symbols
\usepackage{nicefrac}       % compact symbols for 1/2, etc.
\usepackage{microtype}      % microtypography
\usepackage{xcolor}         % colors
\usepackage{graphicx}
\usepackage{subcaption}

\usepackage{comment}
\usepackage{bbm}
\usepackage{bm}
\usepackage{xspace}
\usepackage{diagbox}
\hypersetup{
    colorlinks=true, % Enables coloring of links
    linkcolor=blue,  % Color for internal links
    filecolor=magenta, % Color for file links
    urlcolor=blue,    % Color for URLs
    citecolor=blue,       % For citation links 
}

\usepackage{enumitem}
\usepackage{booktabs,tabularx} 
\usepackage{color,soul}
\usepackage{xcolor}  
\usepackage[export]{adjustbox}
\usepackage{pifont}
\usepackage{amsmath, amssymb}
\usepackage{tikz}
\usepackage{graphicx}
\usepackage{pgfplots}
\usepgfplotslibrary{groupplots}
\definecolor{lineBlue}{RGB}{57,106,177}
\definecolor{lineOrange}{RGB}{218,124,48}
\definecolor{lineGreen}{RGB}{62,150,81}
\definecolor{lineRed}{RGB}{204,37,41}
\definecolor{lineGray}{RGB}{83,81,84}
\definecolor{linePurple}{RGB}{107,76,154}
\definecolor{lineMaroon}{RGB}{146,36,40}
\definecolor{barBlue}{RGB}{114,147,203}
\definecolor{barOrange}{RGB}{225,151,76}
\definecolor{barGreen}{RGB}{132,186,91}
\definecolor{barRed}{RGB}{211,94,96}
\definecolor{barGray}{RGB}{128,133,133}
\definecolor{barPurple}{RGB}{144,103,167}
\definecolor{barMaroon}{RGB}{171,104,81}
\definecolor{bluepigment}{RGB}{51,51,153}

\pgfplotsset{compat=1.18}
\usetikzlibrary{3d,decorations.text,shapes.arrows,positioning,fit,backgrounds}
\usetikzlibrary{intersections, pgfplots.fillbetween}

\usepackage{multirow}

\usetikzlibrary{arrows.meta}
\usetikzlibrary{calc}

\makeatletter
\DeclareRobustCommand\onedot{\futurelet\@let@token\@onedot}
\def\@onedot{\ifx\@let@token.\else.\null\fi\xspace}

\def\eg{\emph{e.g}\onedot} 
\def\ie{\emph{i.e}\onedot} 
 
\def\etc{\emph{etc}\onedot}

\def\etal{\emph{et al}\onedot}

\makeatother
\newcommand{\RR}{\ensuremath{\mathbb{R}}}
\newcommand{\Rtwo}{\ensuremath{\RR^2}}

\newcommand{\set}[1]{\ensuremath{{\{#1\}}}}

\newcommand{\ellTwo}{\ell_2}

\definecolor{brown}{RGB}{165,42,42}

\bmdefine\ba{a}
\bmdefine\bb{b}
\bmdefine\bc{c}
\bmdefine\bd{d}
\bmdefine\be{e}
\bmdefine\boldf{f}
\bmdefine\bg{g}
\bmdefine\bh{h}
\bmdefine\bi{i}
\bmdefine\bj{j}
\bmdefine\bk{k}
\bmdefine\bl{l}
\bmdefine\bm{m}
\bmdefine\bn{n}
\bmdefine\bo{o}
\bmdefine\bp{p}
\bmdefine\bq{q}
\bmdefine\br{r}
\bmdefine\bs{s}
\bmdefine\bt{t}
\bmdefine\bu{u}
\bmdefine\bv{v}
\bmdefine\bw{w}
\bmdefine\bx{x}
\bmdefine\by{y}
\bmdefine\bz{z}
\bmdefine\bA{A}
\bmdefine\bB{B}
\bmdefine\bC{C}
\bmdefine\bD{D}
\bmdefine\bE{E}
\bmdefine\bF{F}
\bmdefine\bG{G}
\bmdefine\bH{H}
\bmdefine\bI{I}
\bmdefine\bJ{J}
\bmdefine\bK{K}
\bmdefine\bL{L}
\bmdefine\bM{M}
\bmdefine\bN{N}
\bmdefine\bO{O}
\bmdefine\bP{P}
\bmdefine\bQ{Q}
\bmdefine\bR{R}
\bmdefine\bS{S}
\bmdefine\bT{T}
\bmdefine\bU{U}
\bmdefine\bV{V}
\bmdefine\bW{W}
\bmdefine\bX{X}
\bmdefine\bY{Y}
\bmdefine\bZ{Z}

\bmdefine\balpha{\alpha}
\bmdefine\bbeta{\beta}
\bmdefine\bgamma{\gamma}

\bmdefine\bdelta{\delta}
\bmdefine\btheta{\theta}
\bmdefine\blambda{\lambda}
\bmdefine\bphi{\phi}
\bmdefine\bxi{\xi}
\bmdefine\bzeta{\zeta}
\bmdefine\boldeta{\eta}
\bmdefine\bpi{\pi}

\bmdefine\bmu{\mu}
\bmdefine\brho{\rho}
\bmdefine\bomega{\omega}
\bmdefine\bOmega{\Omega}
\bmdefine\bPi{\Pi}

\bmdefine\bvarepsilon{\varepsilon}
\bmdefine\bepsilon{\epsilon}

\bmdefine\bDelta{\Delta}

\bmdefine\bTheta{\Theta}

\bmdefine\bsigma{\sigma}
\bmdefine\bSigma{\Sigma}
\bmdefine\bPsi{\Psi}
\bmdefine\bLambda{\Lambda}

\bmdefine\bzero{0}
\bmdefine\bone{1}
\bmdefine\binfty{\infty}

\newcommand{\Lcal}{\mathcal{L}}

\newcommand{\Rcal}{\mathcal{R}}

\newcommand{\xmark}{\ding{55}}%

\newlength{\adaptWidth}
\usepackage{import}

\renewcommand{\figureautorefname}{Fig.}

\renewcommand{\figureautorefname}{Fig.}

\newcommand{\FigureRef}[1]{%
  \begingroup
  \renewcommand{\figureautorefname}{Figure}%
  \autoref{#1}%
  \endgroup
}

\usepackage{array,tabularx}
\newcolumntype{Y}{>{\centering\arraybackslash}X}

\title{Video Stitching from Multiple Moving Cameras}

\author{
  Shira Ifergane \qquad Roy Amoyal \qquad Shahar Benishay\qquad Oren Freifeld \\[1.5mm]
  \small The Stein Faculty of Computer and Information Science, \\
  \small Ben-Gurion University of the Negev, Israel\\[1mm]
  \texttt{\small \{shiraif,amoyalr,shbeni\}@post.bgu.ac.il}, \texttt{\small orenfr@bgu.ac.il}}

\begin{document}

\maketitle

\begin{abstract}
We address geometric video alignment for the same scene captured by multiple freely moving cameras. Existing methods, which are based on image stitching or video stabilization, 
are typically limited to two cameras or must assume the restrictive setting of a fixed linear ordering of the cameras. We propose MViS, a lightweight test-time optimization framework for joint multi-view alignment with temporal consistency. 
MViS uses a spatial-transformer-inspired net to jointly process frames from all views within a short temporal window and predict per-view, per-time homographies.
Optimization proceeds in two stages: geometry-guided alignment followed by appearance-based refinement, with both stages enforcing multi-view and temporal consistency across the sequence. The method requires no scene-specific hyperparameter tuning, handles arbitrary and time-varying view-intersection graphs, and remains robust under rapid camera motion. On a standard two-view video benchmark, MViS matches or surpasses the state of the art. To facilitate evaluation in more complex scenarios, we introduce an annotated dataset with more than two freely moving cameras and establish the first benchmark for this setting. Project page: \url{https://bgu-cs-vil.github.io/MViS-project/}. 
\end{abstract}

\section{Introduction}\label{sec:intro}
Geometric alignment of multiple videos of the same scene captured by freely moving cameras is a challenging and underexplored task arising in multi-user handheld capture, drone footage, or mobile multi-camera devices. 
The goal is to estimate, for every frame in every view, transformations that bring all videos into a spatially aligned and temporally coherent configuration. 

Applying Multiple Image Stitching (MIS) independently at each time yields spatially coherent but temporally inconsistent panoramas. 
The resulting video exhibits spatially dependent shaking that standard stabilization cannot correct, since different regions require conflicting temporal corrections that themselves may vary from frame to frame. This calls for a video-specific formulation that enforces spatial and temporal consistency jointly. Although pairwise alignment of moving-camera videos has been studied, extending it beyond two cameras is problematic: without global consistency, errors compound across views and time. Moreover, camera ordering and view intersections may vary over time, violating assumptions of fixed layouts or persistent overlaps.

\begin{figure}[t]
  \centering
  \resizebox{0.65\textwidth}{!}{
  \subimport{../figures/tikz/}{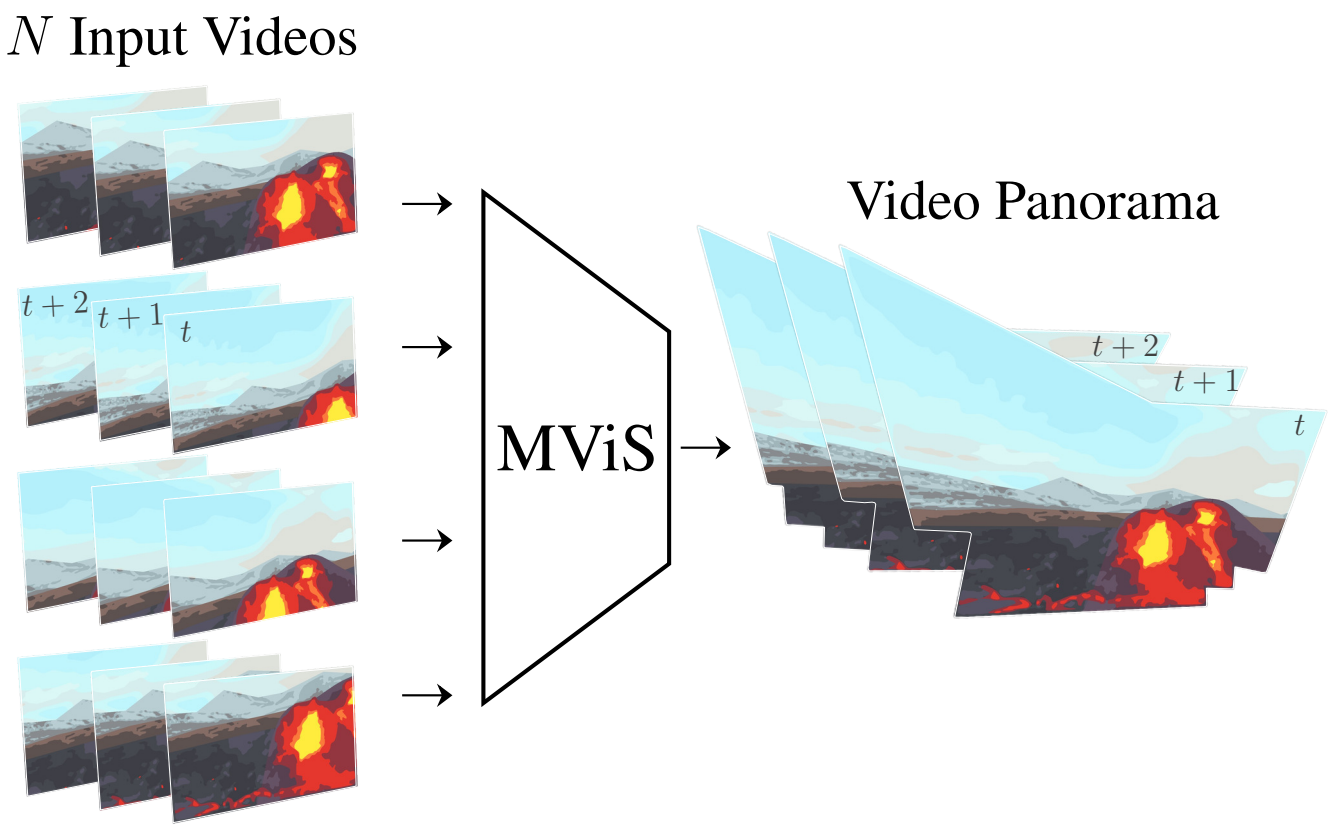}
  }
  \caption{MViS
  aligns $N$ moving-camera videos, without assuming a known/fixed topology of the
view-intersection graph
and requires no scene-specific hyperparameter tuning.}
  \label{Fig:intro}
\end{figure}

Video stitching has long been treated as a hybrid of image stitching and video stabilization, either handling the two tasks sequentially in constrained setups or addressing them jointly. Most moving-camera methods still follow these pipelines and are limited to two cameras.

A prime example is StabStitch~\cite{Nie:ECCV:2024:stabstitch}, which handles two moving cameras, and  StabStitch++~\cite{Nie:TPAMI:2025:stabstitch++}, which extends to three cameras but only under a fixed linear spatial ordering. No existing method supports more than two freely moving cameras, where the view-intersection graph  (defined by pairwise field-of-view overlaps) is arbitrary or time-varying. More broadly, existing approaches operate pairwise and lack a principled mechanism for global consistency across views and time.

We propose Multiple Video Stitching (MViS), an unsupervised optimization method that aligns videos from multiple freely moving cameras (see~\autoref{Fig:intro}) without requiring fixed or known camera ordering. It only assumes that the view-intersection graph is connected at each time. MViS uses a lightweight alignment network (inspired by, but different from, the Spatial Transformer Network (STN)~\cite{Jaderberg:NIPS:2015:STN}) that takes a local multi-view temporal window as input and jointly predicts the corresponding per-frame transformations. These predictions are optimized using losses that enforce multi-view and temporal consistency through relative compositions of transformations (\ie, composing one per-frame transformation with the inverse of another) for pairs of frames, either across views at the same time or across consecutive times in the same view. This yields a scalable optimization that enforces global consistency over cameras and time, rather than independent pairwise alignments.

 The optimization proceeds in two stages: geometry-guided alignment based on noisy pairwise transformations, followed by appearance-based refinement. The resulting regularization-free losses enable stable stitching under fast motion and time-varying view-intersection graphs without per-scene hyperparameter tuning. Operating on short time windows allows MViS to scale to hundreds or thousands of frames while adapting to diverse capture conditions.

\textbf{Our key contributions are:}

1)
A method for stitching videos from multiple freely moving cameras, the first in this setting to handle more than two cameras with arbitrary, time-varying view-intersection graphs.

2)
A scalable unsupervised formulation that jointly aligns all views over time, while avoiding per-scene hyperparameter tuning through regularization-free losses.

3)
A new annotated multi-view video dataset and benchmark for video stitching under unconstrained multi-camera motion.

4)
A variance-based geometric metric for assessing alignment quality in multi-view video stitching.

\section{Related Work}\label{sec:related}
\textbf{Multiple Image Stitching (MIS).}  
Classical MIS methods often follow an incremental pairwise strategy, stitching one image at a time in a predefined order, which makes them prone to error accumulation and drift.
A popular alternative, AutoStitch~\cite{Brown:IJCV:2007:AutoStitch}, addresses this by jointly optimizing the transformations of all images. Yet, despite many MIS methods (\eg,~\cite{Chen:ECCV:2016:GSP,Liao:TIP:2019:SPW,Du:CVPR:2022:GESGSP,Cai:AAAI:2025:OBJGSP}), most classical and learning-based methods still handle only two input images. Methods supporting more than two images often rely on predefined sequences or spatial layout assumptions, such as known left-right ordering or a fixed reference image~\cite{Zaragoza:CVPR:2013:APAP,Li:TMM:2018:ELA}. In multiple video stitching from freely moving cameras, MIS methods produce good panoramas at individual time steps but yield temporally incoherent results when applied frame by frame.
This failure, along with the difficulty of post-hoc video stabilization, stems from the fact that, due to independent camera motions, different panoramic regions usually move differently between times $t$ and $t+1$, which per-frame stitching cannot capture.

\noindent \textbf{Video Stitching.}  
Early works handled the task sequentially, first stitching and then stabilizing videos~\cite{HAMZA:2015:Stabilization}, but only in restricted setups with stationary cameras or fixed rigs. Lin \etal~\cite{Lin:2016:seamless} instead stabilized videos before stitching, but showed results only for two or three cameras (and, to date, have not released code).
Several methods~\cite{Guo:TIP:2016:joint, Nie:TIP:2017:dynamic, Su:2016:dynamic1} incorporate video stabilization techniques such as bundled camera paths~\cite{Liu:SINGGRAPH:2013:BundledPaths}. 
Among these, Guo \etal~\cite{Guo:TIP:2016:joint} first jointly optimized stitching and stabilization for moving cameras.
Although they suggested extending the method beyond two cameras, it relied on camera-by-camera optimization with an alternating bundled-path scheme, resulting in high computational cost and limited scalability. Subsequent works (\eg, \cite{Nie:TIP:2017:dynamic}) noted that such optimization is prohibitively hard.
Moreover, the authors of~\cite{Guo:TIP:2016:joint} released no code, and the multi-camera extension was not demonstrated experimentally.
In contrast, MViS is simple and scalable: each stage optimizes its loss over all views and sliding windows with standard solvers (\eg, ADAM), without alternating procedures or hand-crafted schedules.
Perazzi \etal~\cite{Perazzi:2015:panoramic} stitch videos from linearly arranged wide-baseline cameras, seemingly using calibrated camera parameters. Similarly, \cite{jiang:CVPRW:2015:video} stitch videos from fixed camera rigs.
The restrictive assumptions of both methods make them incompatible with unconstrained setups of multiple freely moving cameras, as also noted in~\cite{Nie:ECCV:2024:stabstitch}.

The methods above improve short-term coherence, but remain limited to two-camera setups, fixed arrays, or stationary capture, and are inapplicable to more general, practical settings.
Other methods target constrained capture setups. Lai \etal~\cite{lai2019video} propose a learning-based approach for linear camera arrays, training a pushbroom-style interpolation network that learns dense flow between adjacent views. Lo \etal~\cite{lo2022efficient} focus on dual-fisheye 360° cameras with geometric calibration and seam optimization. Both assume rigid, synchronized hardware and do not extend to freely moving cameras.
The state-of-the-art (SOTA) video stitching method, StabStitch~\cite{Nie:ECCV:2024:stabstitch}, is the most relevant to our work. It stitches two videos and jointly performs stabilization in an unsupervised, online framework based on UDIS++~\cite{Nie:2023:udis++}. Its slower extension, StabStitch++~\cite{Nie:TPAMI:2025:stabstitch++}, improves alignment and stabilization via a differentiable bidirectional decomposition and a warp-smoothing model. StabStitch++ can handle three cameras, but only with a fixed linear ordering, and does not generalize to unknown, dynamic topology or beyond three videos. In contrast, MViS handles arbitrary numbers of cameras and flexible view-intersection graphs.

\noindent\textbf{Image Alignment.}   
Several methods~\cite{Ofri:CVPR:2023:neuracongealingl,Gupta:ICCV:2023:ASIC,Barel:ECCV:2024:JA} jointly align tens of still images using high-dimensional DINO features~\cite{Caron:ICCV:2021:Dino}. 
They act on still images with masked objects and do not scale to videos: even \cite{Barel:ECCV:2024:JA}, the most efficient, runs out of memory when the number of frames exceeds 70.
While MViS shares with~\cite{Barel:ECCV:2024:JA} the idea of comparing images via relative transformations, its formulation, architecture, and losses differ: MViS acts on videos and processes multi-view temporal windows rather than single images.
Moreover, \cite{Barel:ECCV:2024:JA} is quadratic in the number of images, while MViS is linear in the number of frames and quadratic only in the (typically small) number of cameras.

\noindent
\textbf{Keypoint matchers} such as RoMa~\cite{Edstedt:CVPR:2024:RoMa} or LoFTR~\cite{sun:2021:loftr} operate on image pairs and do not account for temporal or multi-view consistency. MViS uses such matchers solely to generate noisy observed relative transformations that guide the geometric stage.
 
\section{Method}
\begin{figure}[t]
\centering
\resizebox{\textwidth}{!}{
\subimport{../figures/tikz/}{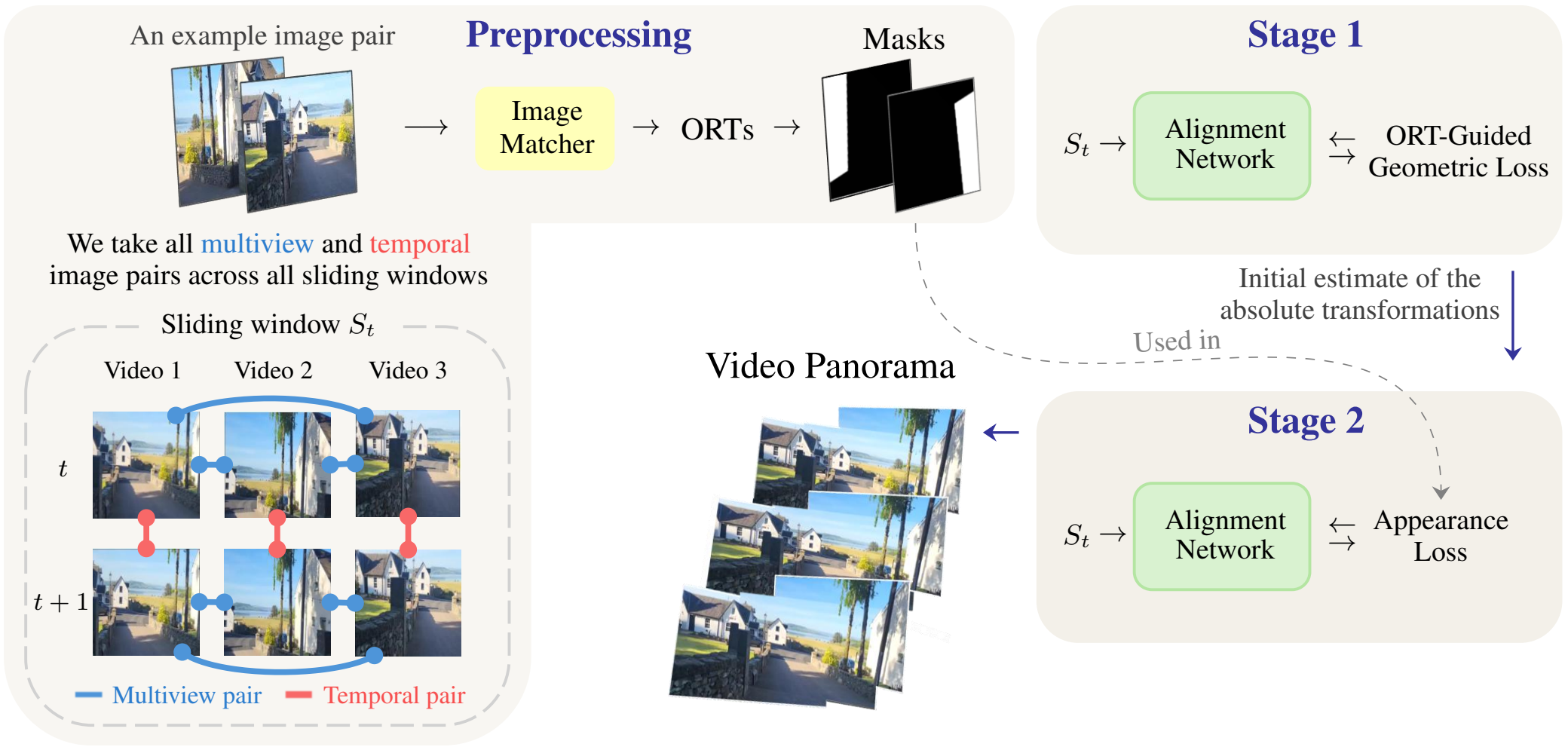} 
}
  \caption{An overview of the proposed method, Multiple Video Stitching (MViS).}
  \label{Fig:MethodOverview}
\end{figure} 
Consider $N$ approximately synchronized videos, $(V_i)_{i=1}^N$, of the same scene, captured by moving cameras. 
Each $V_i$ has $\tau$ frames, $(F^t_i)_{t=1}^\tau$.
We do not assume that the two views in every pair overlap at time $t$, only that the view-intersection graph, defined by field-of-view overlaps, is connected at each $t$. 
Our task is to estimate, for each  $F_i^t$, a homography $\bH_i^t$ that aligns it to a single panoramic coordinate system shared across all views and times. 
We now describe MViS, outlined in \autoref{Fig:MethodOverview}. 
After preprocessing (\autoref{Sec:Method:Subsec:Preprocessing}), which includes 
the extraction of noisy pairwise homographies, we use an alignment net (\autoref{Sec:Method:Subsec:AlignmentNetwork}) to optimize view- and time-specific transformations in two stages: geometric-loss minimization (\autoref{Sec:Method:Subsec:GeometricLoss}) followed by appearance-based refinement (\autoref{Sec:Method:Subsec:FeatureLoss}). 
Both losses are defined using absolute per-view, per-time transformations but depend only on relative compositions, making them invariant to a global transformation and regularization-free, as explained in \autoref{Sec:Method:Subsec:IC}. 
Finally, we generate a panoramic video as described in \autoref{Sec:Method:Subsec:CreatingPanoram}, while \autoref{Sec:Metohd:Subsec:Stablization} explains the implicit temporal stabilization.

\subsection{Preprocessing}\label{Sec:Method:Subsec:Preprocessing}

The preprocessing has four steps: resizing, RGB normalization, Observed Relative Transformation (ORT) estimation, and mask extraction. We first \textbf{resize} each frame $F_i^t$ to a fixed square resolution $r \times r$ ($r=256$ in all experiments), making runtime independent of the original resolution. We then \textbf{normalize} each RGB channel by subtracting its sample mean and dividing by its sample standard deviation, computed across all frames in all videos. This simple normalization handles inter-view color variations effectively. Let $I_i^t \in \RR^{r \times r \times 3}$ denote the resulting image.
Next, we estimate noisy pairwise homographies, which we call \textbf{Observed Relative Transformations (ORTs)}, for view pairs $(I_i^t,I_j^t)$, $i\neq j$, and temporal pairs $(I_i^t,I_i^{t+1})$. 
For each such pair, we extract keypoint correspondences using an image matcher and then use  RANSAC to estimate the homography.
Let $\widehat{\bH}_{i \rightarrow j}^t$ denote the ORT from $I_i^t$ to $I_j^t$, let $\widehat{\bH}_i^{t \rightarrow t+1}$ denote the ORT from $I_i^t$ to $I_i^{t+1}$, and let $\widehat{\bH}_i^{t+1 \rightarrow t}$ denote the ORT from $I_i^{t+1}$ to $I_i^t$. 
Due to estimation noise, typically $\widehat{\bH}_i^{t+1 \rightarrow t}\neq(\widehat{\bH}_i^{t \rightarrow t+1})^{-1}$.

The ORTs and the overlap masks they define affect the losses below. At a given $t$, $I_i^t$ and $I_j^t$ may have an empty or negligible spatial intersection. We detect such cases using matcher-provided certainty scores: if the average score for an image pair is below a threshold (\eg, 0.3 for RoMa), the losses ignore that pair; otherwise, we assume a nonempty intersection.
Accordingly, let $M^t_{i\to j}$ be the \textbf{binary mask} obtained by warping an $r$-by-$r$ all-ones mask from the domain of $I_i^t$ to that of $I_j^t$ using $\widehat{\bH}^t_{i\to j}$. This mask represents the overlapping image region between $I_j^t$ and the warped $I_i^t$. Similarly, $M^{t \to t+1}_{i}$ and $M^{t+1 \to t}_{i}$ are defined via $\widehat{\bH}^{t \to t+1}_{i}$ and $\widehat{\bH}^{t+1 \to t}_{i}$.

\subsection{The Architecture of the Alignment Network}\label{Sec:Method:Subsec:AlignmentNetwork}
\begin{figure}[h]
  \centering
  \usetikzlibrary{positioning,arrows.meta,shapes.geometric,calc,decorations.pathreplacing,fit,backgrounds}

\begin{tikzpicture}[
  node distance = 1cm and 1cm,
  every node/.style = {font=\small},
  box/.style  = {draw, minimum width=1.5cm, minimum height=3mm, align=center},
  fea/.style  = {draw, rounded corners=2pt, minimum width=0.8cm,
                 minimum height=2mm, align=center},
  layer/.style = {draw, minimum width=0.5cm, minimum height=2.6cm, align=center},
  arrow/.style = {-{Stealth[length=1mm,width=1mm]}, shorten >=0.35mm},
]

% ---- First set of inputs (time t) ----
\node[text width=0.5cm, align=center] (u1)
  {\footnotesize $I_{1}^{t}$};
\node[below=2mm of u1, text width=0.5cm, align=center] (uN)
  {\footnotesize $I_{N}^{t}$};
\node[text width=0.15cm, align=right] at ($(u1)!0.4!(uN)$) {\scalebox{0.75} {\vdots}};

% ---- Second set of inputs (time t+1) ----
\node[below=0.15mm of uN, text width=0.5cm, align=center] (v1)
  {\footnotesize $I_{1}^{t+1}$};
\node[below=2mm of v1, text width=0.5cm, align=center] (vN)
  {\footnotesize $I_{N}^{t+1}$};
\node[text width=0.15cm, align=right] at ($(v1)!0.4!(vN)$) {\scalebox{0.75} {\vdots}};

% ---- CNNs for time t ----
\node[box, right=0.5cm of u1] (cnn1) {CNN};
\node[box, right=0.5cm of uN] (cnn2) {CNN};
\draw[arrow] (u1.east) -- (cnn1.west);
\draw[arrow] (uN.east) -- (cnn2.west);
\node at ($(cnn1)!0.4!(cnn2)$) {\scalebox{0.75} {\vdots}};

% ---- CNNs for time t+1 ----
\node[box, right=0.5cm of v1] (cnn3) {CNN};
\node[box, right=0.5cm of vN] (cnn4) {CNN};
\draw[arrow] (v1.east) -- (cnn3.west);
\draw[arrow] (vN.east) -- (cnn4.west);
\node at ($(cnn3)!0.4!(cnn4)$) {\scalebox{0.75} {\vdots}};

% ---- Features for time t ----
\node[fea, right=0.5cm of cnn1] (f1) {\scriptsize $f_{1}^{t}$};
\node[fea, right=0.5cm of cnn2] (fN) {\scriptsize $f_{N}^{t}$};
\draw[arrow] (cnn1.east) -- (f1.west);
\draw[arrow] (cnn2.east) -- (fN.west);
\node at ($(f1)!0.4!(fN)$) {\scalebox{0.75} {\vdots}};

% ---- Features for time t+1 ----
\node[fea, right=0.5cm of cnn3] (g1) {\scriptsize $f_{1}^{t+1}$};
\node[fea, right=0.5cm of cnn4] (gN) {\scriptsize $f_{N}^{t+1}$};
\draw[arrow] (cnn3.east) -- (g1.west);
\draw[arrow] (cnn4.east) -- (gN.west);
\node at ($(g1)!0.4!(gN)$) {\scalebox{0.75} {\vdots}};

% ---- Brace ----
\draw[decorate,decoration={brace,amplitude=4pt}]
  ($(f1.north east)+(0.14,0)$) -- ($(gN.south east)+(0.14,0)$);

% ---- Flatten point ----
\coordinate (flatten) at
  ($ ($(f1.north east)+(0.4,0)$) !0.5! ($(gN.south east)+(0.4,0)$) $)
  +(8mm,0);

% ---- FC stack ----
\node[layer, right=1.2cm of flatten] (fc1)  {\rotatebox{90}{\small FC}};
\node[layer, right=1.5mm of fc1]    (gelu) {\rotatebox{90}{\small GELU}};
\node[layer, right=1.5mm of gelu]   (fc2)  {\rotatebox{90}{\small FC}};

% ---- Arrow into FC stack ----
\draw[arrow] (flatten) -- (fc1)
  node[midway, above=3pt, font=\footnotesize] {concat\,\&}
  node[midway, below=3pt, font=\footnotesize] {flatten};

% ---- Theta outputs: individual nodes at exact y of each input ----
\path let \p1=(fc2.east), \p2=(u1.center) in 
  node[anchor=west] (th1) at (\x1+0.5cm, \y2) {\footnotesize $\boldsymbol{\theta}_{1}^{t}$};
\path let \p1=(fc2.east), \p2=(uN.center) in 
  node[anchor=west] (thN) at (\x1+0.5cm, \y2) {\footnotesize $\boldsymbol{\theta}_{N}^{t}$};
\node at ($(th1)!0.4!(thN)$) {\scalebox{0.75} {\vdots}};

\path let \p1=(fc2.east), \p2=(v1.center) in 
  node[anchor=west] (th1p) at (\x1+0.5cm, \y2) {\footnotesize $\boldsymbol{\theta}_{1}^{t{+}1}$};
\path let \p1=(fc2.east), \p2=(vN.center) in 
  node[anchor=west] (thNp) at (\x1+0.5cm, \y2) {\footnotesize $\boldsymbol{\theta}_{N}^{t{+}1}$};
\node at ($(th1p)!0.4!(thNp)$) {\scalebox{0.75} {\vdots}};

% ---- Single arrow from fc2 to theta column ----
\draw[arrow] (fc2.east) -- (th1.west |- fc2.east);

% ---- Outer box ----
\coordinate (leftMid) at ($(v1.east)!0.4!(cnn3.west)$);
\coordinate (rightMid) at ($(fc2.east)+(0.15,0)$);
\begin{scope}[on background layer]
  \draw[rounded corners=2pt]
    ([yshift=2pt]  leftMid |- u1.north) rectangle
    ([yshift=-2pt] rightMid |- vN.south);
\end{scope}

\end{tikzpicture} 
  \caption{The alignment network for $N$ cameras. 
  The weights of the CNNs are shared.}
  \label{Fig:AlignmentNetwork}
\end{figure}
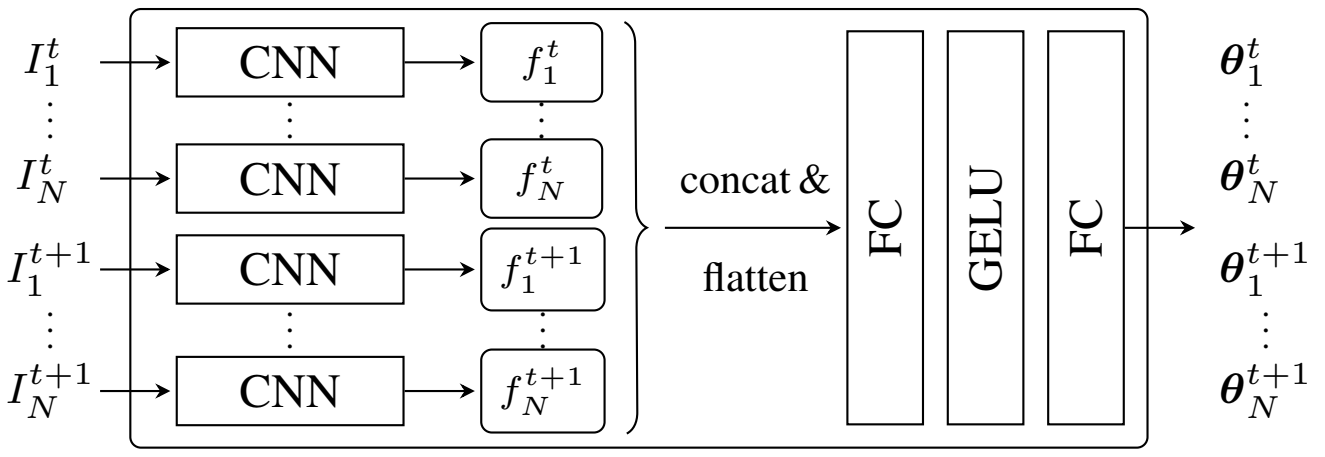

MViS uses an alignment network (\autoref{Fig:AlignmentNetwork}) inspired by the Spatial-Transformer Net (STN)~\cite{Jaderberg:NIPS:2015:STN}, but adapted to multi-view video alignment.
In a standard STN, the input is a single image
and it predicts a single transformation. 
In contrast, here the net's input is a sliding window $(I_1^t, \dots, I_N^t, I_1^{t+1}, \dots, I_N^{t+1})$, and it predicts the parameters of $2N$ homographies, $(\btheta_1^t, \dots, \btheta_N^t, \btheta_1^{t+1}, \dots, \btheta_N^{t+1})$. A shared lightweight CNN first processes each image independently. A fully-connected module then aggregates information across the window and jointly predicts, for each frame in the window, an 8-dimensional vector $\btheta_i^t$ parameterizing a homography $\bH_i^t=\mathrm{func}(\btheta_i^t)$, a 3-by-3 invertible matrix with determinant 1.
That well-known parameterization (used, \eg, in~\cite{mei2006homography,Arrigoni:IJCV:2020:Sync,Barel:ECCV:2024:JA, Hirsch:NeurIPS:2025:FastJAM}) is based on the Special Linear group $\mathrm{SL}(3)$ and detailed in ~\autoref{sec:sl}. It guarantees matrix invertibility; this invertibility improves optimization stability~\cite{Skafte:CVPR:2018:DDTN} and is key to our relative-composition losses below.
The fully-connected module has two linear layers separated by a GELU activation~\cite{Hendrycks:2016:gelu}.
The full architecture appears in ~\autoref{sec:arch}. 
We note two differences from existing STN-based joint-alignment methods for images (\eg,~\cite{Ofri:CVPR:2023:neuracongealingl,Barel:ECCV:2024:JA}). 
1) These methods process one image at a time, so images interact only through the loss. 
In contrast, MViS processes all $2N$ images in the sliding window jointly, allowing cross-view and temporal interactions already inside the net. 
2) While STN~\cite{Jaderberg:NIPS:2015:STN} has an image-warping module, MViS uses image warping only in the second-stage appearance loss; the geometric loss acts on sparse grid points so it requires no image warping.

\subsection{Stage 1: The ORT-guided Geometric Loss}\label{Sec:Method:Subsec:GeometricLoss}

\begin{figure*}[b]
\centering
\resizebox{\textwidth}{!}{%
\subimport{../figures}{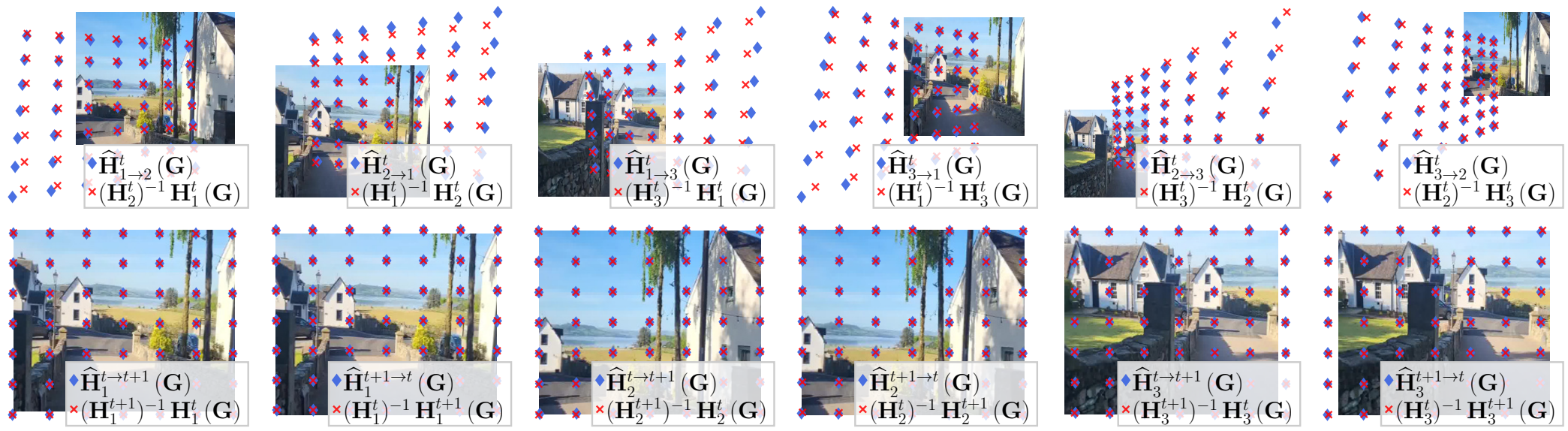}}
\caption{\textbf{Stage 1's ORT-guided Geometric Loss}, $\mathcal{L}_{\mathrm{geo}}$. Illustration using a sliding window $S_t$ with $N=3$. Top row: the inter-view pairs. 
Bottom: the temporal pairs. In each plot, the grid warped by the predicted transformation is overlaid on the grid warped by the ORT estimate. Each image corresponds to the target frame of the transformation (\eg, top-left: $I^t_1 \to I^t_2$, visualized over $I^t_2$).
} 
\label{Fig:fig_stage1}
\end{figure*}
Minimizing the appearance-based loss in the next section
(\autoref{Sec:Method:Subsec:FeatureLoss}) is highly effective, but typically requires a good initialization.
We therefore first optimize a geometric loss (depicted in~\autoref{Fig:fig_stage1}) guided by the ORTs from~\autoref{Sec:Method:Subsec:Preprocessing}. Unlike the appearance loss, this loss yields reasonable results even when the initial transformations are set to the identity.
Let $\bG \in \RR^{d^{2} \times 2}$ contain the $d^2$ 2D points of a regular $d \times d$ grid over $[-1,1] \times [-1,1]$, with the image center as the origin. 
In our experiments, $d=7$. 
For a 3-by-3 homography matrix $\bH$, we slightly abuse notation and let $\bH(\bG)\in \RR^{d^{2} \times 2}$ denote the result, in inhomogeneous coordinates, of applying $\bH$ to $\bG$.

Our geometric loss, for sliding window $S_t = \{ I^s_{i} \mid s \in \{t,t+1\}, i \in \{1,\dots,N\} \}$,

is
$\Lcal_{\mathrm{geo}} = \Lcal_{\mathrm{geo}}^{\mathrm{time}} + \Lcal_{\mathrm{geo}}^{\mathrm{views}}$
 where these terms are
 {\setlength{\abovedisplayskip}{4pt}
 \setlength{\belowdisplayskip}{4pt}\begin{align}
&\Lcal_{\mathrm{geo}}^{\mathrm{views}}
= %\nonumber
\tfrac{1}{2(N^2 - N)} \sum_{i \neq j} \sum_{k \in \{0,1\}}
     \Lcal^{\mathrm{smooth\text{-}\ell_1}}\Bigl(\underbrace{\widehat{\bH}^{t+k}_{i \to j}}_{\text{ORT}}(\bG) - 
     ((\bH^{t+k}_j)^{-1}\bH^{t+k}_i)(\bG)
\Bigr) %\raisetag{6pt}
\\
%\end{align}\begin{align}
&\Lcal_{\mathrm{geo}}^{\mathrm{time}} = 
\Lcal^{\mathrm{fwd}}_{\mathrm{geo}} + \Lcal^{\mathrm{bwd}}_{\mathrm{geo}}\ , 
\\
&\Lcal^{\mathrm{fwd}}_{\mathrm{geo}} = \tfrac{1}{2N} \sum\nolimits_{i=1}^{N} \Lcal^{\mathrm{smooth\text{-}\ell_1}}\Bigl(
  \underbrace{\widehat{\bH}^{t \rightarrow t+1}_{i}}_{\text{ORT}}  (\bG) - 
  ((\bH^{t+1}_i)^{-1} 
  \bH^{t}_i)(\bG) \Bigr) ,  
\\
&\Lcal^{\mathrm{bwd}}_{\mathrm{geo}} = \tfrac{1}{2N} \sum\nolimits_{i=1}^{N} \Lcal^{\mathrm{smooth\text{-}\ell_1}}\Bigl(
\underbrace{\widehat{\bH}^{t+1 \rightarrow t}_{i}}_{\text{ORT}}  (\bG) -((\bH^t_i)^{-1} 
\bH^{t+1}_i  )(\bG) \Bigr) \, 
\end{align}
}
and $\Lcal^{\mathrm{smooth\text{-}\ell_1}}$
is the smooth $\ell_1$ loss. 
For each image pair, $\Lcal_{\mathrm{geo}}$ compares two transformations of $\bG$:
1) the ORT and 2) the composition of the predicted transformation for one image with the inverse of the transformation predicted for the other.   
During optimization, 
the overall geometric loss 
is averaged over all sliding windows in a batch.

\subsection{Stage 2: The Appearance Loss}\label{Sec:Method:Subsec:FeatureLoss}
\begin{figure}[t]
\centering
\resizebox{\textwidth}{!}{%
\subimport{../figures/tikz/}
{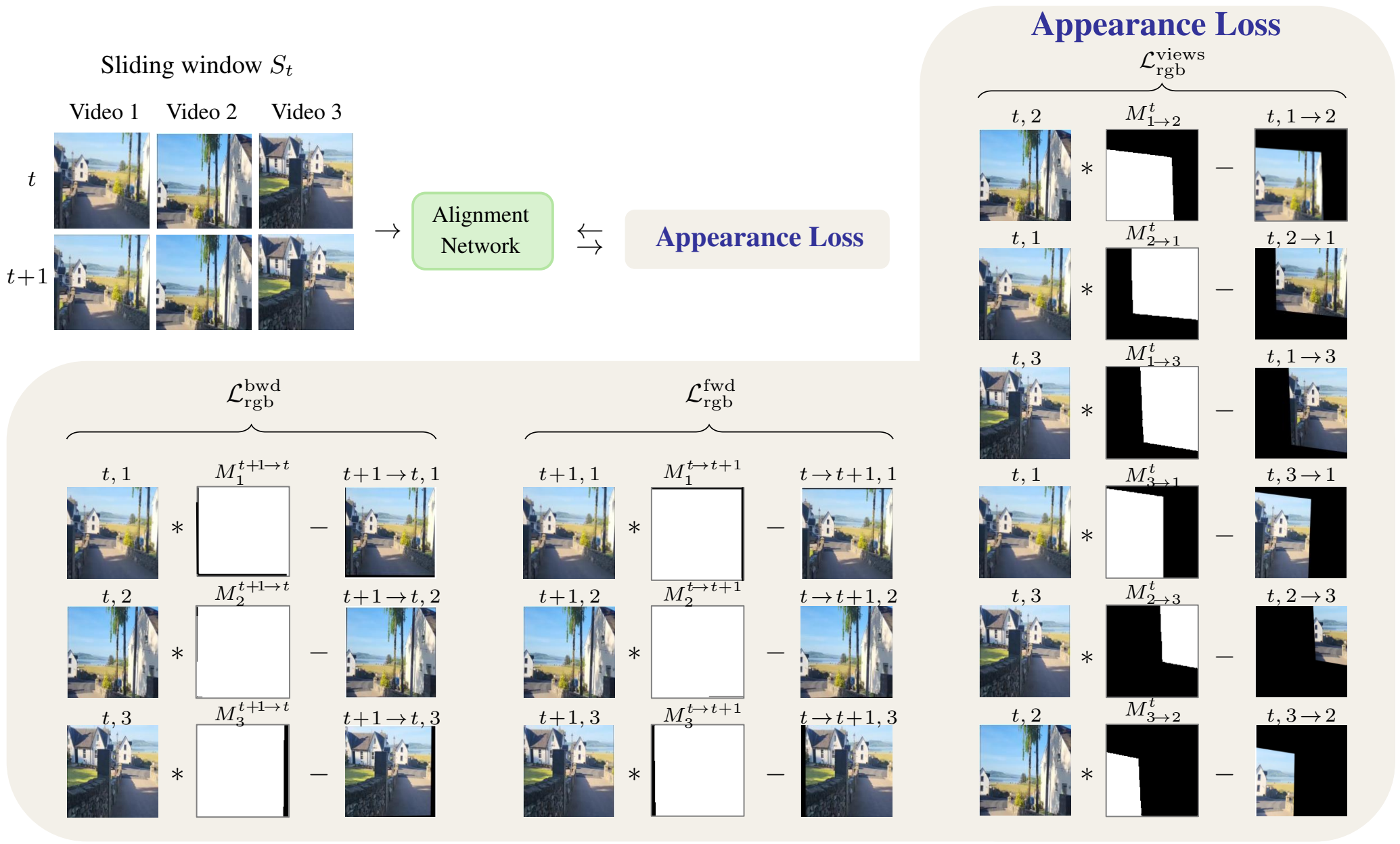} 
}
\caption{\textbf{Stage 2's Appearance Loss $\Lcal_{\text{rgb}}$}. 
  In this figure, $N=3$.
  } 
  \label{Fig:Stage2Fig}
\end{figure}

The second stage uses an appearance loss based on (normalized) RGB values rather than transformed grid points. 
In another slight abuse of notation, for a homography matrix $\bH$ and image $I$, let $\bH(I)$ denote the result of warping $I$ by $\bH$. 
For each sliding window $S_t$, the proposed appearance-based loss is
$\Lcal_{\mathrm{rgb}} = \Lcal_{\mathrm{rgb}}^{\mathrm{time}} + \Lcal_{\mathrm{rgb}}^{\mathrm{views}}$,
where 
 {\setlength{\abovedisplayskip}{4pt}
 \setlength{\belowdisplayskip}{4pt}
 \begin{align}
&
\Lcal_{\mathrm{rgb}}^{\mathrm{views}} =   \tfrac{1}{2(N^2-N)}\sum_{i \neq j} \sum_{k \in \{0,1\}} || I_j^{t+k} -
((\bH^{t+k}_j)^{-1}\bH^{t+k}_i)(I_i^{t+k}) ||_{\ellTwo}^2\, ,
\label{Eqn:Loss:Feature:Multiview}
\\
&\Lcal_{\mathrm{rgb}}^{\mathrm{time}} = 
\Lcal^{\mathrm{fwd}}_{\mathrm{rgb}} + \Lcal^{\mathrm{bwd}}_{\mathrm{rgb}}\, ,
\\
&\Lcal^{\mathrm{fwd}}_{\mathrm{rgb}} = \tfrac{1}{2N} \sum\nolimits_{i=1}^{N} || I_i^{t+1} - 
((\bH^{t+1}_i)^{-1}
 \bH^t_i)(I_i^t) ||_{\ellTwo}^2\, ,\label{Eqn:Loss:Feature:Fwd}
\\
&\Lcal^{\mathrm{bwd}}_{\mathrm{rgb}} = \tfrac{1}{2N} \sum\nolimits_{i=1}^{N} || I_i^t - 
({(\bH^t_i)}^{-1}\bH^{t+1}_i)(I_i^{t+1}) ||_{\ellTwo}^2\,  ,
\label{Eqn:Loss:Feature:Bwd}
\end{align}}
and $||\cdot||_{\ellTwo}$ is the $\ellTwo$ norm. 
As in~\autoref{Sec:Method:Subsec:GeometricLoss}, we average the losses across all sliding windows in a batch. 
In each summand of \autoref{Eqn:Loss:Feature:Multiview}, \autoref{Eqn:Loss:Feature:Fwd}, and \autoref{Eqn:Loss:Feature:Bwd}, the image difference is computed only in the overlap region determined by the appropriate mask, which in turn is induced by the relevant ORT (see \autoref{Sec:Method:Subsec:Preprocessing}). 
We omit the masks from the notation to avoid clutter; \autoref{Fig:Stage2Fig} depicts the loss and the masked regions.

\subsection{Regularization-free Losses and Global Invariance}\label{Sec:Method:Subsec:IC}
The losses from \autoref{Sec:Method:Subsec:GeometricLoss} and \autoref{Sec:Method:Subsec:FeatureLoss} are defined via compositions of predicted absolute transformations; \eg, for views $i$ and $j$, a loss term depends on the relative transformation $\bH_j^{-1}\bH_i$. In the geometric stage, this transformation is applied to corresponding points; in the appearance stage, to image content.
A central consequence is that no explicit regularization over the predicted transformations is required. Each transformation must explain observed data, either geometric correspondences or image appearance. Degenerate transformations (\eg, excessive shrinking or severe distortion) would increase inter-view misalignment and hence the loss. Thus, the loss cannot be trivially reduced through collapse or uncontrolled growth of the transformations. Consequently, regularization terms (\eg, on transformation size or lack of smoothness) are unneeded, freeing us from scene-specific hyperparameter tuning and the optimization complexity such regularization entails (see also~\autoref{sec:reg}).
Note that regularization-free losses and their associated benefits have also been demonstrated in several of our earlier joint-alignment works in various other settings, including~\cite{Shapira:ICML:2023:RFDTAN, Barel:ECCV:2024:JA, Hirsch:NeurIPS:2025:FastJAM, Naaman:ICCV:2025:TPL}.

A second property is that no explicit atlas or canonical panorama is needed: all comparisons are made directly between views rather than to a learned template. This is important in multi-camera video, where the scene evolves over time and maintaining a stable global reference is nontrivial. Crucially, although the loss terms are defined over pairs of views, MViS jointly predicts the transformations of all views in the sliding window, yielding a globally consistent alignment rather than a collection of independent pairwise alignments.

Lastly, the losses are invariant to a common global transformation: if all predicted transformations are left-composed with some transformation $\bG$, then their relative transformations are unchanged, since $(\bG \bH_j)^{-1}(\bG \bH_i)=\bH_j^{-1}\bH_i$. Thus, the minimizers are defined only up to a global transformation and form an equivalence class under such compositions. While this implies infinitely many solutions in terms of \emph{transformations} (a phenomenon shared with~\cite{Barel:ECCV:2024:JA, Hirsch:NeurIPS:2025:FastJAM} and works on synchronization over Lie groups, \eg,~\cite{Rosen:IJRS:2019:SESync,Arrigoni:IJCV:2020:Sync}, or SfM), all members of a given equivalence class induce the same inter-frame \emph{correspondences}.
In practice, for visualization, we apply to all predicted transformations the inverse of the predicted transformation of an arbitrarily-chosen frame (\eg, the first frame of the first video). Since this choice is post-hoc, it has no effect on alignment quality. Finally, while relative-transformation losses were recently used in~\cite{Barel:ECCV:2024:JA, Hirsch:NeurIPS:2025:FastJAM}, their roots go back to classical methods such as~\cite{Cox:ICCV:2009:LS} and even~\cite{Lucas:IJCAI:1981:iterative}.

\subsection{From the Prediction to a Panoramic Video}\label{Sec:Method:Subsec:CreatingPanoram}
After optimization, we create a panoramic video by applying the predicted $\bH_i^t$ to the frames $F_i^t$, while accounting for the resizing. 
Since the solution is defined up to a global transformation, visualization requires choosing a reference. 
We either apply, at each $t$, the inverse of $\bH_1^t$ to all frames at that time, or apply the inverse of $\bH_1^1$ to all frames across all times and views. 
This postprocessing is purely for visualization: MViS gives no special weight to the first video or to its first frame.

 \subsection{Temporal Stabilization}\label{Sec:Metohd:Subsec:Stablization}
MViS implicitly obtains temporal stabilization through three mechanisms: 
1) overlapping windows ($S_t$ contains times $t,t\!+\!1$, while $S_{t\!+\!1}$ contains $t+1,t+2$, \etc); 
2) the alignment net's inductive smoothness bias, whereby similar images yield similar predicted transformations; and 
3) the temporal alignment terms in the losses.
\section{Multiview Moving Camera Dataset (MMCD)}
To our knowledge, no available dataset supports multiple video stitching with $N>2$ freely moving cameras. 
Thus, we created MMCD, a new multiview dataset.
MMCD contains 53 scenes, each with 3--6 approximately synchronized handheld videos captured on diverse mobile devices under varied environmental and lighting conditions. 
Camera models, resolutions, and motions vary substantially, providing a broad evaluation of MViS. 
Overall, MMCD contains 221 videos and 47,040 frames. 
It is divided into 4 categories based on $N$, the number of cameras: 22 scenes with $N=3$, 10 with $N=4$, 11 with $N=5$, and 10 with $N=6$. 
Same-scene videos have the same length, with recordings between 5 and 10 seconds. 
For video-length statistics across different values of $N$ and overlap statistics, see \autoref{sec:mmcd_stats}. 
To facilitate quantitative evaluation, we manually annotated keypoints in the overlap regions of the first frames and propagated them through the sequences using CoTracker3~\cite{Karaev:ICCV:2025:cotracker3}. 
The annotations, used solely for evaluation, are available with the dataset.

\section{Experiments and Results}

In all experiments, we optimize MViS using ADAM with a ReduceLROnPlateau scheduler and learning rates of $5\!\times\!10^{-4}$ (Stage~1) and $5\!\times\!10^{-5}$ (Stage~2). 
Batch size is 6 for $N=2$ and 4 for $N>2$, and each stage is optimized for 600 epochs on an RTX 6000 GPU. 
We use RoMa~\cite{Edstedt:CVPR:2024:RoMa} as the image matcher; MViS is not tied to this choice, and in \autoref{sec:matchers} we report results using LoFTR~\cite{sun:2021:loftr}. 
In Stage~2, we use normalized RGB rather than deep features; DINO~\cite{Caron:ICCV:2021:Dino} features did not improve results in our experiments (likely due to their lower spatial resolution) and increased memory cost.

\begin{table}[t]
\caption{$N=2$. Comparison across the 4 categories
in the StabStitch-D dataset~\cite{Nie:ECCV:2024:stabstitch}. The PSNR (\textuparrow)  and SSIM (\textuparrow) values 
were averaged over all of the frames in the category.}
\label{Tab:Pairs}
\centering
\footnotesize
\renewcommand{\arraystretch}{0.93} 

\resizebox{0.7\linewidth}{!}{
\begin{tabular}{
@{}l
cc
cc
cc
cc
@{}} 
\toprule
\multirow{2}{*}{Method}
& \multicolumn{2}{c}{Regular}
& \multicolumn{2}{c}{Low-Light}
& \multicolumn{2}{c}{Low-Texture}
& \multicolumn{2}{c}{Fast-Moving} \\
& {\scriptsize PSNR} & {\scriptsize SSIM} 
& {\scriptsize PSNR} & {\scriptsize SSIM} 
& {\scriptsize PSNR} & {\scriptsize SSIM} 
& {\scriptsize PSNR} & {\scriptsize SSIM} \\
\midrule
LCP~\cite{Jia:CVPR:2021:LCP} & {\scriptsize 24.22} & {\scriptsize .812} & {\scriptsize -} & {\scriptsize -} & {\scriptsize -} & {\scriptsize -} & {\scriptsize 23.88} & {\scriptsize .813} \\
UDIS\texttt{++}~\cite{Nie:2023:udis++}   & {\scriptsize 23.19} & {\scriptsize .785} & {\scriptsize 31.09} & {\scriptsize .936} & {\scriptsize 29.98} & {\scriptsize .906} & {\scriptsize 21.56} & {\scriptsize .756} \\
StabStitch~\cite{Nie:ECCV:2024:stabstitch}  & {\scriptsize 24.64} & {\scriptsize .832} & {\scriptsize 34.51} & {\scriptsize .958} & {\scriptsize 33.63} & {\scriptsize .927} & {\scriptsize 23.36} & {\scriptsize .787} \\
StabStitch\texttt{++}~\cite{Nie:TPAMI:2025:stabstitch++} & {\scriptsize \textbf{25.51}} & {\scriptsize .837} & {\scriptsize 35.10} & {\scriptsize .958} & {\scriptsize \textbf{34.23}} & {\scriptsize .928} & {\scriptsize 25.64} & {\scriptsize .830} \\
\midrule
\textbf{MViS (Ours)}& {\scriptsize 25.47} & {\scriptsize \textbf{.846}} & {\scriptsize \textbf{36.51}} & {\scriptsize \textbf{.960}} & {\scriptsize 30.80} & {\scriptsize \textbf{.932}} & {\scriptsize \textbf{25.76}} & {\scriptsize \textbf{.850}} \\ 
\bottomrule
\end{tabular}
}
\end{table}

We also provide videos in the Supplemental Material 
and at our project page, \url{https://bgu-cs-vil.github.io/MViS-project/}, 
to demonstrate MViS's temporal coherence and handling of time-varying view-intersection graphs.

\textbf{For $\mathbf{N=2}$}, we evaluate on the StabStitch-D benchmark~\cite{Nie:ECCV:2024:stabstitch} and compare with LCP~\cite{Jia:CVPR:2021:LCP}, UDIS++~\cite{Nie:2023:udis++}, StabStitch~\cite{Nie:ECCV:2024:stabstitch}, and StabStitch++~\cite{Nie:TPAMI:2025:stabstitch++}. 
We report Peak Signal-to-Noise Ratio (PSNR) and Structural Similarity (SSIM). 
\autoref{Tab:Pairs} shows that MViS achieves the best SSIM in all categories and the best PSNR in 2/4; all non-MViS numbers are taken from~\cite{Nie:ECCV:2024:stabstitch,Nie:TPAMI:2025:stabstitch++}. 
The `-' symbol indicates that LCP failed in that category.

\textbf{For $\mathbf{N>2}$}, no existing video stitching method handles freely moving cameras: StabStitch++ supports only $N=3$ and requires fixed, known camera ordering. 
We therefore construct a new benchmark for this setting, based on the annotated MMCD.

\begin{table}[b]
\caption{
The Multi-Camera Benchmark. Results on the MMCD Dataset. $N$ is the \# of videos
per scene.  
The values shown, 
 Stab. (\textuparrow), 
 GV (\textdownarrow) and 
 CMSA (\textdownarrow), are averaged over the scenes (on which the respective method succeeded without collapsing)
with the same $N$.
 StabStitch++'s high GV (27.8) on $N=3$ reflects severe misalignment in scenes where the assumed camera ordering is violated. StabStitch++'s code does not support $N>3$.
}
\renewcommand{\arraystretch}{0.97}
\label{Table:Multiview}
\centering
\footnotesize
\resizebox{\linewidth}{!}{
\begin{tabular}{
  w{c}{0.6cm}
  l
  c@{\hskip 3pt}c@{\hskip 3pt}c@{\hskip 3pt}c
  @{\hskip 10pt}
  c@{\hskip 3pt}c@{\hskip 3pt}c@{\hskip 3pt}c
  @{\hskip 10pt}
  c@{\hskip 3pt}c@{\hskip 3pt}c@{\hskip 3pt}c
  @{\hskip 10pt}
  c@{\hskip 3pt}c@{\hskip 3pt}c@{\hskip 3pt}c
}
\toprule
&\multirow{2}{*}{\normalsize Method}
& \multicolumn{4}{c}{\normalsize $N=3$}
& \multicolumn{4}{c}{\normalsize $N=4$}
& \multicolumn{4}{c}{\normalsize $N=5$}
& \multicolumn{4}{c}{\normalsize $N=6$} \\
& & {\scriptsize GV} &{\scriptsize Stab.} & {\scriptsize CMSA}&
{\scriptsize Succ.}  & {\scriptsize GV} &{\scriptsize Stab.} & {\scriptsize CMSA}
& {\scriptsize Succ.}  & {\scriptsize GV} &{\scriptsize Stab.} & {\scriptsize CMSA}
& {\scriptsize Succ.}  & {\scriptsize GV} &{\scriptsize Stab.} & {\scriptsize CMSA}  &{\scriptsize Succ.}  \\

\midrule
\multirow{6}{*}{\rotatebox[origin=c]{90}{\scriptsize MIS}} 
& {AutoStitch ~\cite{Brown:IJCV:2007:AutoStitch}} & 3.3 & .847 & 635 & \scriptsize 14/22 & \textbf{2.4} & .901 & 217 & \scriptsize 6/10 & 3.1 & .834 & {\scriptsize >1k} & \scriptsize 9/11 & 3.3 & .924 & 111 & \scriptsize 5/10 \\
& {APAP ~\cite{Zaragoza:CVPR:2013:APAP}} & 2.9 & .850 & {\scriptsize >1k} & \scriptsize 19/22 & 5. & .837 & {\scriptsize >1k} & \scriptsize 9/10 & 2.4 & .846 & {\scriptsize >1k} & \scriptsize \textbf{11/11} & 4.0 & .875 & {\scriptsize >1k} & \scriptsize 8/10 \\
& {GSP ~\cite{Chen:ECCV:2016:GSP}} & 2.8 & .648 & {\scriptsize >1k} & \scriptsize 19/22 & 2.5 & .649 & {\scriptsize >1k} & \scriptsize 9/10 & \textbf{2.1} & .625 & {\scriptsize >1k} & \scriptsize \textbf{11/11} & 3.0 & .672 & {\scriptsize >1k} & \scriptsize 9/10 \\
& {SPW ~\cite{Liao:TIP:2019:SPW}} & 4.8 & .807 & {\scriptsize >1k} & \scriptsize \textbf{22/22} & 10.4 & .716 & {\scriptsize >1k} & \scriptsize \textbf{10/10} & 2.5 & .835 & {\scriptsize >1k} & \scriptsize 10/11 & 14.8 & .745 & {\scriptsize >1k} & \scriptsize \textbf{10/10} \\
& {GES-GSP ~\cite{Du:CVPR:2022:GESGSP}} & 2.9 & .618 & {\scriptsize >1k} & \scriptsize 19/22 & 2.5 & .668 & {\scriptsize >1k} & \scriptsize 9/10 & 2.4 & .614 & {\scriptsize >1k} & \scriptsize 10/11 & 3.2 & .670 & {\scriptsize >1k} & \scriptsize 9/10 \\
& {OBJ-GSP ~\cite{Cai:AAAI:2025:OBJGSP}} & 2.9 & .621 & {\scriptsize >1k} & \scriptsize 19/22 & 2.5 & .634 & {\scriptsize >1k} & \scriptsize 9/10 & 2.2 & .683 & {\scriptsize >1k} & \scriptsize 8/11 & 3.1 & .681 & {\scriptsize >1k} & \scriptsize 6/10 \\
\midrule
\multirow{3}{*}{\rotatebox[origin=c]{90}{\scriptsize MIS\texttt{+}TR}} 
& {GSP\texttt{+}TR} & 3.1 & .863 & 62.9 & \scriptsize 19/22 & 2.8 & .898 & 52.2 & \scriptsize 9/10 & 3.2 & .817 & 93.6 & \scriptsize \textbf{11/11} & 3.5 & .879 & 38.1 & \scriptsize 9/10 \\
& {GES-GSP\texttt{+}TR} & 3.3 & .849 & 69.4 & \scriptsize 19/22 & 3.1 & .884 & 57.2 & \scriptsize 9/10 & 3.8 & .800 & 100 & \scriptsize \textbf{11/11} & 3.9 & .866 & 39.5 & \scriptsize 9/10 \\
& {OBJ-GSP\texttt{+}TR} & 3.2 & .853 & 75.3 & \scriptsize 19/22 & 3.0 & .879 & 71.6 & \scriptsize 9/10 & 3.5 & .808 & 106 & \scriptsize \textbf{11/11} & 3.7 & .872 & 42.0 & \scriptsize 9/10 \\
\midrule
\multirow{3}{*}{\rotatebox[origin=c]{90}{\scriptsize MIS\texttt{+}PS}} 
& {GSP\texttt{+}PS} & \textbf{2.7} & .816 & 216 & \scriptsize 19/22 & \textbf{2.4} & .841 & 251 & \scriptsize 9/10 & \textbf{2.1} & .783 & 577 & \scriptsize \textbf{11/11} & \textbf{2.8} & .867 & 174 & \scriptsize 9/10 \\
& {GES-GSP\texttt{+}PS} & 2.8 & .802 & 206 & \scriptsize 19/22 & 2.5 & .853 & 154 & \scriptsize 9/10 & 2.3 & .796 & 644 & \scriptsize 10/11 & 3.0 & .887 & 90.7 & \scriptsize 9/10 \\
& {OBJ-GSP\texttt{+}PS} & 2.8 & .814 & 177 & \scriptsize 19/22 & \textbf{2.4} & .836 & 196 & \scriptsize 9/10 & \textbf{2.1} & .822 & 73.6 & \scriptsize 8/11 & 2.9 & .884 & 101 & \scriptsize 6/10 \\
\midrule
\multirow{2}{*}{\rotatebox[origin=c]{90}{\scriptsize VS}} 
& {StabStitch\texttt{++}~\cite{Nie:TPAMI:2025:stabstitch++}} & 27.8 & .799 & 20.0 & \scriptsize \textbf{22/22} & {N/A} & {N/A} & {N/A} & {N/A} & {N/A} & {N/A} & {N/A} & {N/A} & {N/A} & {N/A} & {N/A} & {N/A} \\
& \textbf{MViS (Ours)} & 3.1 & \textbf{.917} & \textbf{2.5} & \scriptsize \textbf{22/22} & 2.8 & \textbf{.925} & \textbf{14.2} & \scriptsize \textbf{10/10} & 2.3 & \textbf{.874} & \textbf{11.} & \scriptsize \textbf{11/11} & 3.0 & \textbf{.930} & \textbf{2.4} & \scriptsize \textbf{10/10} \\
\bottomrule
\end{tabular}
}
\end{table}

\subsection{The Multi-Camera Benchmark}

{\textbf{Metrics.} 
For \(N>2\), where PSNR and SSIM are not directly applicable, we report the stability score~\cite{Nie:TIP:2017:dynamic}, computed on annotated keypoint trajectories, and our new \emph{Geometric Variance} (GV) metric, which quantifies cross-view consistency of aligned keypoints (details in~\autoref{sec:geovar}). 
We measure temporal smoothness using the Mean Square Acceleration of video-corner trajectories (CMSA).

As \textbf{baselines}, we evaluate MIS methods and video-adapted extensions we built. 
Since some methods, but not MViS, collapse on certain scenes, we also report the number of successful scenes (Succ.) in each category. 
For each method, metrics are averaged only over successful scenes, slightly favoring baselines because their failures are typically on hard scenes; full results and the detailed evaluation protocol appear in the Appendix. 
We now analyze the results summarized in \autoref{Table:Multiview}.}

\textbf{1. Comparison with MIS:}
Per-frame MIS methods~\cite{Brown:IJCV:2007:AutoStitch,Zaragoza:CVPR:2013:APAP,Chen:ECCV:2016:GSP,Liao:TIP:2019:SPW,Du:CVPR:2022:GESGSP,Cai:AAAI:2025:OBJGSP} may produce plausible single-frame panoramas, but their videos exhibit severe temporal inconsistency: their stability is substantially lower than MViS and their CMSA (\textdownarrow) is orders of magnitude higher. This shows that strong per-frame alignment does not imply temporal coherence.

\textbf{2. Comparison with temporally extended MIS variants:}
We augment key MIS methods (GSP, GES-GSP, OBJ-GSP) with two stabilization strategies: (1) temporal regularization penalizing changes in estimated transformations between consecutive frames (TR), and (2) post-hoc smoothing of per-frame transformations (PS). 
Both improve stability over per-frame baselines but still lose to MViS. 
This is expected: in multiple moving-camera videos, the optimal stabilization strength is not uniform, since different cameras, and hence different panoramic regions, can have different and possibly time-dependent motions.

\textbf{3. Comparison with Video Stitching (VS):}
Among existing video stitching methods, StabStitch++ is the only one supporting more than two inputs, but it is limited to $N=3$ and requires a fixed spatial ordering of the cameras, as noted in its GitHub repository: \textit{``video1 should overlap with video2, and video2 should overlap with video3''}. 
This assumes that the cameras' relative ordering remains fixed, whereas in freely moving multi-camera capture the view-intersection graph can change over time. 
Thus, StabStitch++ is not fully applicable to three moving cameras; in practice, when cameras swap order, its output degrades. 
MViS outperforms StabStitch++ in both alignment and stability metrics. 
For $N>3$, no existing video stitching method is applicable, so we report results only for MViS.
\captionsetup[subfigure]{skip=1pt}
\begin{figure}[t]
   \centering

  % -------- Row 1 --------
  
  \begin{subfigure}[b]{0.2435\linewidth}
    \includegraphics[width=\linewidth,height=1.7cm]{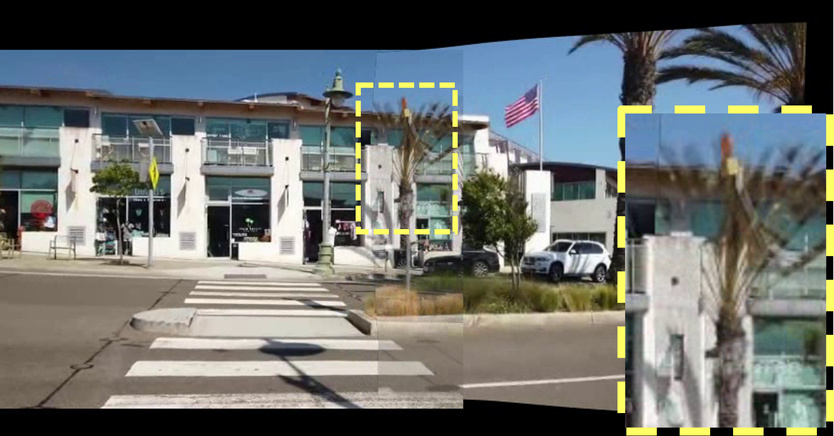}
    \caption{\footnotesize StabStitch++}
    \label{fig:vis_a}
  \end{subfigure}
  \begin{subfigure}[b]{0.2435\linewidth}
    \includegraphics[width=\linewidth,height=1.7cm]{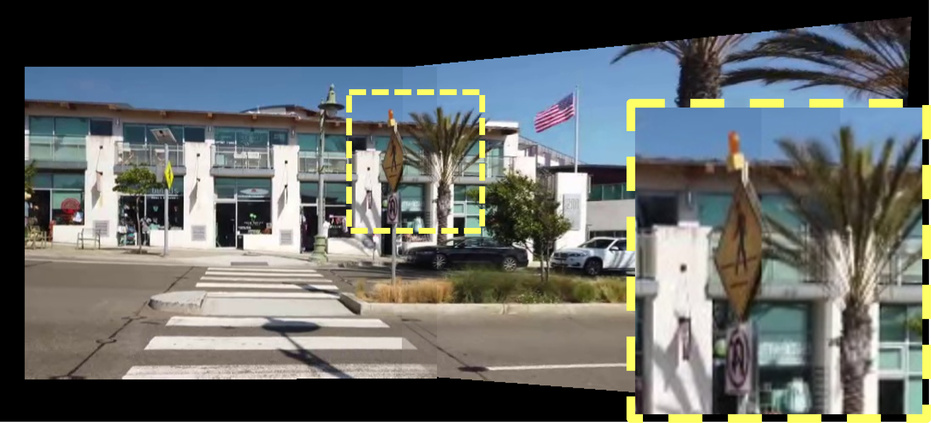}
    \caption*{\footnotesize MViS (Ours)}
  \end{subfigure}
  \begin{subfigure}[b]{0.2435\linewidth}
    \includegraphics[width=\linewidth,height=1.7cm]{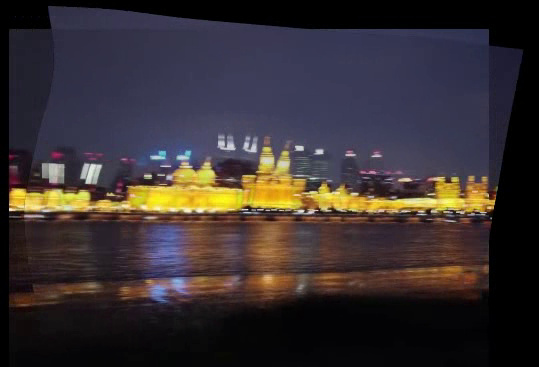}
    \caption{\footnotesize StabStitch++}
    \label{fig:vis_b}
  \end{subfigure}
  \begin{subfigure}[b]{0.2435\linewidth}
    \includegraphics[width=\linewidth,height=1.7cm]{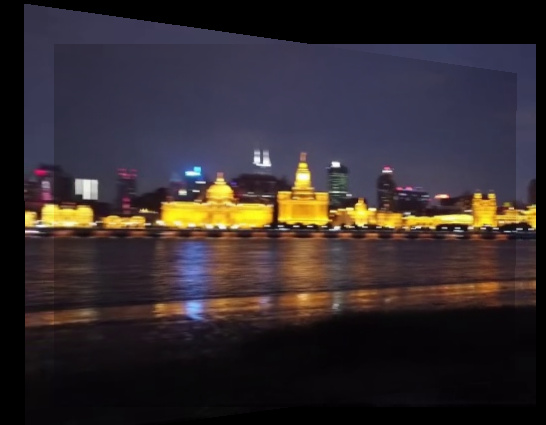}
    \caption*{\footnotesize MViS (Ours)}
  \end{subfigure}

  % -------- Row 2 --------
  \begin{subfigure}[b]{0.2435\linewidth}
    \includegraphics[width=\linewidth,height=1.7cm]{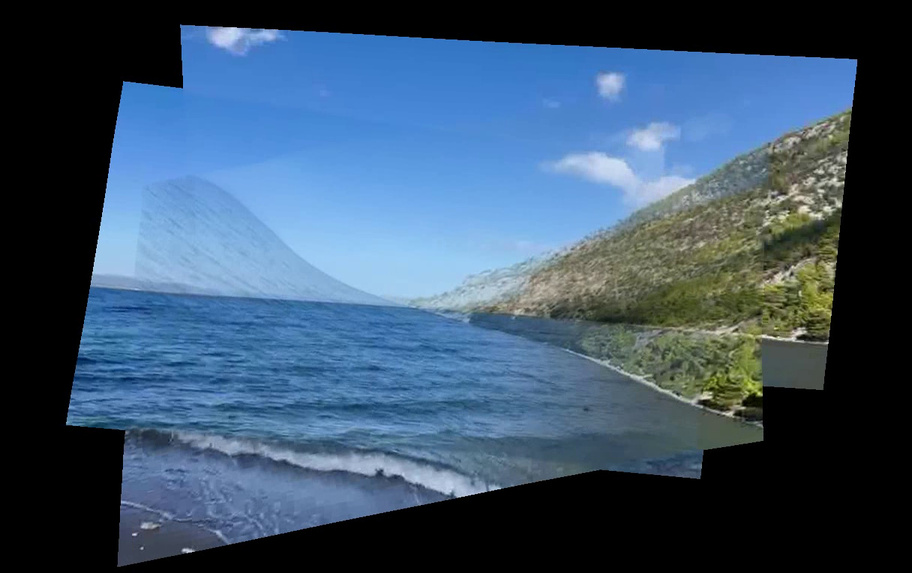}
    \caption{\footnotesize StabStitch++}
    \label{fig:vis_c}
  \end{subfigure}
  \begin{subfigure}[b]{0.2435\linewidth}
    \includegraphics[width=\linewidth,height=1.7cm]{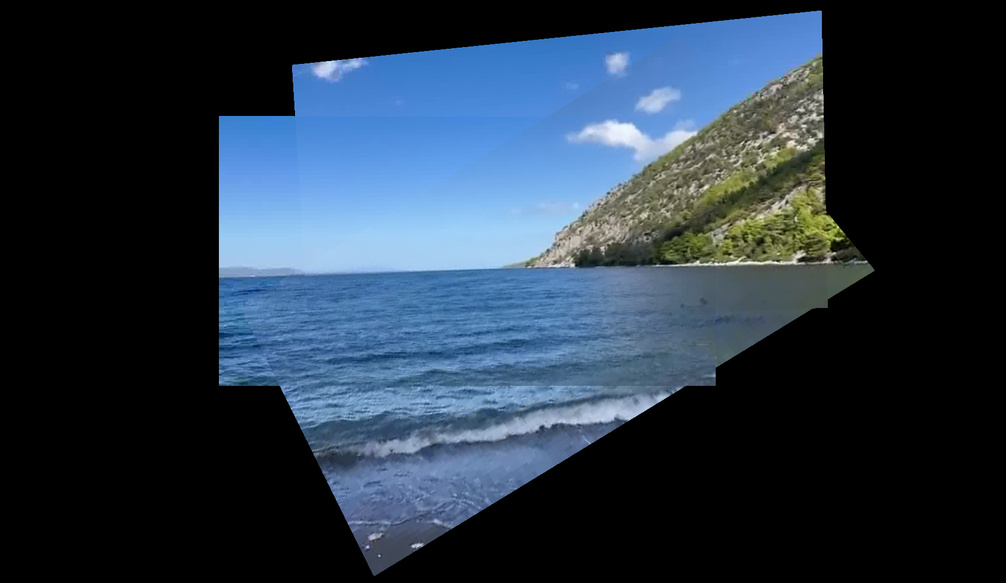}
    \caption*{\footnotesize MViS (Ours)}
  \end{subfigure}
  \begin{subfigure}[b]{0.2435\linewidth}
    \includegraphics[width=\linewidth,height=1.7cm]{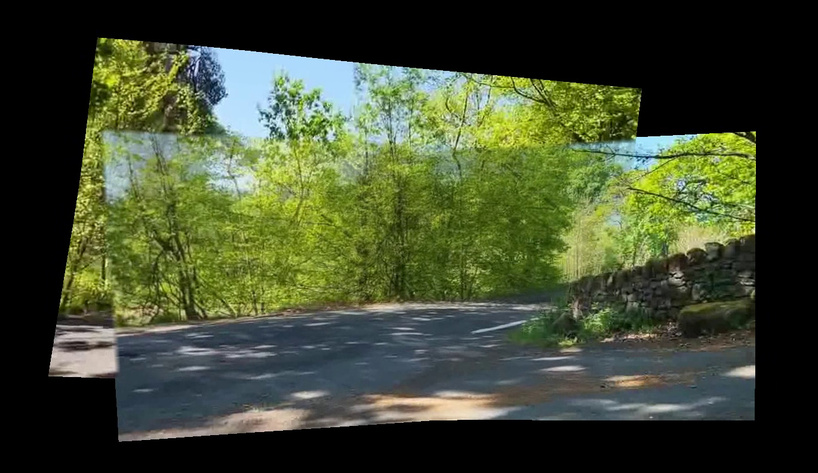}
    \caption{\footnotesize StabStitch++}
    \label{fig:vis_d}
  \end{subfigure}
  \begin{subfigure}[b]{0.2435\linewidth}
    \includegraphics[width=\linewidth,height=1.7cm]{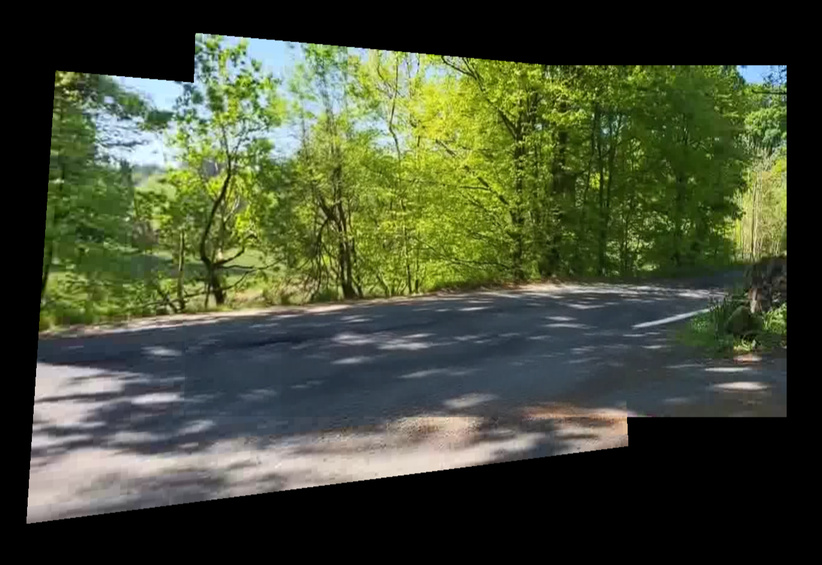}
    \caption*{\footnotesize MViS (Ours)}
  \end{subfigure}
  %\vspace{-2pt}
    % -------- Row 3  --------
  \begin{subfigure}[b]{0.2435\linewidth}
    \includegraphics[width=\linewidth,height=1.7cm]{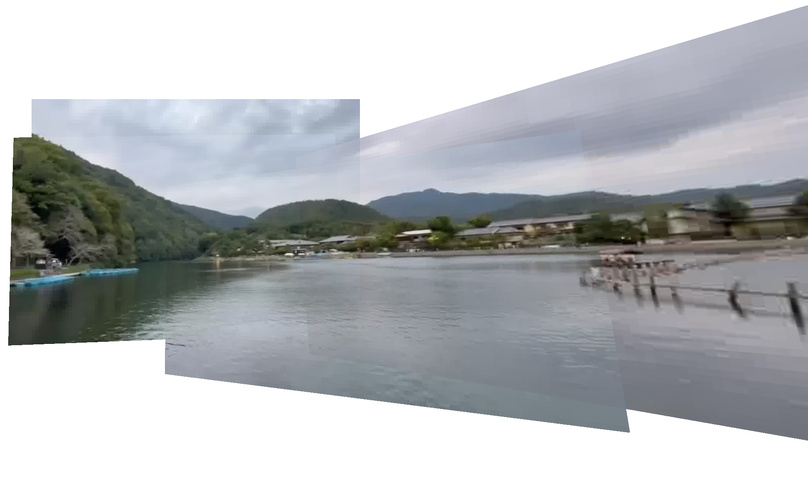}
    \caption{\footnotesize $N=4$ (Ours)}
    \label{fig:vis_e}
  \end{subfigure}
  \begin{subfigure}[b]{0.2435\linewidth}
    \includegraphics[width=\linewidth,height=1.7cm]{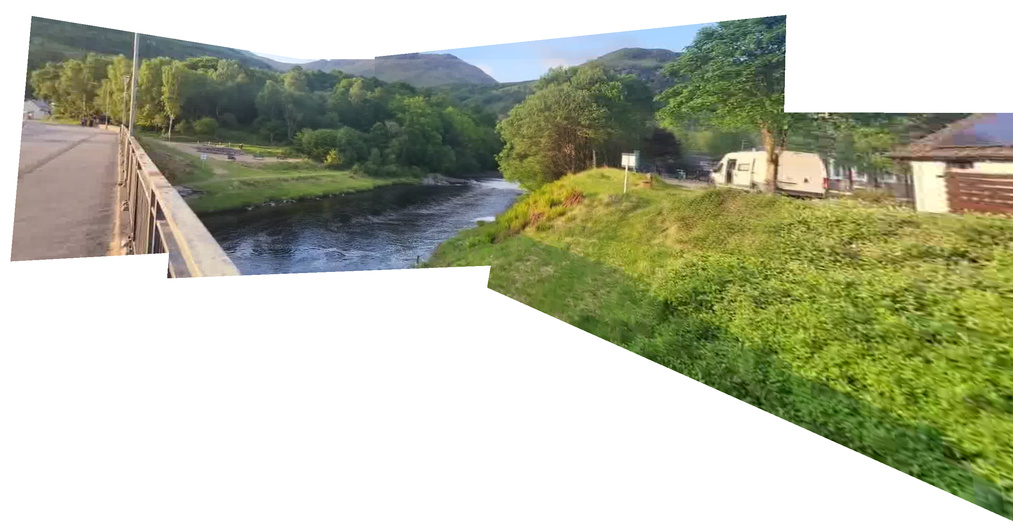}
    \caption{\footnotesize $N=5$ (Ours)}
    \label{fig:vis_f}
  \end{subfigure}
  \begin{subfigure}[b]{0.2435\linewidth}
    \includegraphics[width=\linewidth,height=1.7cm]{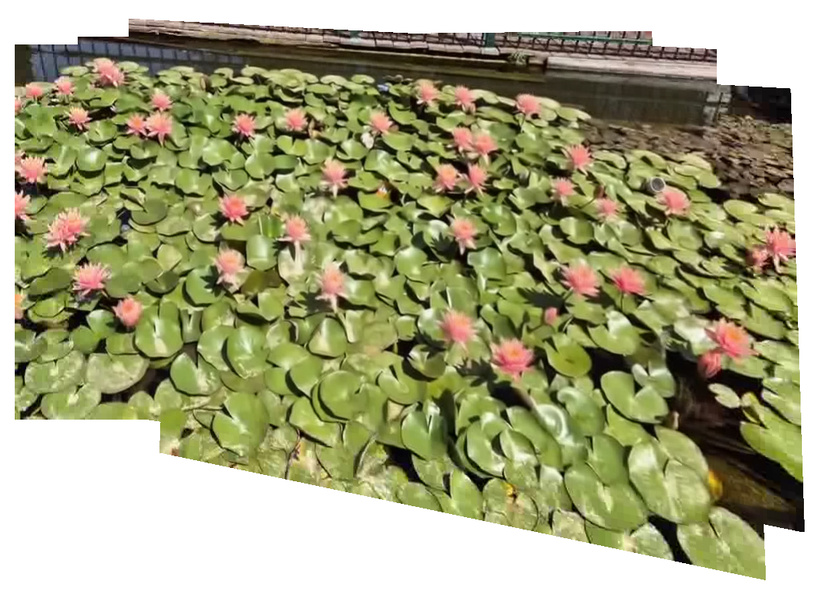}
    \caption{\footnotesize $N=6$ (Ours)}
    \label{fig:vis_g}
  \end{subfigure}
  \begin{subfigure}[b]{0.2435\linewidth}
    \includegraphics[width=\linewidth,height=1.7cm]{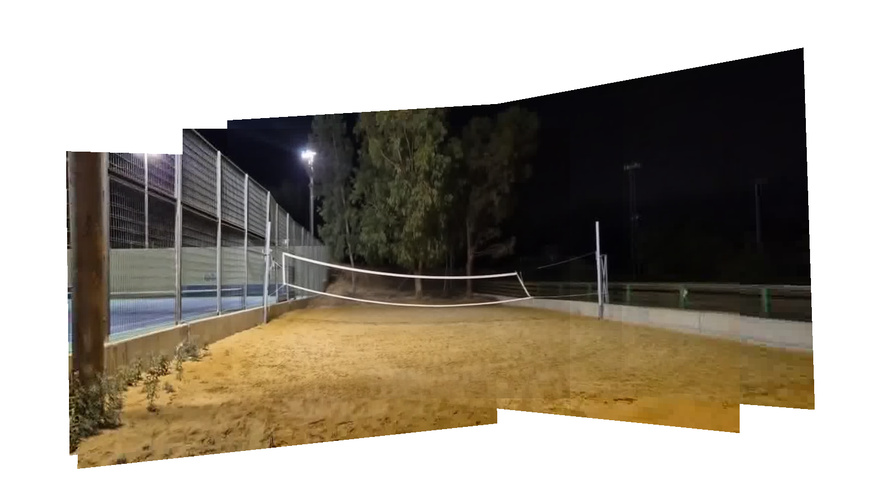}
    \caption{\footnotesize $N=8$ (Ours)}
    \label{fig:vis_h}
  \end{subfigure}
  \caption{ \textbf{Visual comparisons.} \emph{Top row:} $N=2$ cases from the StabStitch-D dataset~\cite{Nie:ECCV:2024:stabstitch}.
  {StabStitch++'s severe misalignment makes the traffic sign vanish (\subref{fig:vis_a}) and causes ghosting of the buildings (\subref{fig:vis_b}).}
  \emph{Middle:} $N=3$ cases from the MMCD dataset. \emph{Bottom:} 
  MViS results on MMCD cases with
  $N=4,5,6,8$. The $N=8$ is an extended scene from the $N=6$ category. 
  }
  \label{Fig:VisualCompare}
\end{figure}

\subsection{Qualitative Results, Runtime and Memory Usage}
\begin{figure}[t]
\centering
\setlength{\tabcolsep}{2pt} 
\resizebox{\linewidth}{!}{
\begin{tabular}{*{4}{c}}   

\includegraphics[width=3.8cm, height=2.0cm,    viewport=220 510 1790 1390,   clip]{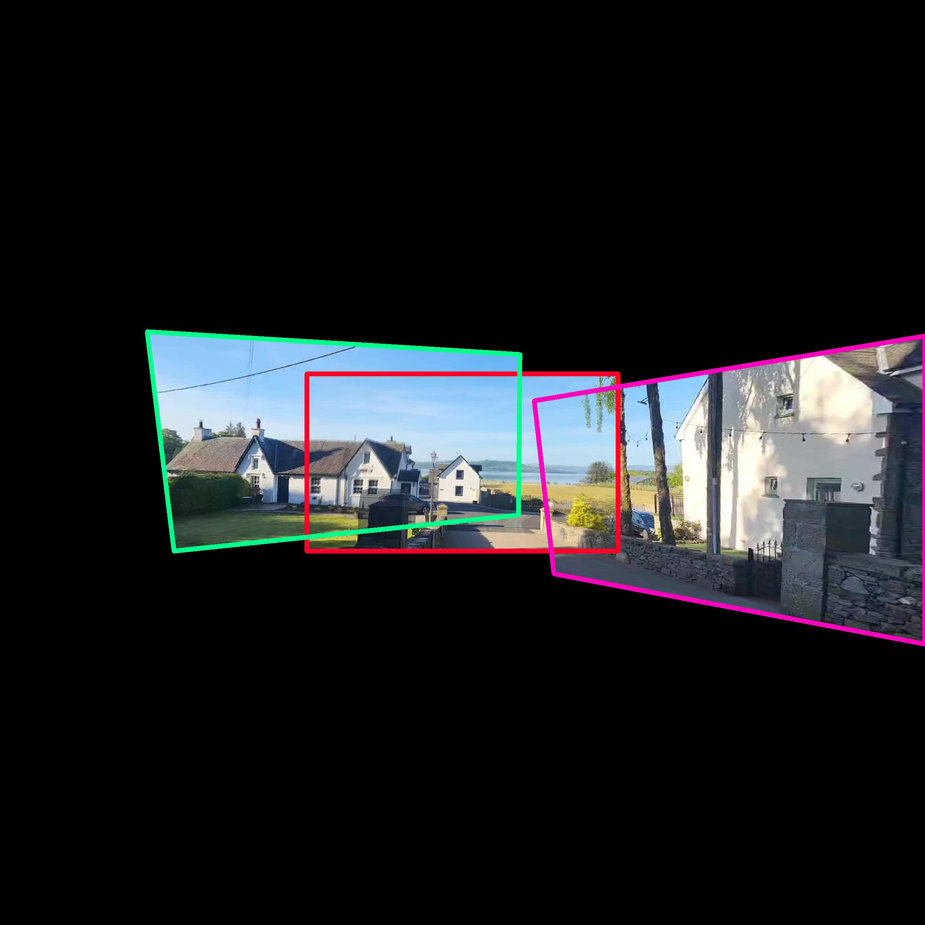} &
\includegraphics[width=3.8cm, height=2.0cm,    viewport=220 510 1790 1390,   clip]{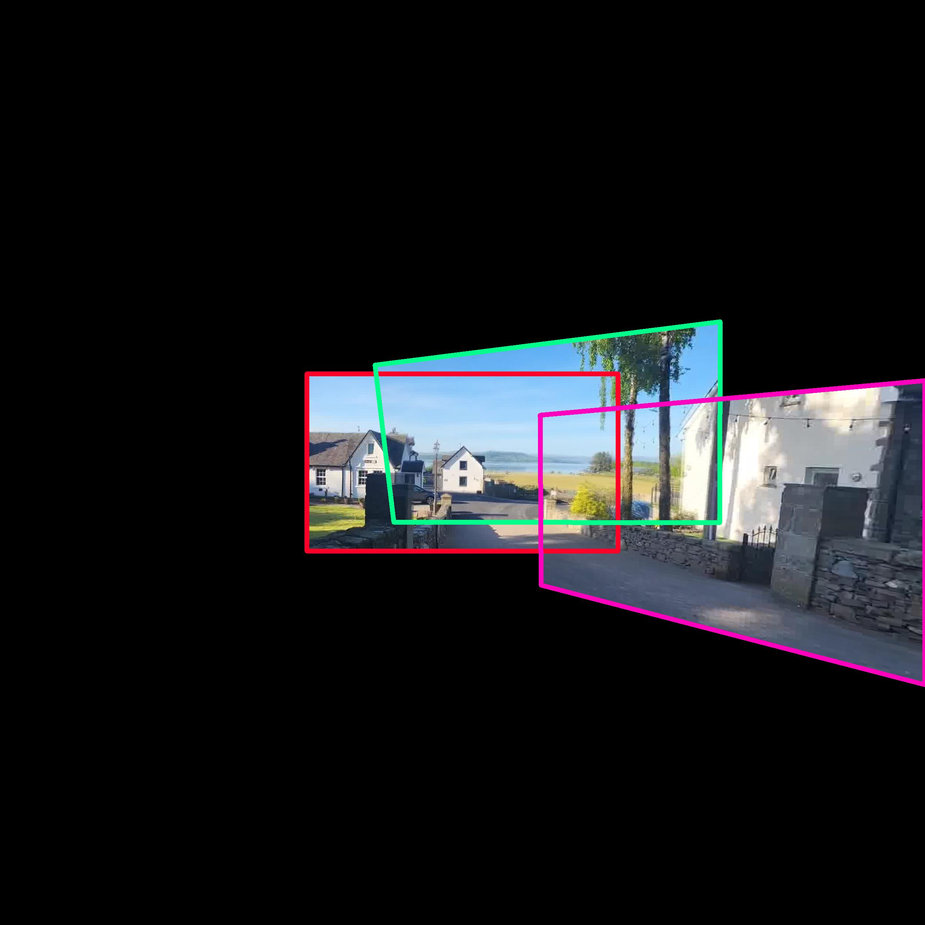} &
\includegraphics[width=3.8cm, height=2.0cm,    viewport=220 510 1790 1390,   clip]{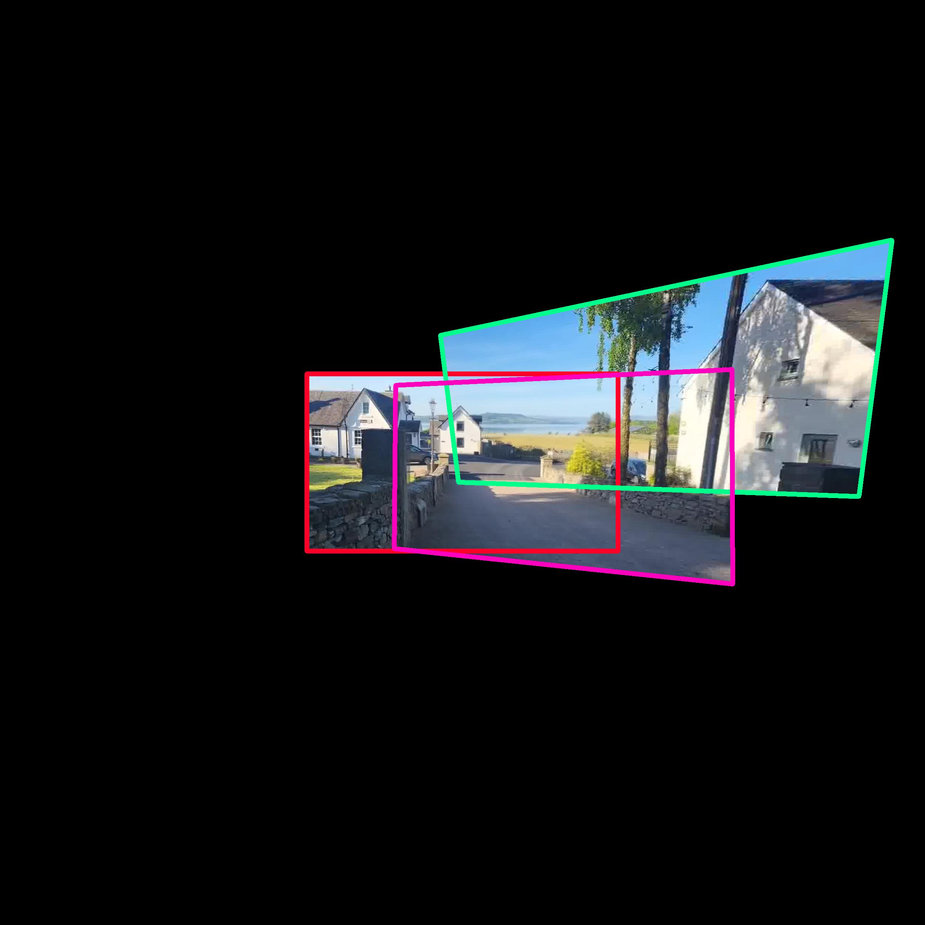} &

\includegraphics[width=3.8cm, height=2.0cm,    viewport=220 510 1790 1390,   clip]{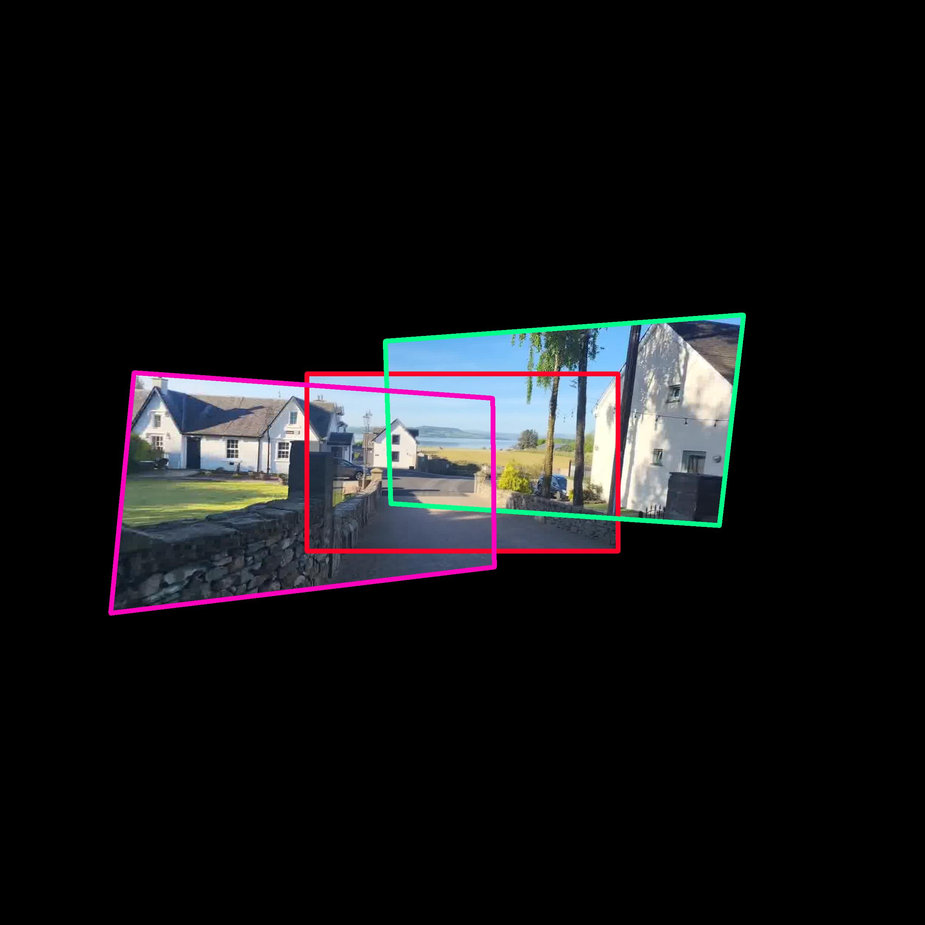} \\[-1.3mm]

\footnotesize{$t=0$} & 
\footnotesize{$t=80$} & 
\footnotesize{$t=120$} & 
\footnotesize{$t=180$} \\

\includegraphics[width=3.8cm, height=2.0cm,    viewport=180 330 1702 1222,   clip]{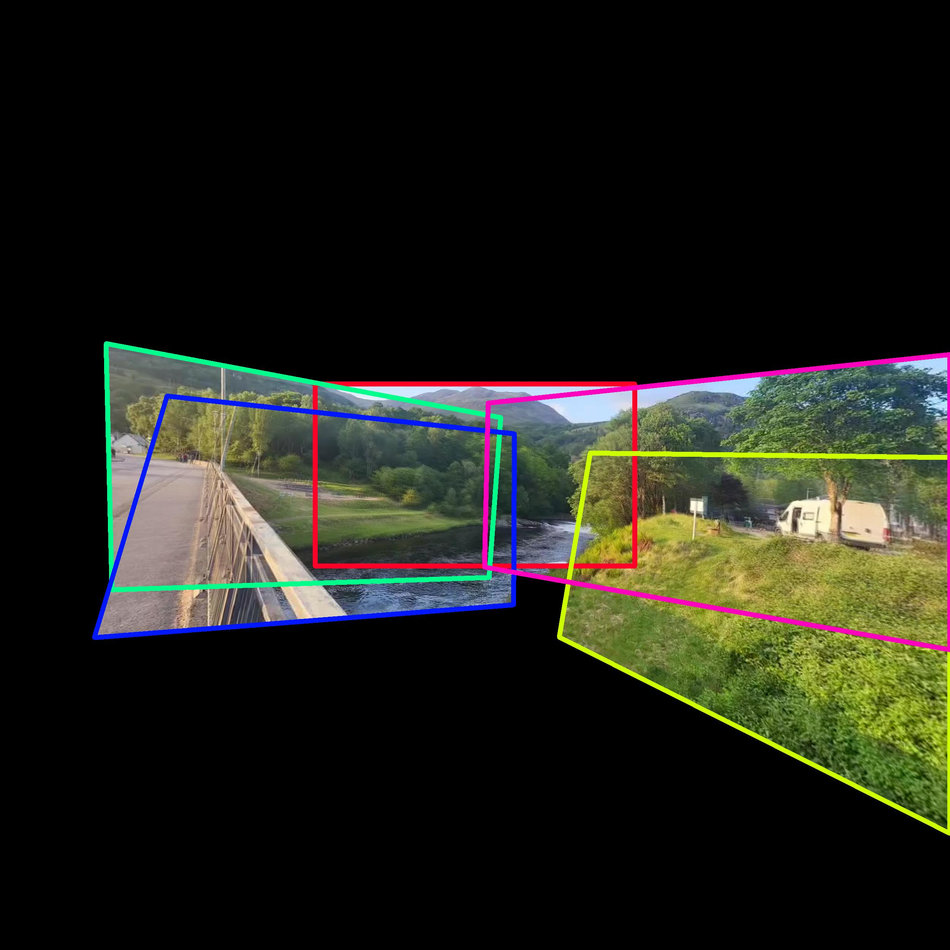} &

\includegraphics[width=3.8cm, height=2.0cm,    viewport=180 330 1702 1222,   clip]{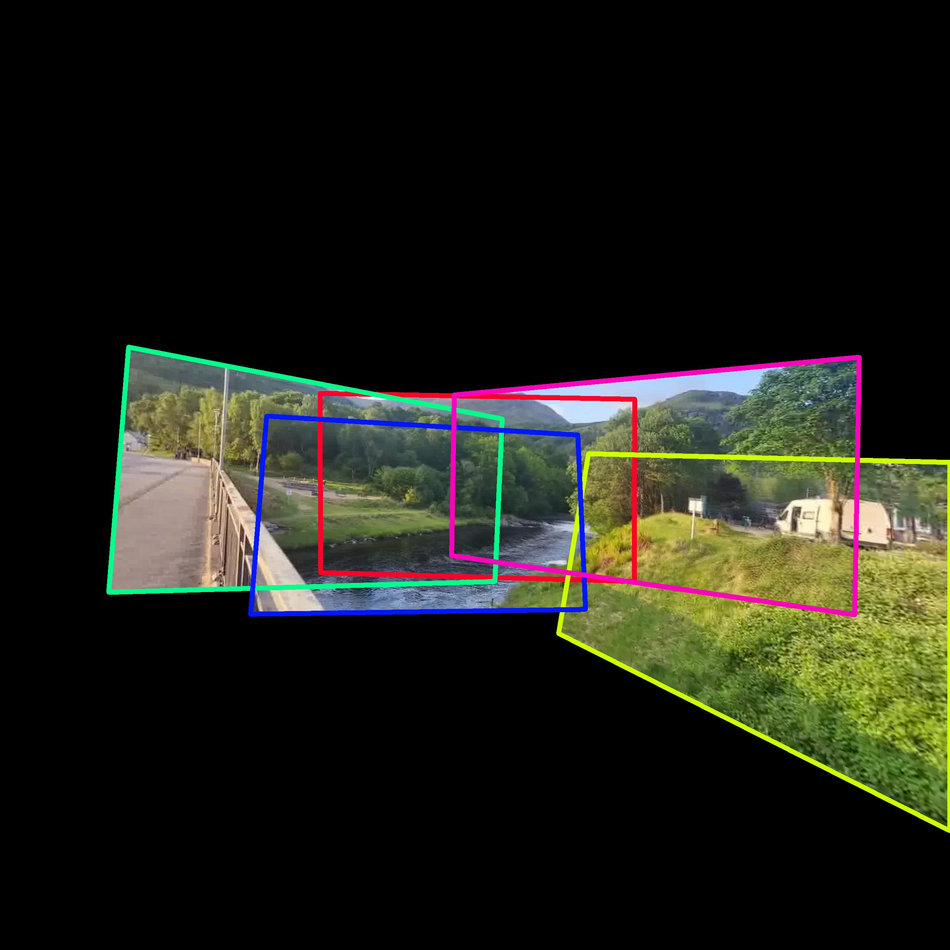} &

\includegraphics[width=3.8cm, height=2.0cm,    viewport=180 330 1702 1192,   clip]{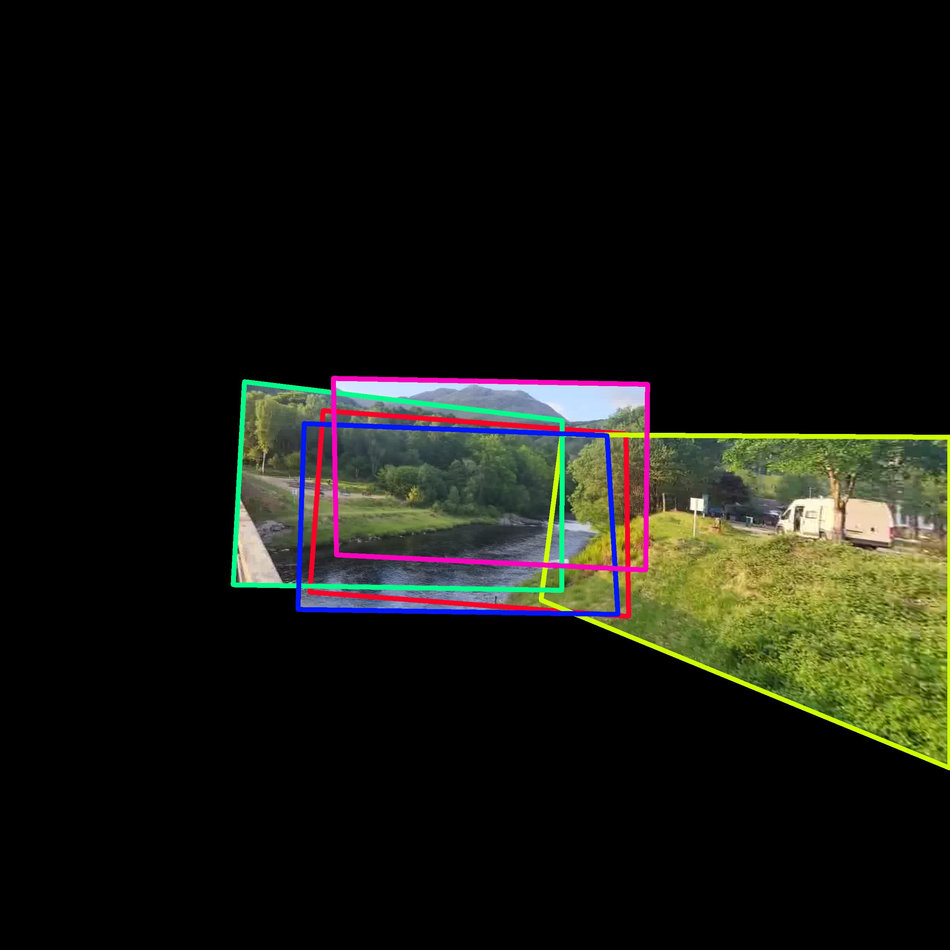} &

\includegraphics[width=3.8cm, height=2.0cm,    viewport=180 330 1702 1222,   clip]{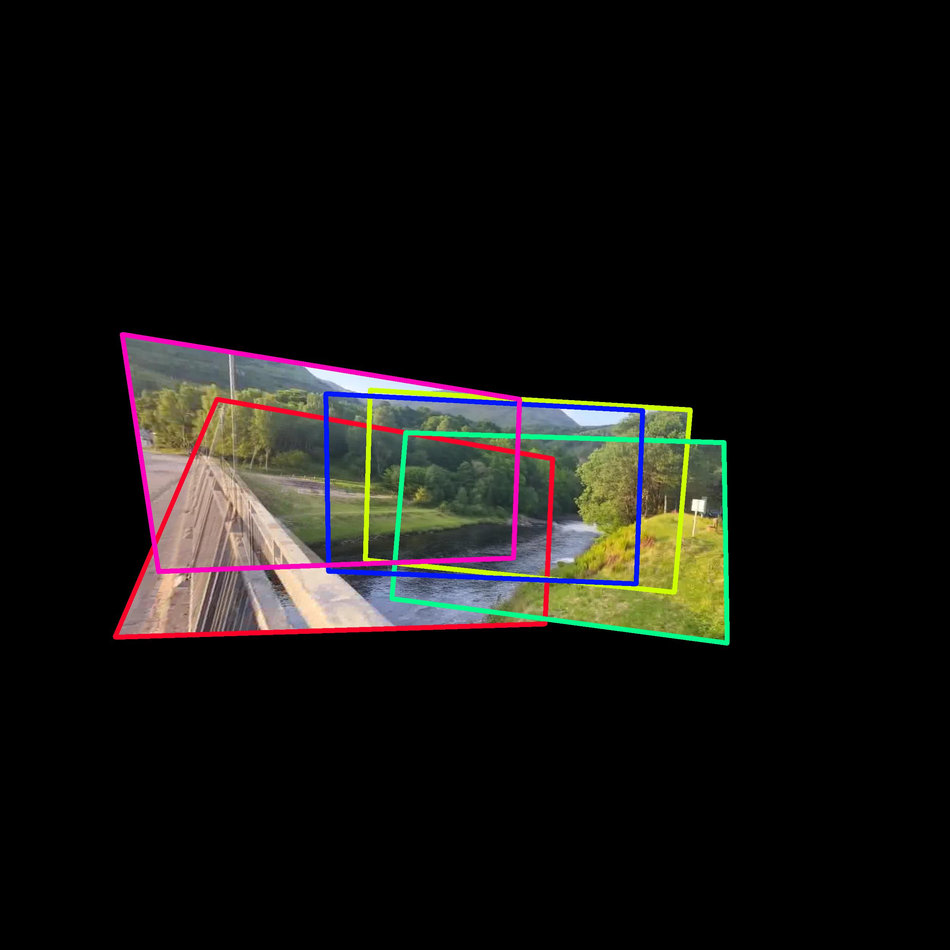} \\[-1.2mm]

\footnotesize{$t=0$} & 
\footnotesize{$t=30$} & 
\footnotesize{$t=60$} &
\footnotesize{$t=120$} \\

\end{tabular}
}
\vspace{-2.5mm}  
\caption{Dynamically changing view-intersection graph, MViS (Ours). Camera outlines are color-coded to highlight overlaps. Each camera is assigned a consistent color throughout the video.}
  \label{fig:view_graph}
\end{figure}

\FigureRef{Fig:VisualCompare} compares MViS with StabStitch++ ($N\in\set{2,3}$) and shows that MViS maintains visual consistency even under rapid motion. 
\FigureRef{fig:view_graph} shows two scenes in which the view-intersection graph changes over time: as the cameras move, overlaps between views appear and disappear, and MViS adapts the panorama accordingly while keeping every camera consistently placed throughout the video.
Additional visualizations, including $N>3$ cases, appear in the Supplemental Material and on our project webpage.

\textbf{Runtime.} We report time per frame, computed as total runtime divided by $N$ times the number of frames: Stage~1 takes 0.5--0.6 s, Stage~2 takes 0.7--0.8 s, and preprocessing takes 0.4--0.5 s. 
These times are stable across different $N$ values.

\textbf{Memory.} Memory usage is independent of the original image resolution and scales linearly with $N$.

\subsection{Ablation Study and Direct Optimization without a Net}
\autoref{tab:ablation_n3} summarizes an \textbf{Ablation Study} on StabStitch-D ($N=2$) and MMCD ($N>2$). 
Removing Stage~1 causes frequent failures, confirming its importance. 
Removing Stage~2 degrades alignment, especially when inter-view overlap is small. 
Optimizing the Stage~1 and Stage~2 losses simultaneously usually harms performance, highlighting the need for our two-stage approach. 
Using unnormalized RGB also hurts performance. 
As expected, the temporal loss terms mainly affect video stabilization.

To validate the role of the network, we also directly optimize the proposed losses over the homographies, without a neural network. 
We test three variants: geometric loss only, appearance loss only, and geometric-then-appearance sequential optimization.
The resulting videos are spatially aligned but exhibit severe temporal jitter. 
Adding regularization over homographies could reduce this jitter, but would require per-scene hyperparameter tuning and, more problematically, supervision for selecting appropriate values. 
In contrast, MViS obtains temporal smoothness implicitly, partly through the inductive smoothness bias of the network (see \autoref{Sec:Metohd:Subsec:Stablization}): consecutive frames tend to be similar, and so are the predictions. 
Since Stability and GV are computed only in overlap regions, where temporal instability is less visible, their numerical differences remain small despite the clear qualitative difference; CMSA captures it well. 
See the Appendix for visual and quantitative results.

\begin{table}[h]
  \caption{Ablation Study. 
  See text for details. 
  }
  \label{tab:ablation_n3}
  \centering
  \footnotesize

  \renewcommand{\arraystretch}{0.85}
  \resizebox{0.85\linewidth}{!}{
    \begin{tabular}{@{}
  l
  @{\hskip 3pt} c
  @{\hskip 3pt} c
  @{\hskip 6pt} c
  @{\hskip 3pt} c
  @{\hskip 3pt} c
  @{\hskip 6pt} c
  @{\hskip 3pt} c
  @{\hskip 3pt} c
  @{\hskip 6pt} c
  @{\hskip 3pt} c
  @{\hskip 3pt} c
  @{\hskip 6pt} c
  @{\hskip 3pt} c
  @{\hskip 3pt} c
@{}}
    \toprule
    \multirow{1}{*}{Ablation}
& \multicolumn{2}{c}{$N=2$}
& \multicolumn{3}{c}{$N=3$}
& \multicolumn{3}{c}{$N=4$}
& \multicolumn{3}{c}{$N=5$}
& \multicolumn{3}{c}{$N=6$} \\
     & {\scriptsize PSNR}& {\scriptsize SSIM} & {\scriptsize GV} & {\scriptsize Stab.} & {\scriptsize CMSA}& {\scriptsize GV} & {\scriptsize Stab.} & {\scriptsize CMSA}& {\scriptsize GV} & {\scriptsize Stab.} & {\scriptsize CMSA}& {\scriptsize GV} & {\scriptsize Stab.} & {\scriptsize CMSA}\\
    \midrule

    No stage 1 & 25.22 & .794 & 879 & .886 & {\scriptsize >1k} & {\scriptsize >1k} & .888 & {\scriptsize >1k} & {\scriptsize >1k} & .872 & {\scriptsize >1k} & {\scriptsize >1k} & .909 & {\scriptsize >1k}\\
    No stage 2 & 28.93 & .897 & 3.5 & .915 & {\scriptsize >1k} & 12.5 & \textbf{.928} & 275 & 2.6 & .873 & 264 & 3.3 & \textbf{.930} & 10.8\\
    No temporal loss & 30.13 & .903 & \textbf{3.1} & .911 & 44.9 & \textbf{2.8} & .917 & {\scriptsize >1k} & 2.4 & .870 & 48.4 & \textbf{3.0} & .926 & 31.3\\
    Both stages simultaneously & 25.40 & .779 & 12.9 & .905 & 42.0 & 11.7 & .920 & 131 & 4.6 & .871 & 12.7 & 11.0 & \textbf{.930} & \textbf{0.8}\\
    Unnormalized RGB & 29.18 & .891 & \textbf{3.1} & .914 & 5.8 & 13.9 & .907 & 31.9 & 2.6 & \textbf{.874} & \textbf{7.0} & 3.1 & \textbf{.930} & 2.5\\
    \midrule
    MViS & \textbf{30.24} & \textbf{.904} & \textbf{3.1} & \textbf{.917} & \textbf{2.5} & \textbf{2.8} & .925 & \textbf{14.2} & \textbf{2.3} & \textbf{.874} & 11.0 & \textbf{3.0} & \textbf{.930} & 2.4\\
    \bottomrule
  \end{tabular}
  }
\end{table}

\section{Conclusion}
We presented MViS, a method for stitching videos from multiple freely moving cameras. 
MViS combines a regularization-free joint-alignment formulation with a two-stage optimization over geometric and appearance consistency, enabling accurate stitching across arbitrary numbers of views and flexible view-intersection graphs. 
Experiments show clear advantages over existing methods, particularly for $N>2$, where prior video-stitching methods are either inapplicable or require restrictive fixed-ordering assumptions.

\textbf{Limitations.} 
1) Runtime: optimization time would not suffice for real-time applications.
2) Although the relative-composition framework can support richer invertible transformations, such as nonrigid diffeomorphisms, our current implementation uses homographies, a standard and cost-effective choice that already performs well in our experiments. 
3) When the image matcher fails, e.g., due to a dominant textureless region, the ORTs may be unreliable and performance can temporarily degrade, although MViS typically recovers once reliable matches reappear later in the video.

We are not aware of any negative societal impacts associated with this work.

\clearpage
\begin{ack}
\begin{flushleft}
Shira Ifergane was partially supported by the VATAT
National Excellence Scholarship for MSc Students in AI and Data Science.
Roy Amoyal was partially supported by the Kreitman School of Advanced
Graduate Studies.

The authors declare no competing interests.
\end{flushleft}
\end{ack}
\bibliography{refs}
\bibliographystyle{unsrt}

\clearpage
\appendix

\begin{center}
{\LARGE \textbf{Video Stitching from Multiple Moving Cameras}}\\[12pt]
{\large \textbf{Appendix}}
\end{center}

\vspace{1em}

In this appendix, we provide additional qualitative and quantitative results,  implementation details, further information on the proposed metric, and dataset length statistics.

In addition to this document, we include a set of videos under \texttt{videos/}. The structure and content of the videos are described in \autoref{Sup:supp_videos}.

\renewcommand\thesection{\Alph{section}}
\section*{Content:}
\begin{figure}[t]
\centering
\resizebox{\linewidth}{!}{
\setlength{\tabcolsep}{2pt} 
\begin{tabular}{c*{4}{c}}

% Scene_76
\rotatebox{90}{\parbox{2.0cm}{\centering \small APAP ~\cite{Zaragoza:CVPR:2013:APAP}}} & 
\includegraphics[width=3.8cm, height=2.0cm,    viewport=650 850 1850 1700,   clip]{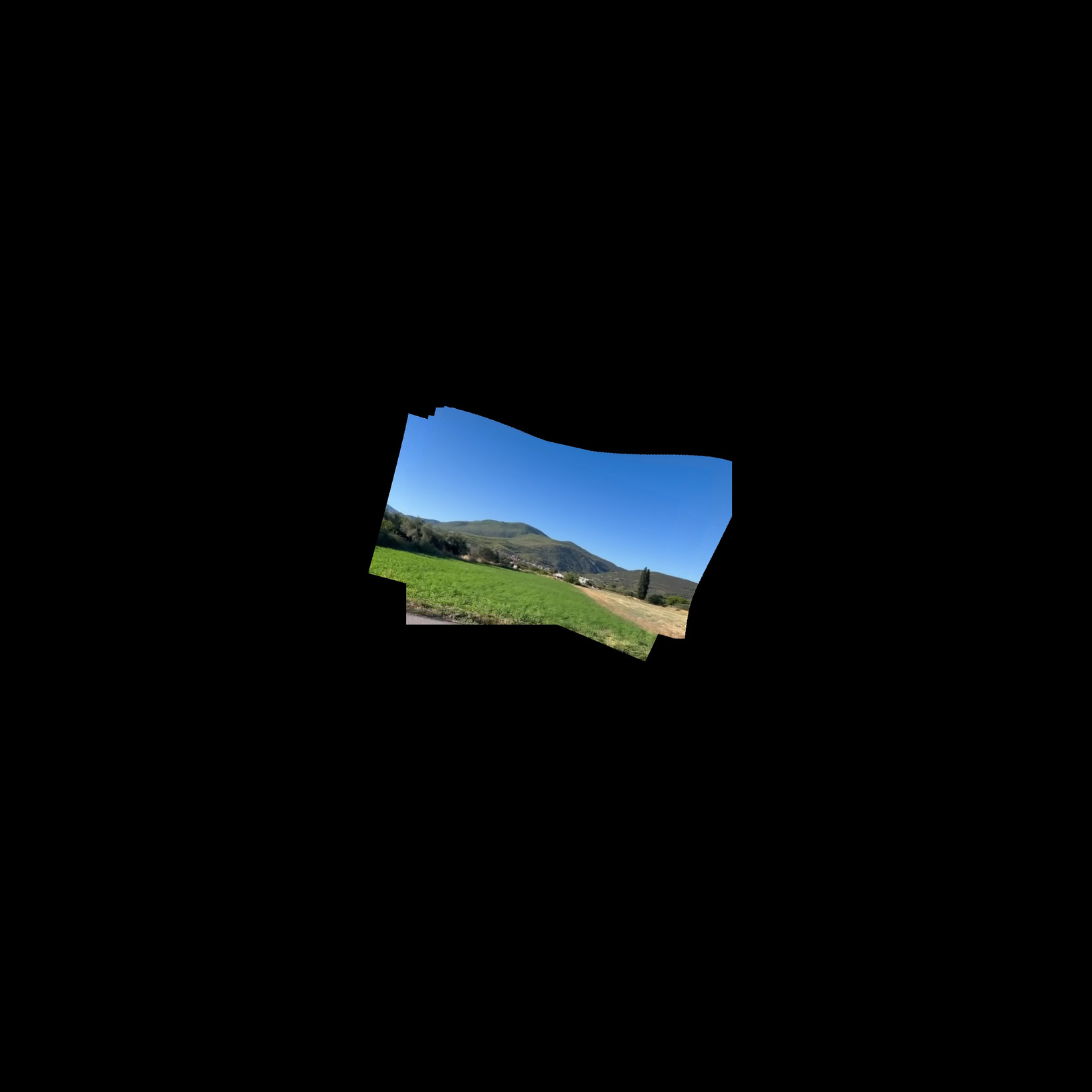} &
\includegraphics[width=3.8cm, height=2.0cm,    viewport=650 850 1850 1700,   clip]{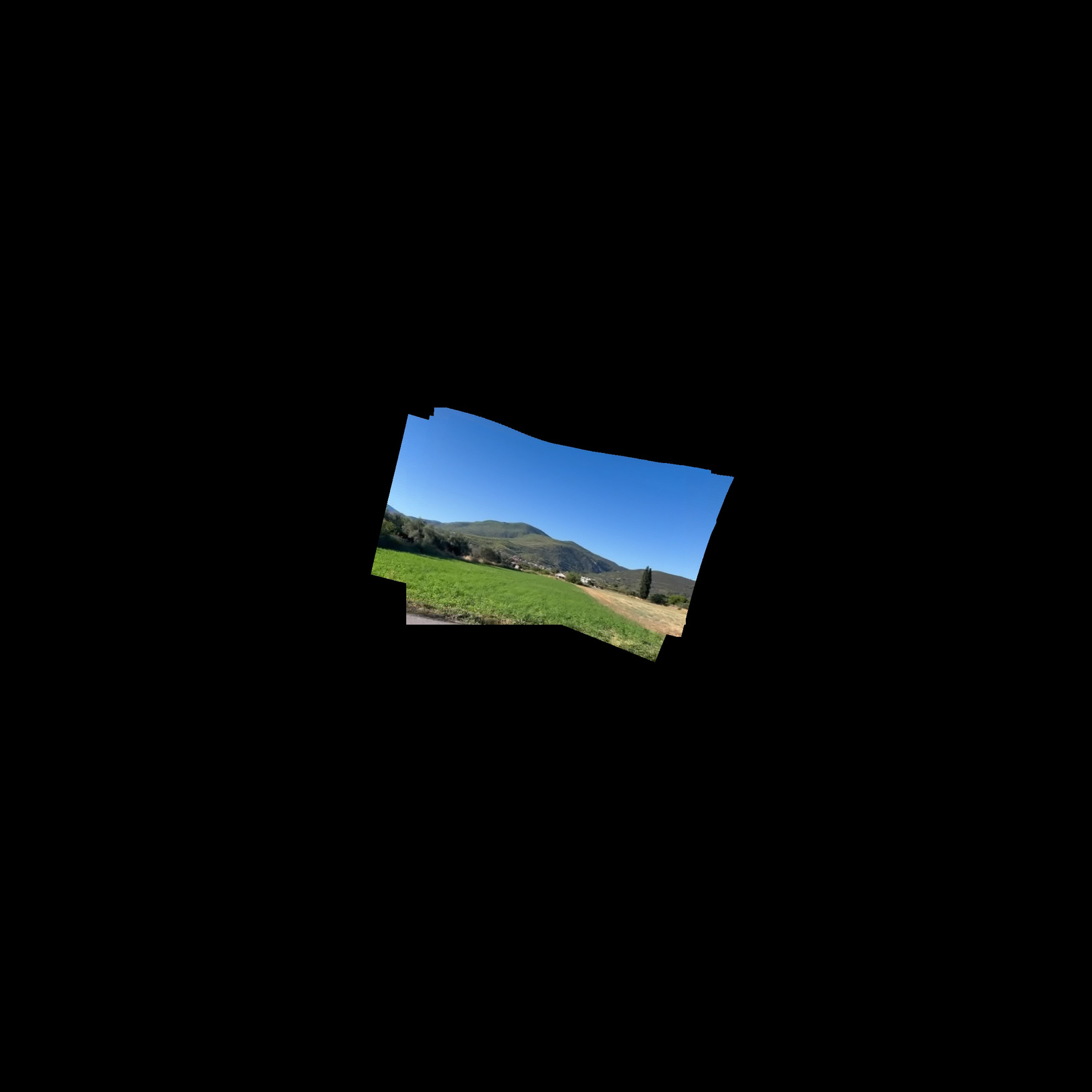} &
\includegraphics[width=3.8cm, height=2.0cm,    viewport=650 850 1850 1700,   clip]{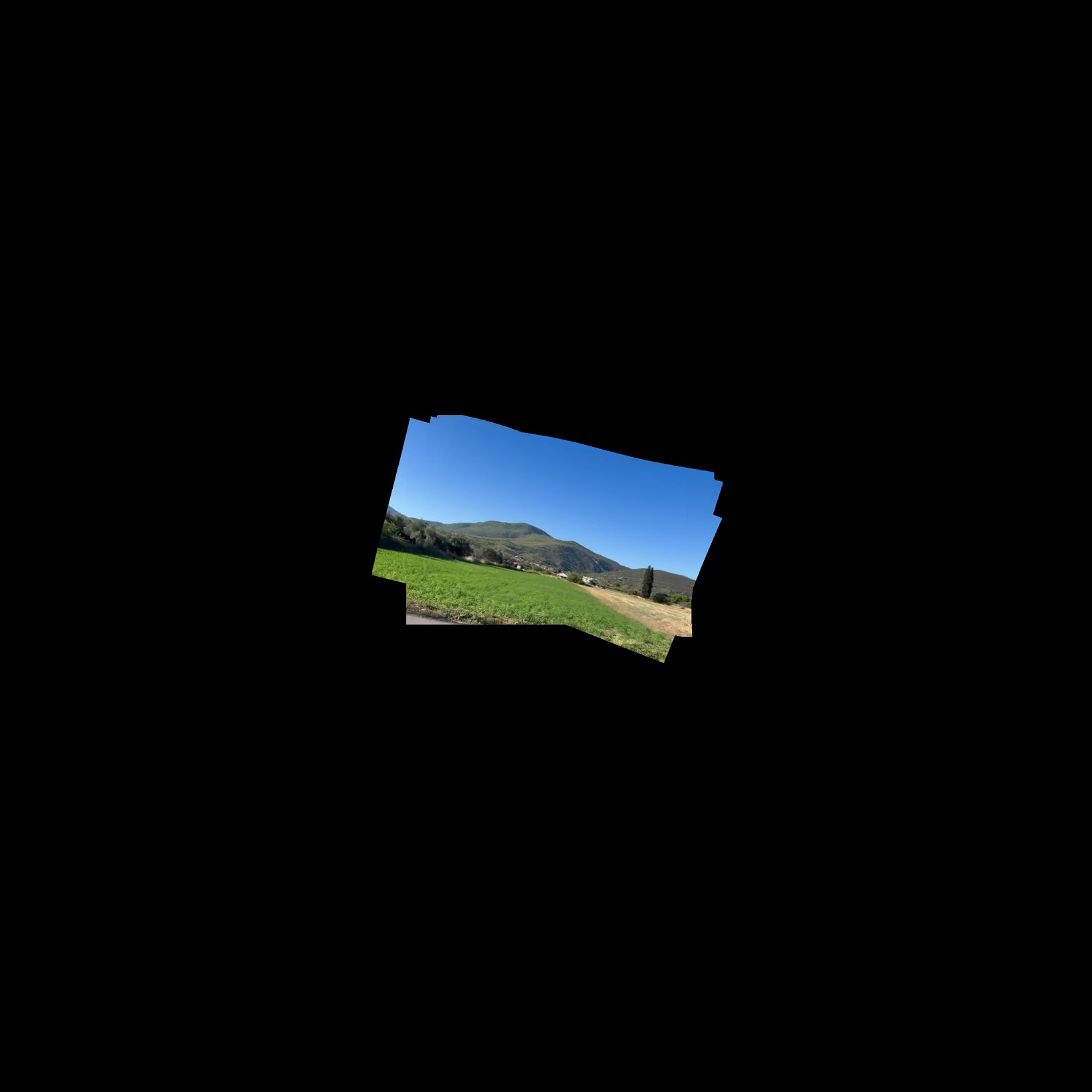}&
\includegraphics[width=3.8cm, height=2.0cm,    viewport=650 850 1850 1700,   clip]{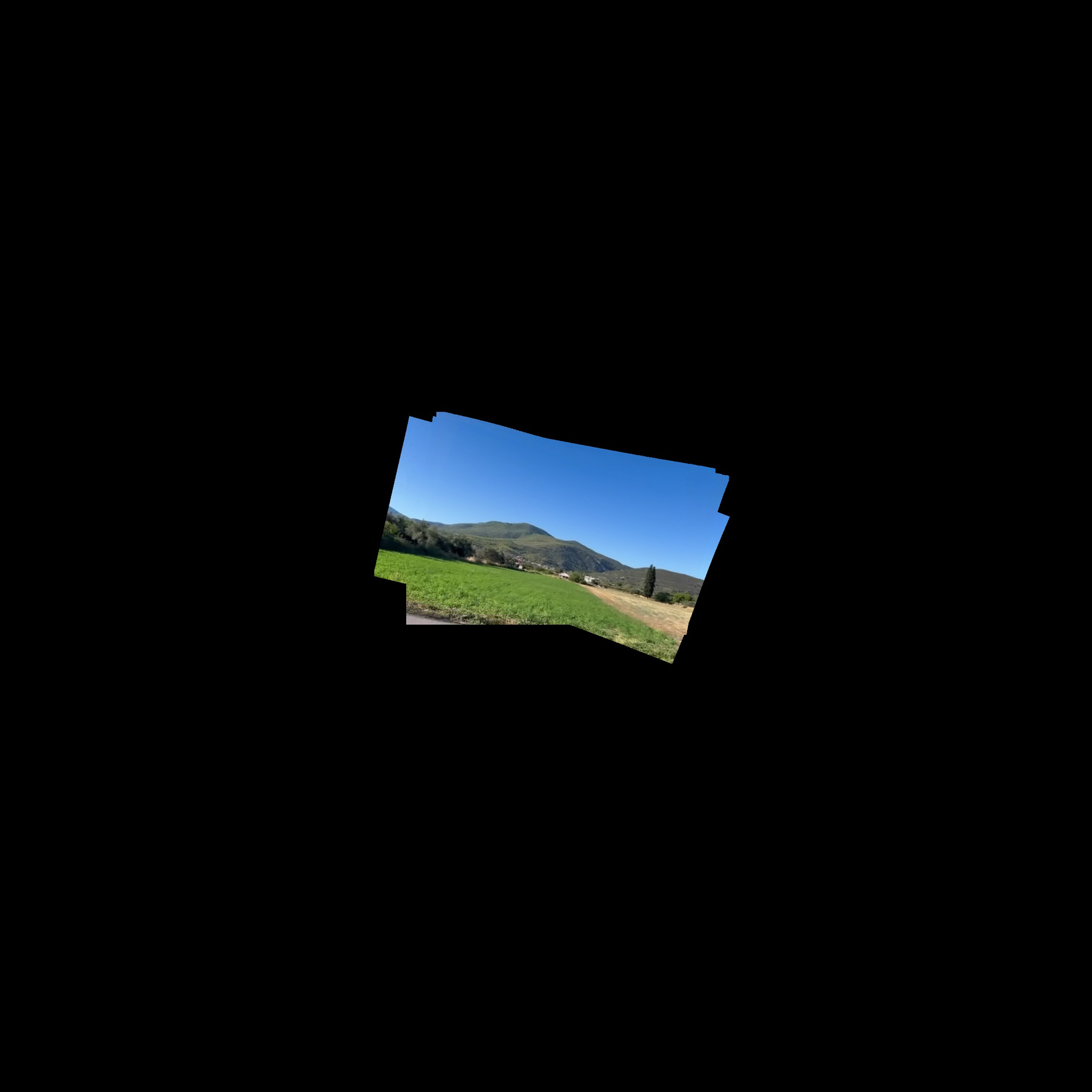} \\[0.5mm]

\rotatebox{90}{\parbox{2.0cm}{\centering \small GSP ~\cite{Chen:ECCV:2016:GSP}}} & 
\includegraphics[width=3.8cm, height=2.0cm,    viewport=530 800 2100 1800,   clip]{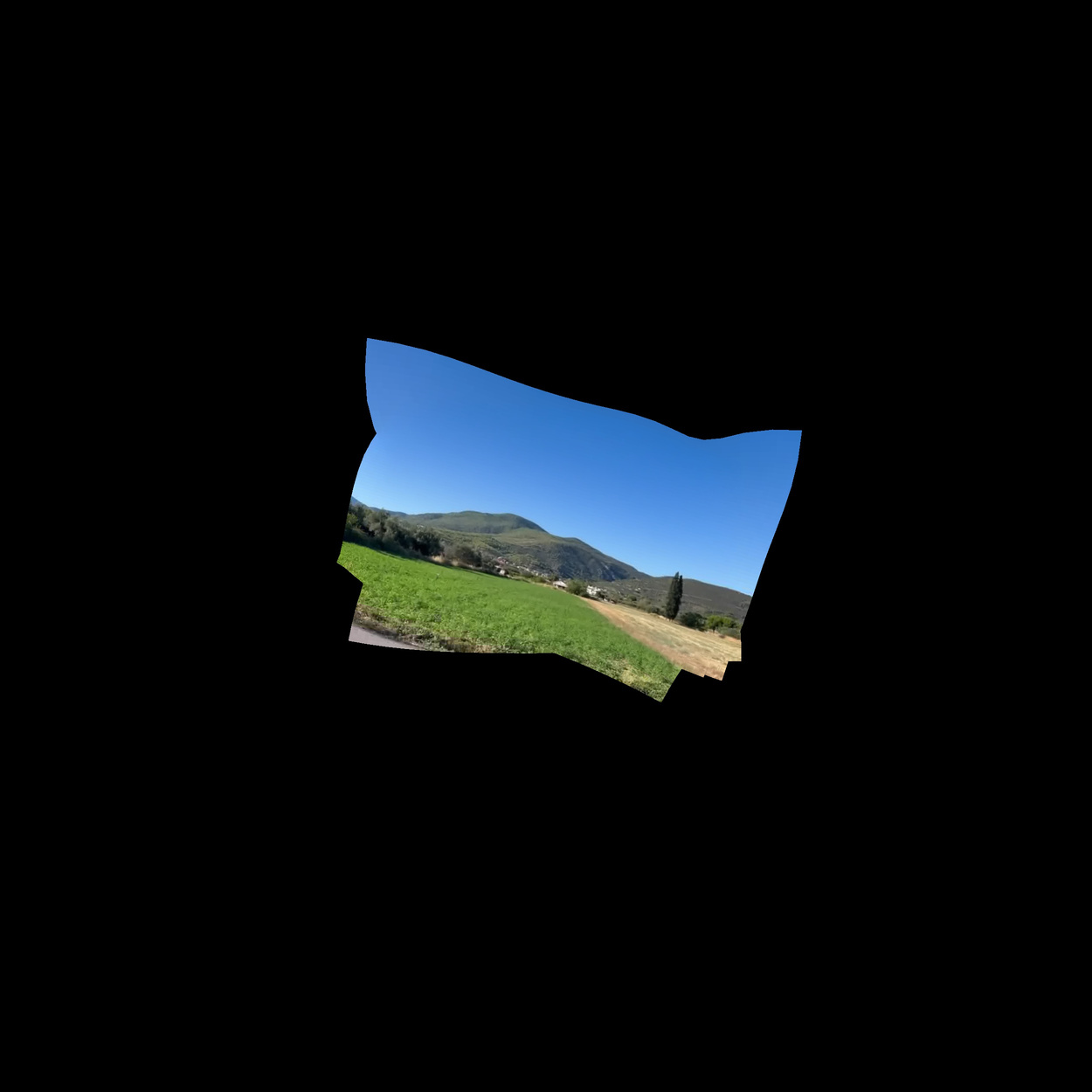} &
\includegraphics[width=3.8cm, height=2.0cm,    viewport=530 800 2100 1800,   clip]{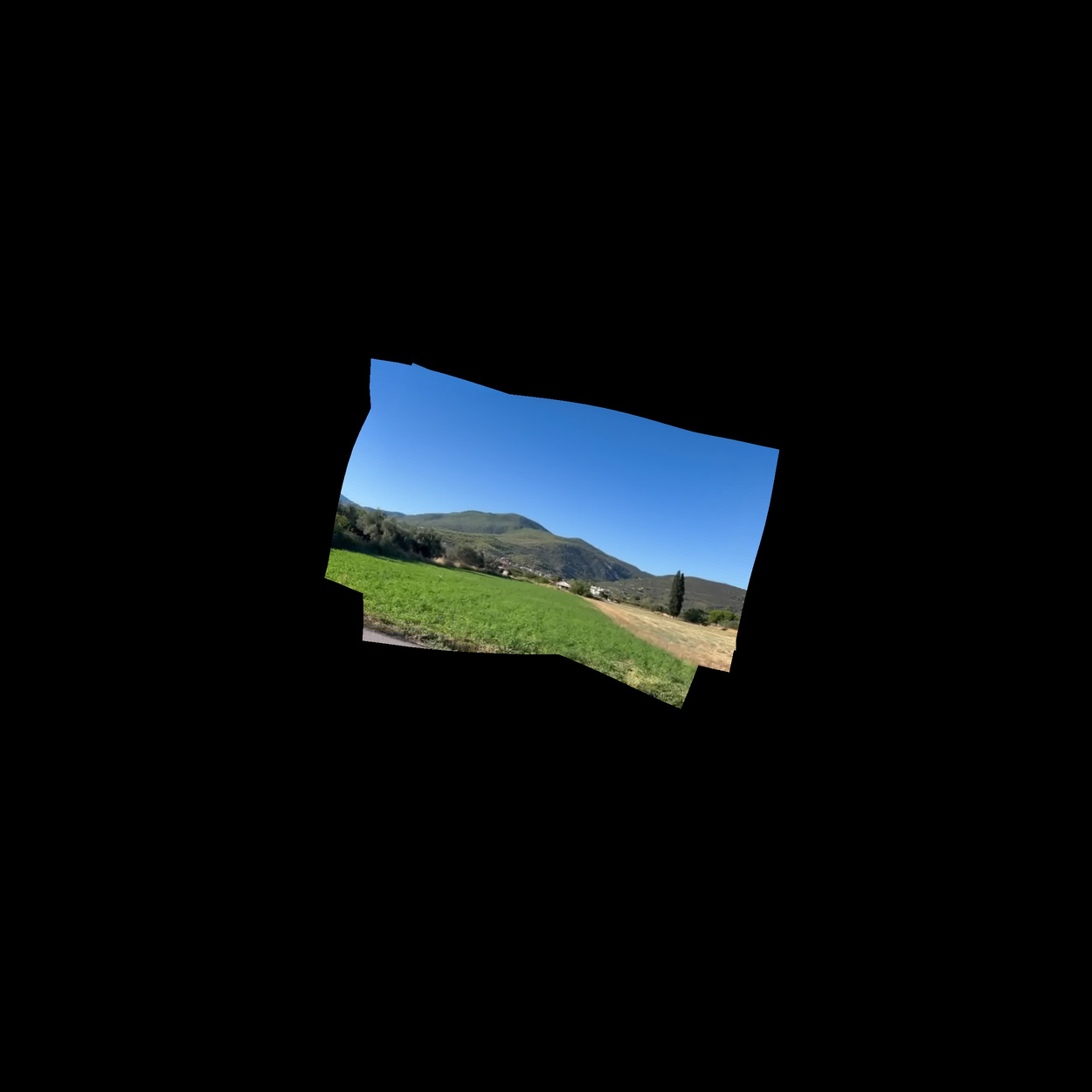} &
\includegraphics[width=3.8cm, height=2.0cm,    viewport=530 800 2100 1800,   clip]{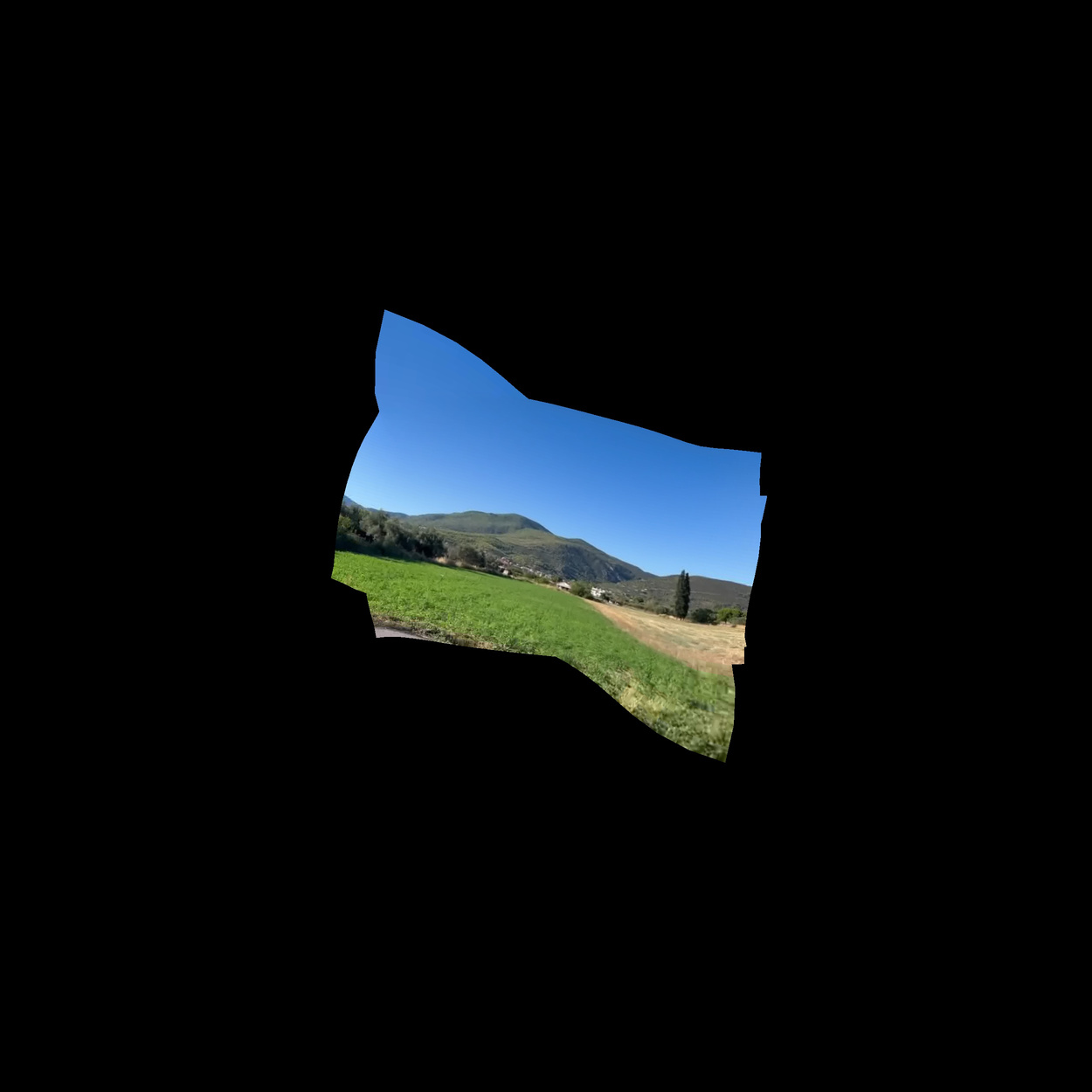}&
\includegraphics[width=3.8cm, height=2.0cm,    viewport=530 800 2100 1800,   clip]{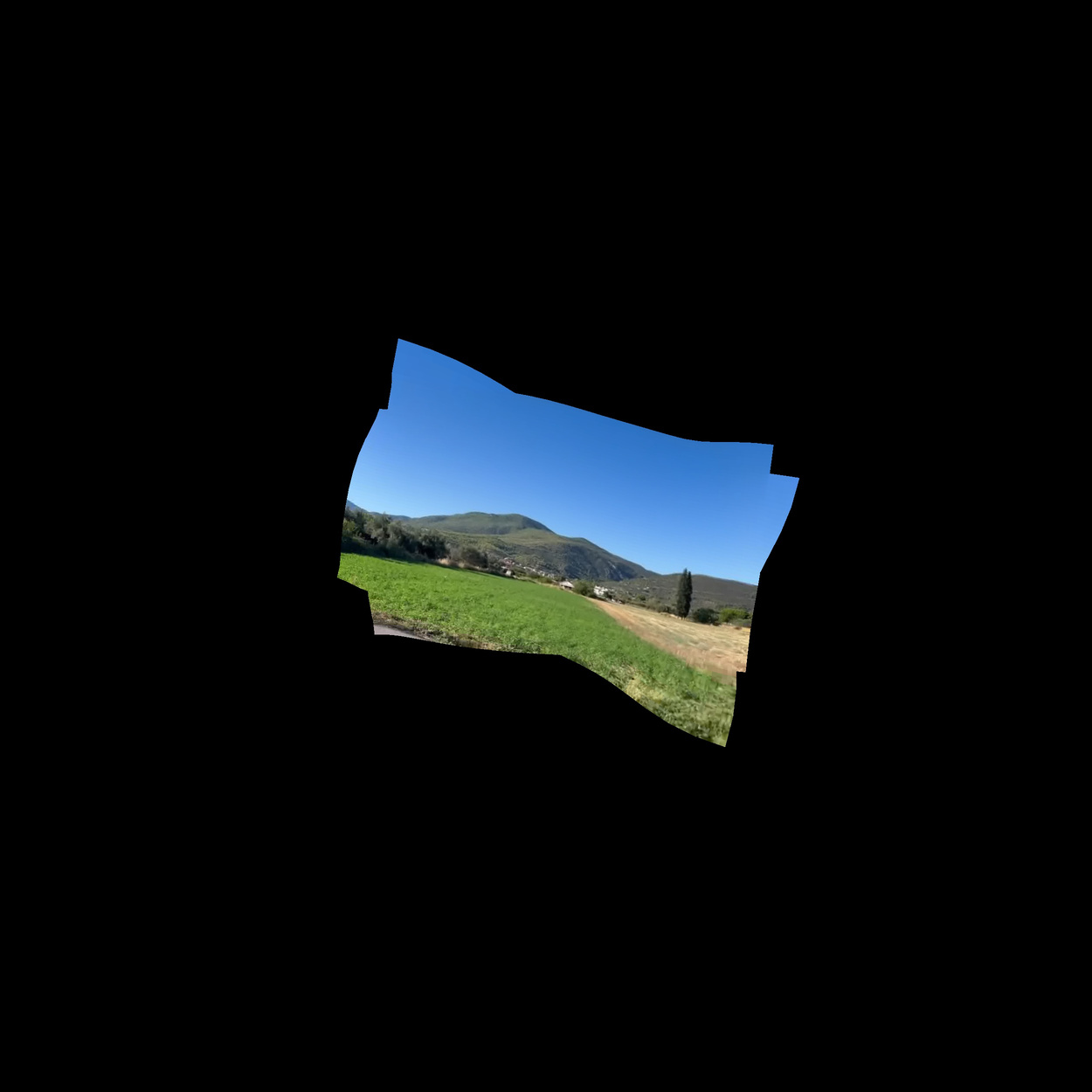} \\[0.3mm]

\rotatebox{90}{\parbox{2.0cm}{\centering \small SPW ~\cite{Liao:TIP:2019:SPW}}} & 
\includegraphics[width=3.8cm, height=2.0cm,    viewport=530 800 2100 1800,   clip]{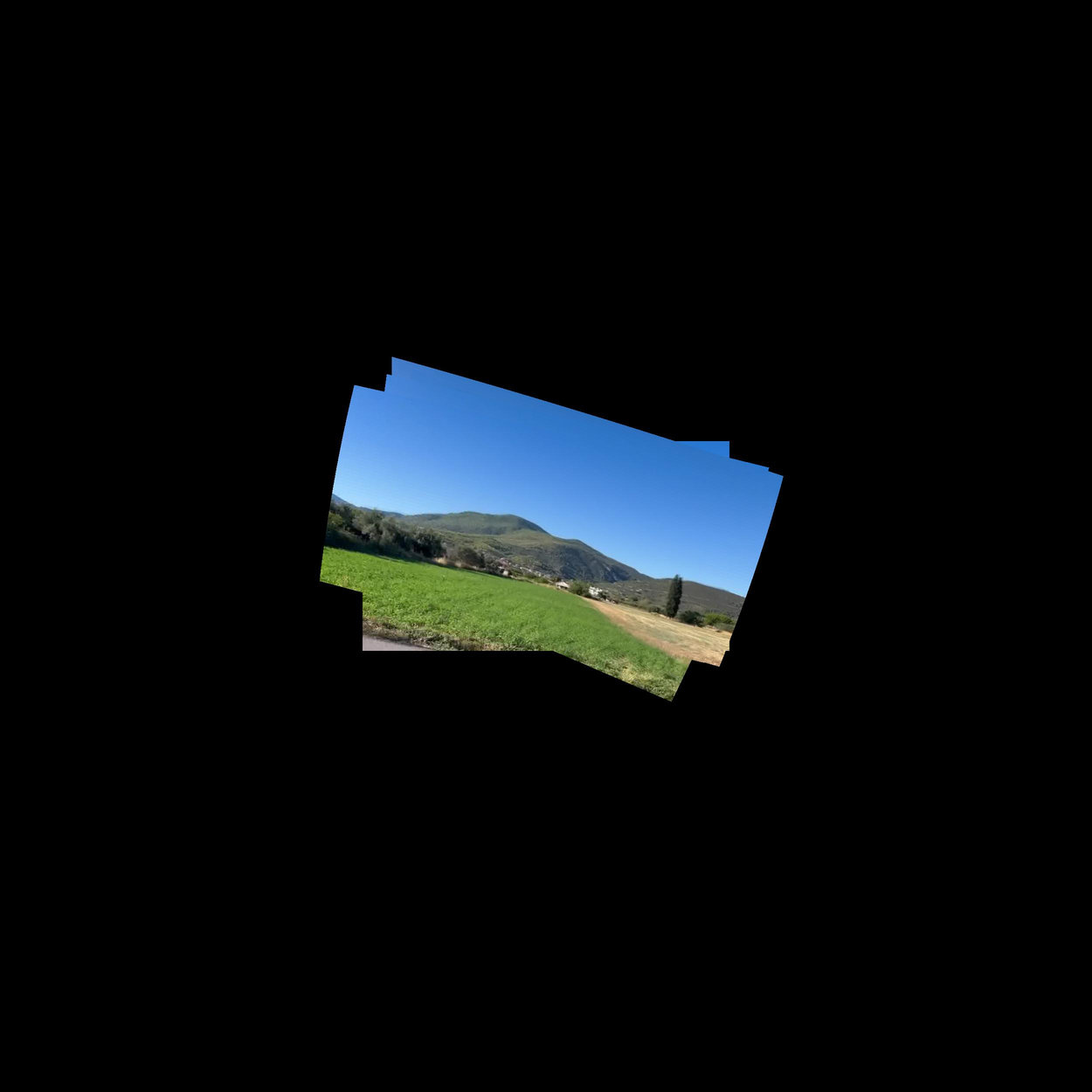} &
\includegraphics[width=3.8cm, height=2.0cm,    viewport=530 800 2100 1800,   clip]{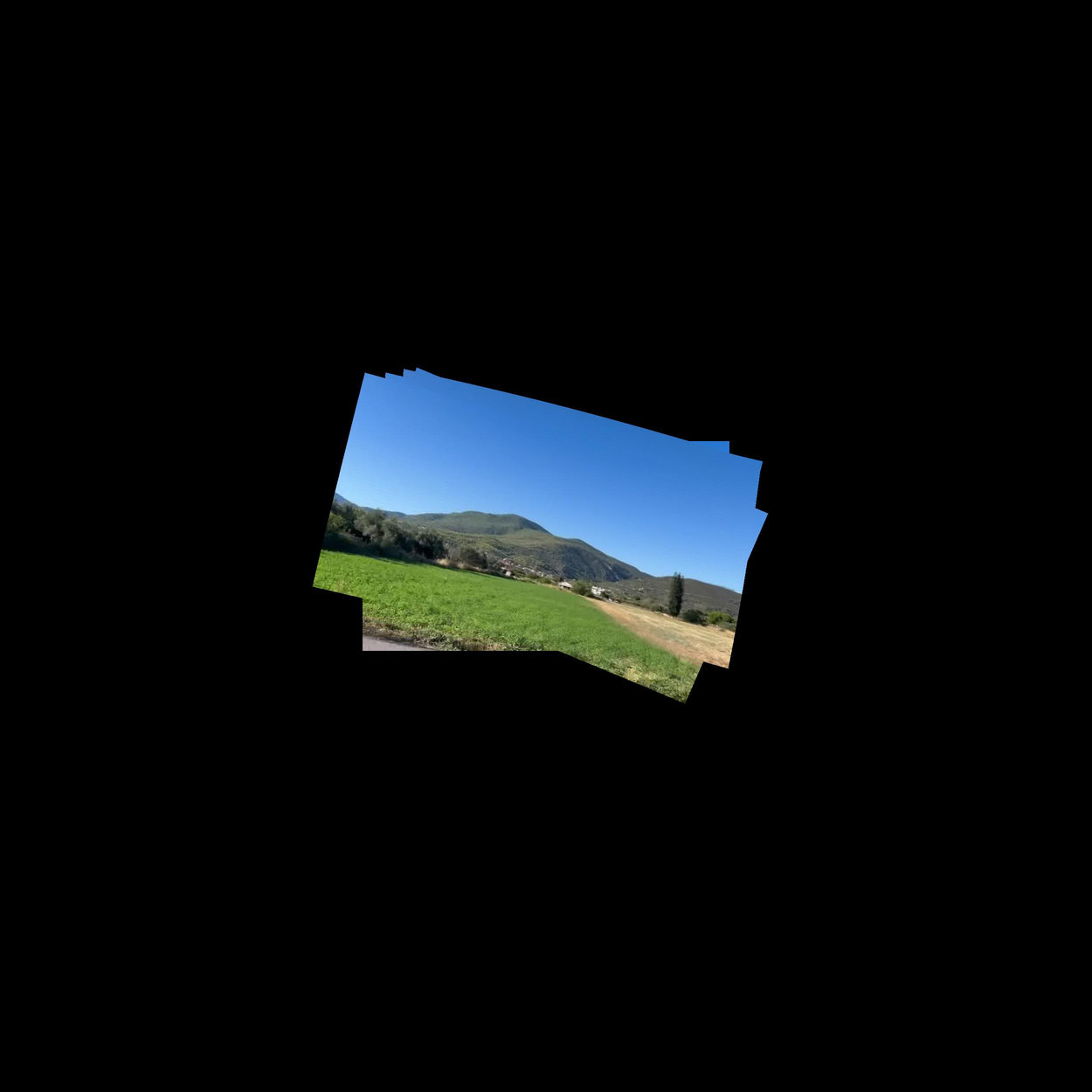} &
\includegraphics[width=3.8cm, height=2.0cm,    viewport=530 800 2100 1800,   clip]{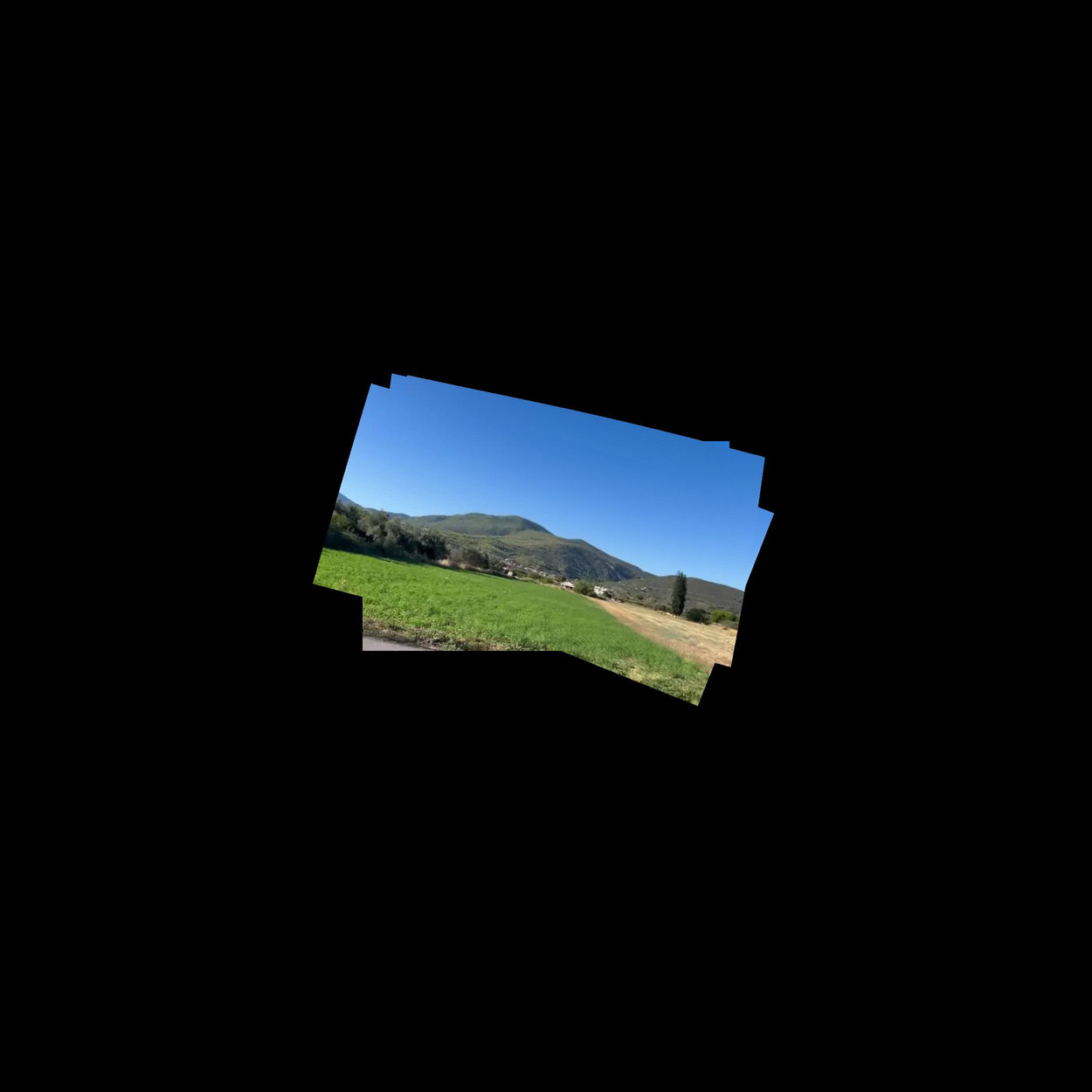}&
\includegraphics[width=3.8cm, height=2.0cm,    viewport=530 800 2100 1800,   clip]{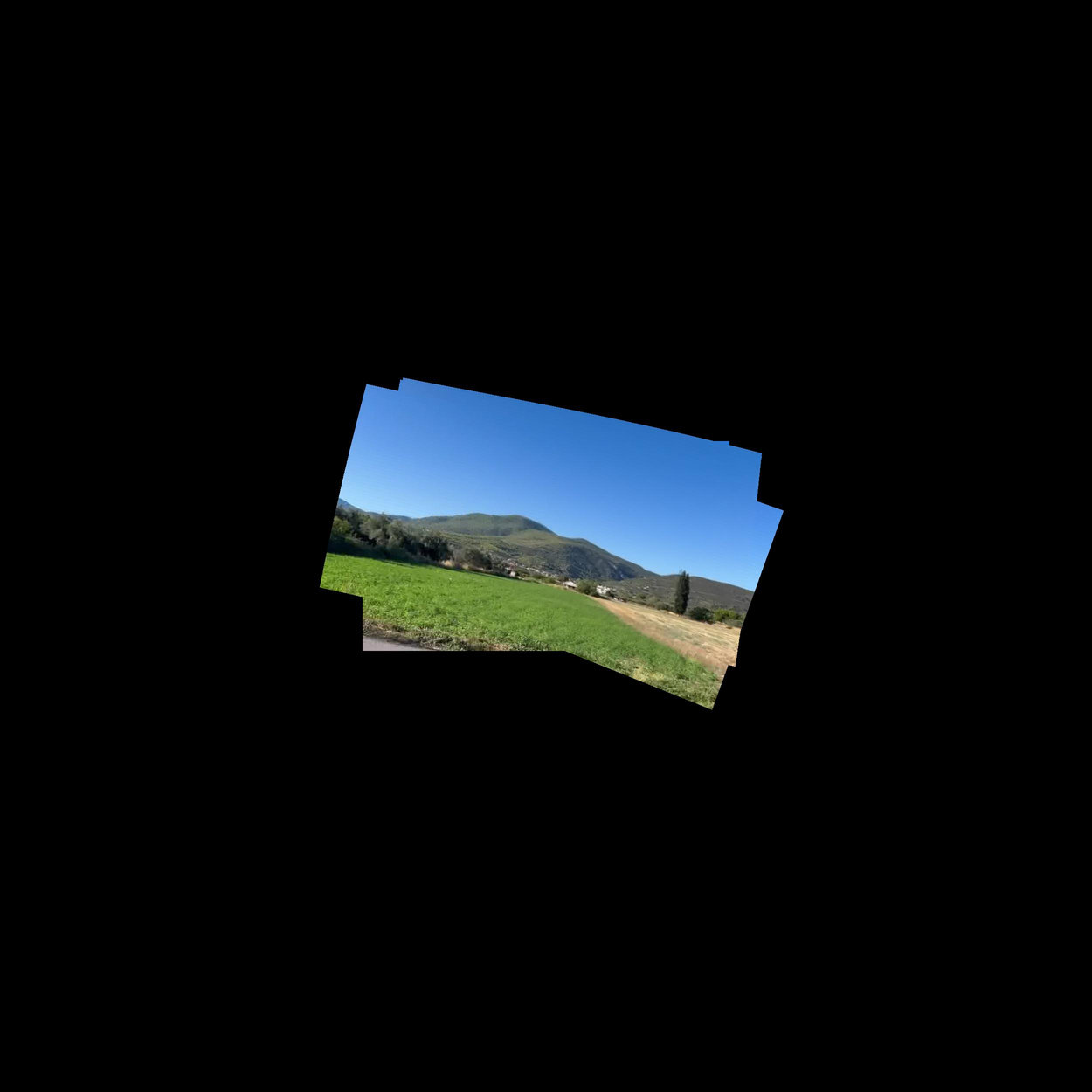} \\[0.3mm]

\rotatebox{90}{\parbox{2.0cm}{\centering \small GES-GSP ~\cite{Du:CVPR:2022:GESGSP}}} & 
\includegraphics[width=3.8cm, height=2.0cm,    viewport=530 800 2100 1800,   clip]{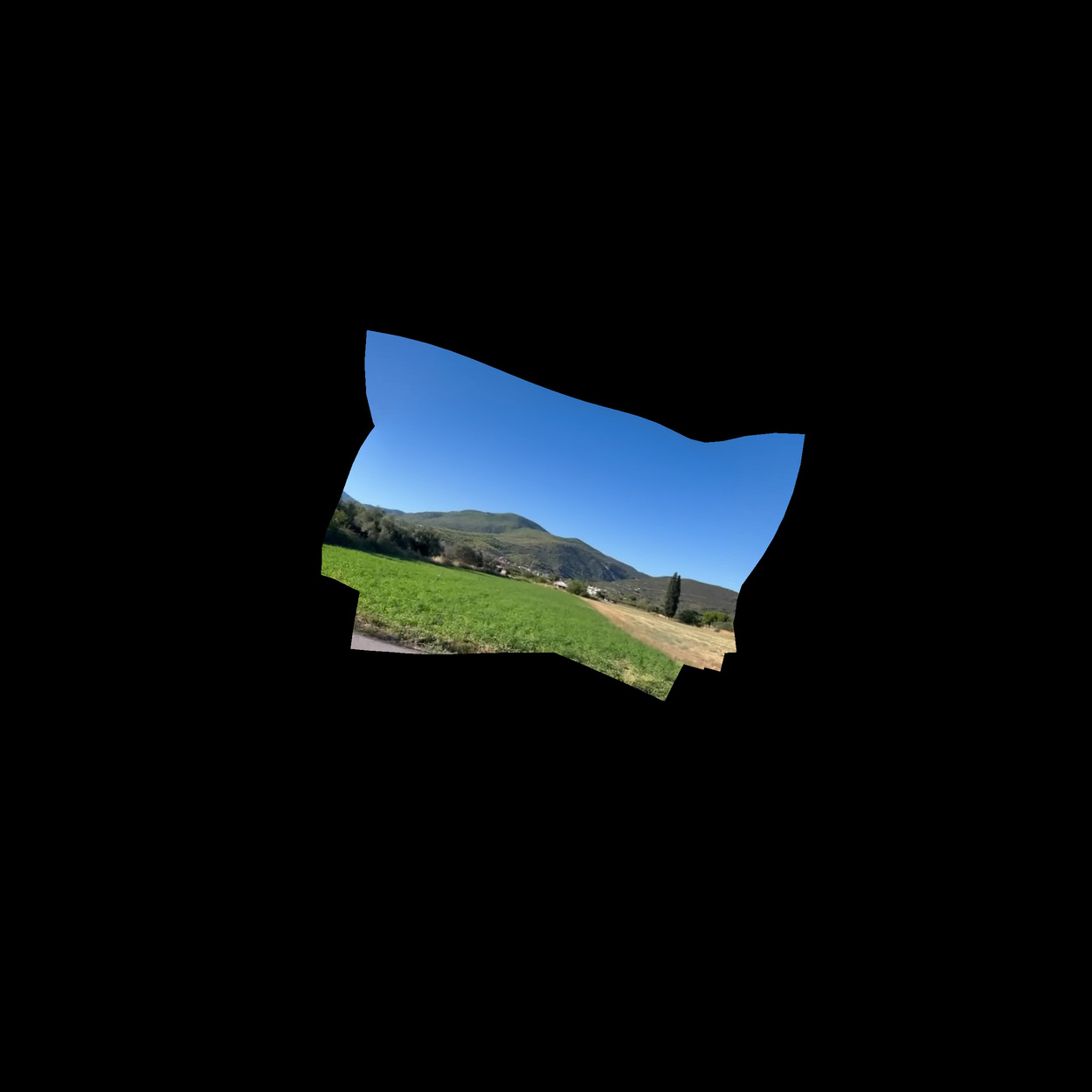} &
\includegraphics[width=3.8cm, height=2.0cm,    viewport=530 800 2100 1800,   clip]{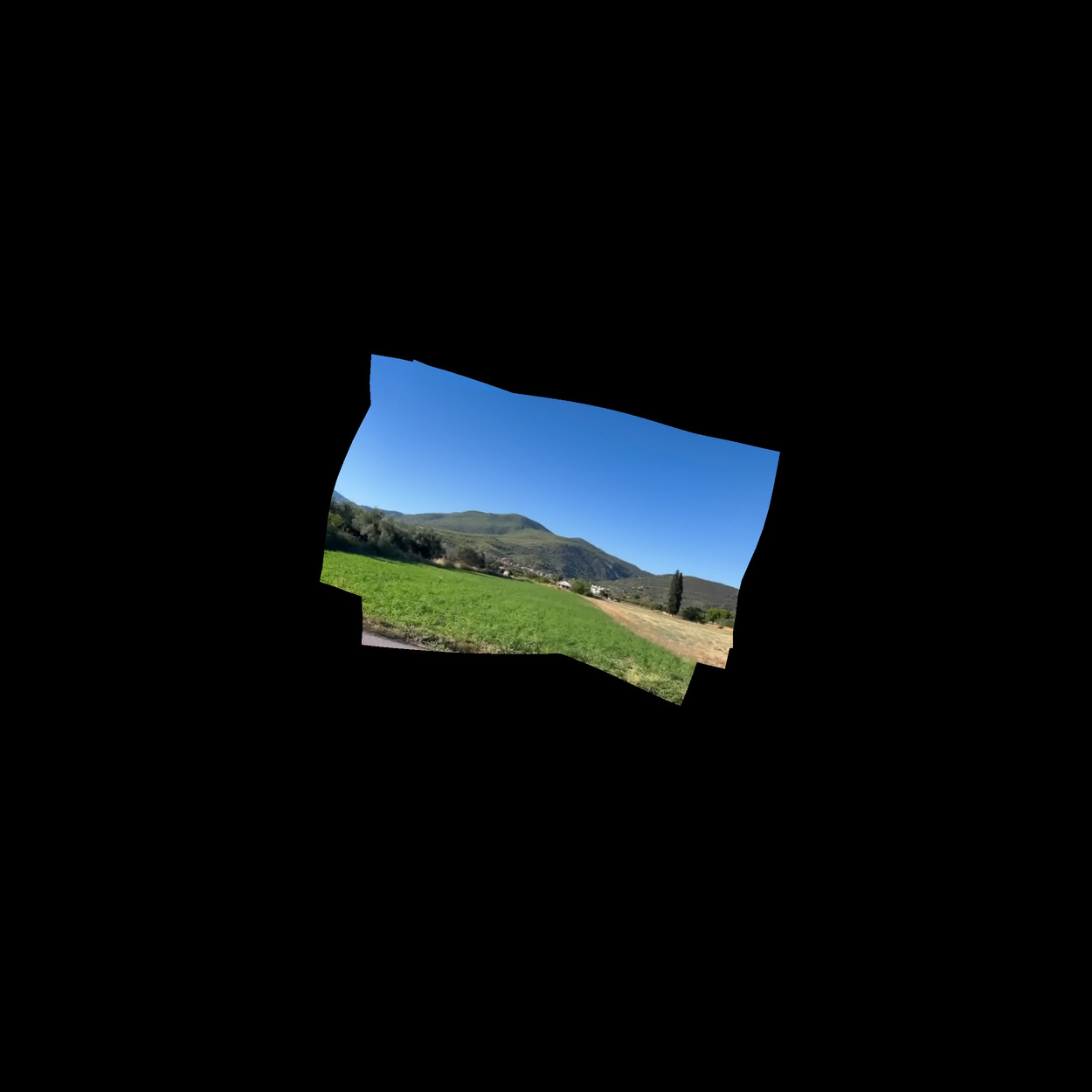} &
\includegraphics[width=3.8cm, height=2.0cm,    viewport=530 800 2100 1800,   clip]{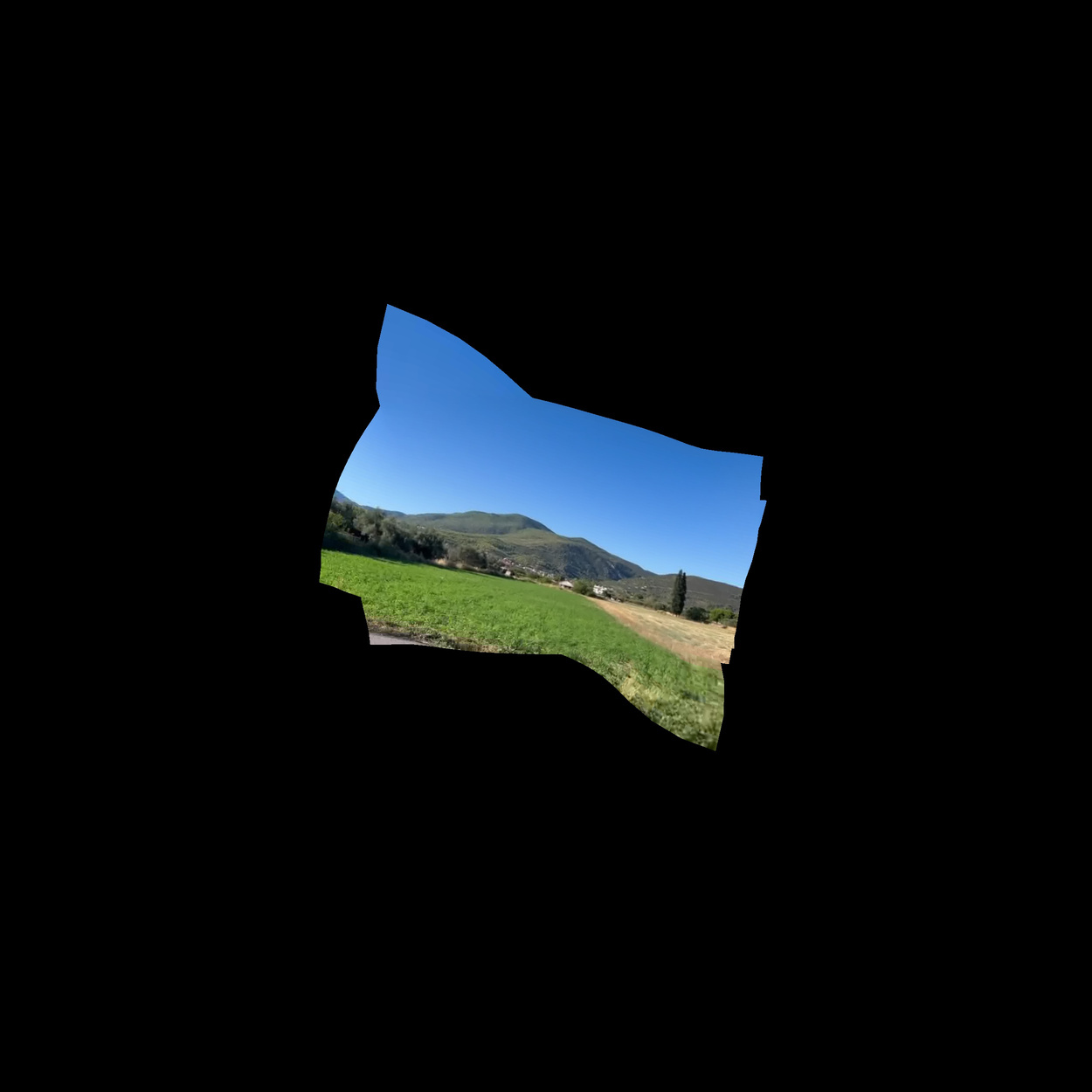}&
\includegraphics[width=3.8cm, height=2.0cm,    viewport=530 800 2100 1800,   clip]{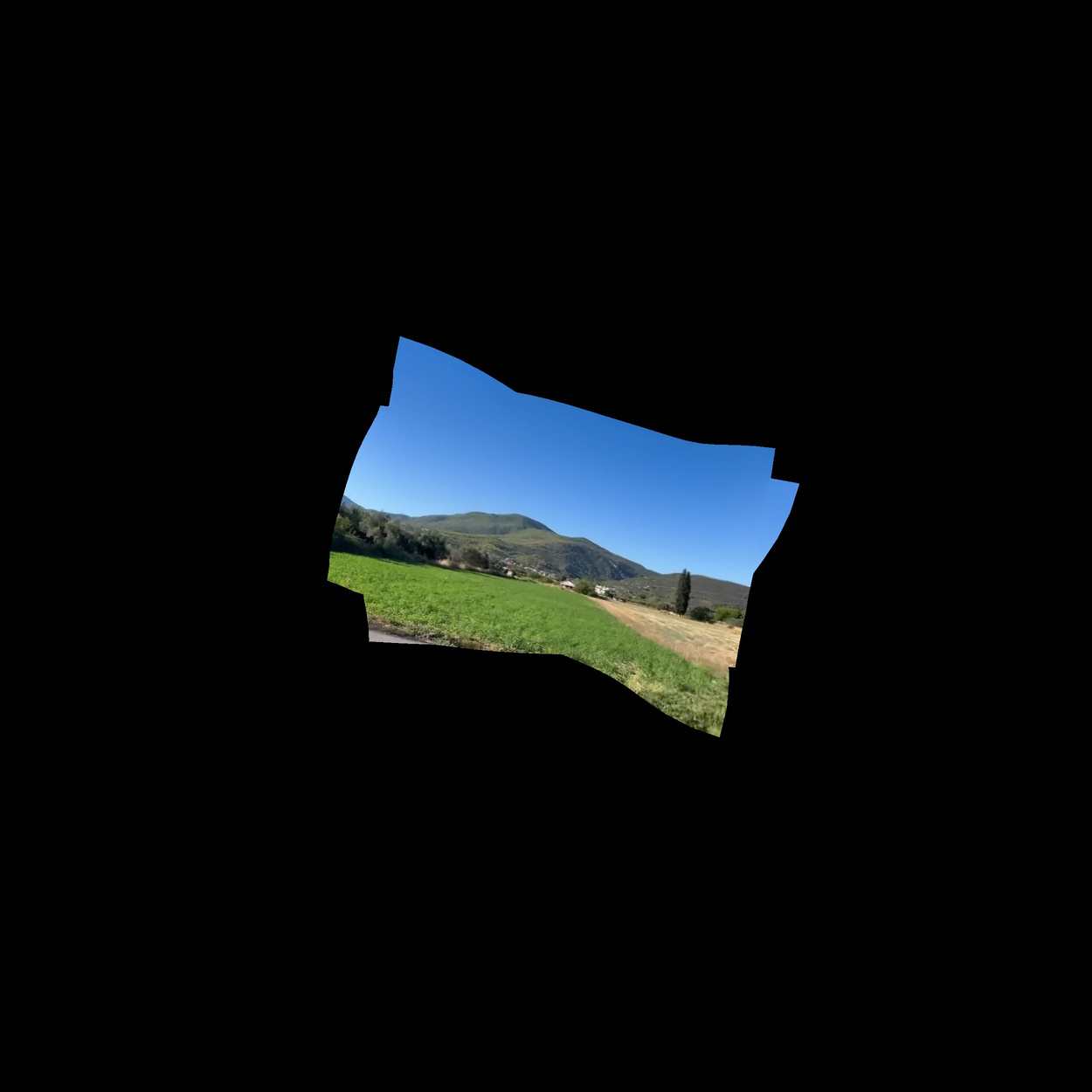} \\[0.3mm]

\rotatebox{90}{\parbox{2.0cm}{\centering \small OBJ-GSP ~\cite{Cai:AAAI:2025:OBJGSP}}} & 
\includegraphics[width=3.8cm, height=2.0cm,    viewport=530 800 2100 1800,   clip]{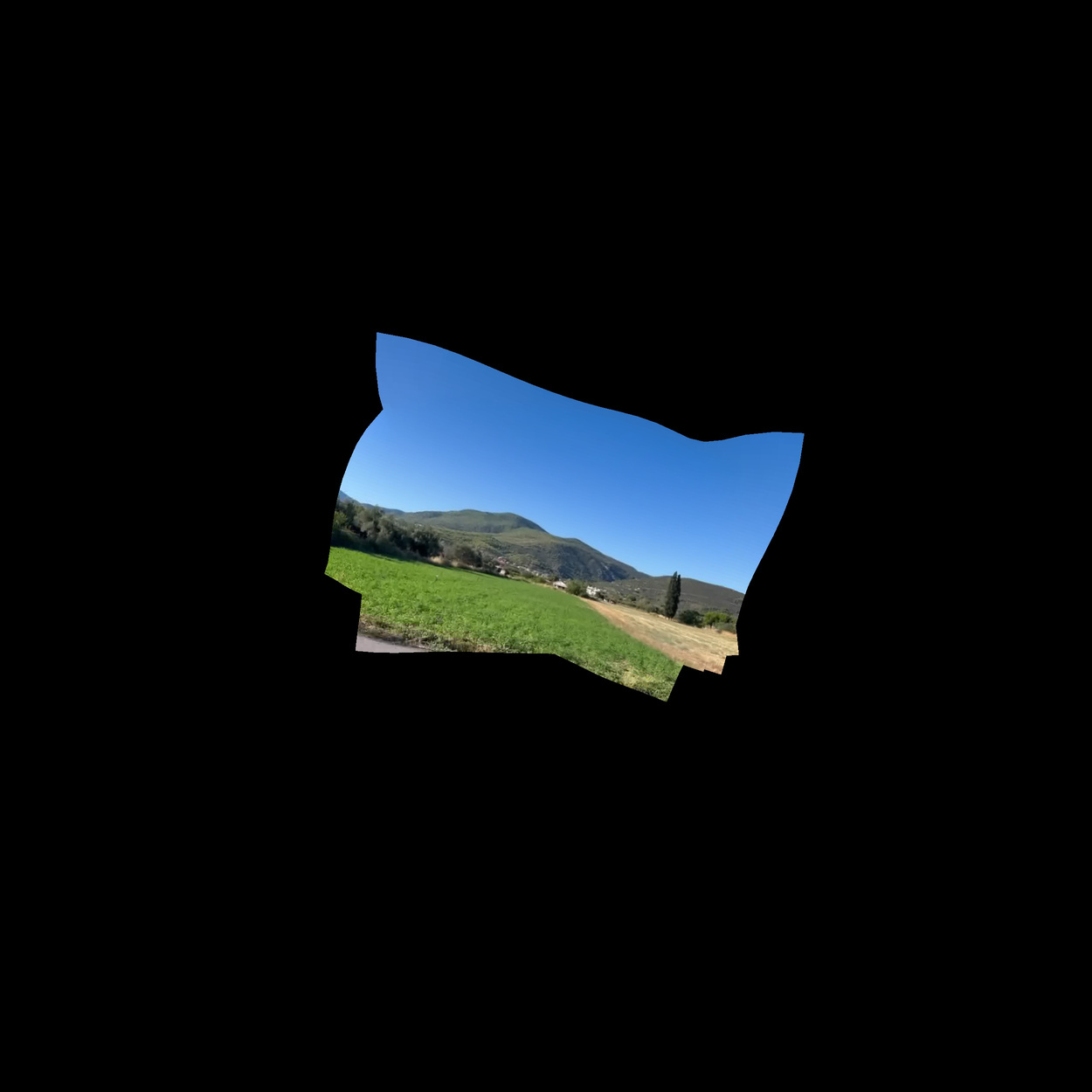} &
\includegraphics[width=3.8cm, height=2.0cm,    viewport=530 800 2100 1800,   clip]{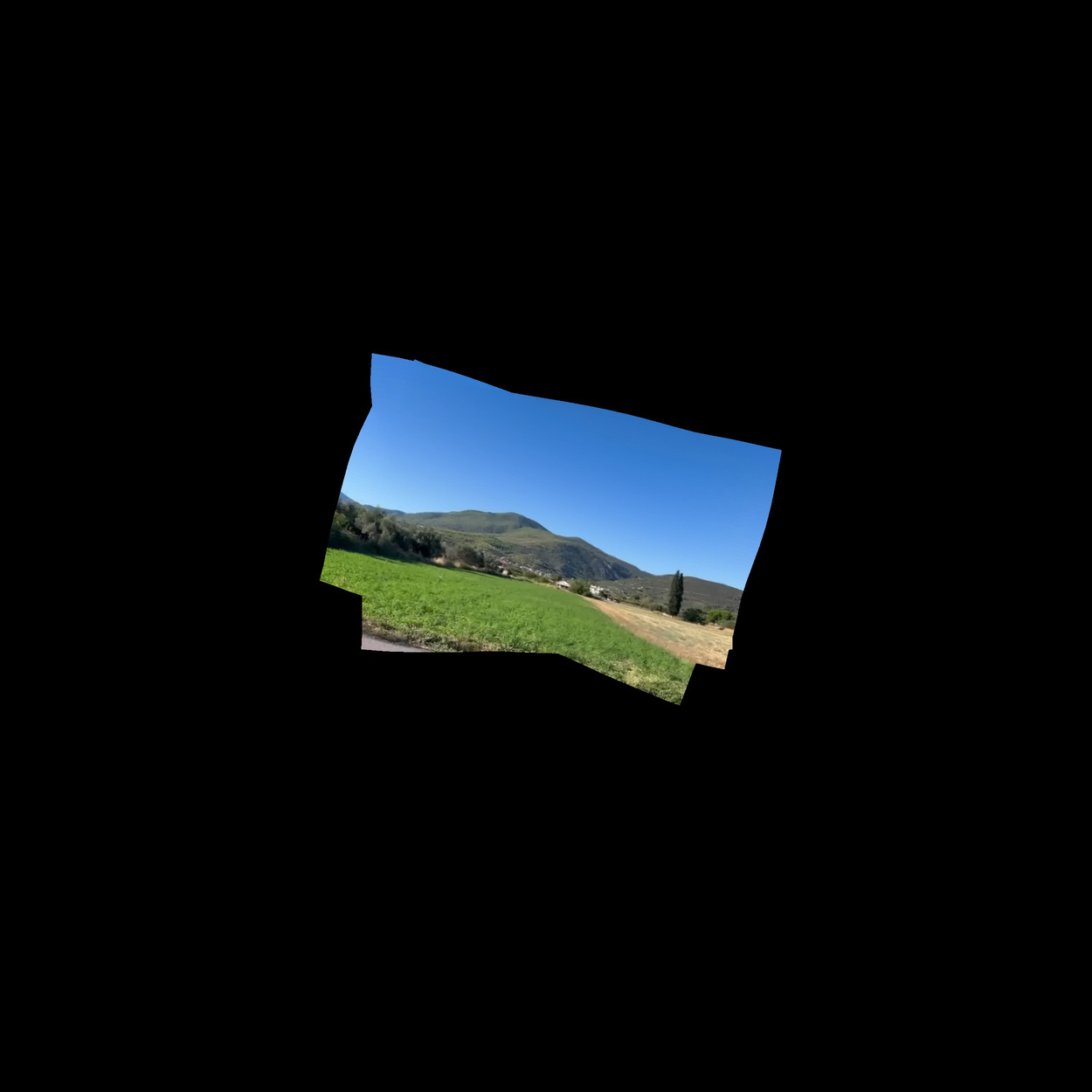} &
\includegraphics[width=3.8cm, height=2.0cm,    viewport=530 800 2100 1800,   clip]{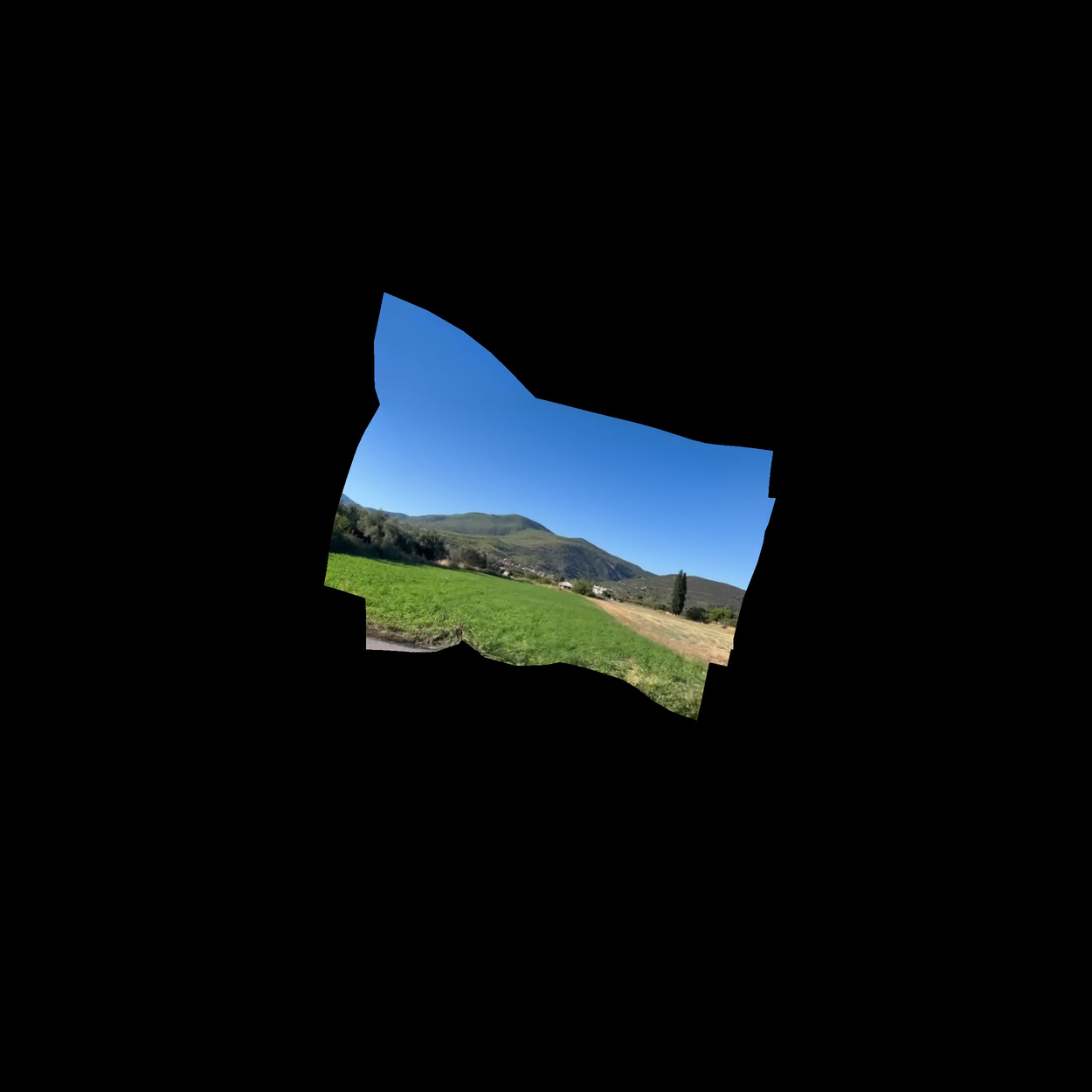}&
\includegraphics[width=3.8cm, height=2.0cm,    viewport=530 800 2100 1800,   clip]{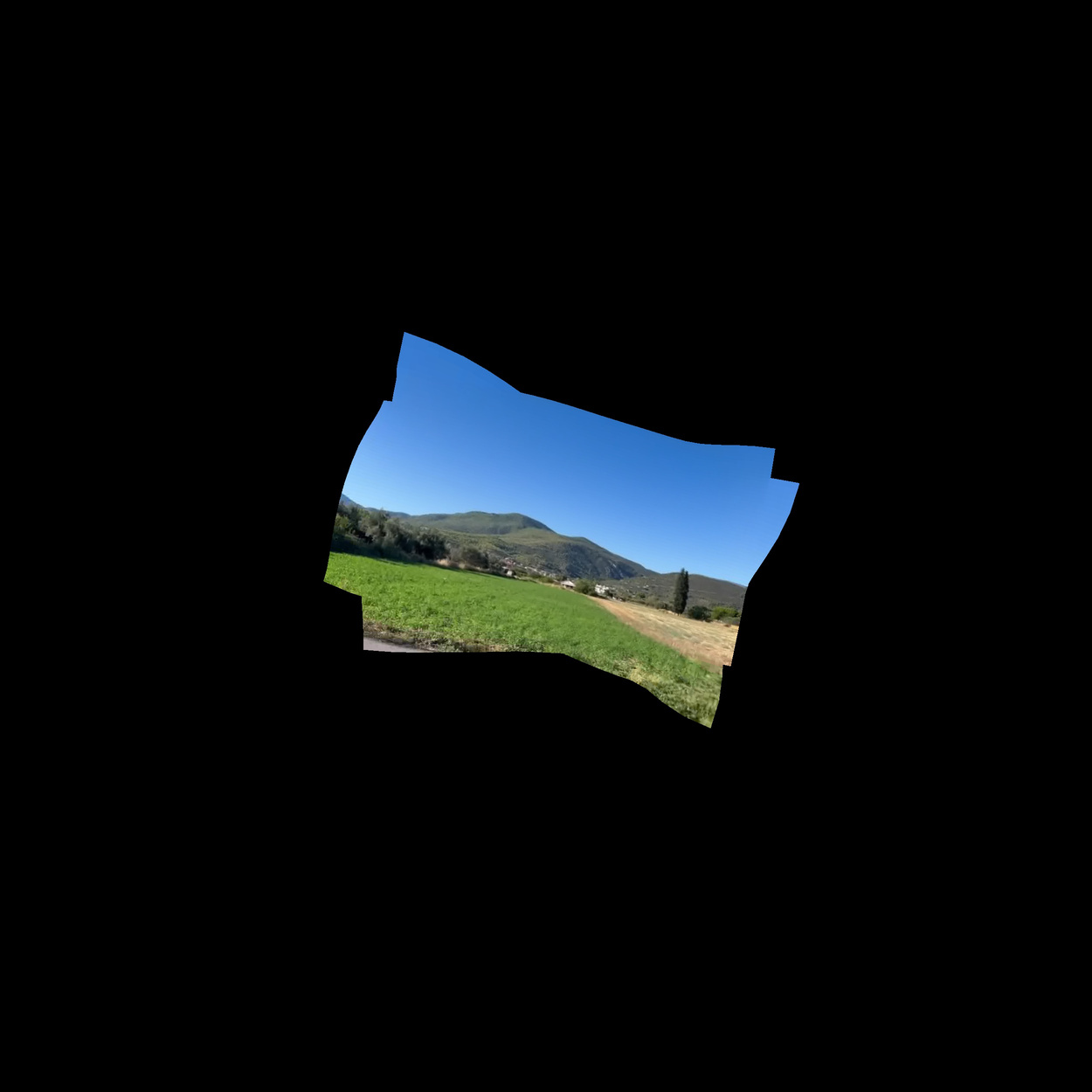} \\[0.3mm]

\rotatebox{90}{\parbox{2.0cm}{\centering \small MViS (Ours)}} & 
\includegraphics[width=3.8cm, height=2.0cm,    viewport=280 580 1692 1262,   clip]{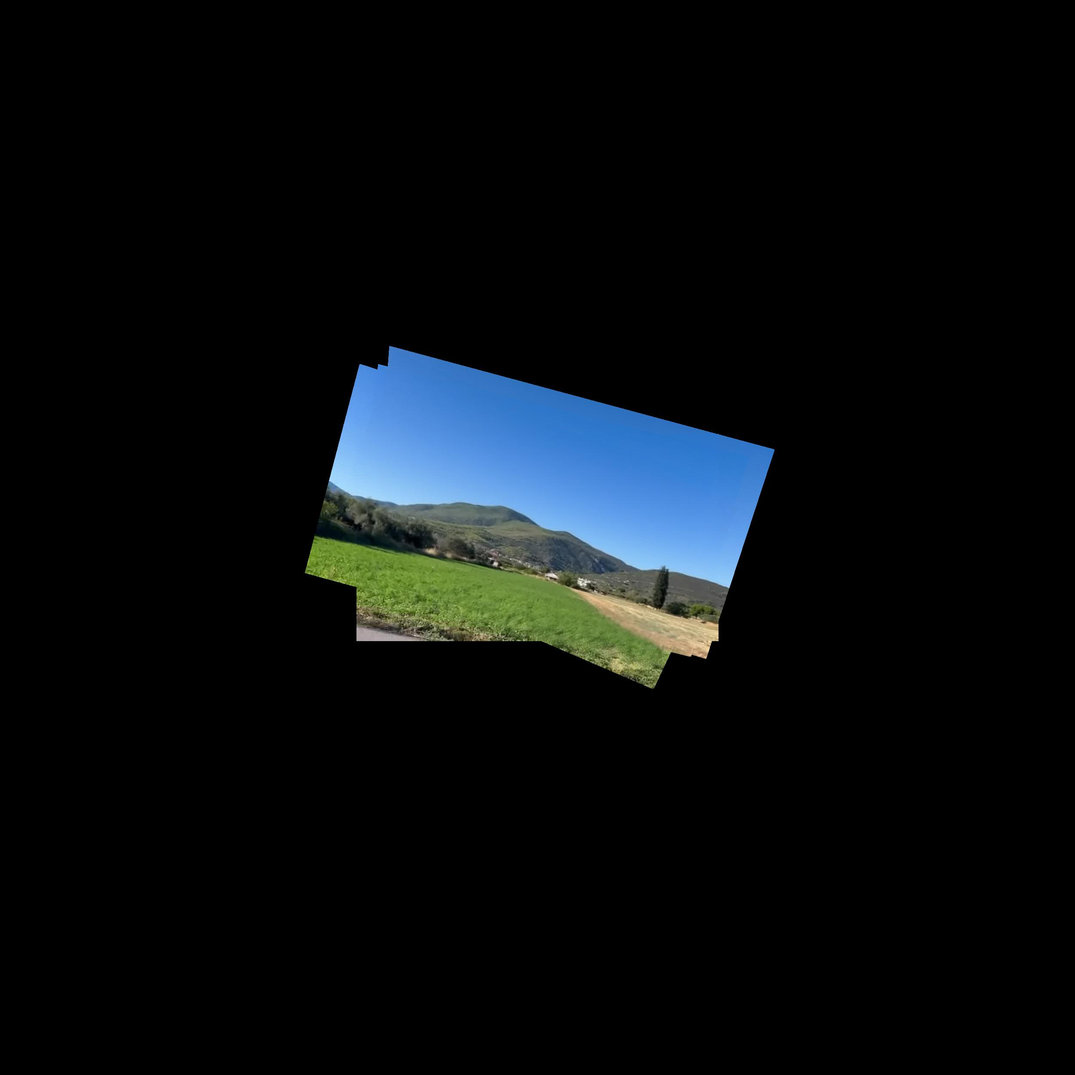} &
\includegraphics[width=3.8cm, height=2.0cm,    viewport=280 580 1692 1262,   clip]{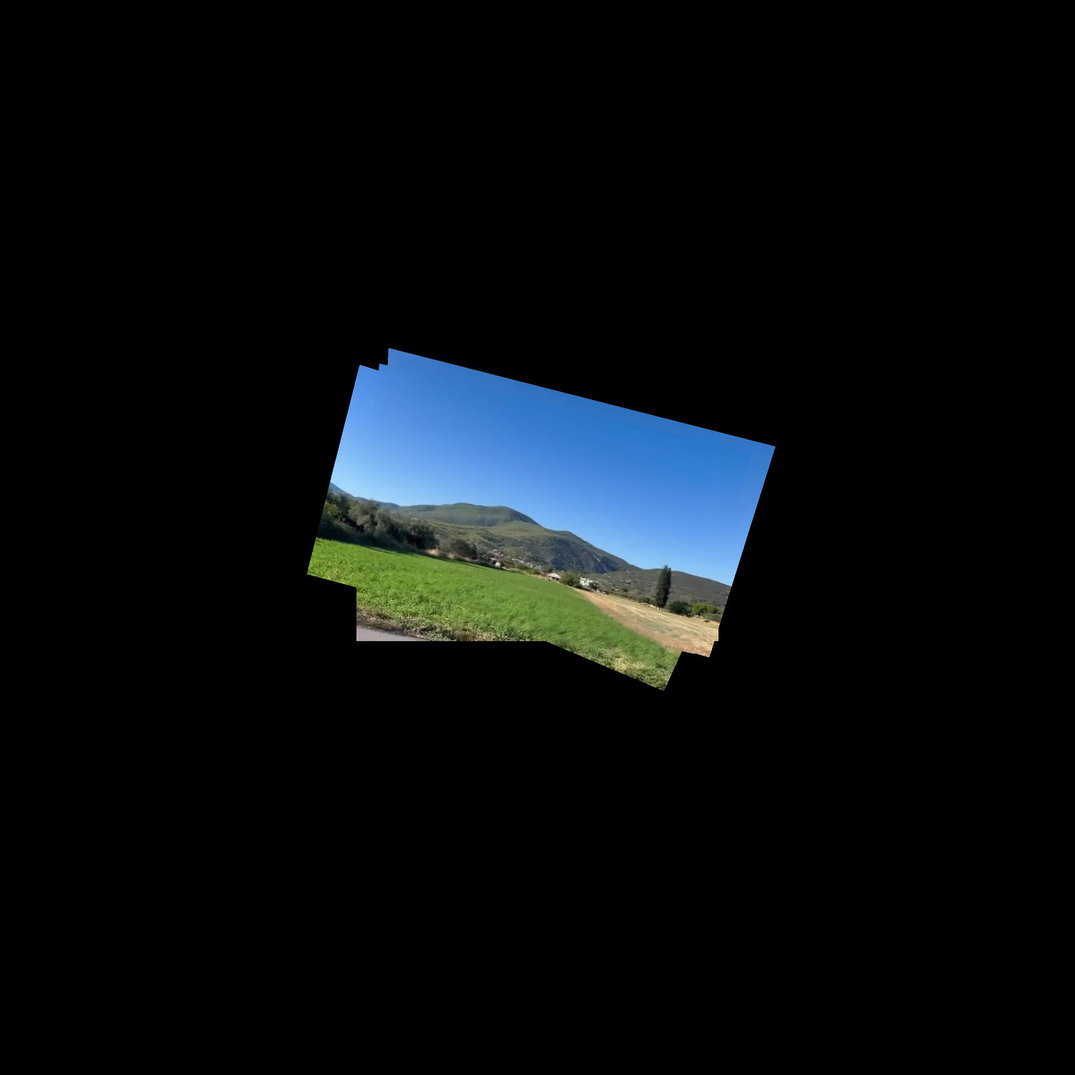} &
\includegraphics[width=3.8cm, height=2.0cm,    viewport=280 580 1692 1262,   clip]{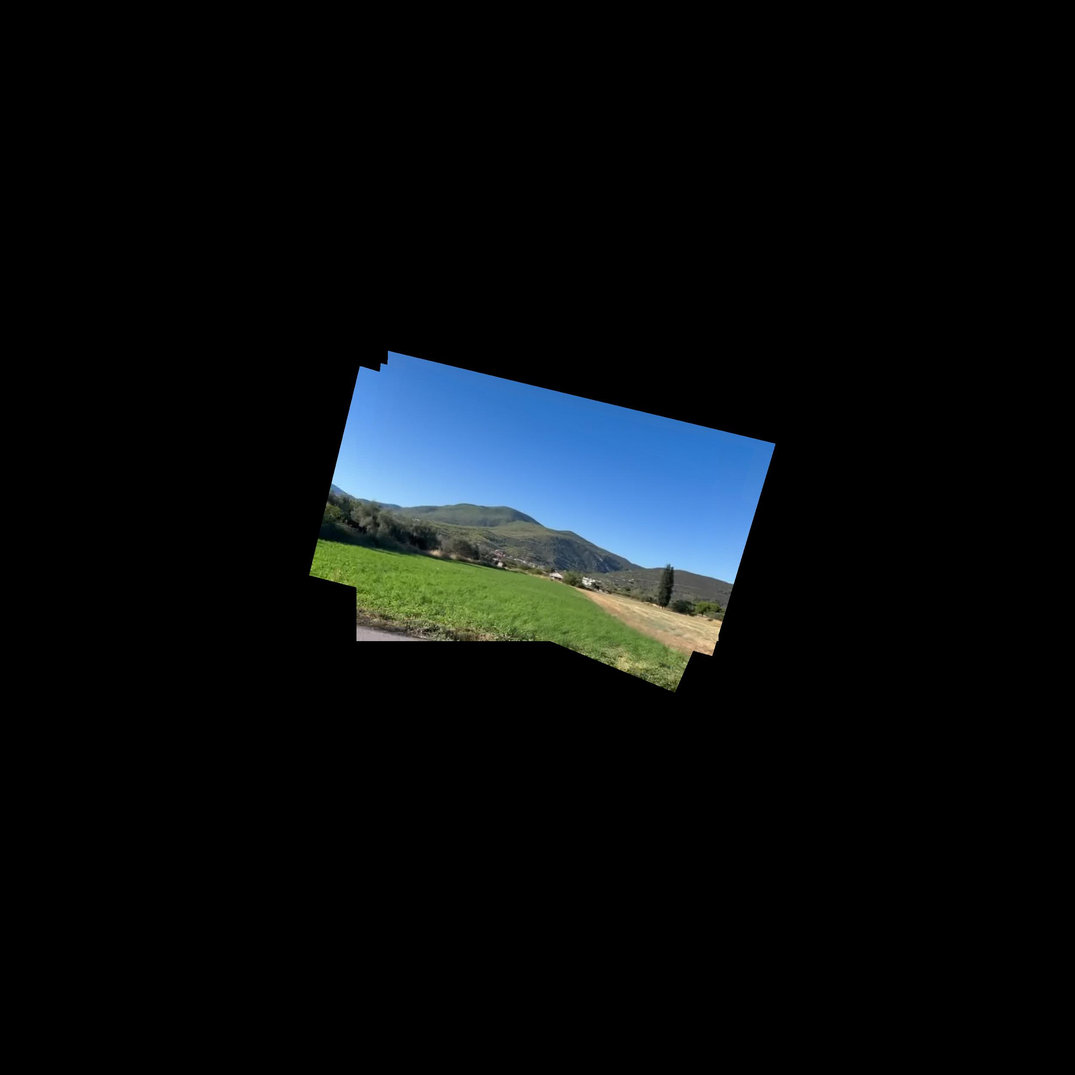}&
\includegraphics[width=3.8cm, height=2.0cm, viewport=280 580 1692 1262,   clip]{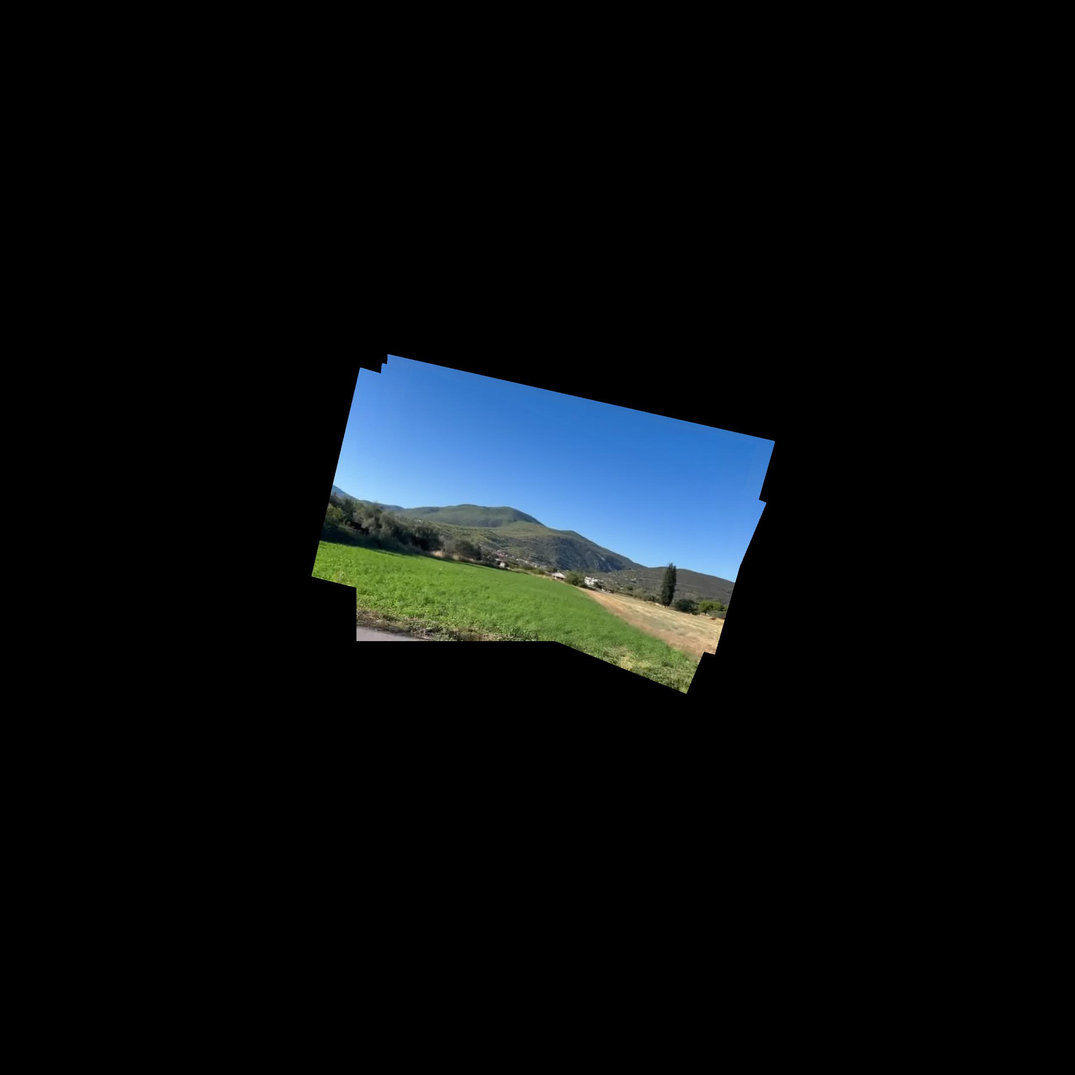} \\[-0.5mm]

& \small{$t=93$} & 
\small{$t=94$} & 
\small{$t=95$} &
\small{$t=96$} \\[2mm]

% Scene_43
\rotatebox{90}{\parbox{2.1cm}{\centering \small GSP ~\cite{Chen:ECCV:2016:GSP}}} & 
\includegraphics[width=3.8cm, height=2.1cm,    viewport=700 700 2300 1700,   clip]{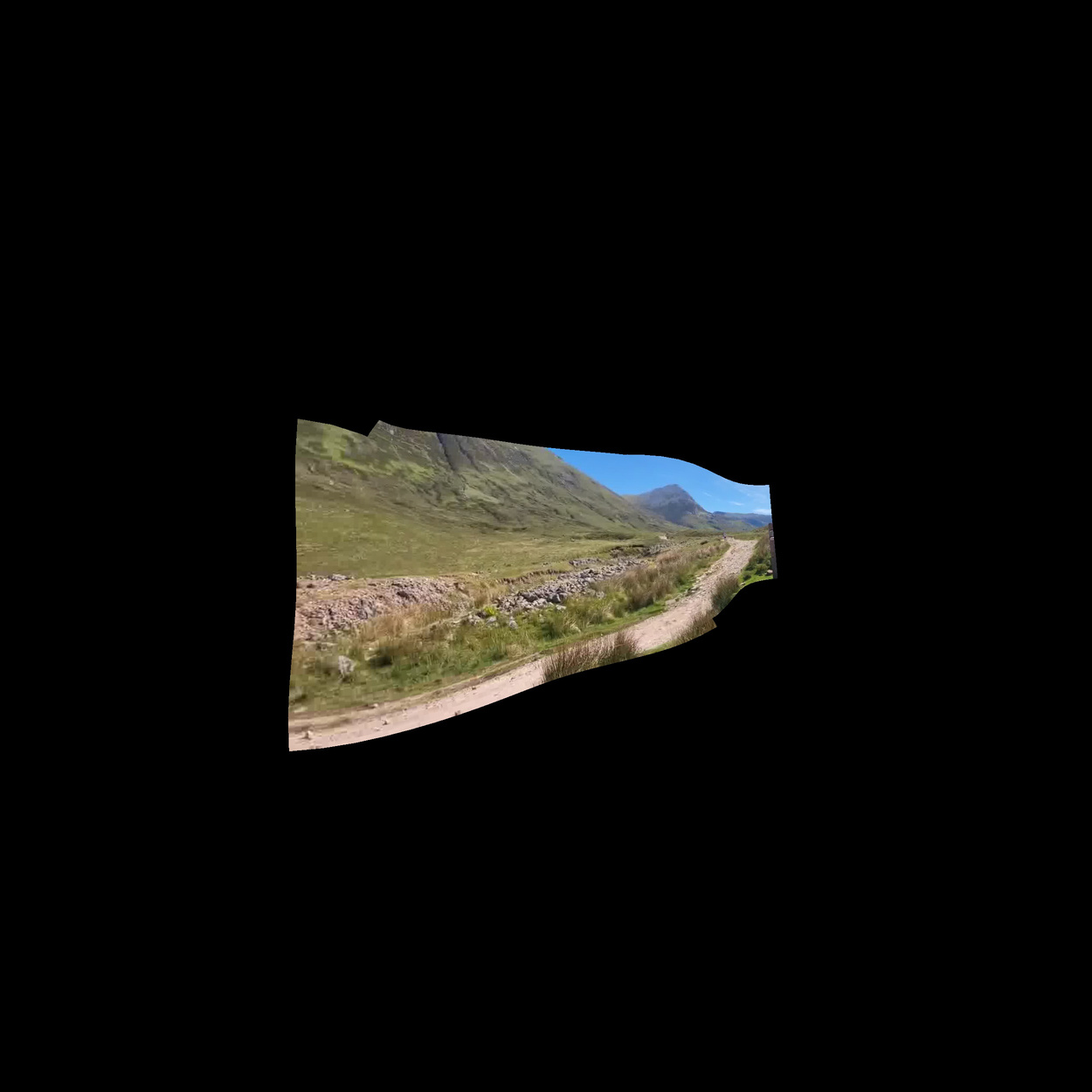} &
\includegraphics[width=3.8cm, height=2.1cm,    viewport=700 700 2300 1700,   clip]{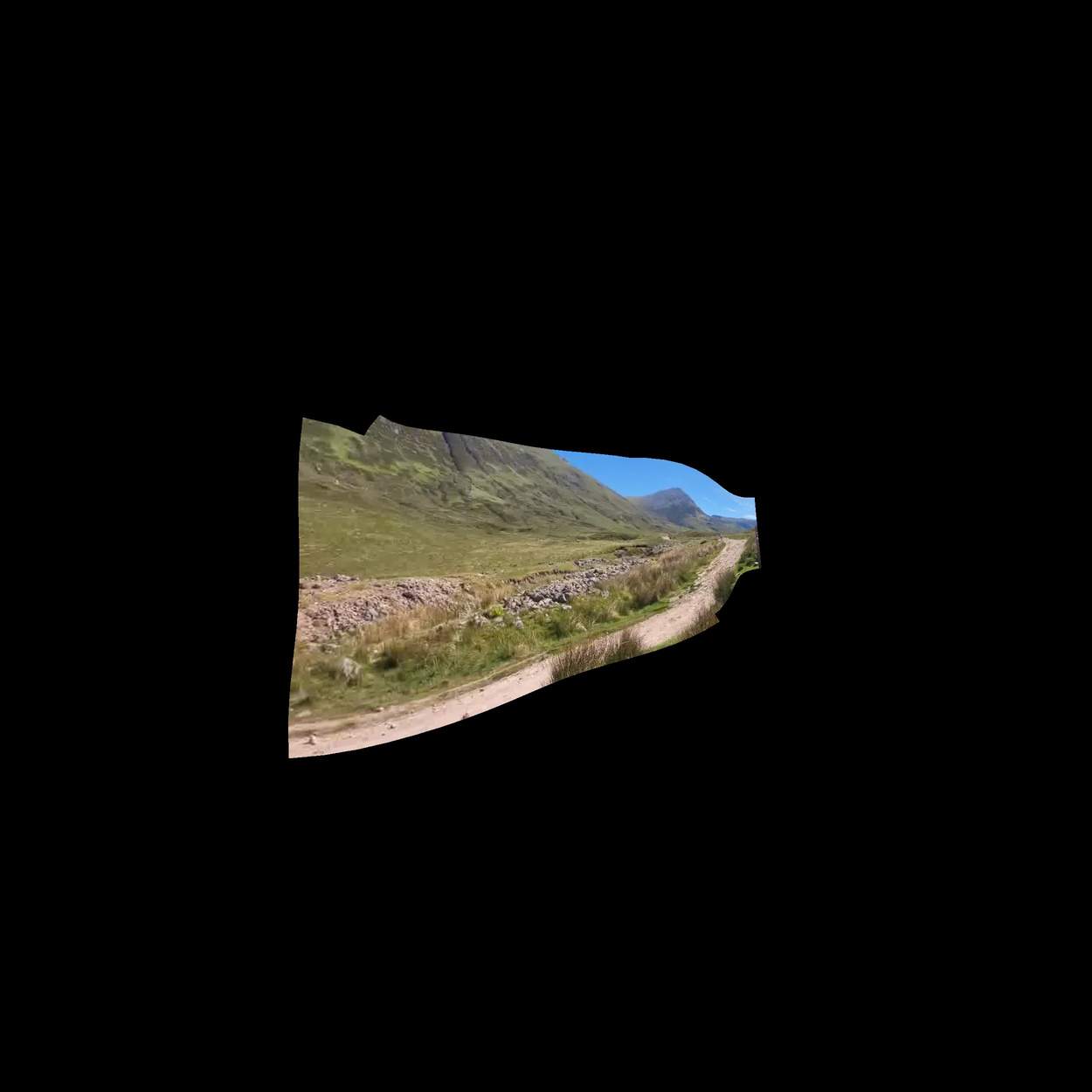} &
\includegraphics[width=3.8cm, height=2.1cm,    viewport=700 700 2300 1700,   clip]{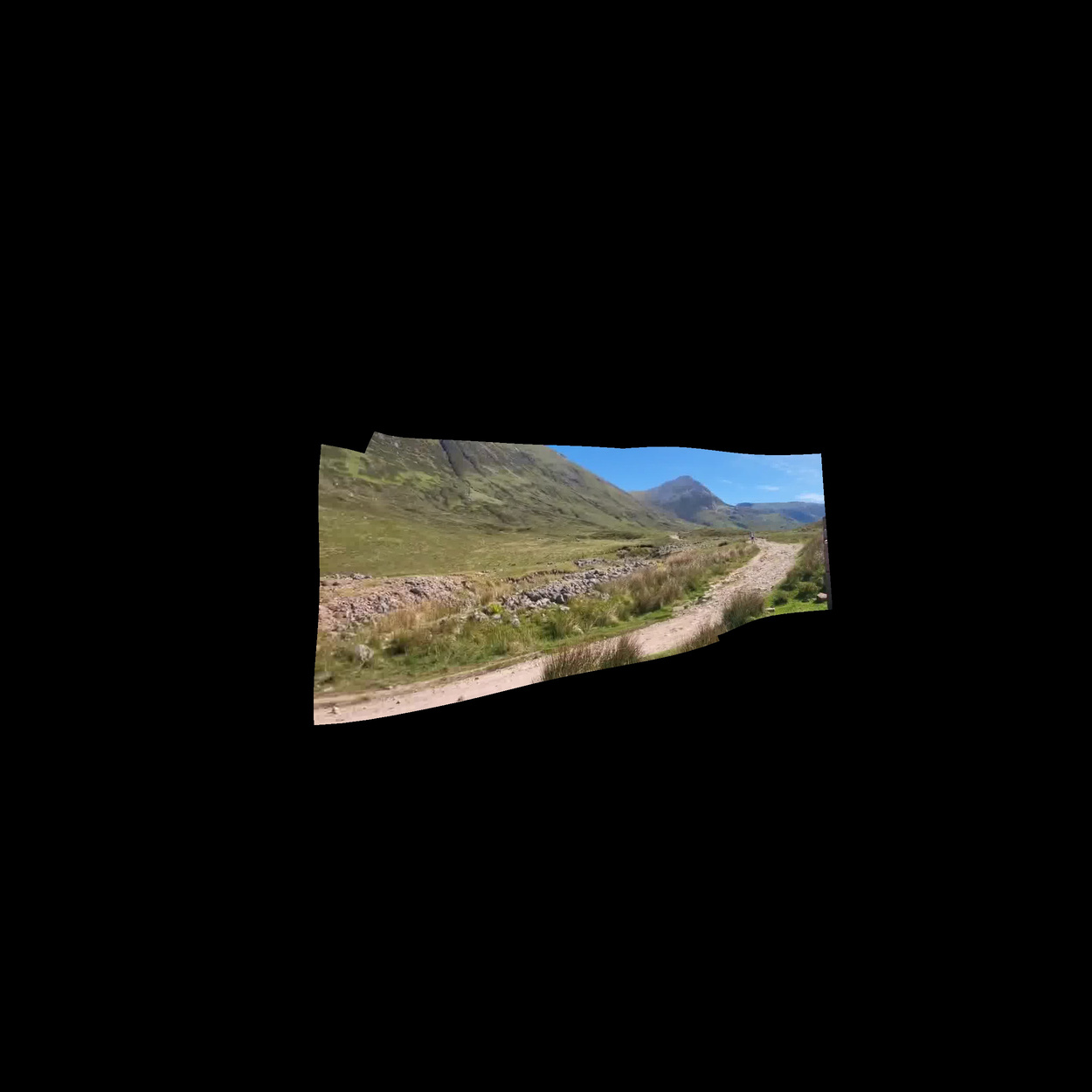}&
\includegraphics[width=3.8cm, height=2.1cm,    viewport=700 700 2300 1700,   clip]{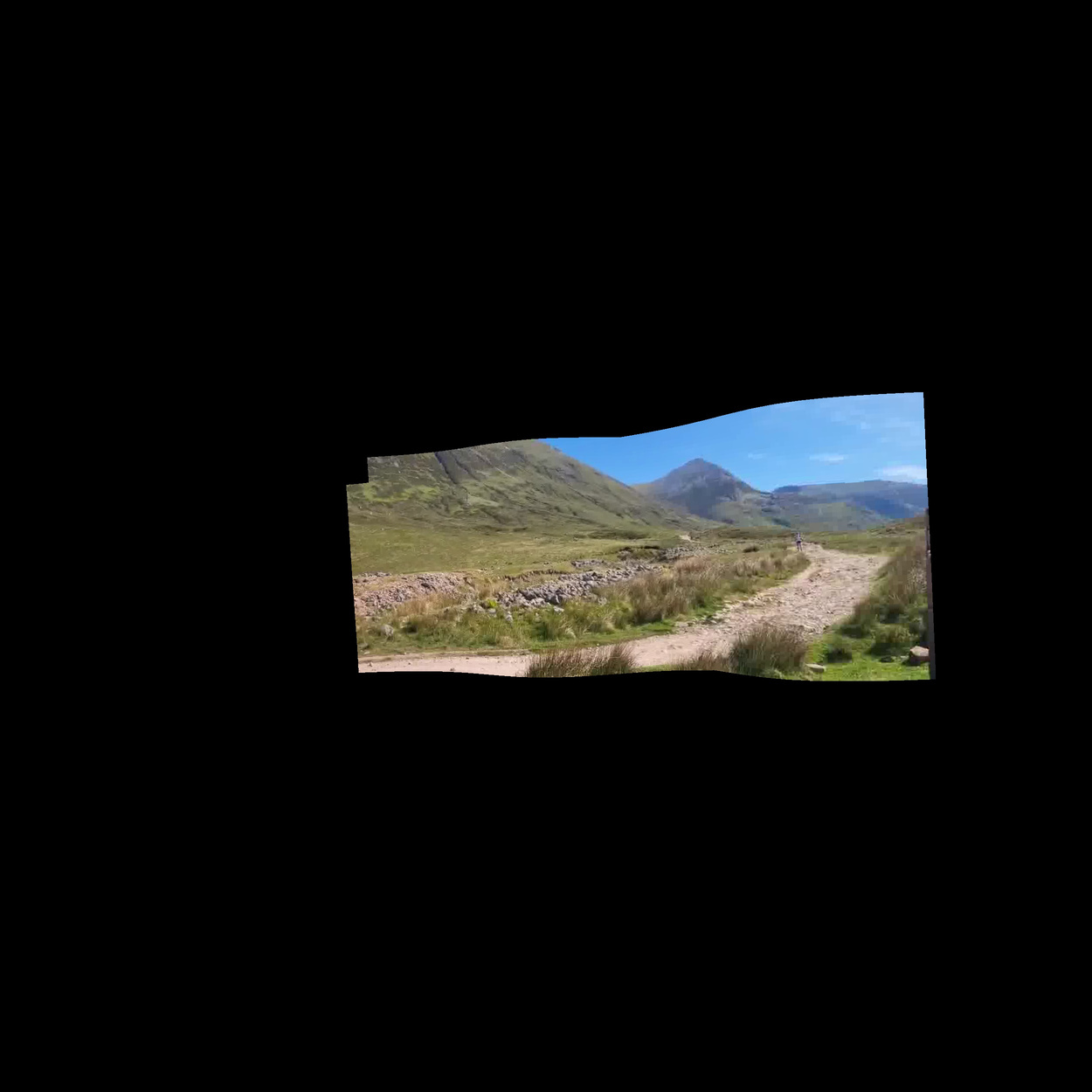} \\[0.5mm]

\rotatebox{90}{\parbox{2.1cm}{\centering \small GES-GSP ~\cite{Du:CVPR:2022:GESGSP}}} & 
\includegraphics[width=3.8cm, height=2.1cm,    viewport=700 700 2300 1700,   clip]{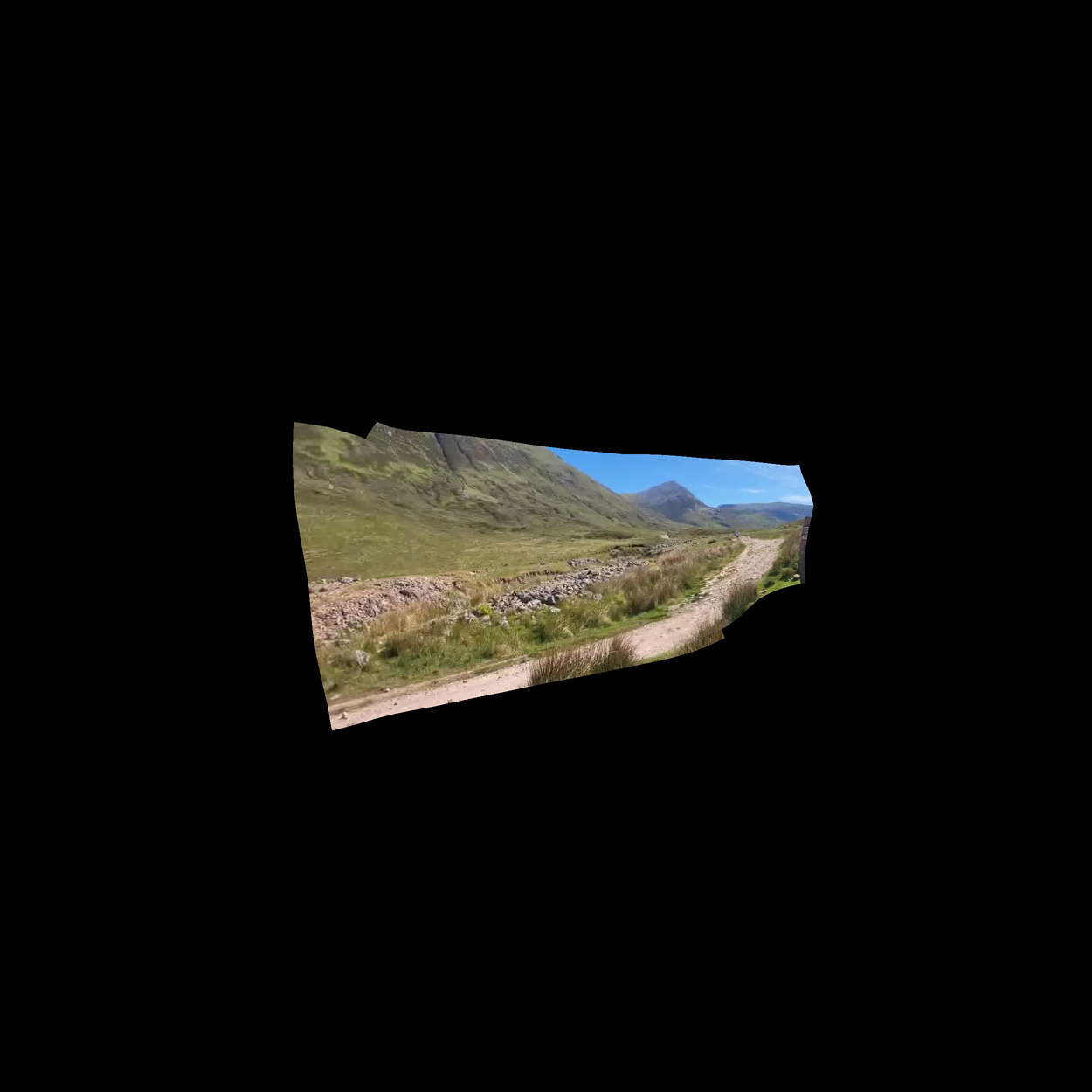} &
\includegraphics[width=3.8cm, height=2.1cm,    viewport=700 700 2300 1700,   clip]{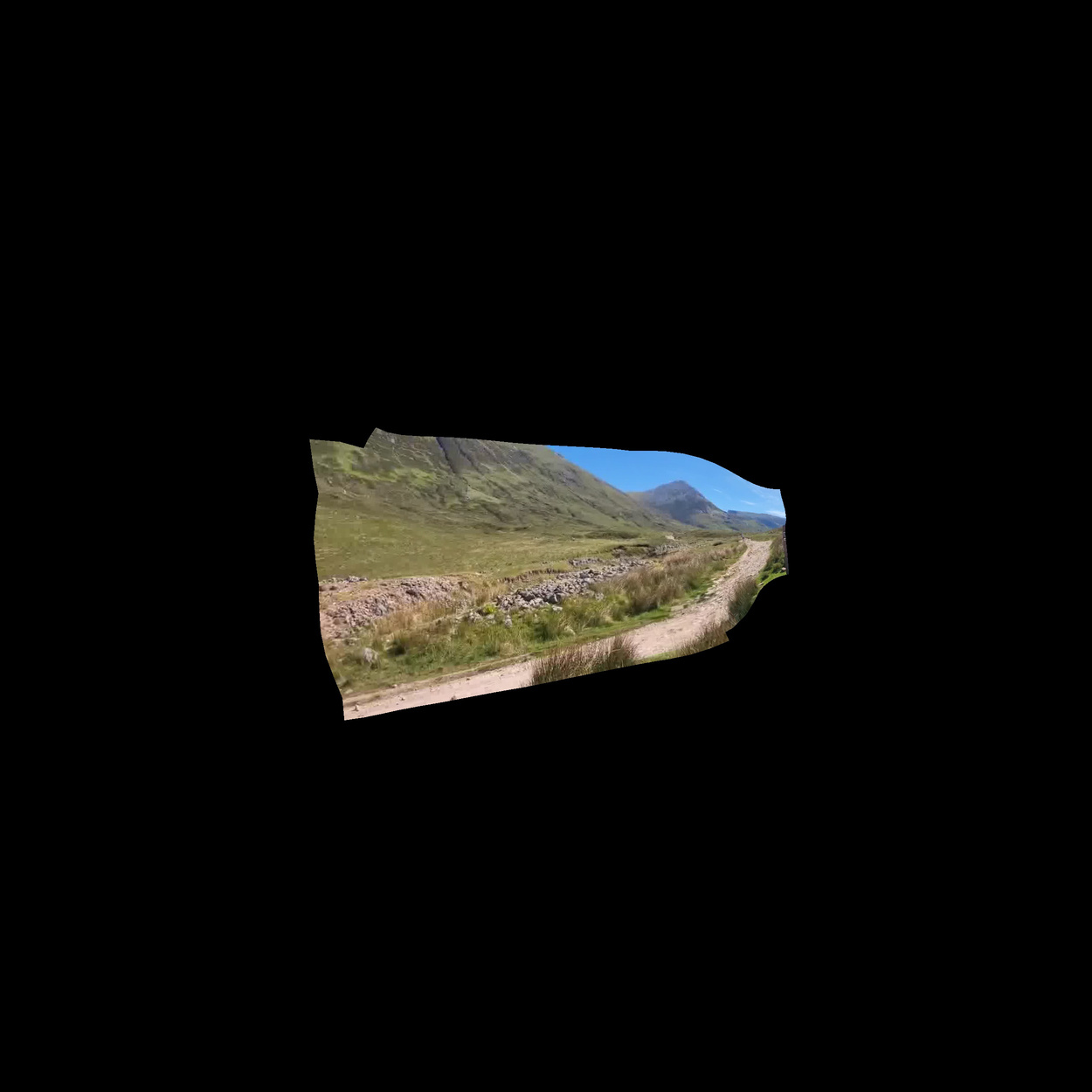} &
\includegraphics[width=3.8cm, height=2.1cm,    viewport=700 700 2300 1700,   clip]{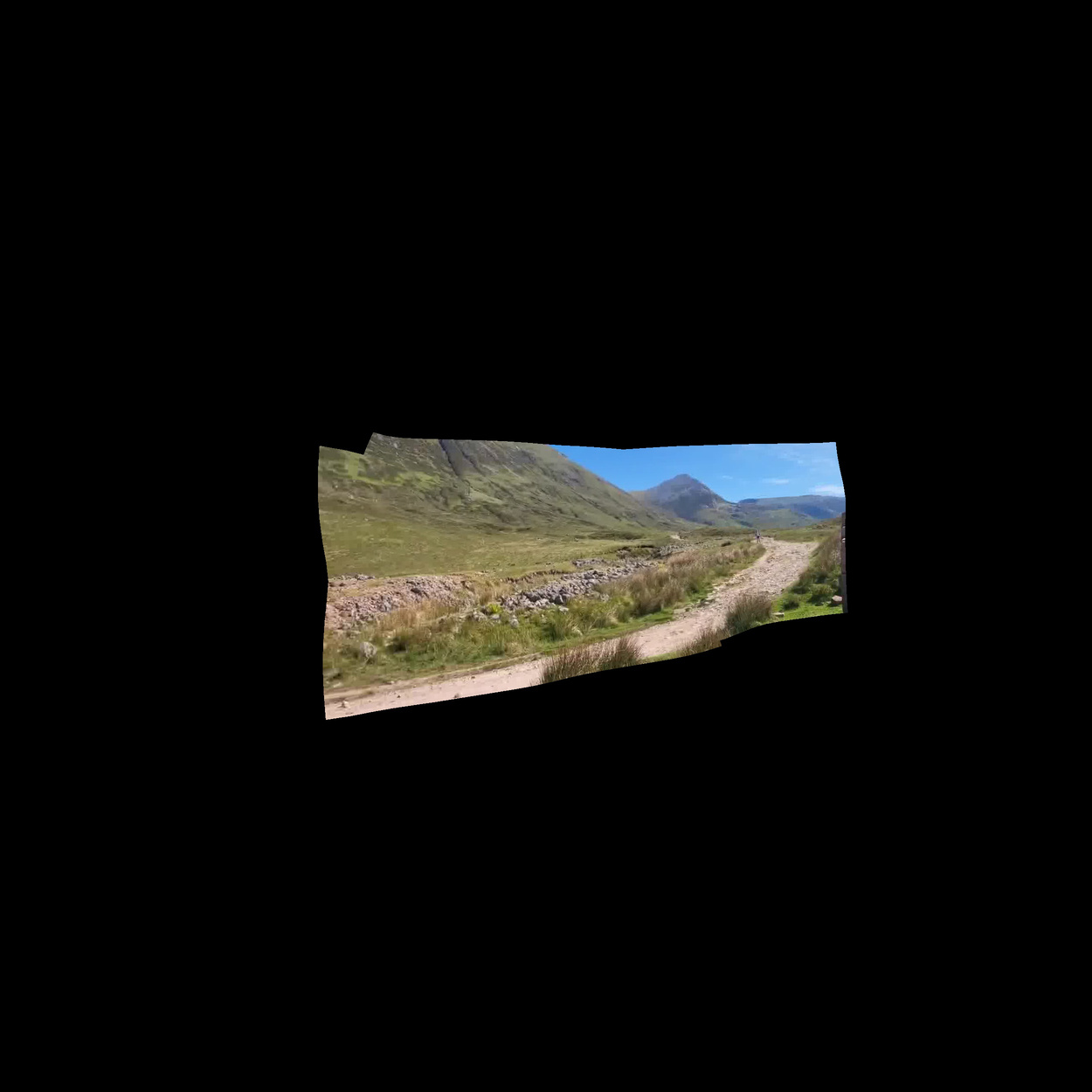}&
\includegraphics[width=3.8cm, height=2.1cm,    viewport=700 700 2300 1700,   clip]{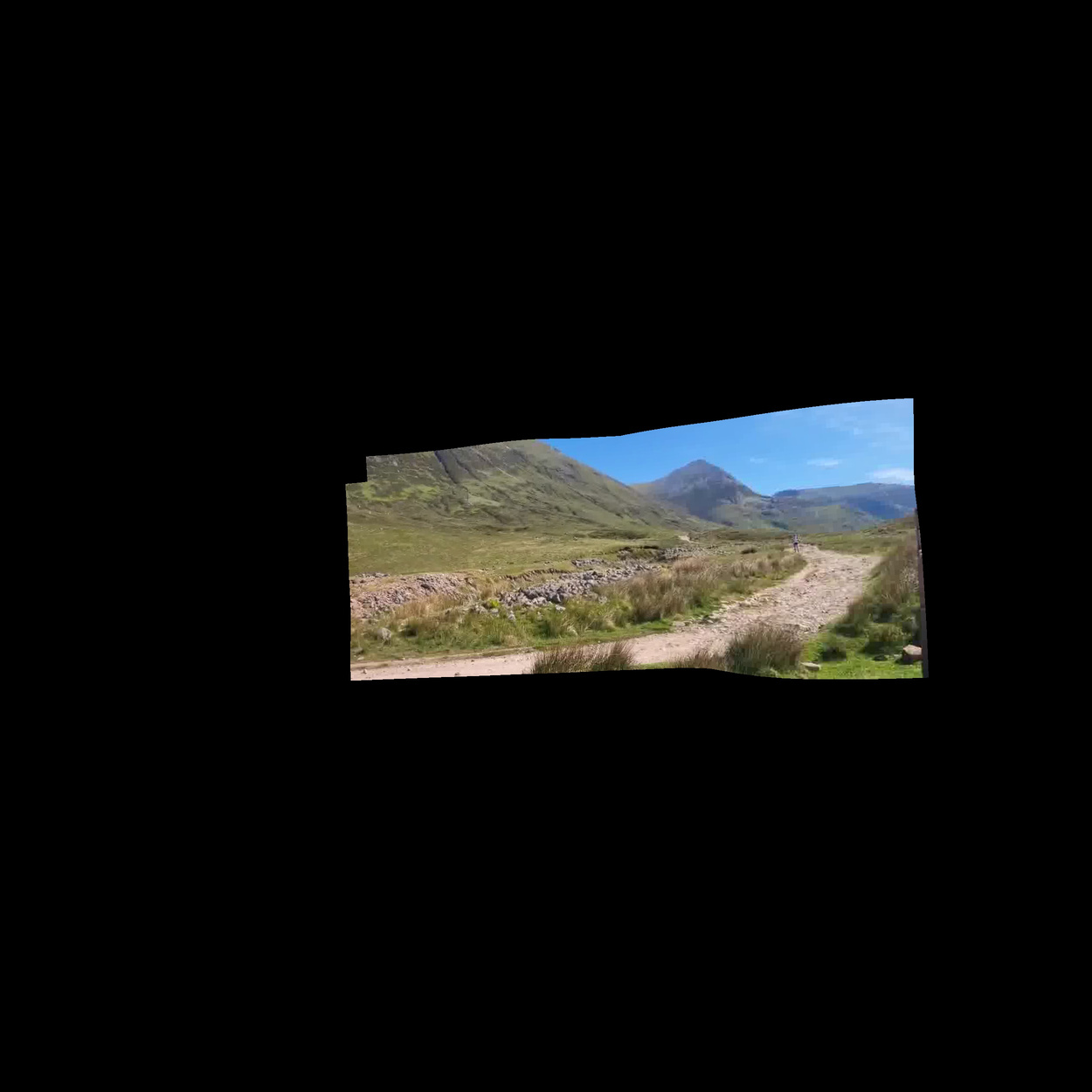} \\[0.5mm]

\rotatebox{90}{\parbox{2.1cm}{\centering \small OBJ-GSP ~\cite{Cai:AAAI:2025:OBJGSP}}} & 
\includegraphics[width=3.8cm, height=2.1cm,    viewport=700 700 2300 1700,   clip]{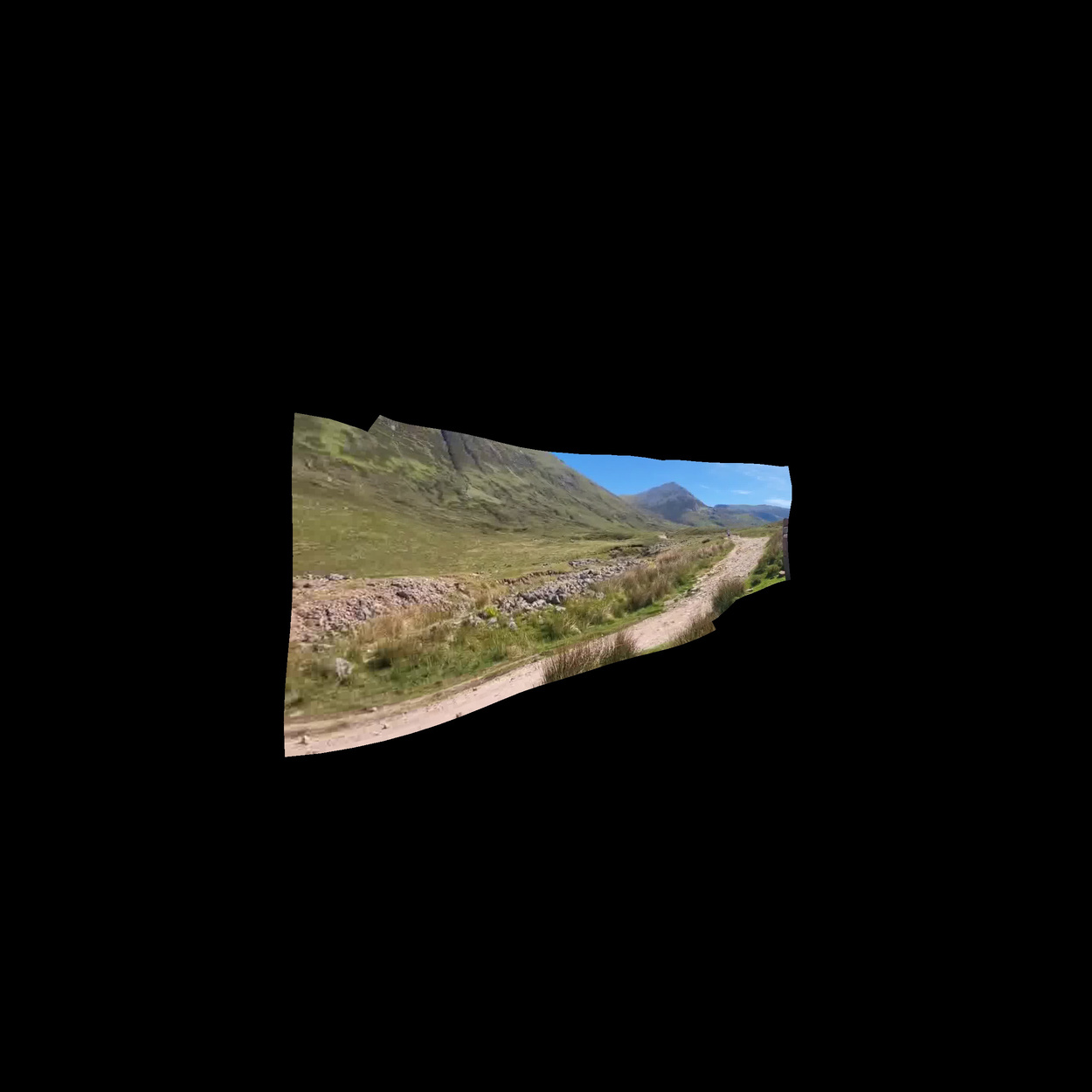} &
\includegraphics[width=3.8cm, height=2.1cm,    viewport=700 700 2300 1700,   clip]{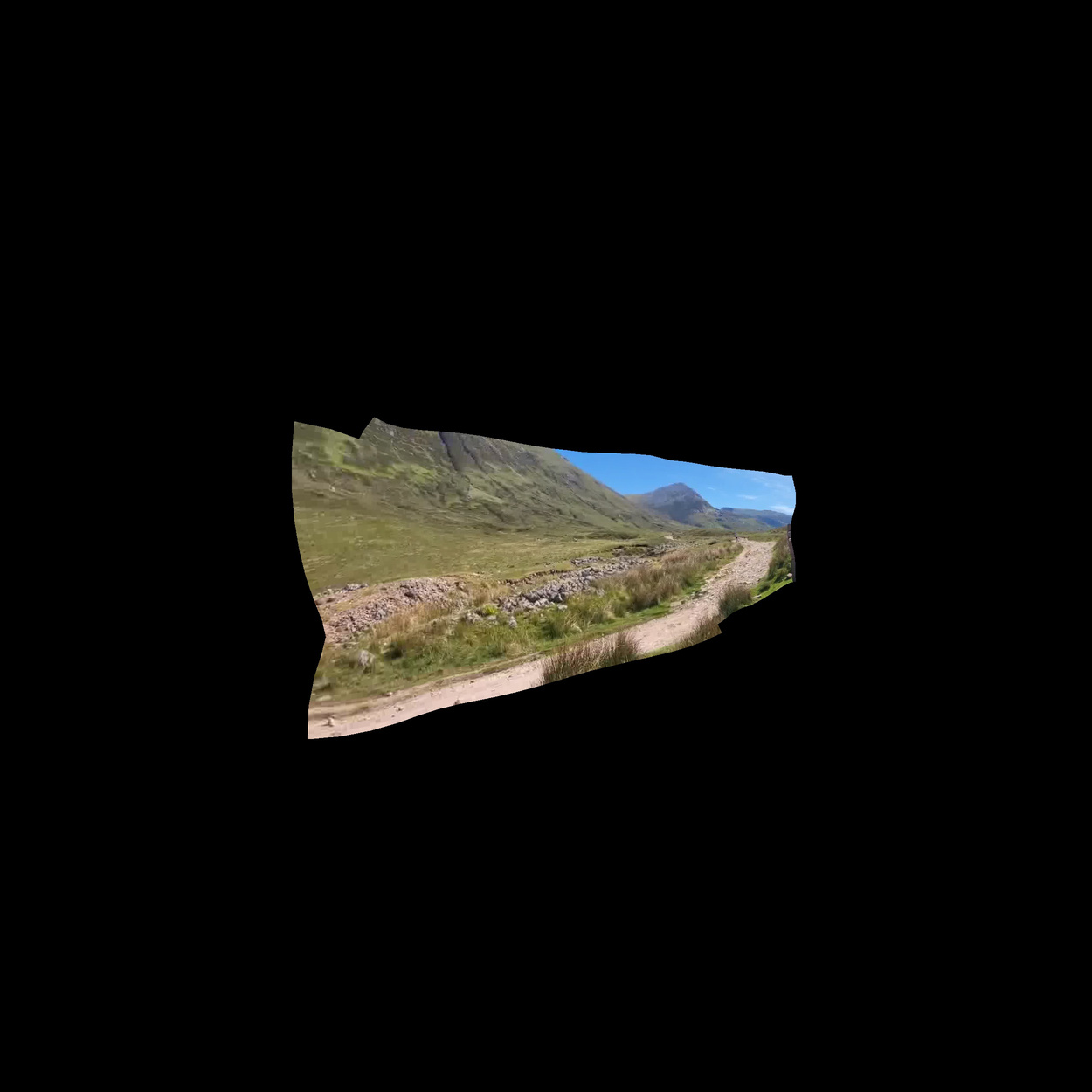} &
\includegraphics[width=3.8cm, height=2.1cm,    viewport=700 700 2300 1700,   clip]{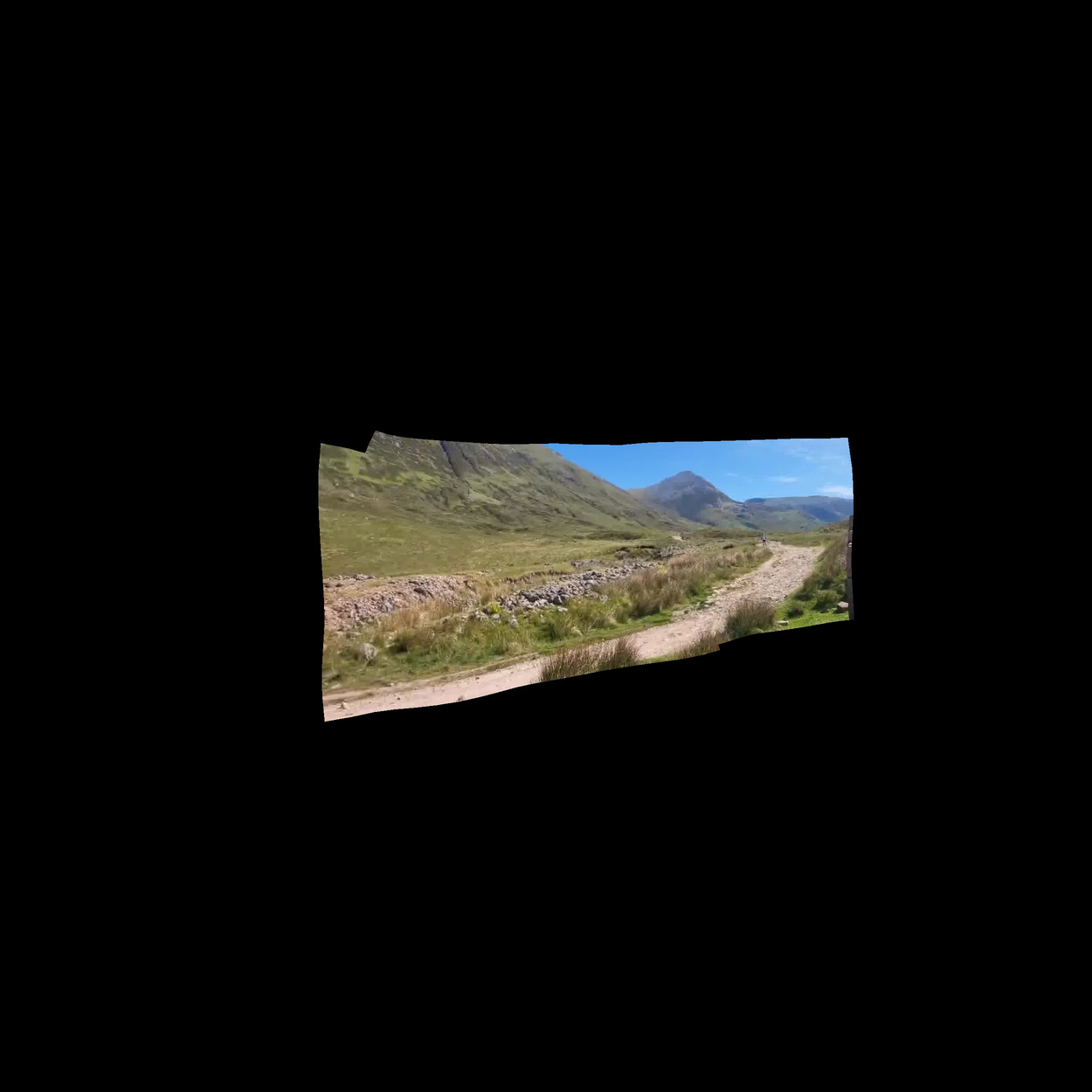}&
\includegraphics[width=3.8cm, height=2.1cm,    viewport=700 700 2300 1700,   clip]{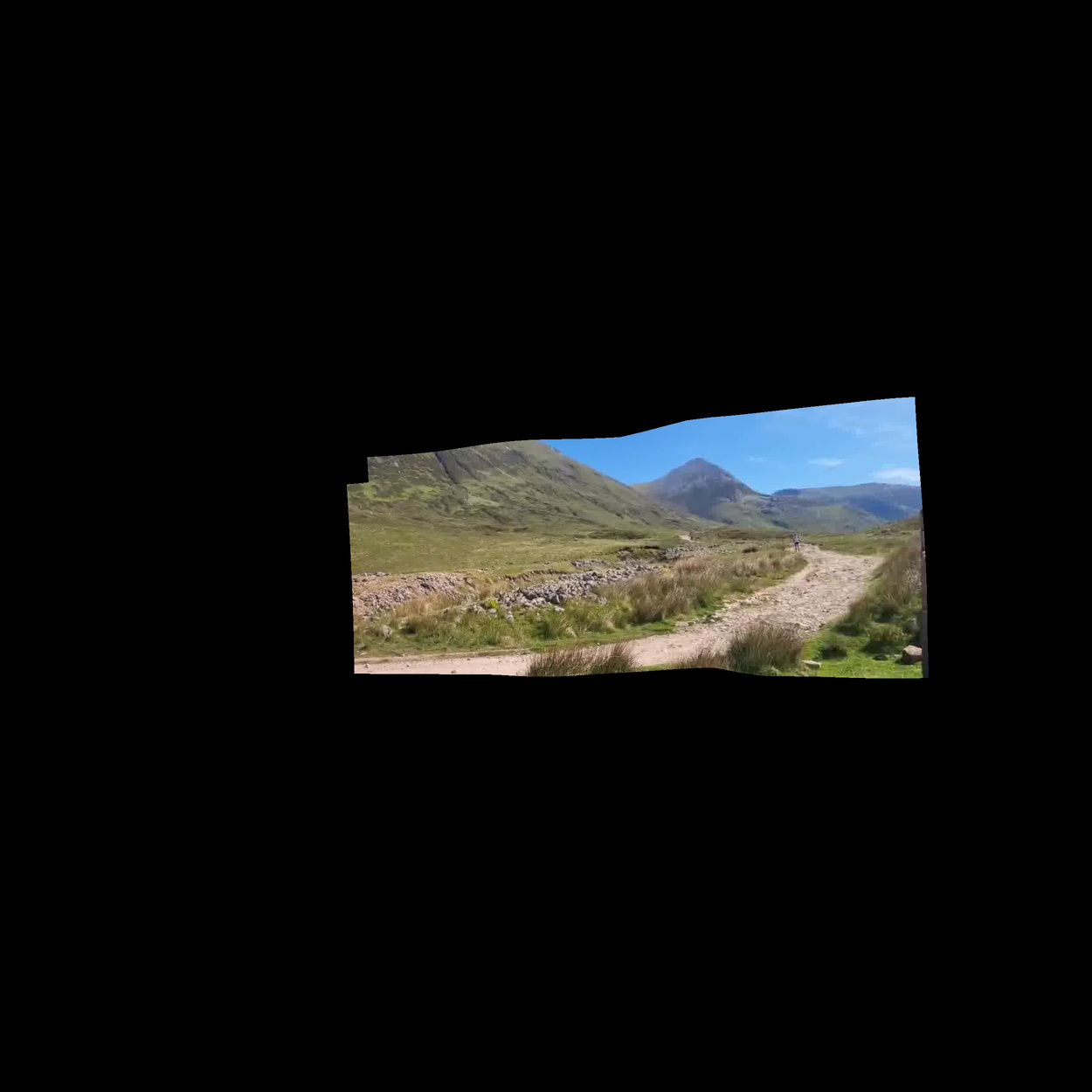} \\[0.5mm]

\rotatebox{90}{\parbox{2.1cm}{\centering \small MViS (Ours)}} & 
\includegraphics[width=3.8cm, height=2.1cm,    viewport=480 510 1792 1192,   clip]{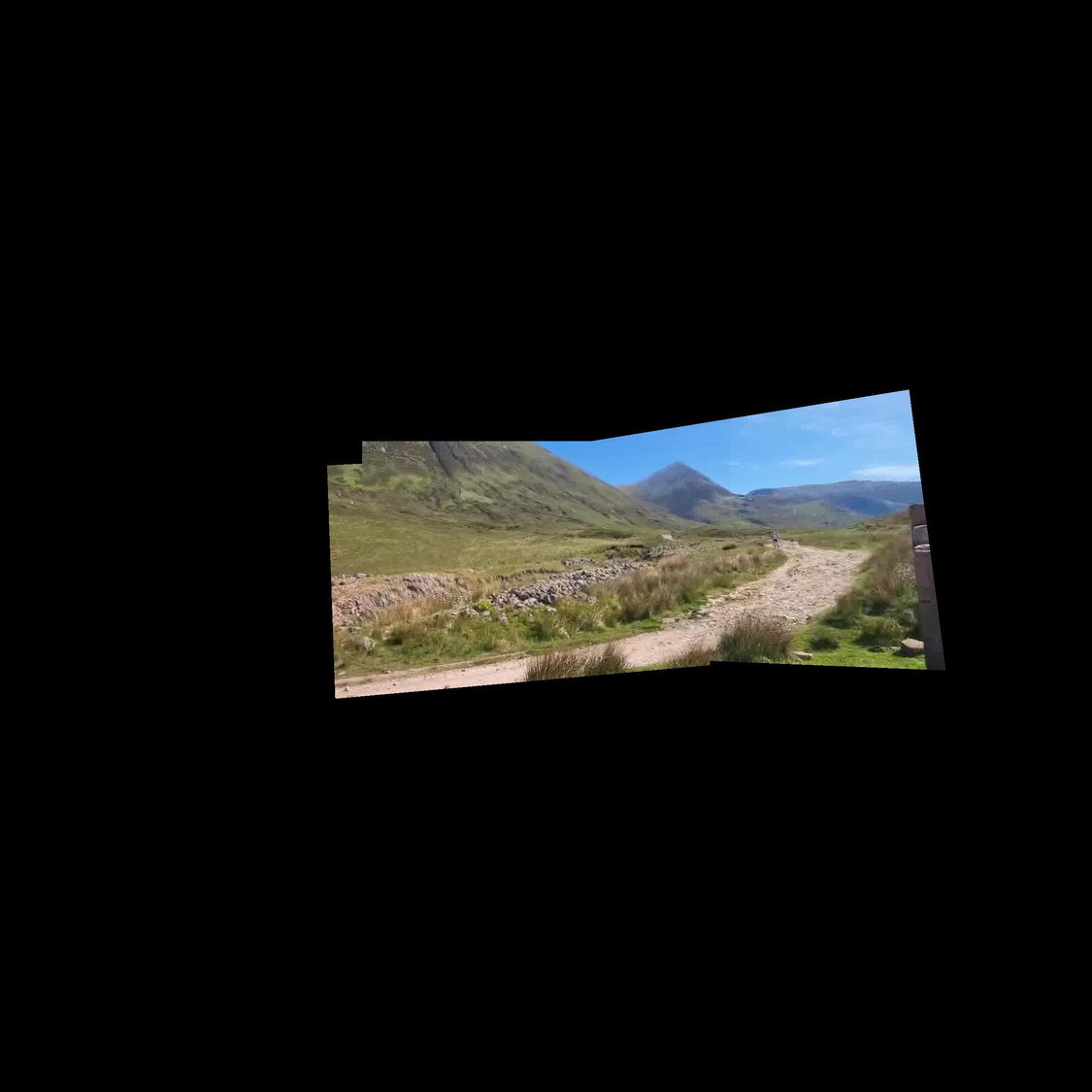} &
\includegraphics[width=3.8cm, height=2.1cm,    viewport=480 510 1792 1192,   clip]{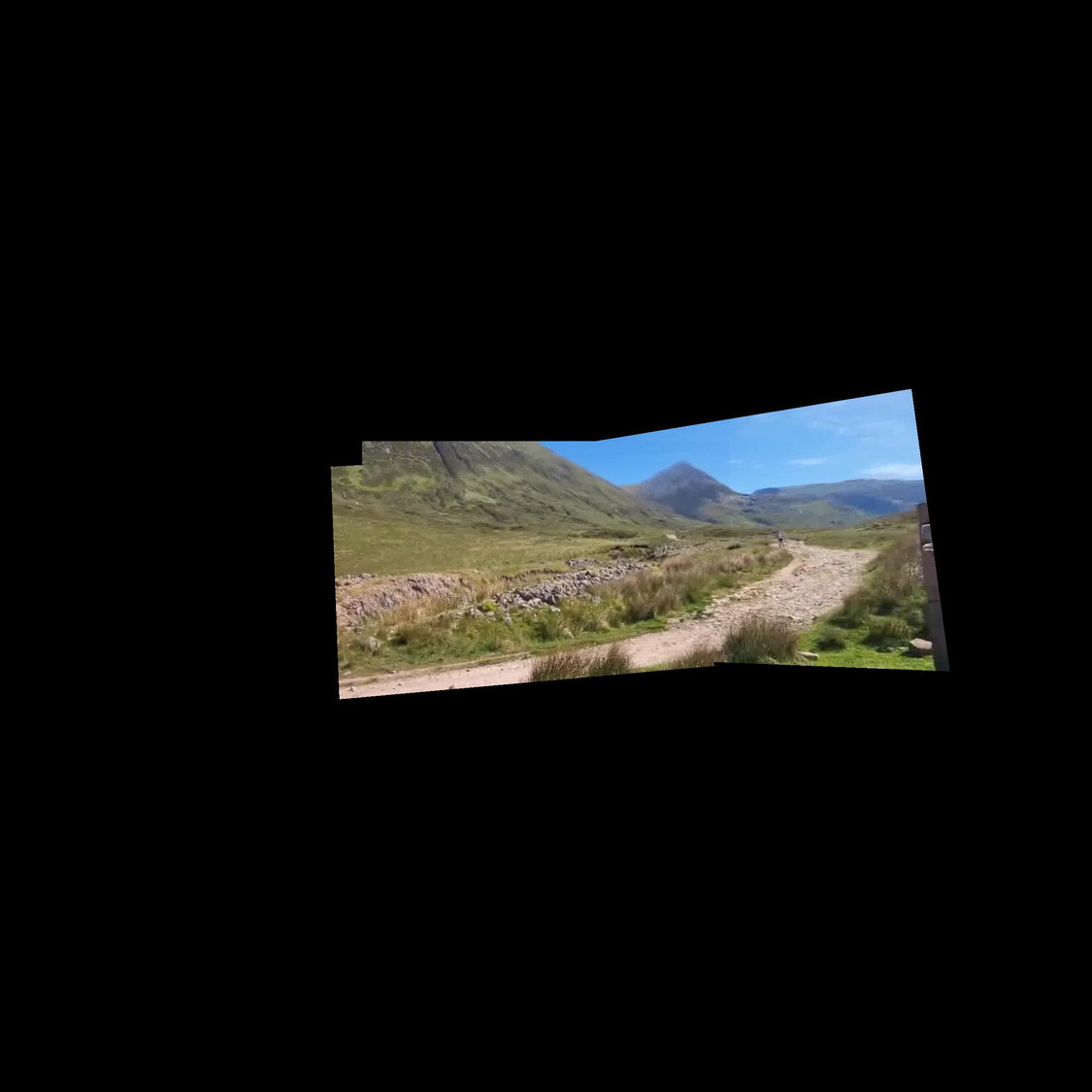} &
\includegraphics[width=3.8cm, height=2.1cm,    viewport=480 510 1792 1192,   clip]{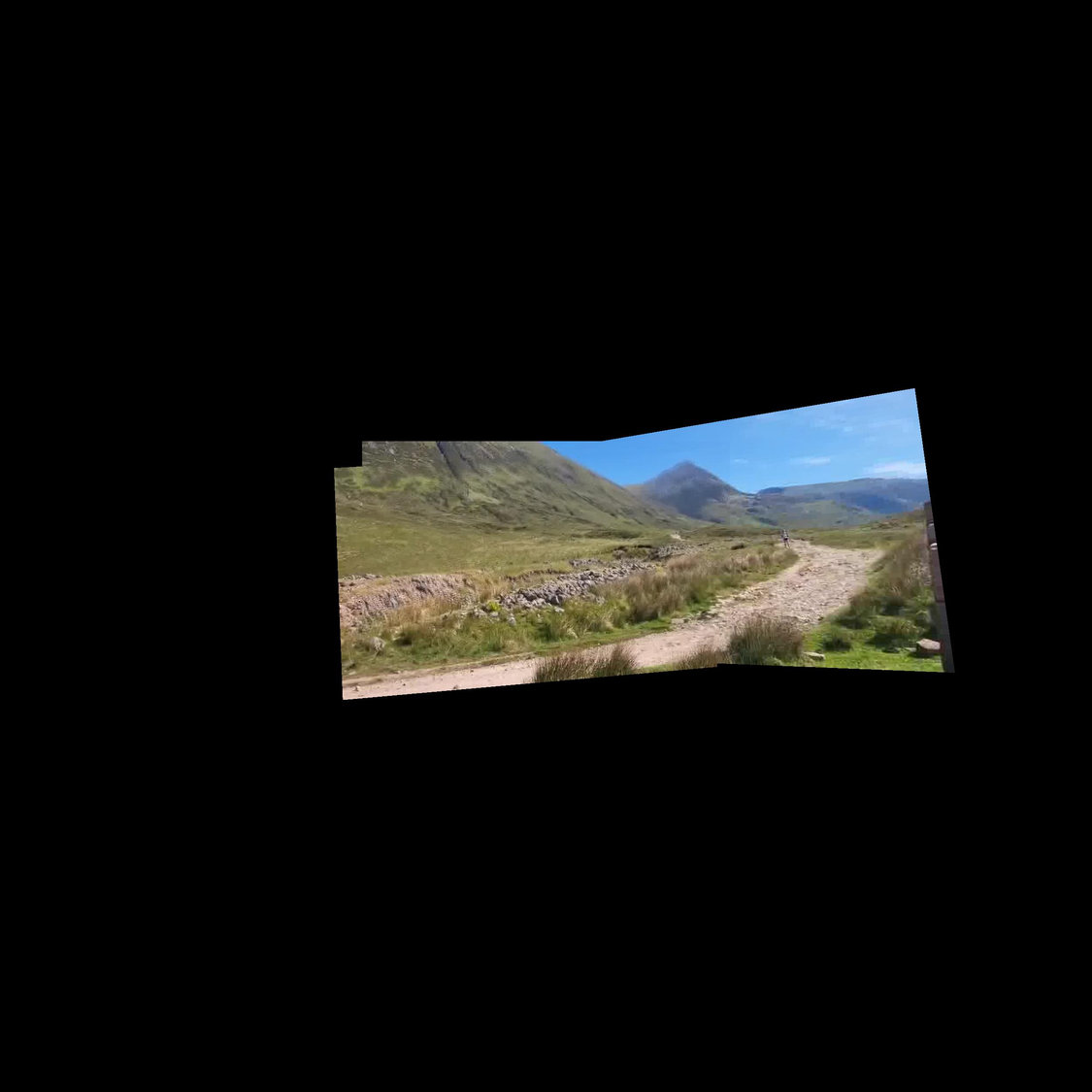}&
\includegraphics[width=3.8cm, height=2.1cm, viewport=480 510 1792 1192,   clip]{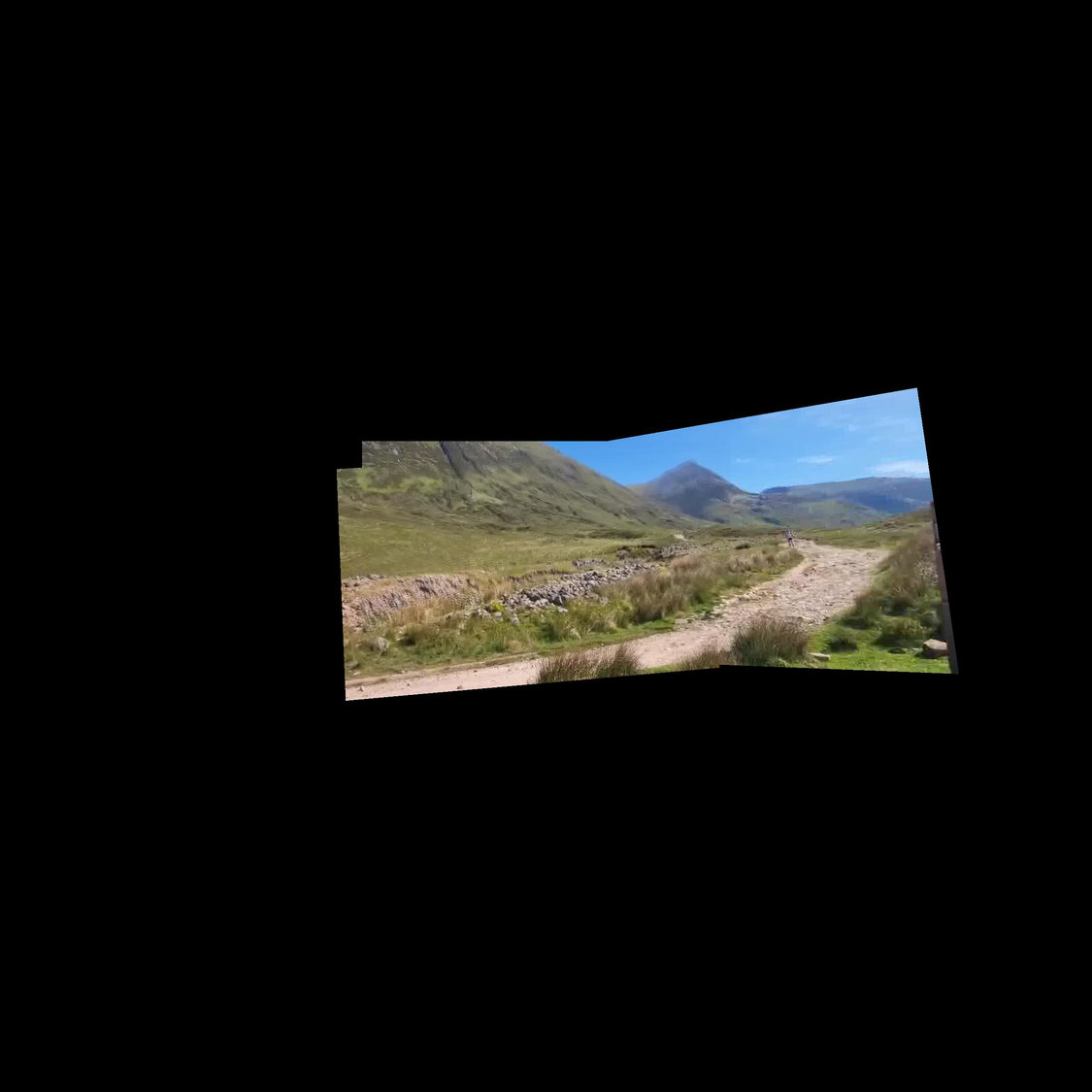} \\[-0.5mm]

& \small{$t=90$} & 
\small{$t=91$} & 
\small{$t=92$} &
\small{$t=93$} \\[3mm]

\end{tabular}
}
\caption{Visual comparison between MIS methods and MViS on scenes with $N>2$ cameras from the MMCD dataset.
We show a sequence of consecutive timestamps to illustrate that, although MIS methods often produce high-quality panoramas at individual frames, they lack temporal consistency.}
  \label{fig:mis_vis}
\end{figure}

This document contains the following:
    \begin{enumerate}[label=\Alph*., leftmargin=5.5em,
itemsep=0.8em] 
    \refstepcounter{enumi}

    \item[\autoref{sec:qual_res}.] Additional Qualitative Results:
    \begin{enumerate}[label=\arabic*), itemsep=0.8em]

    \item An Overview of the Supplementary Videos

    \item Comparisons on the $N=3$ Case (\autoref{fig:triplets})

    \item MViS panoramas with dynamically changing overlap graphs (\autoref{fig:view_graph}).
          
    \item Comparison to Multiple Image Stitching (MIS) in $N>2$ (\autoref{fig:mis_vis})
    \end{enumerate}
                
        \refstepcounter{enumi}
        
        \item[\autoref{sec:sl}.] Parameterizing Homographies via the Special Linear Group
        
        \item[\autoref{sec:exps}.]Additional Experiments:
        \begin{enumerate}[label=\arabic*), itemsep=0.8em]
        
        \item [\autoref{sec:optimization}.] Optimization without a Neural Network
        \item [\autoref{sec:reg}.] Regularization Experiments 
        \item [\autoref{sec:matchers}.] Ablation on Image Matcher Choice
        \item [\autoref{sec:camera_motion}.] Effect of Camera Motion
        
        \item [\autoref{sec:ort_corruption}.] Robustness to ORT Corruption
        \end{enumerate}
        
        \item[\autoref{sec:arch}.] The Alignment Network Architecture
        
        \item[\autoref{sec:details_ben}.]Further Details on the Multi-Camera Benchmark and MMCD Dataset:
        \begin{enumerate}[label=\arabic*), itemsep=0.8em]
        \item[\autoref{sec:geovar}.] The Geometric Variance Metric (GV)
        \item[\autoref{sec:evaluation_protocol}.] The Multi-Camera Benchmark Evaluation Protocol
        \item[\autoref{sec:extended_mis}.] Implementation details of the Extended MIS Variants
        \item[\autoref{sec:mmcd_stats}.] Additional MMCD Statistics (\autoref{fig:dataset_fig}, \autoref{tab:overlap_ratio})
        \item[\autoref{sec:benchmark}.] Full Multi-Camera Benchmark Results (\autoref{tab:gv_mis_table}-\ref{tab:cmsa_exts})
        
        \end{enumerate}
        
    \end{enumerate}
In addition, key notational conventions used across the main paper are summarized in \autoref{tab:notation}.

\clearpage
\section{Additional Qualitative Results}
\label{sec:qual_res}
\subsection*{An Overview of the Supplementary Videos}
\label{Sup:supp_videos}

\begin{figure}[!t]
\centering
\resizebox{\linewidth}{!}{
\setlength{\tabcolsep}{2pt} 
\begin{tabular}{c*{4}{c}}

\multicolumn{5}{l}{The relative left-right ordering of the cameras is not fixed; the cameras swap positions over time.} \\[2mm]
\rotatebox{90}{\small StabStitch++} & 
\includegraphics[width=3.8cm, height=2.1cm]{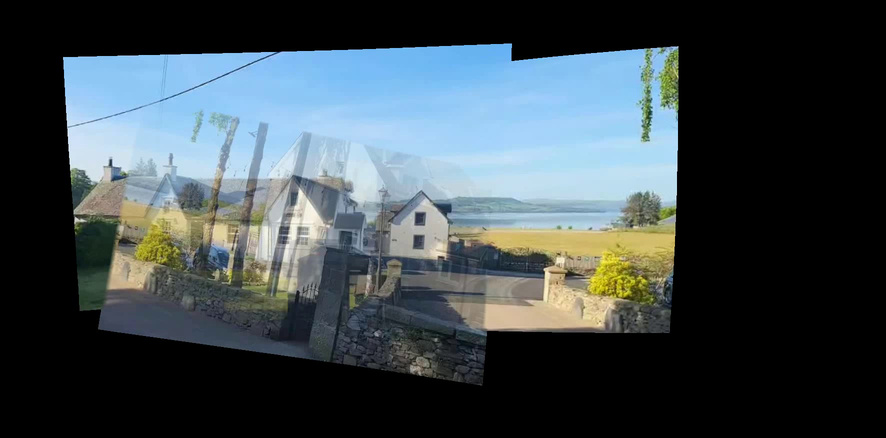} &
\includegraphics[width=3.8cm, height=2.1cm]{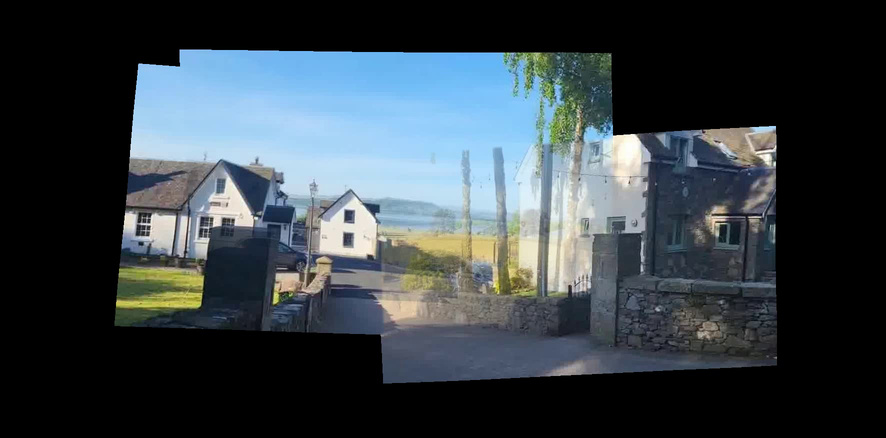} &
\includegraphics[width=3.8cm, height=2.1cm]{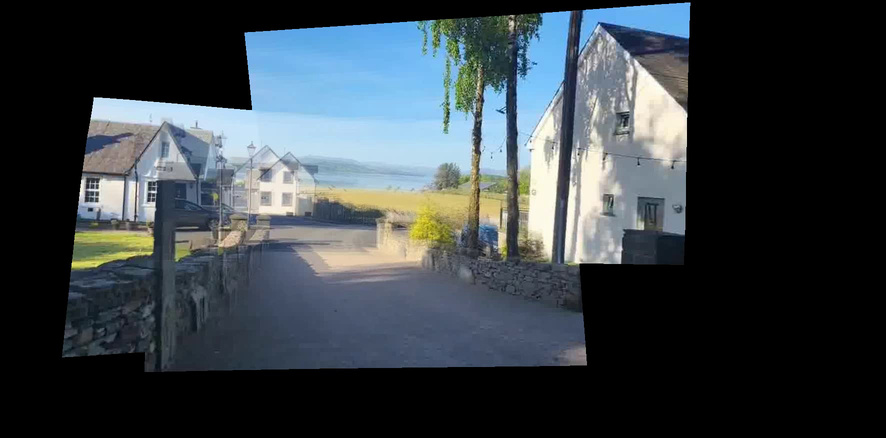}&
\includegraphics[width=3.8cm, height=2.1cm]{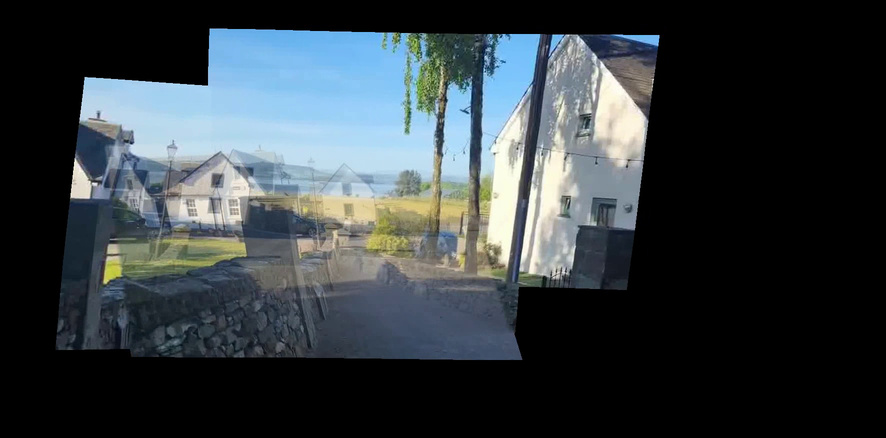} \\[0.2mm]

\rotatebox{90}{\small MViS (Ours)} & 
\includegraphics[width=3.8cm, height=2.1cm]{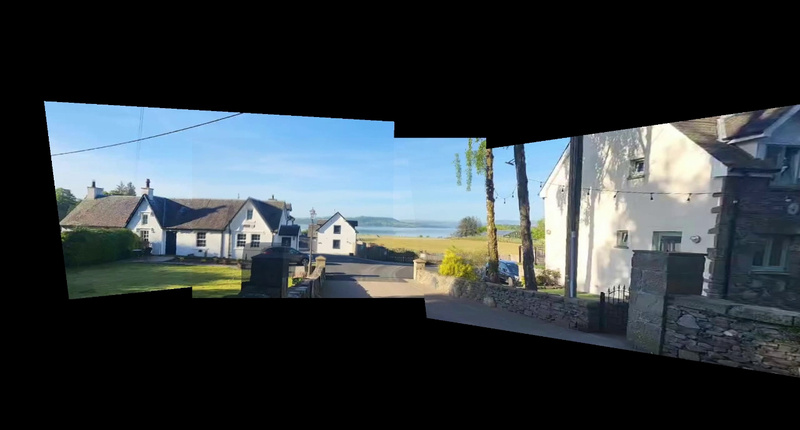} &
\includegraphics[width=3.8cm, height=2.1cm]{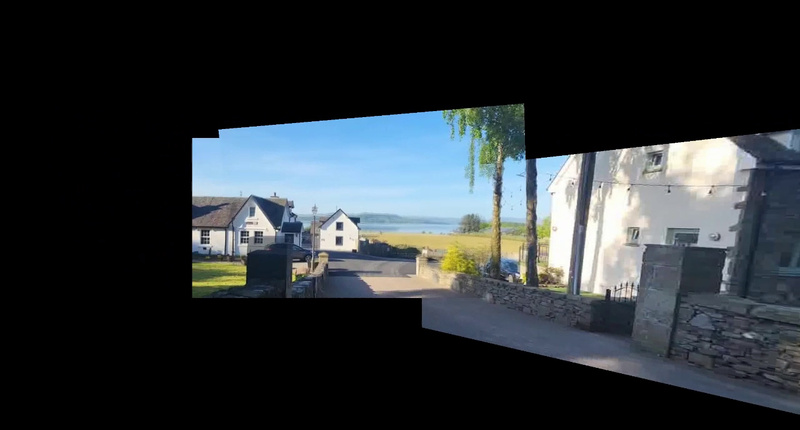} &
\includegraphics[width=3.8cm, height=2.1cm]{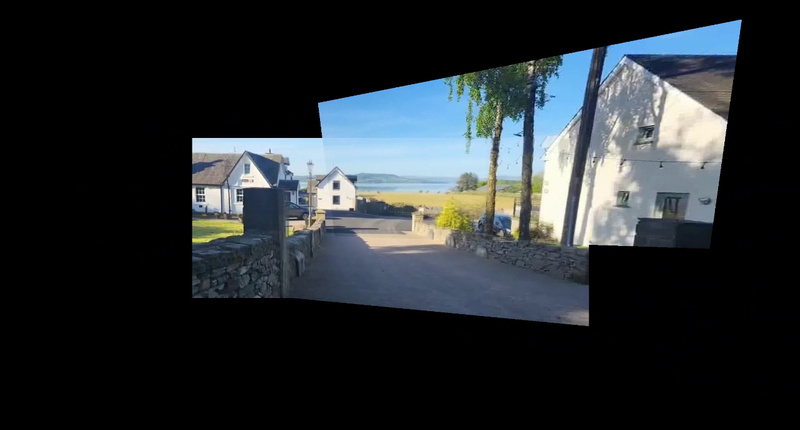}&
\includegraphics[width=3.8cm, height=2.1cm]{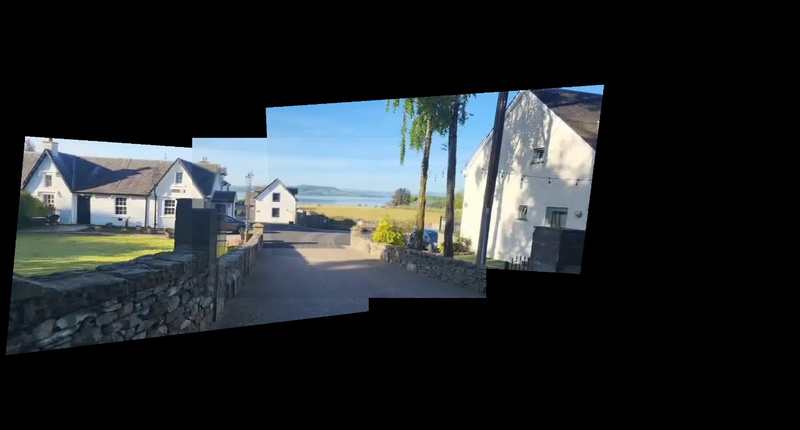} \\[-0.5mm]

& \small{$t=0$} & 
\small{$t=60$} & 
\small{$t=120$} &
\small{$t=180$} \\[3mm]

\multicolumn{5}{l}{Challenging motion from three independent moving cameras and extensive moving water in the scene.} \\[2mm]
\rotatebox{90}{\small StabStitch++} & 
\includegraphics[width=3.8cm, height=2.1cm]{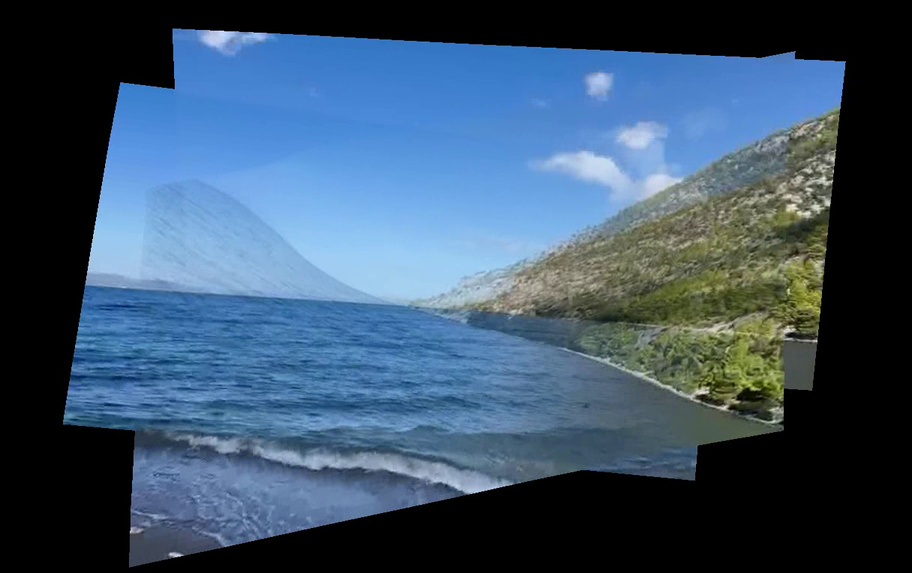} &
\includegraphics[width=3.8cm, height=2.1cm]{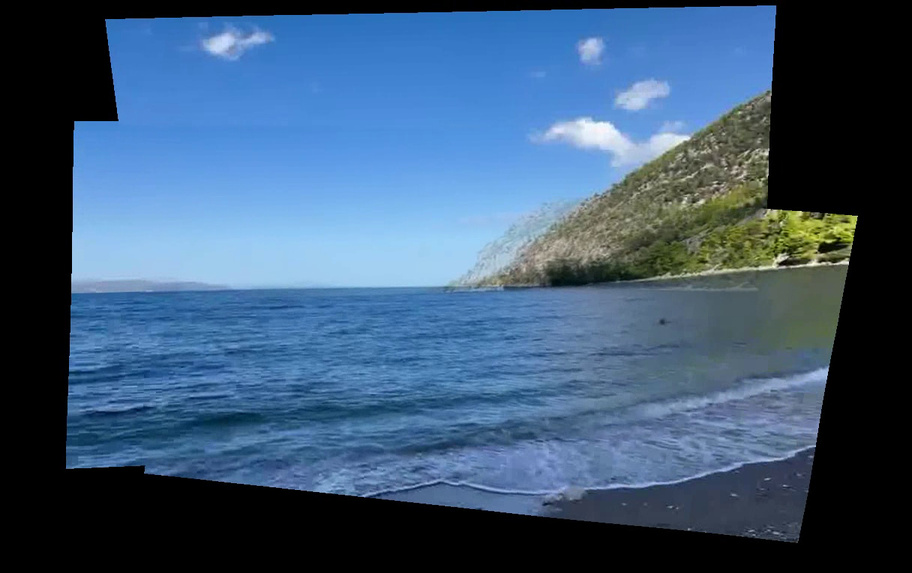} &
\includegraphics[width=3.8cm, height=2.1cm]{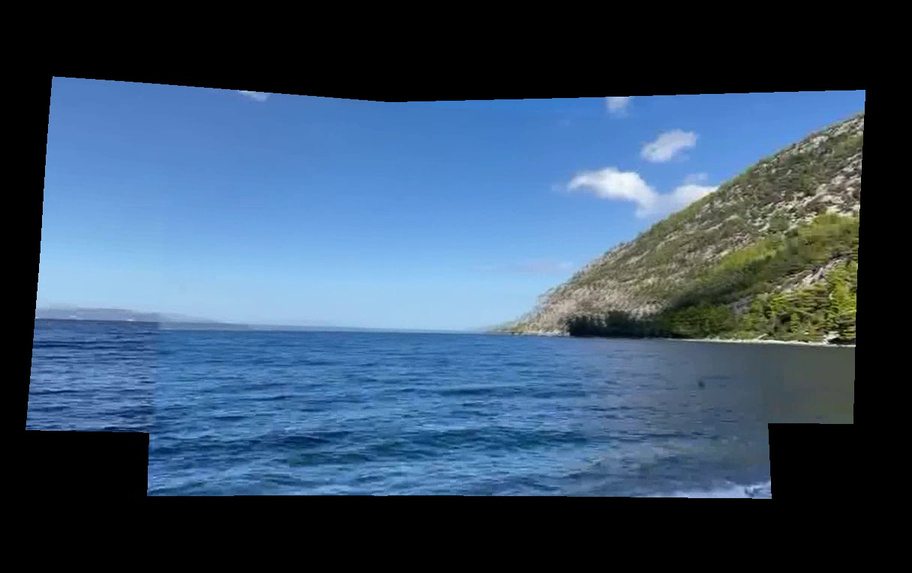} &
\includegraphics[width=3.8cm, height=2.1cm]{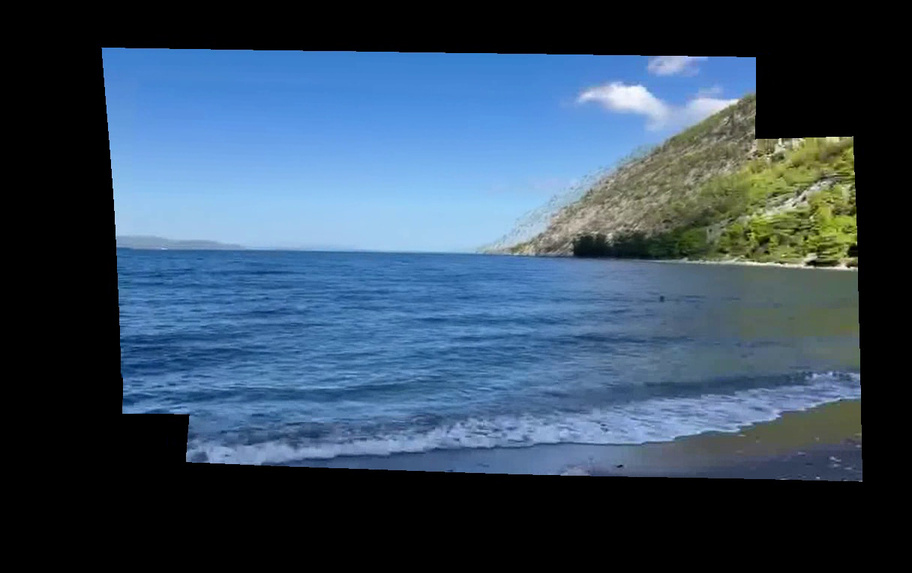} \\[0.2mm]

\rotatebox{90}{\small MViS (Ours)} & 
\includegraphics[width=3.8cm, height=2.1cm]{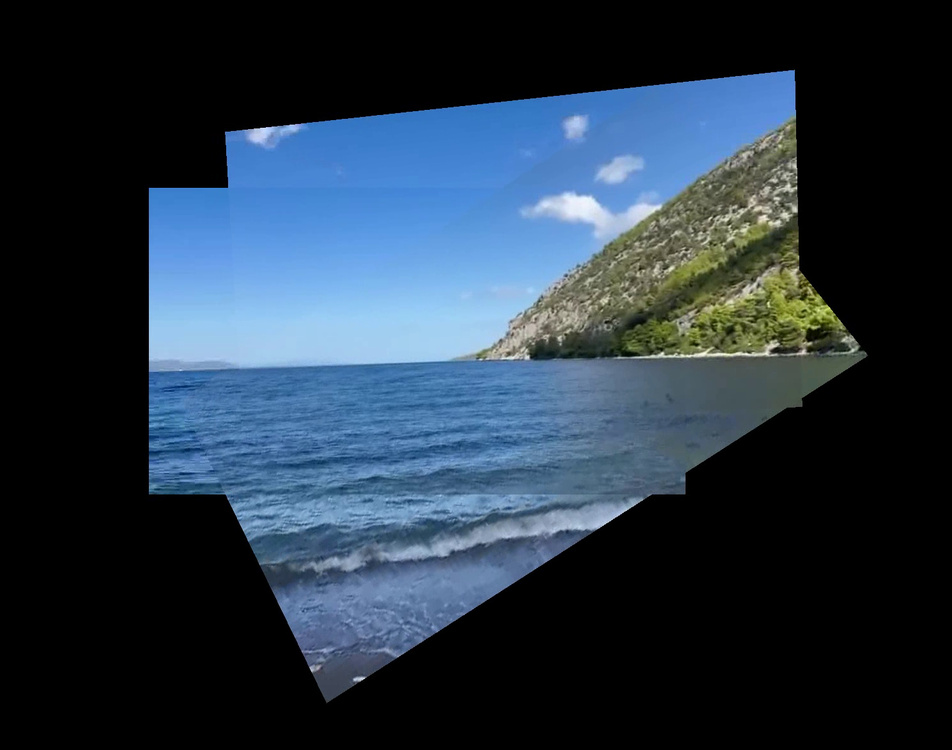} &
\includegraphics[width=3.8cm, height=2.1cm]{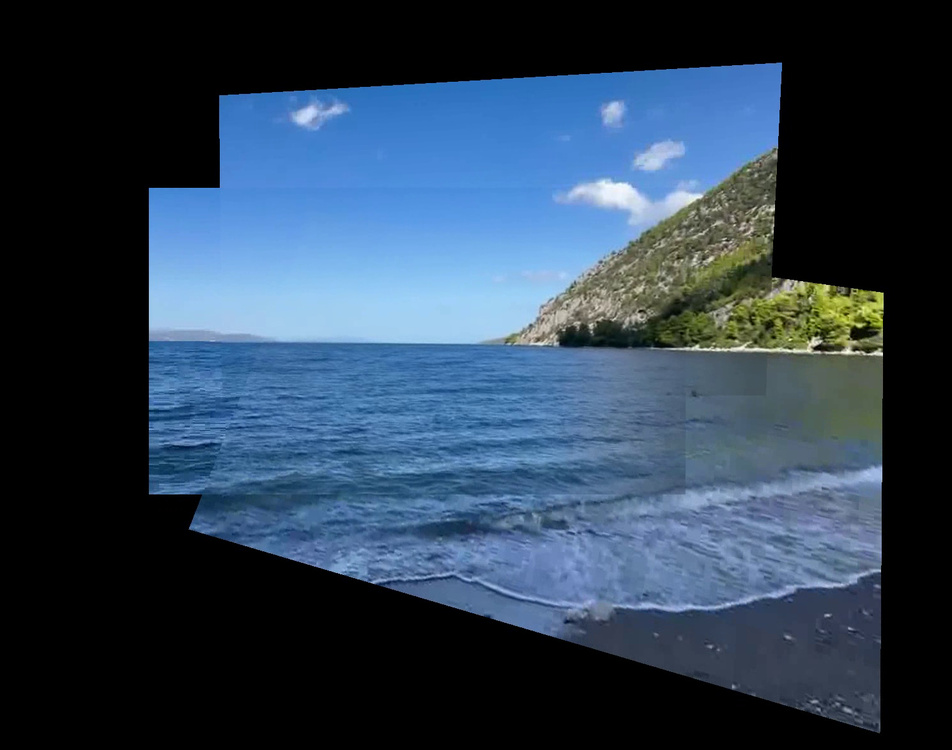} &
\includegraphics[width=3.8cm, height=2.1cm]{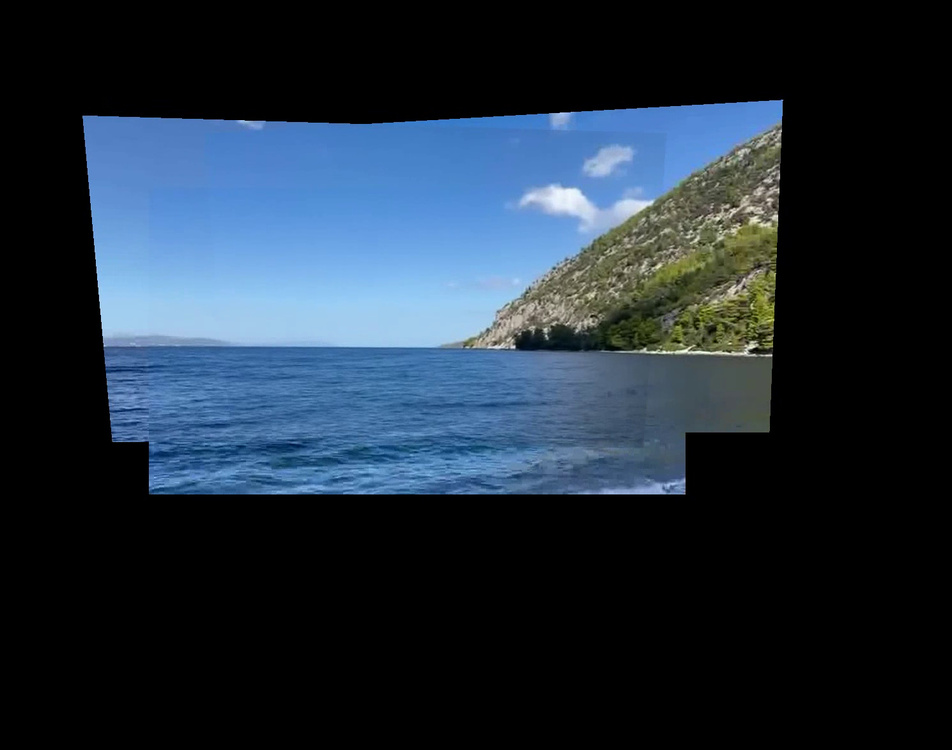} &
\includegraphics[width=3.8cm, height=2.1cm]{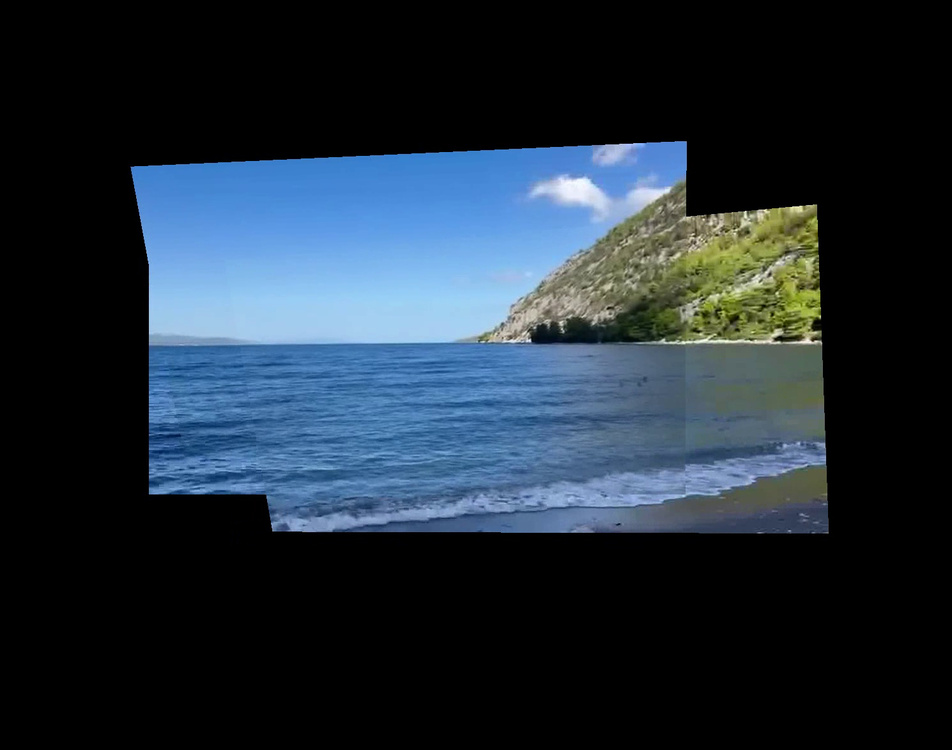} \\[-0.5mm]

& \small{$t=70$} & 
\small{$t=140$} & 
\small{$t=220$} & 
\small{$t=300$} \\[3mm]

\multicolumn{5}{l}{Very low inter-camera overlap and dynamically changing spatial ordering.} \\[2mm]
\rotatebox{90}{\small StabStitch++} & 
\includegraphics[width=3.8cm, height=2.1cm]{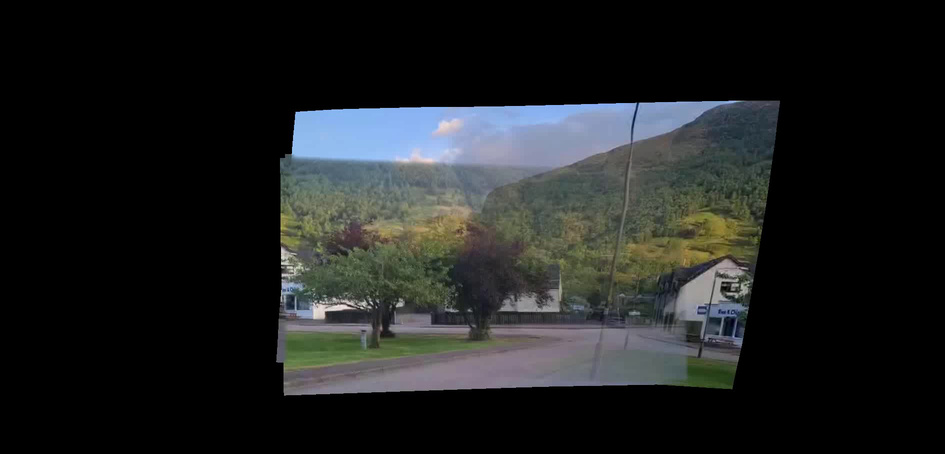} &
\includegraphics[width=3.8cm, height=2.1cm]{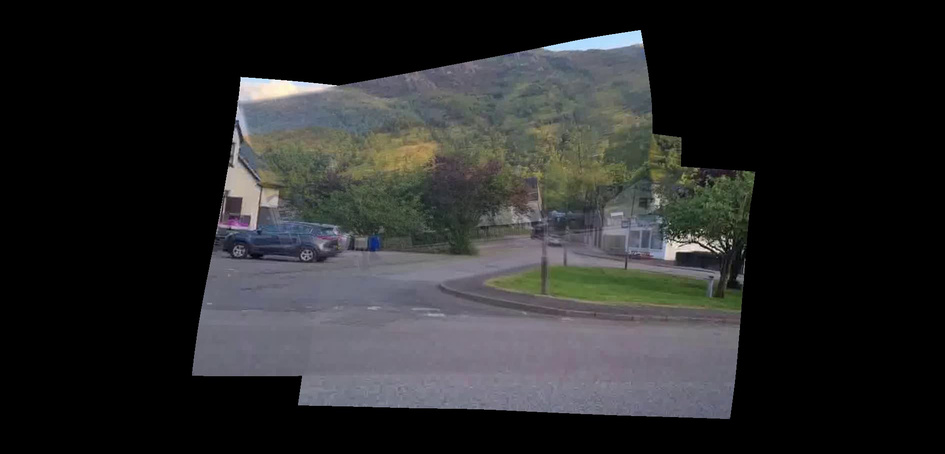} &
\includegraphics[width=3.8cm, height=2.1cm]{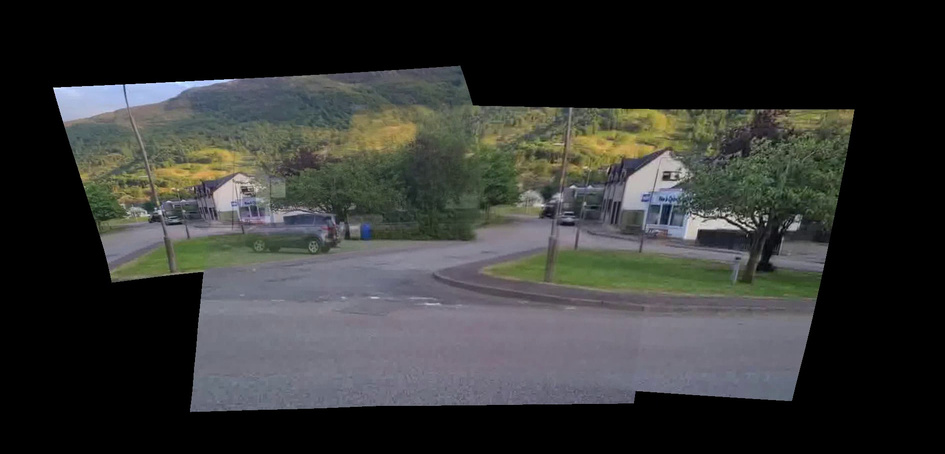} &
\includegraphics[width=3.8cm, height=2.1cm]{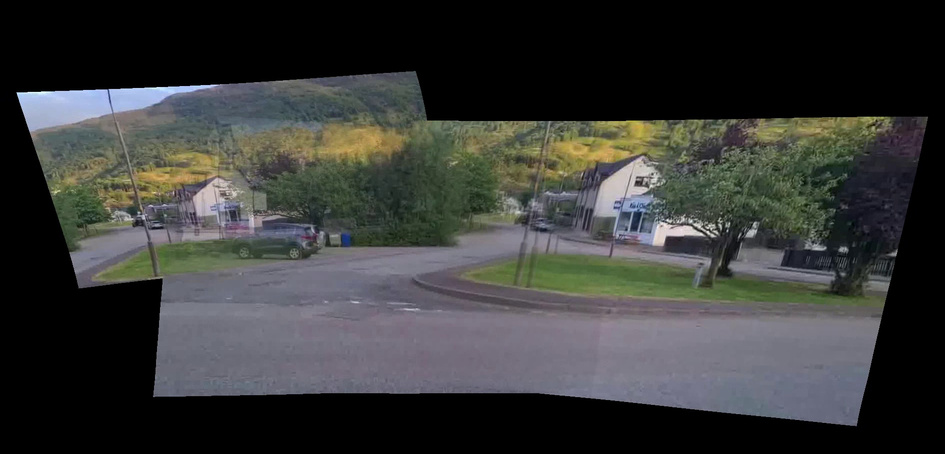} \\[0.2mm]

\rotatebox{90}{\small MViS (Ours)} & 
\includegraphics[width=3.8cm, height=2.1cm]{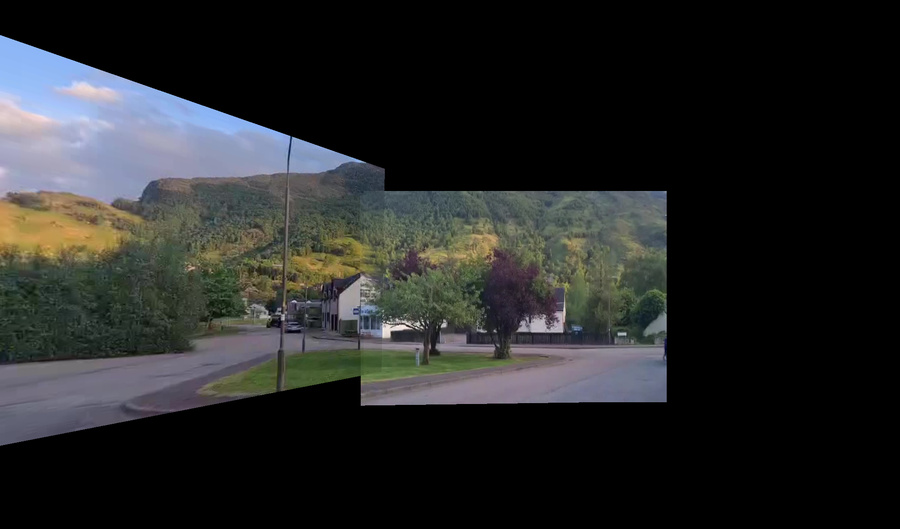} &
\includegraphics[width=3.8cm, height=2.1cm]{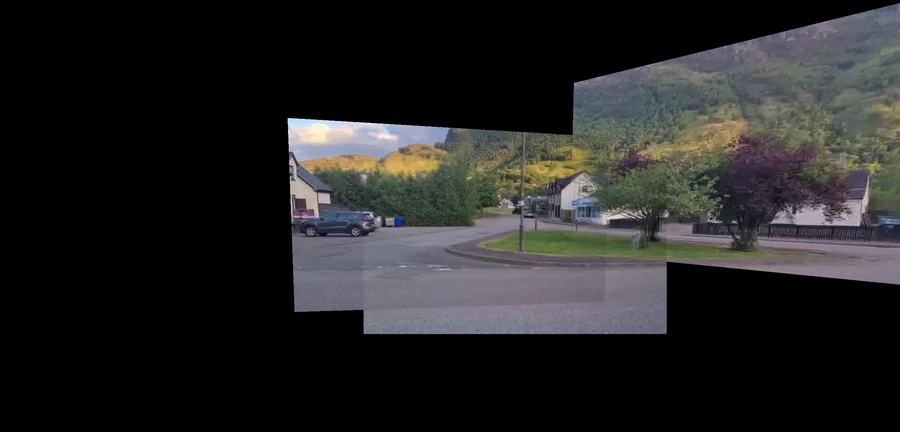} &
\includegraphics[width=3.8cm, height=2.1cm]{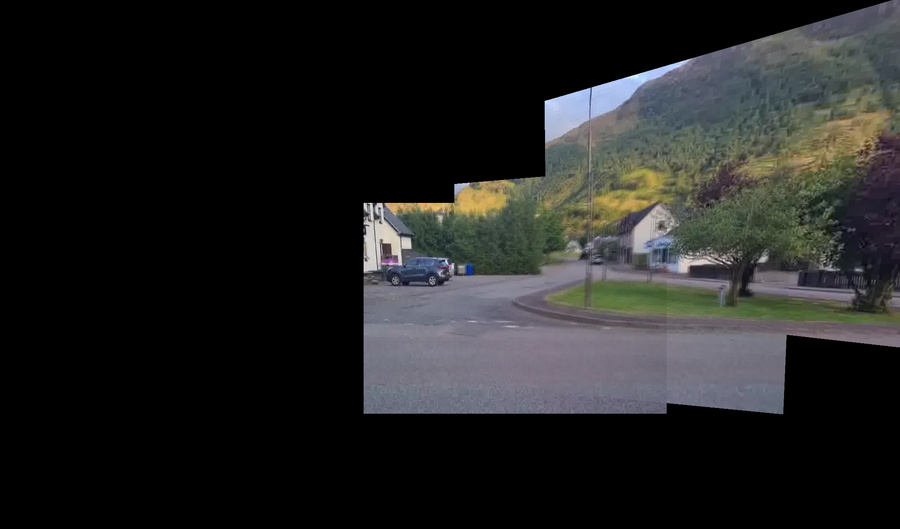} &
\includegraphics[width=3.8cm, height=2.1cm]{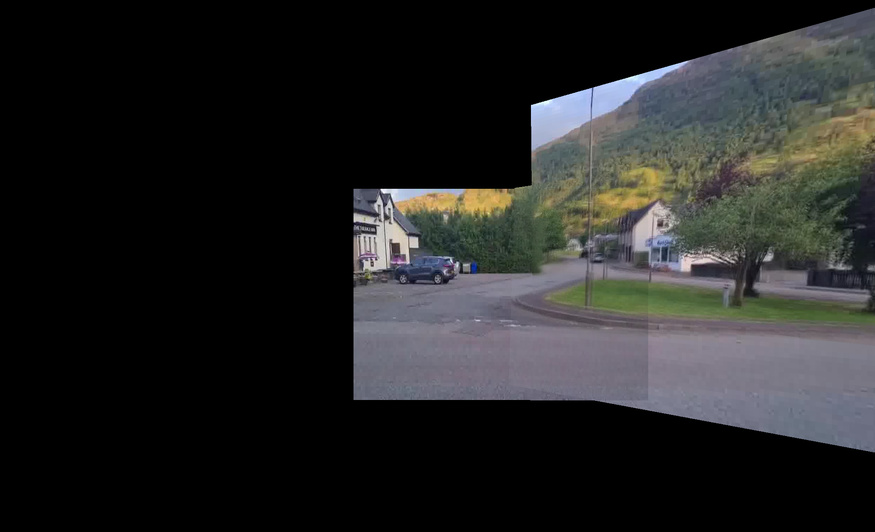} \\[-0.5mm]

& \small{$t=0$} & 
\small{$t=120$} & 
\small{$t=150$} & 
\small{$t=170$}

\end{tabular}
}
\caption{Visual comparison between StabStitch++~\cite{Nie:TPAMI:2025:stabstitch++} 
  and MViS scenes 
  with $N=3$ cameras (from the MMCD dataset).}
  \label{fig:triplets}
\end{figure}

We include a set of videos in the Supplementary Material:

\begin{itemize}[itemsep=0.3em,label=$\triangleright$]

    \item \texttt{2\_stabstitch++\_vs\_mvis.mp4},

    \texttt{3\_stabstitch++\_vs\_mvis.mp4},

    \texttt{4\_stabstitch++\_vs\_mvis.mp4}

    Visual comparisons between StabStitch++ and MViS (Ours) on \(N=3\), MMCD.  
    The specific challenges of these scenes are illustrated in \autoref{fig:triplets} of this appendix.

    \item \texttt{5\_n5\_mis\_methods\_vs\_mvis.mp4}

    \texttt{6\_n4\_mis\_methods\_vs\_mvis.mp4}

    Visual comparison between Multiple Image Stitching (MIS) methods and MViS on scenes from the MMCD dataset with $N=4,5$ cameras. While each MIS method produces visually plausible panoramas per frame, their resulting panoramic video exhibits strong temporal instability.

    \item \texttt{7\_mis\_variants\_vs\_mvis.mp4}
    
    Visual comparison between OBJ-GSP, OBJ-GSP+TR, OBJ-GSP+PS and MViS. Although OBJ-GSP produces visually plausible panoramas per frame, the resulting video is temporally unstable. 
    Both extensions improve temporal stability but still underperform MViS.

    \item \texttt{8\_n5\_mvis.mp4}

    MViS video panorama of an $N=5$ scene  
    (that appears in ~\autoref{Fig:VisualCompare} in the paper and \autoref{fig:view_graph}) from the MMCD dataset, demonstrating robustness to dynamically changing overlap topology.

    \item \texttt{9\_n2\_stabstitch++\_vs\_mvis.mp4},

    \texttt{10\_n2\_stabstitch++\_vs\_mvis.mp4}

    Visual comparisons for $N=2$ on StabStitch-D (corresponding to scenes from ~\autoref{Fig:VisualCompare} in the paper). 
    
    \item \texttt{11\_with\_vs\_without\_net.mp4}

    Visual comparison between MViS and a direct optimization of the homographies without a neural network.
    While it can achieve good results in terms of per-frame panoramas, it yields, usually, much less temporally stable results, as is evident in the video.
    \item \texttt{12\_mmcd\_sample.mp4},

    Samples of annotated scenes from the MMCD dataset.
    \item \texttt{13\_mvis\_failure\_case.mp4}
    
    MViS failure case discussed in the paper, where the matcher fails in a texture-less scene (white wall).
\end{itemize}

\clearpage

\clearpage

\begin{figure}[t]
\centering
\resizebox{\linewidth}{!}{
\setlength{\tabcolsep}{2pt} 
\begin{tabular}{c*{4}{c}}

% Scene_81
\rotatebox{90}{\parbox{2.0cm}{\centering \small GSP ~\cite{Chen:ECCV:2016:GSP}}} & 
\includegraphics[width=3.8cm, height=2.0cm,    viewport=700 800 2300 1600,   clip]{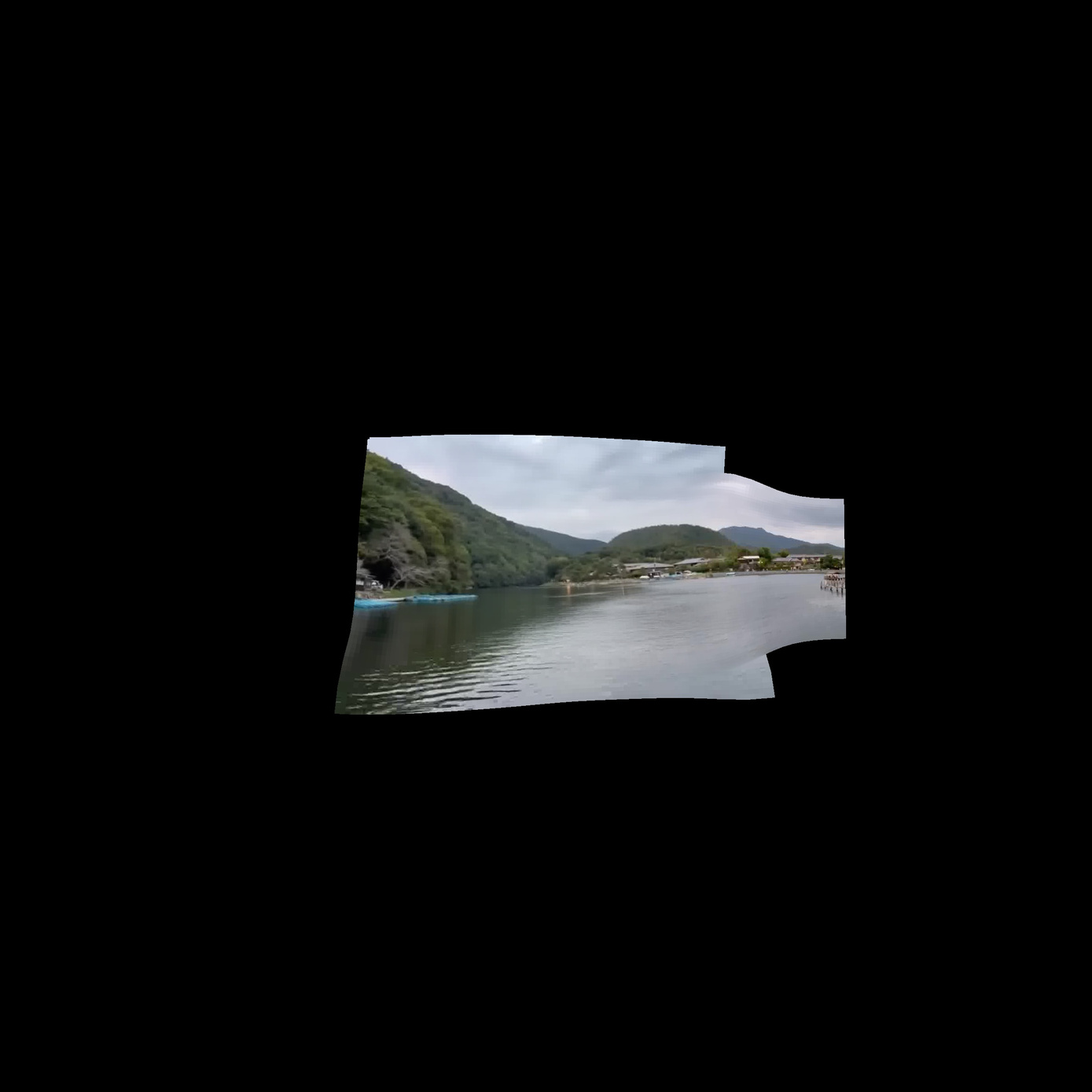} &
\includegraphics[width=3.8cm, height=2.0cm,    viewport=700 800 2300 1600,   clip]{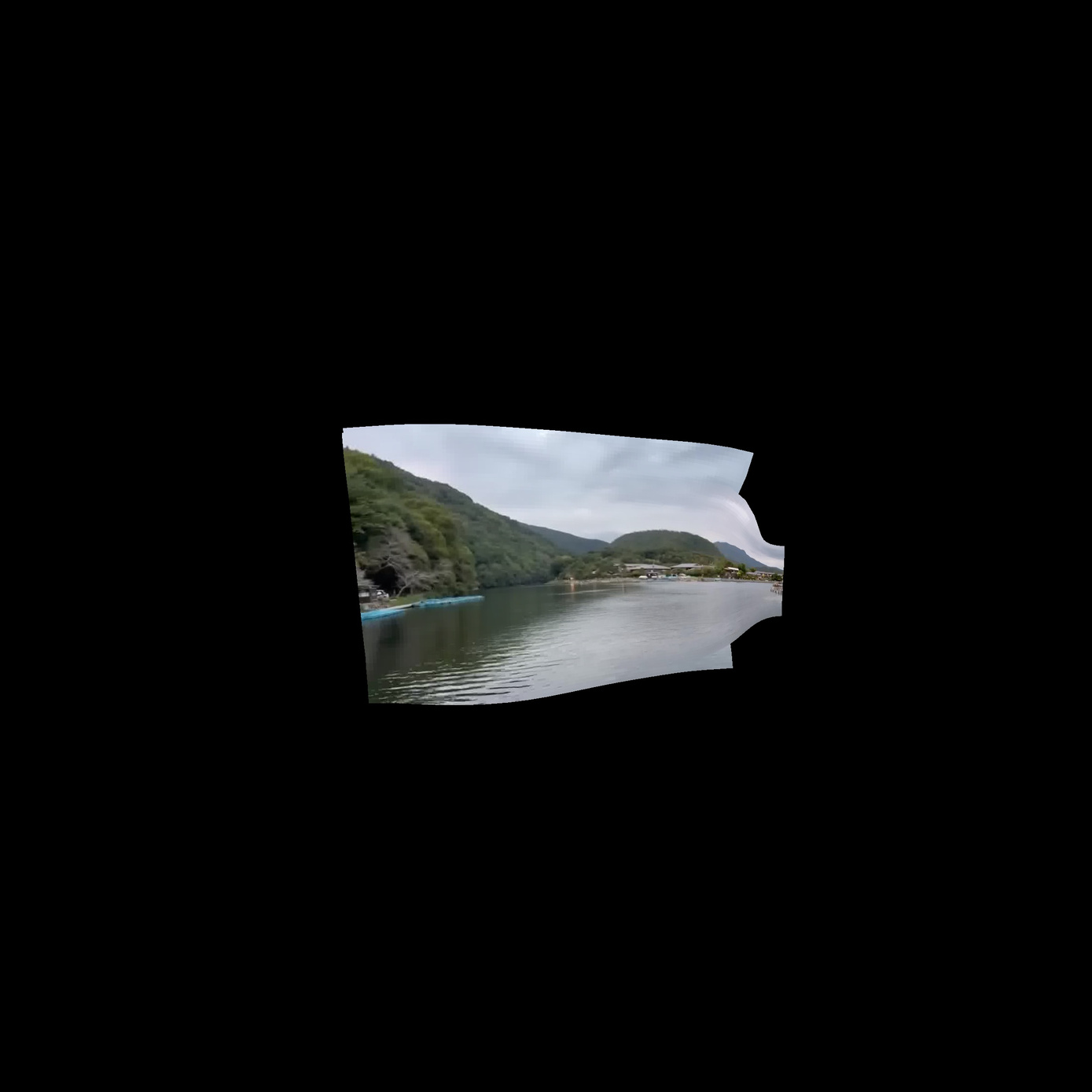} &
\includegraphics[width=3.8cm, height=2.0cm,    viewport=700 800 2300 1600,   clip]{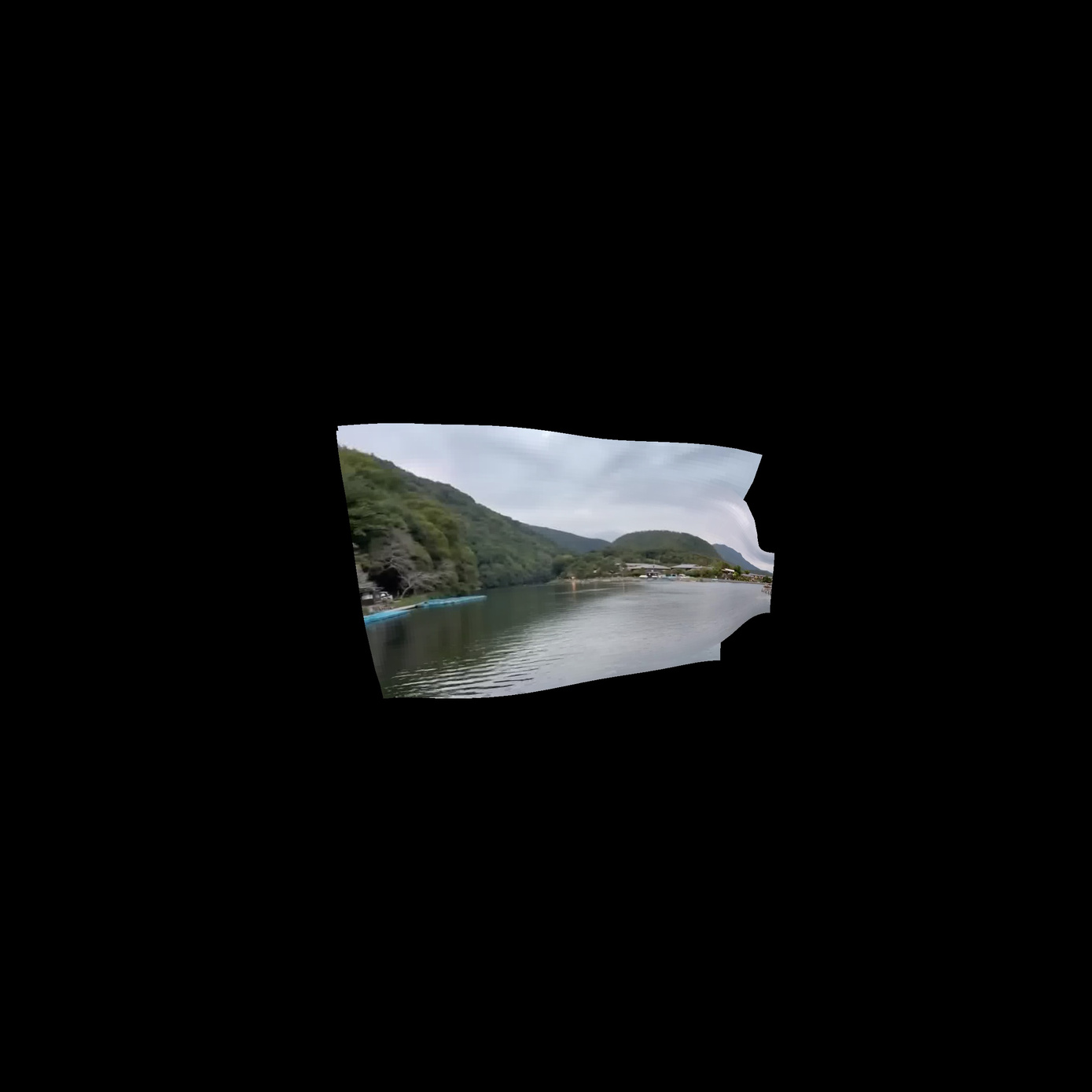}&
\includegraphics[width=3.8cm, height=2.0cm,    viewport=700 800 2300 1600,   clip]{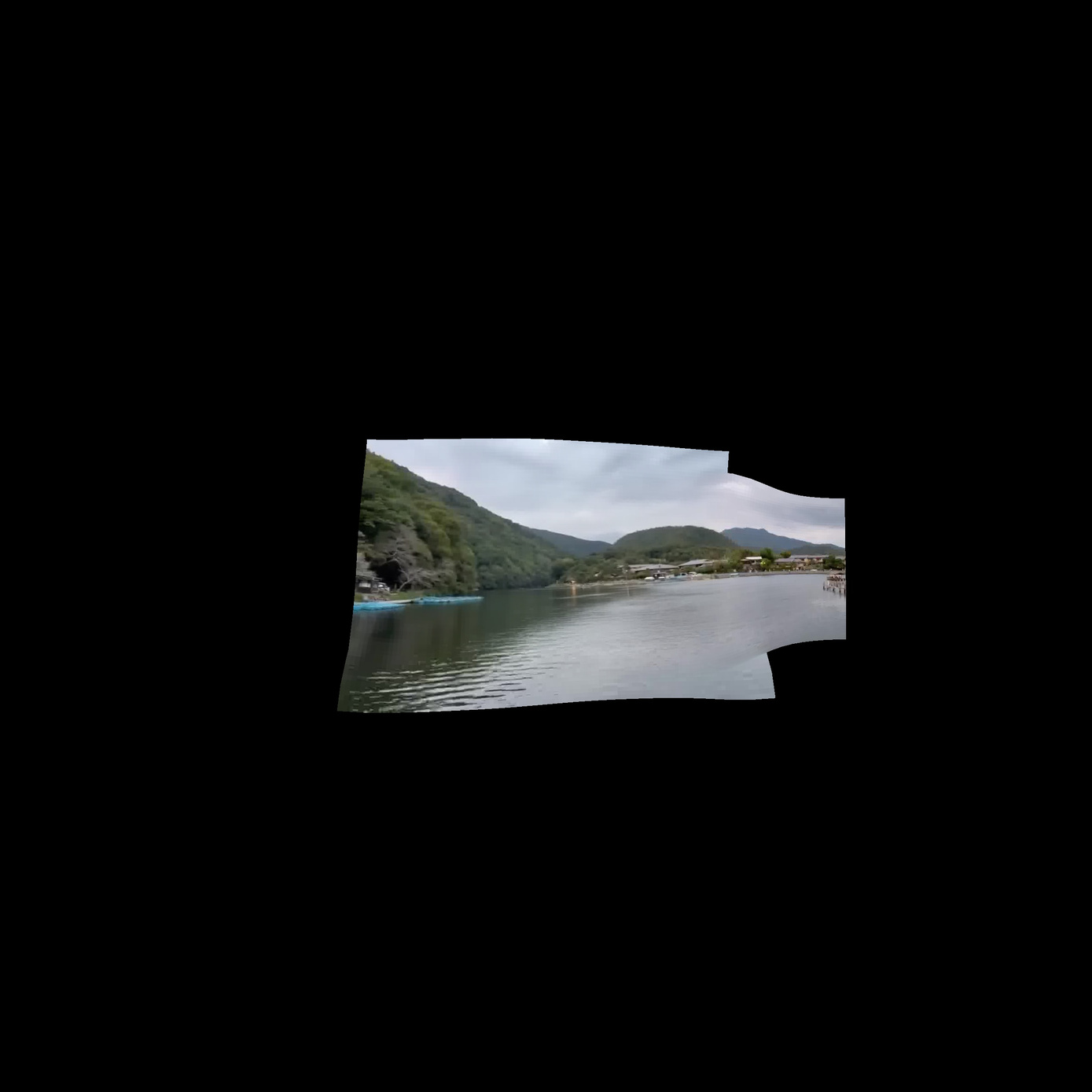} \\[0.3mm]

\rotatebox{90}{\parbox{2.0cm}{\centering \small SPW ~\cite{Liao:TIP:2019:SPW}}} & 
\includegraphics[width=3.8cm, height=2.0cm,    viewport=700 800 2300 1600,   clip]{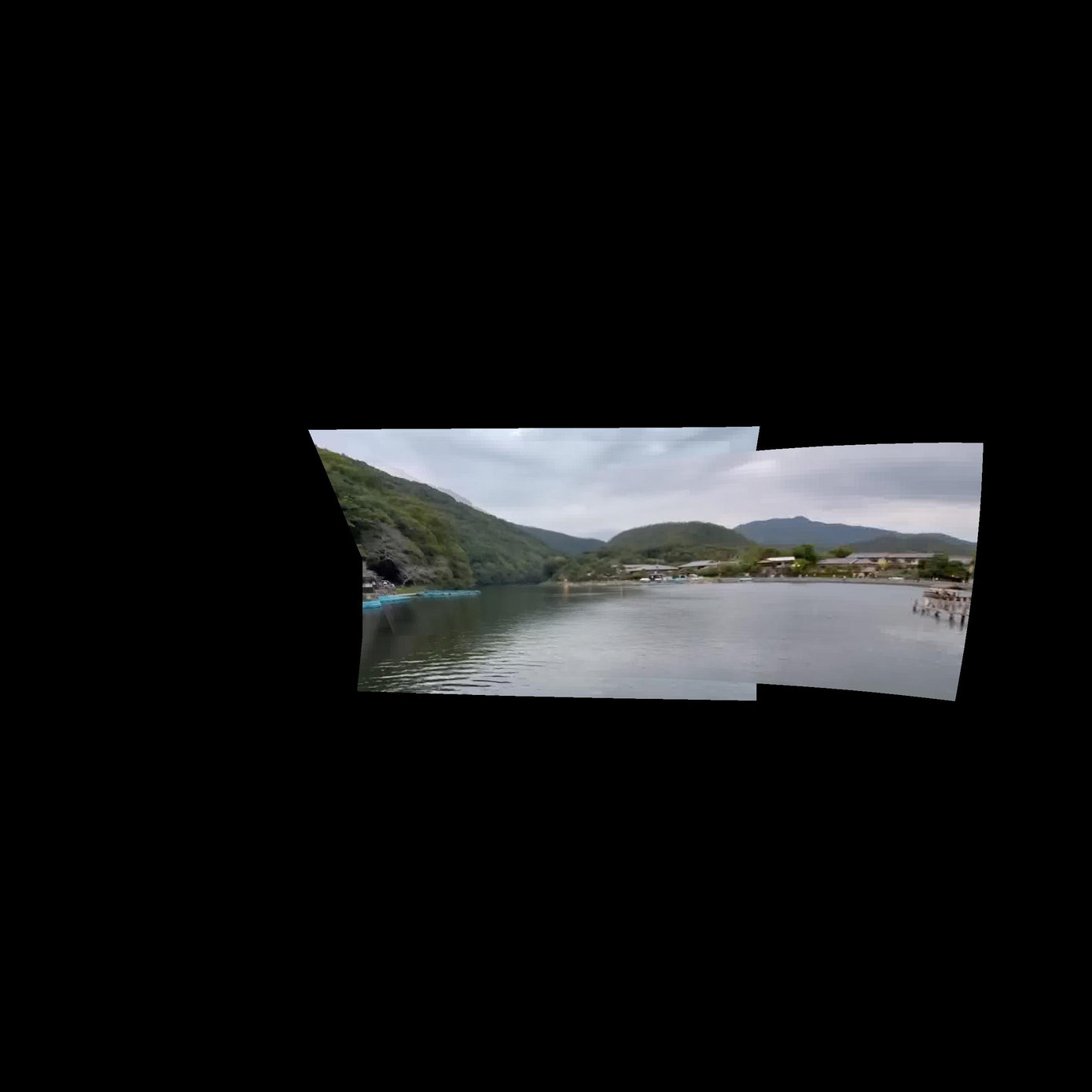} &
\includegraphics[width=3.8cm, height=2.0cm,    viewport=700 800 2300 1600,   clip]{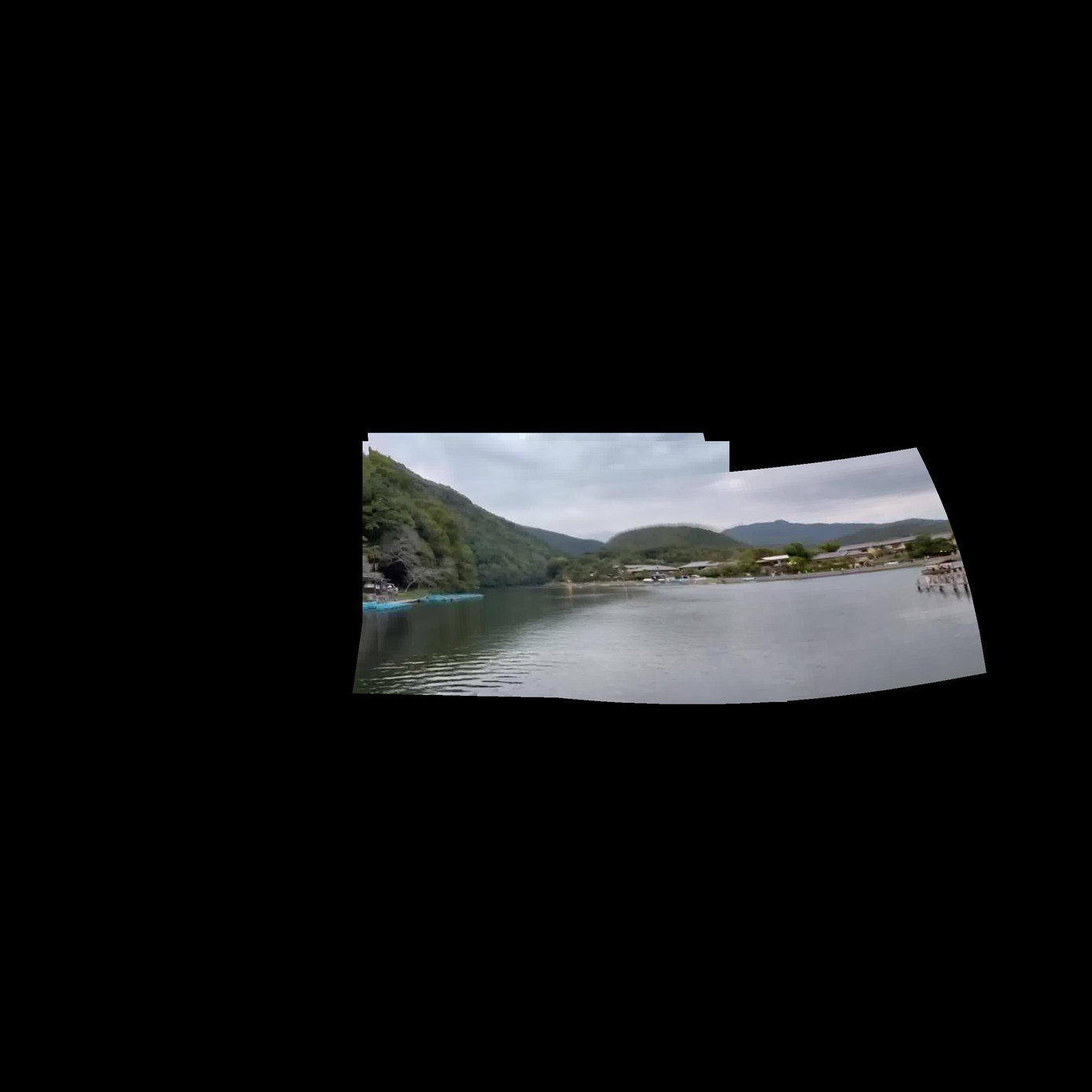} &
\includegraphics[width=3.8cm, height=2.0cm,    viewport=700 800 2300 1600,   clip]{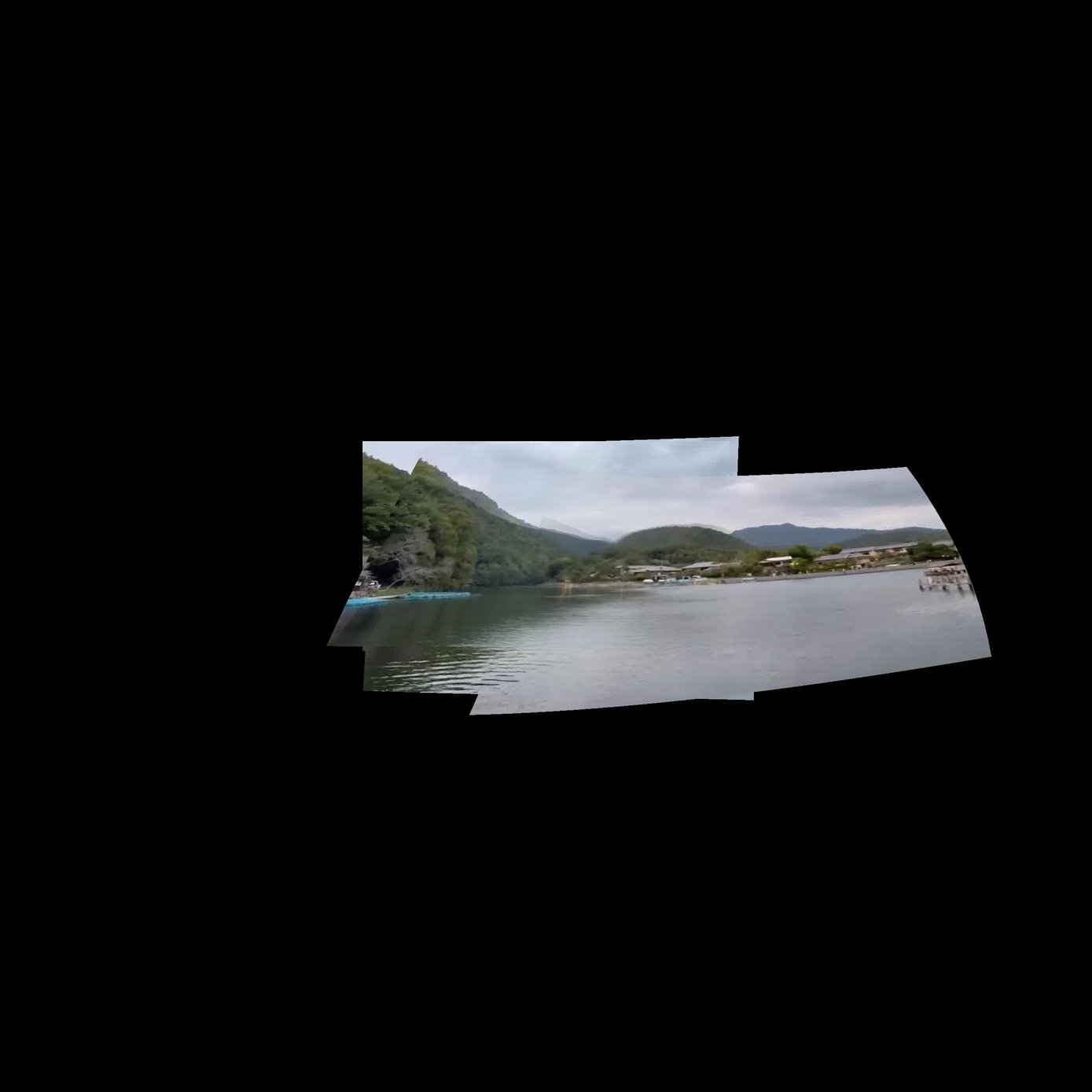}&
\includegraphics[width=3.8cm, height=2.0cm,    viewport=700 800 2300 1600,   clip]{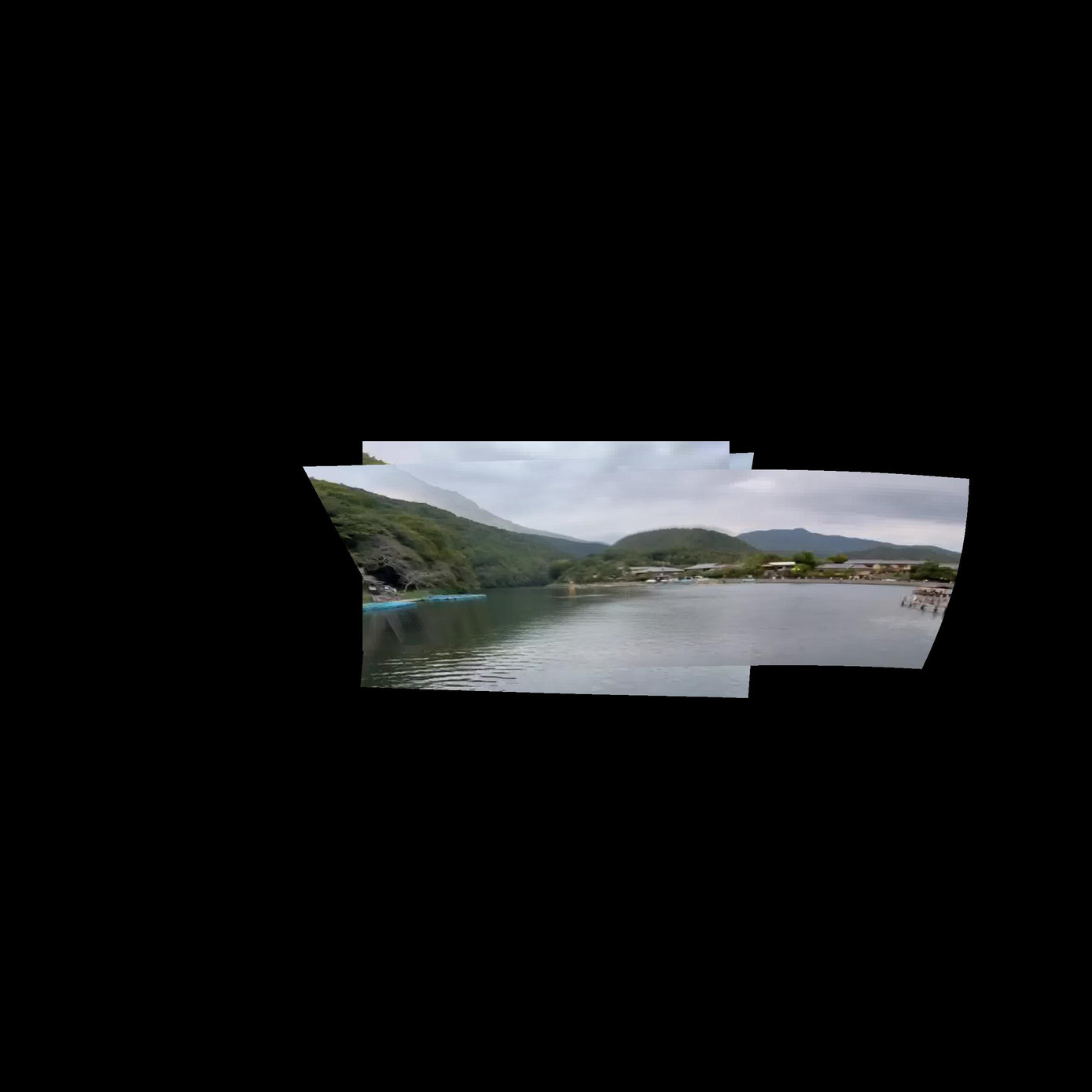} \\[0.5mm]

\rotatebox{90}{\parbox{2.0cm}{\centering \small GES-GSP ~\cite{Du:CVPR:2022:GESGSP}}} & 
\includegraphics[width=3.8cm, height=2.0cm,    viewport=700 800 2300 1600,   clip]{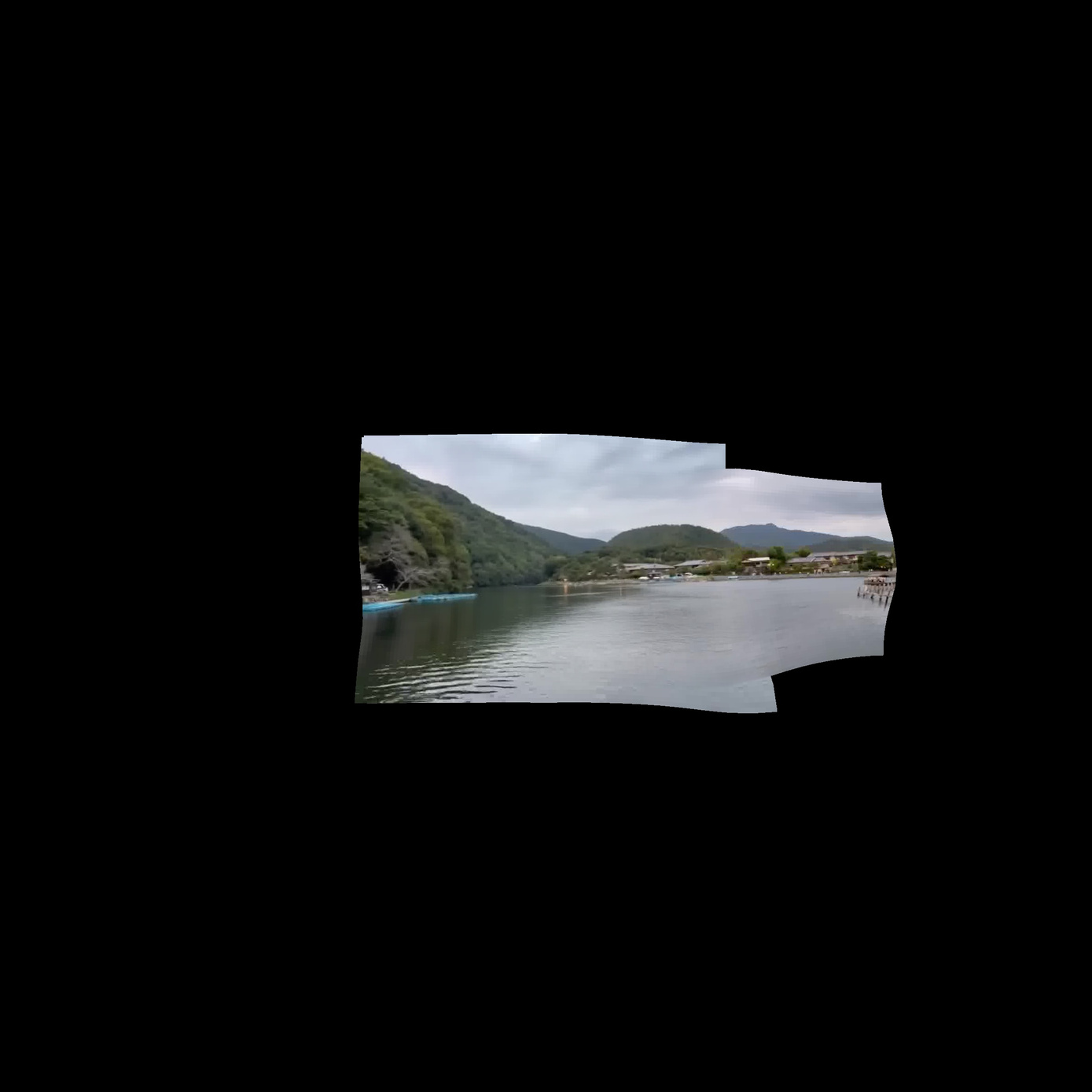} &
\includegraphics[width=3.8cm, height=2.0cm,    viewport=700 800 2300 1600,   clip]{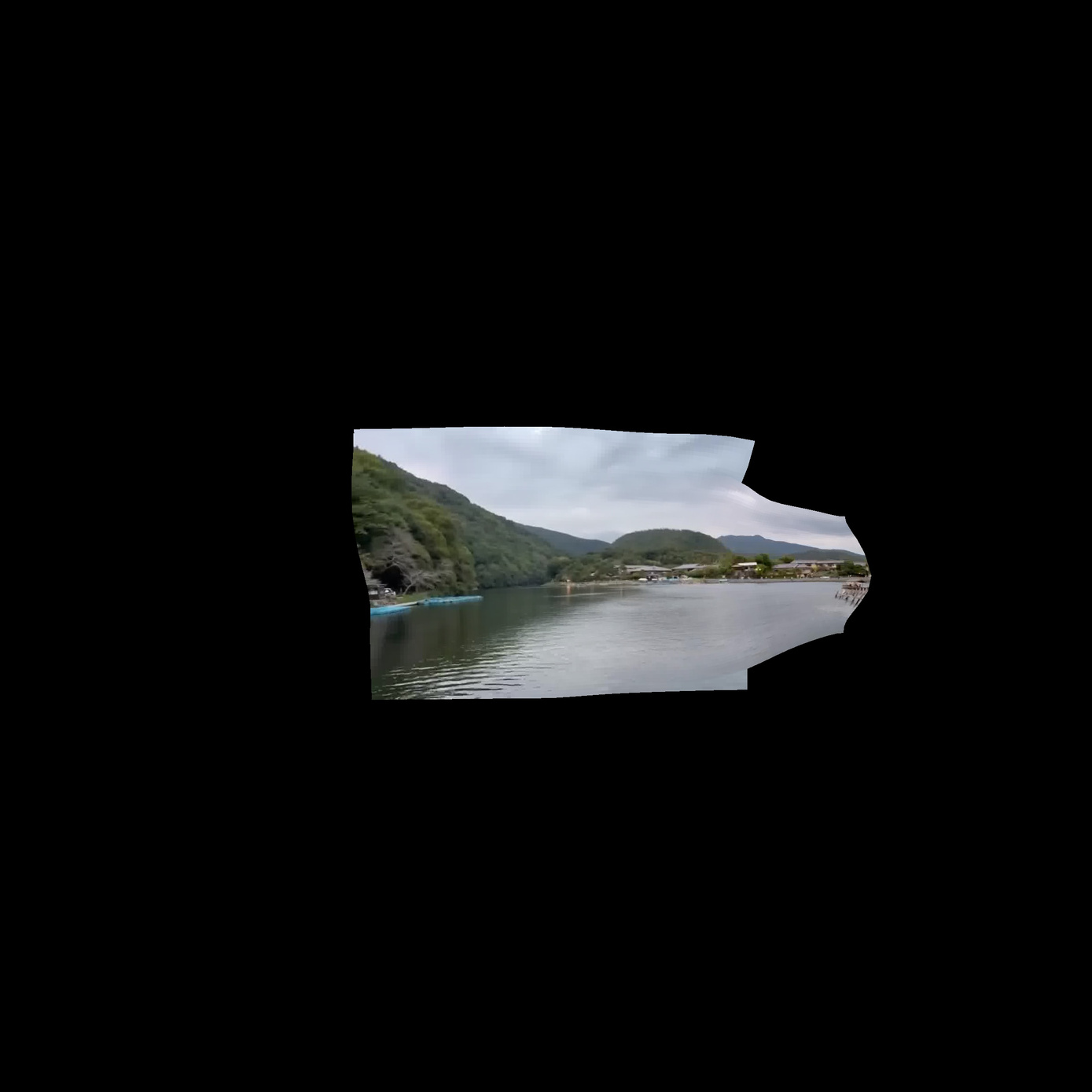} &
\includegraphics[width=3.8cm, height=2.0cm,    viewport=700 800 2300 1600,   clip]{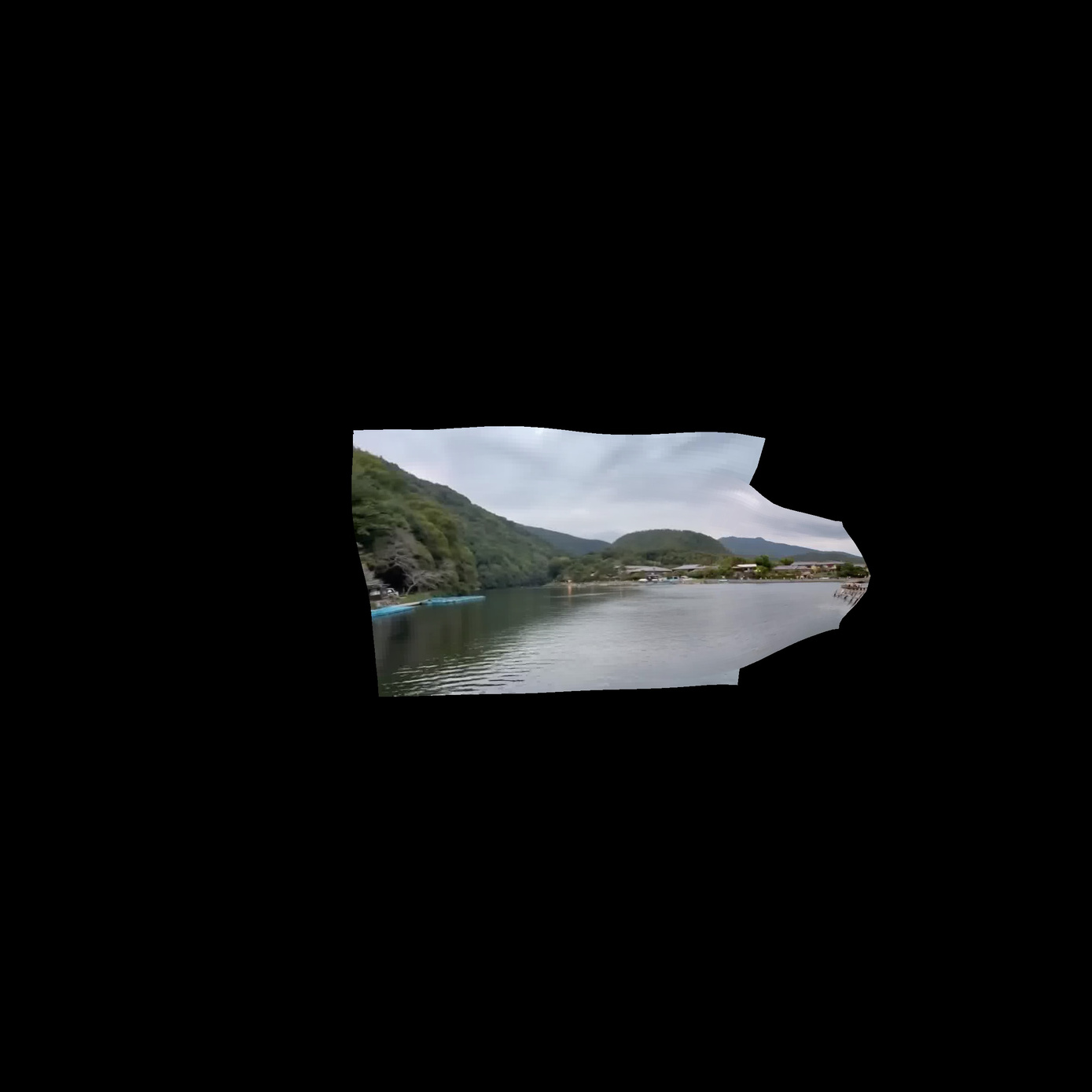}&
\includegraphics[width=3.8cm, height=2.0cm,    viewport=700 800 2300 1600,   clip]{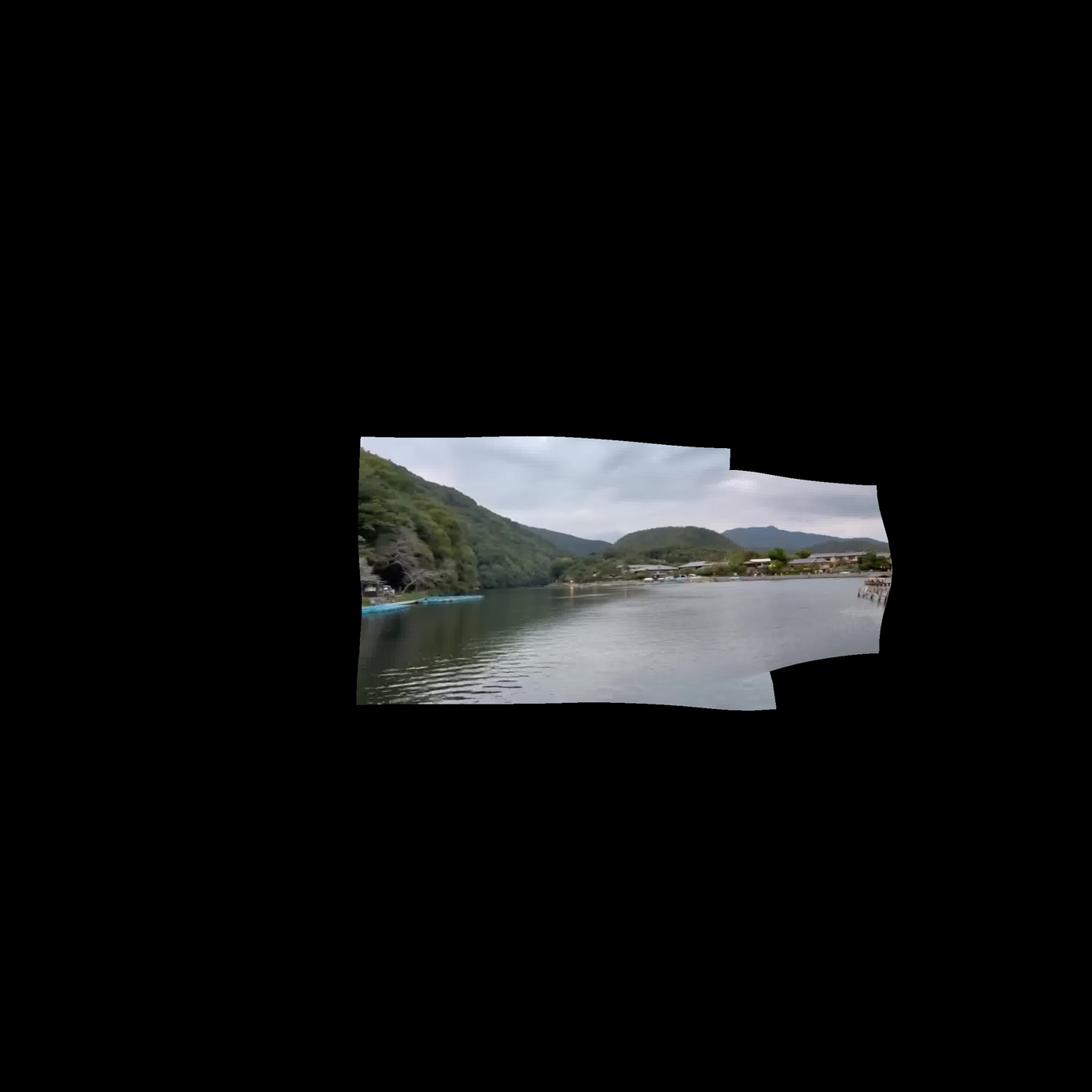} \\[0.3mm]

\rotatebox{90}{\parbox{2.0cm}{\centering \small OBJ-GSP ~\cite{Cai:AAAI:2025:OBJGSP}}} & 
\includegraphics[width=3.8cm, height=2.0cm,    viewport=700 800 2300 1600,   clip]{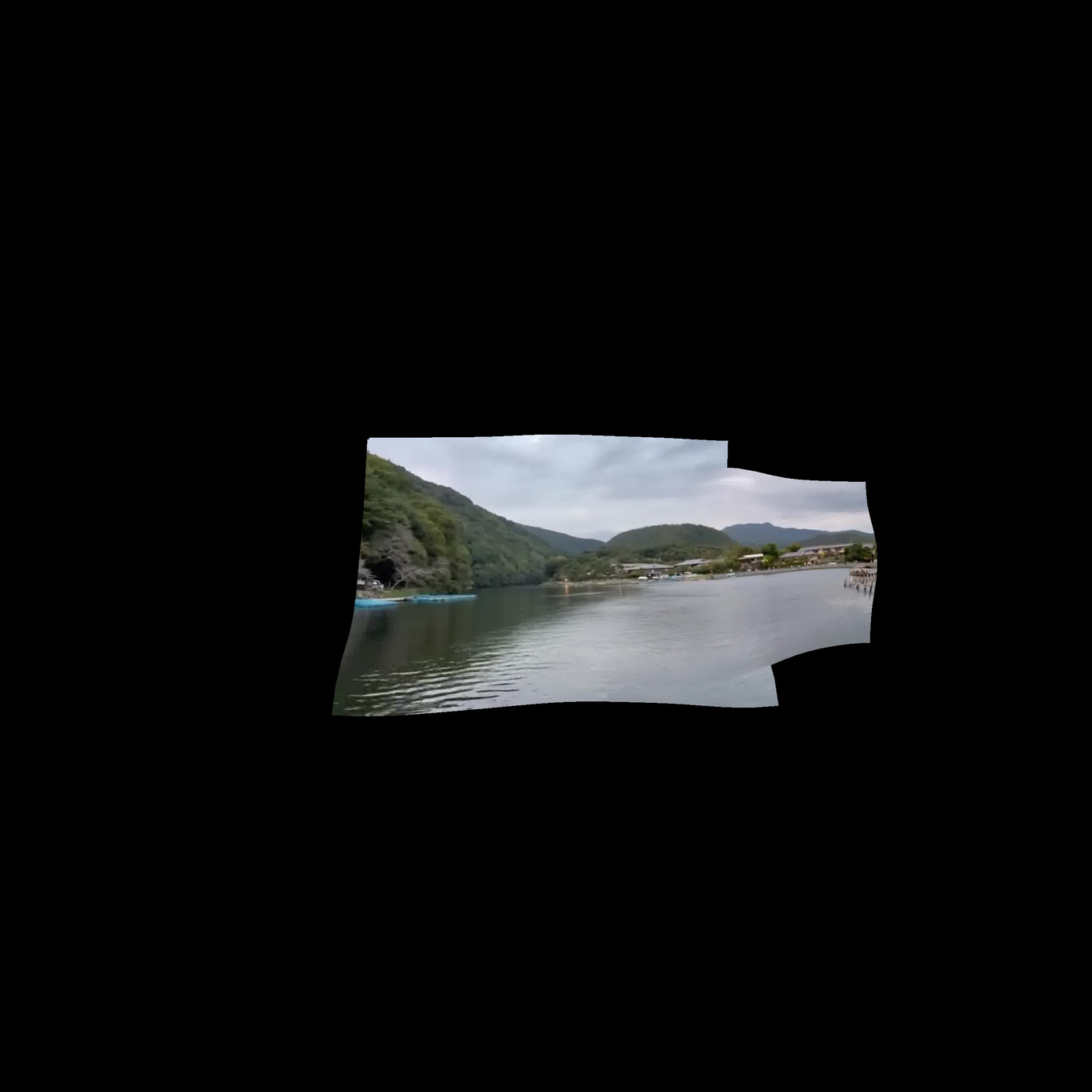} &
\includegraphics[width=3.8cm, height=2.0cm,    viewport=700 800 2300 1600,   clip]{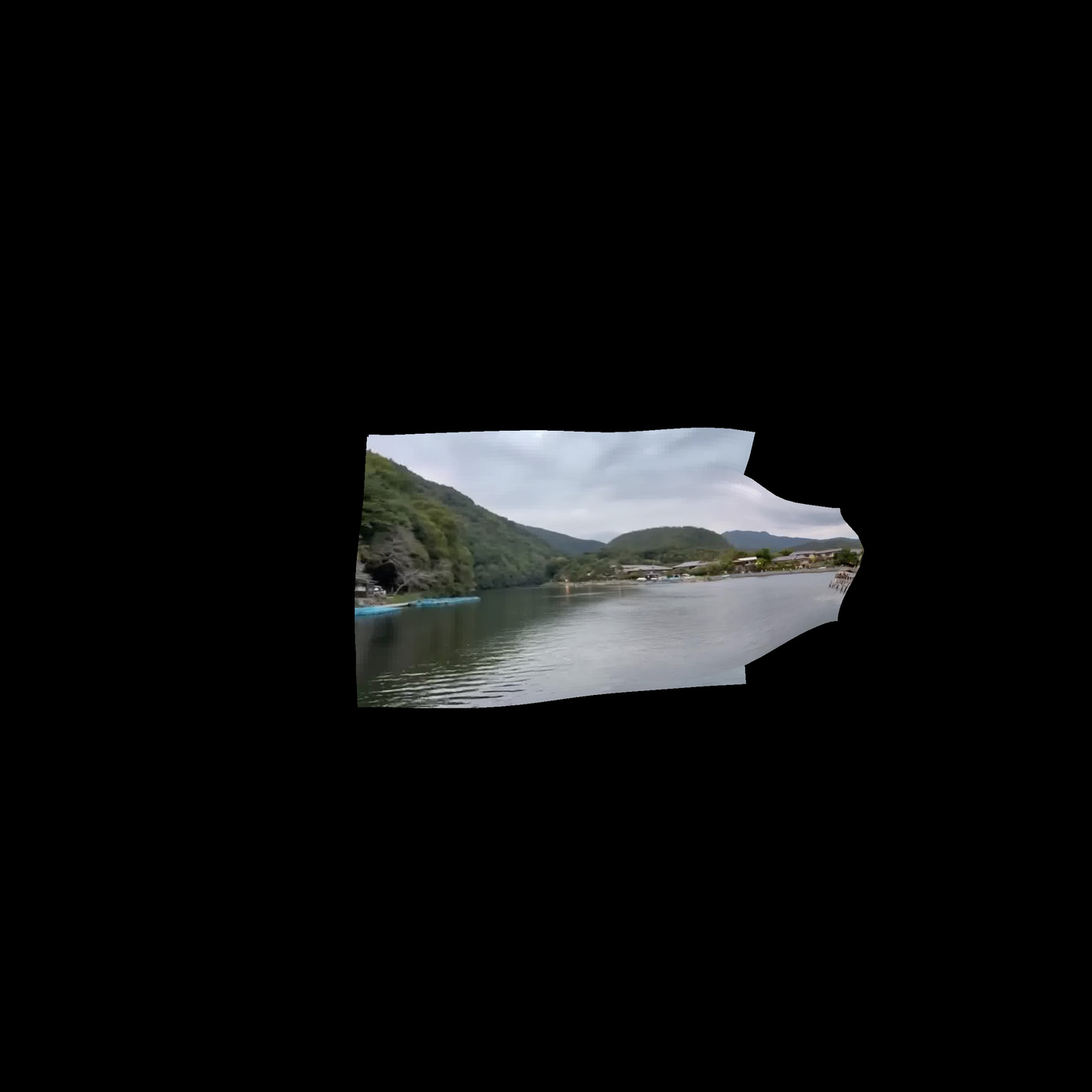} &
\includegraphics[width=3.8cm, height=2.0cm,    viewport=700 800 2300 1600,   clip]{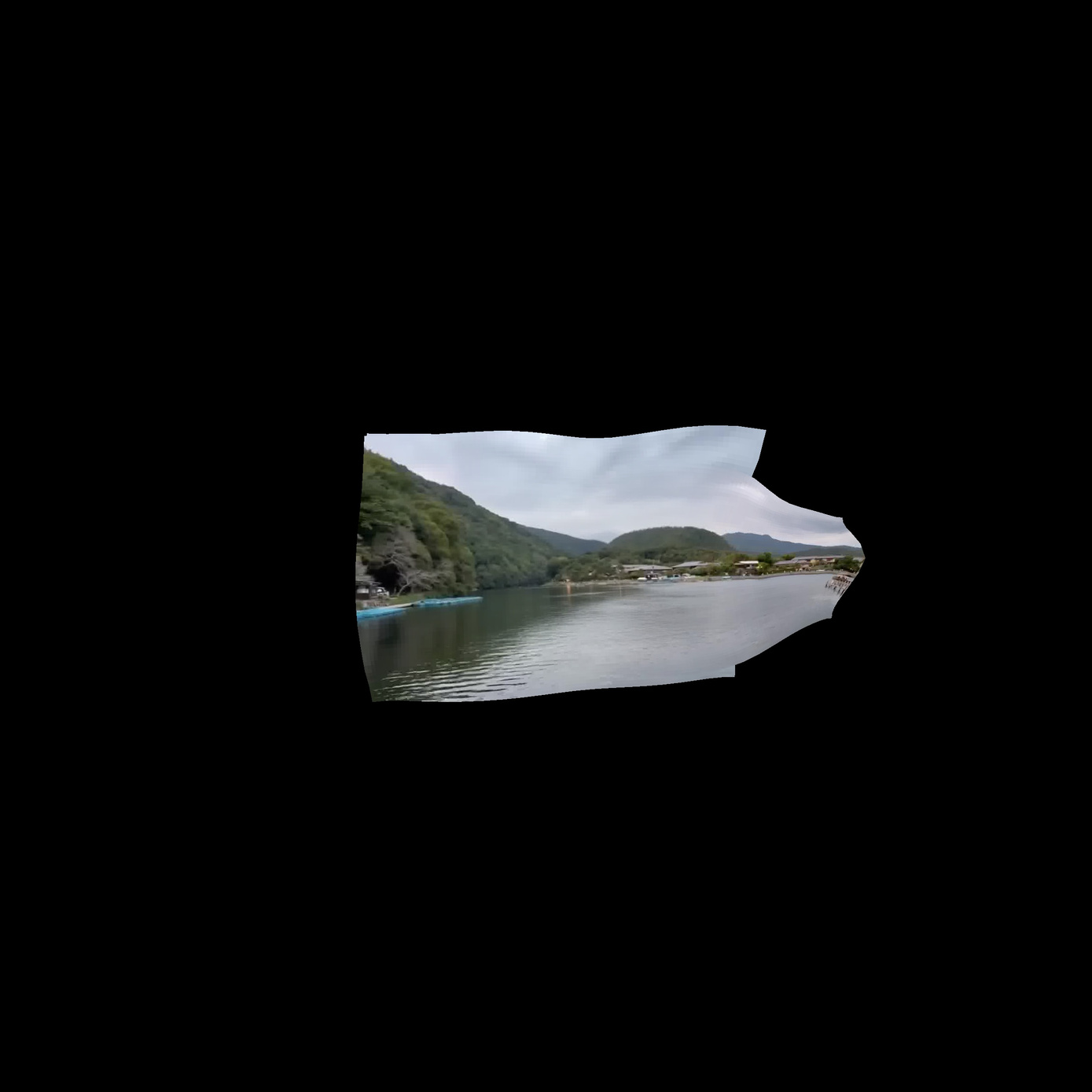}&
\includegraphics[width=3.8cm, height=2.0cm,    viewport=700 800 2300 1600,   clip]{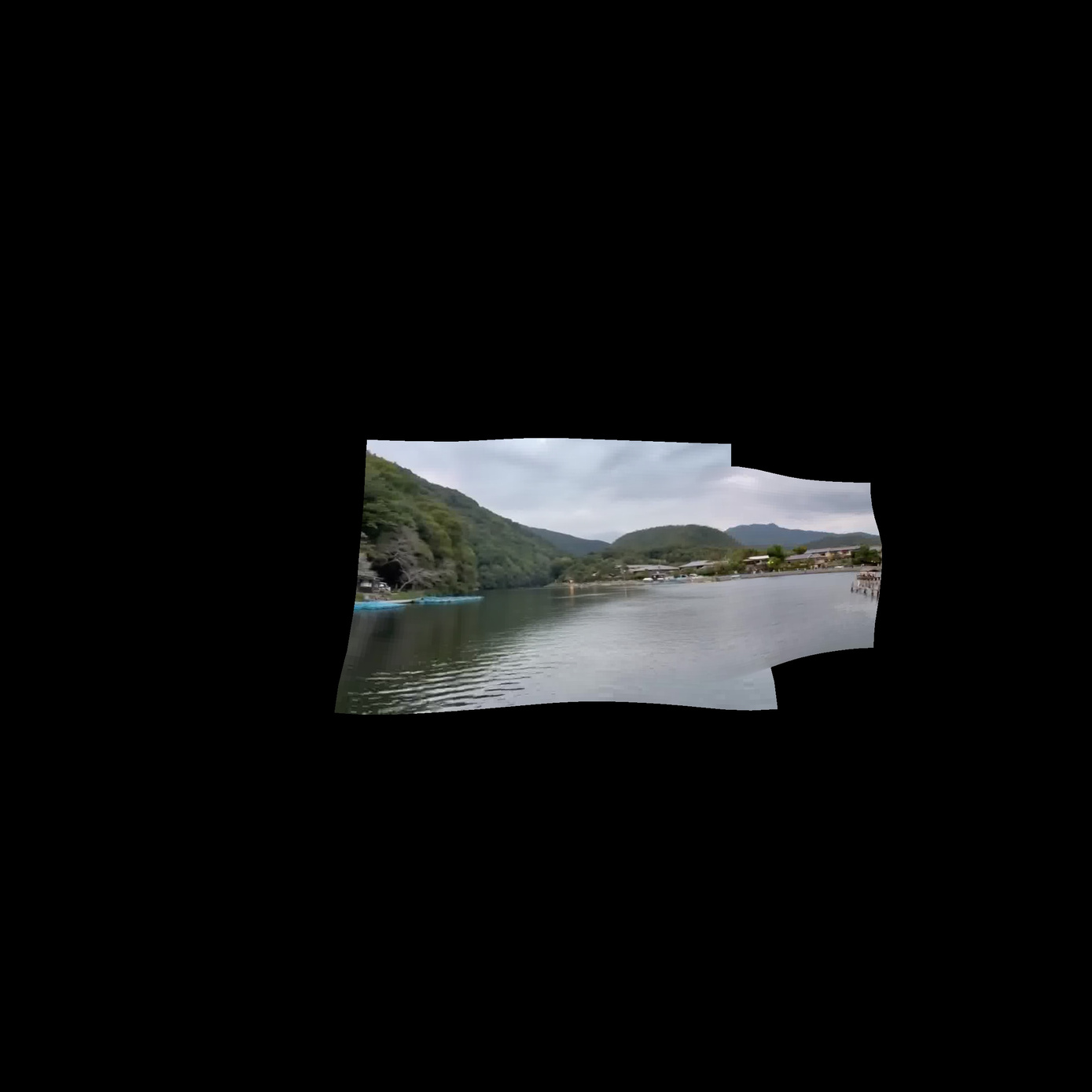} \\[0.3mm]

\rotatebox{90}{\parbox{2.0cm}{\centering \small MViS (Ours)}} & 
\includegraphics[width=3.8cm, height=2.0cm,    viewport=480 510 1792 1192,   clip]{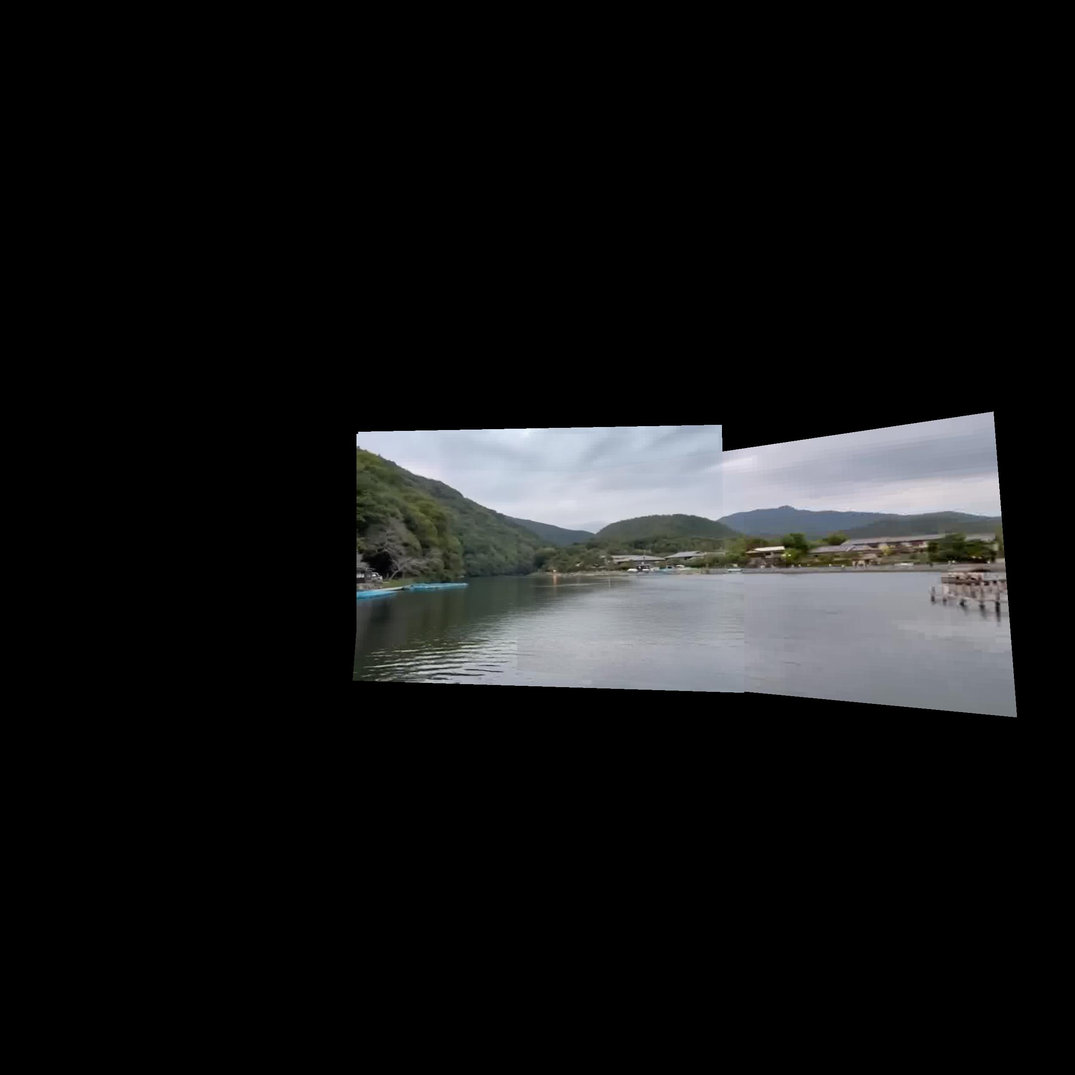} &
\includegraphics[width=3.8cm, height=2.0cm,    viewport=480 510 1792 1192,   clip]{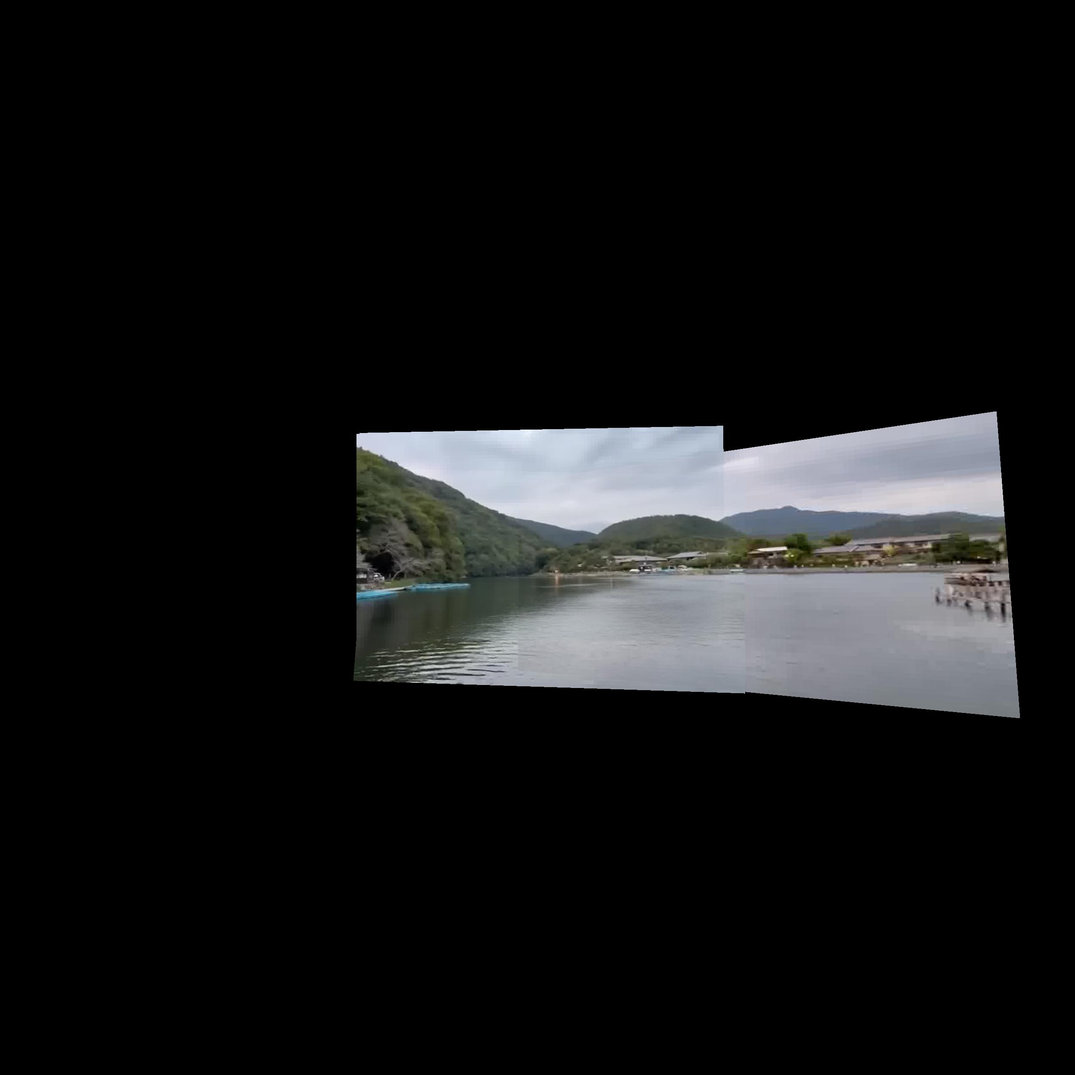} &
\includegraphics[width=3.8cm, height=2.0cm,    viewport=480 510 1792 1192,   clip]{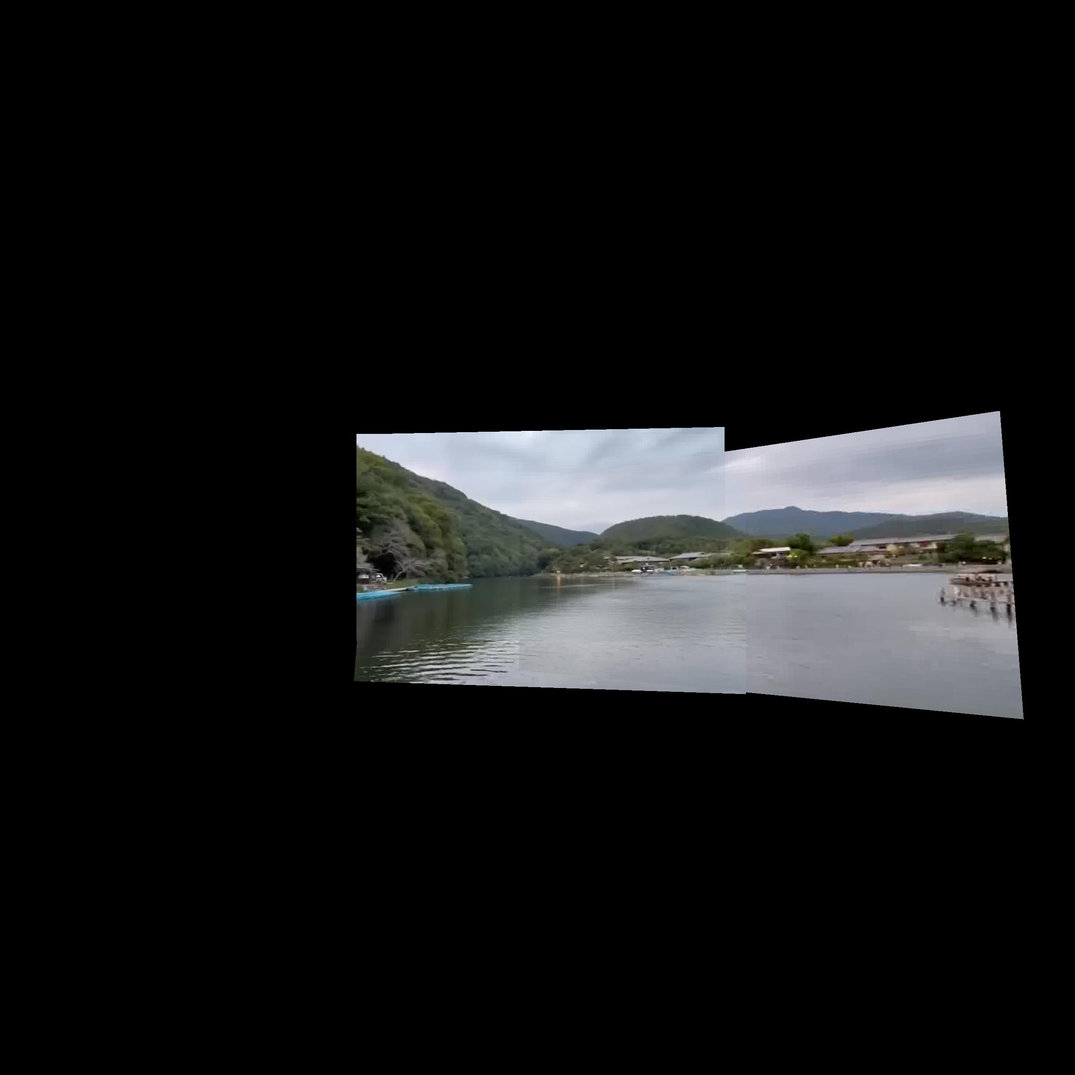}&
\includegraphics[width=3.8cm, height=2.0cm, viewport=480 510 1792 1192,   clip]{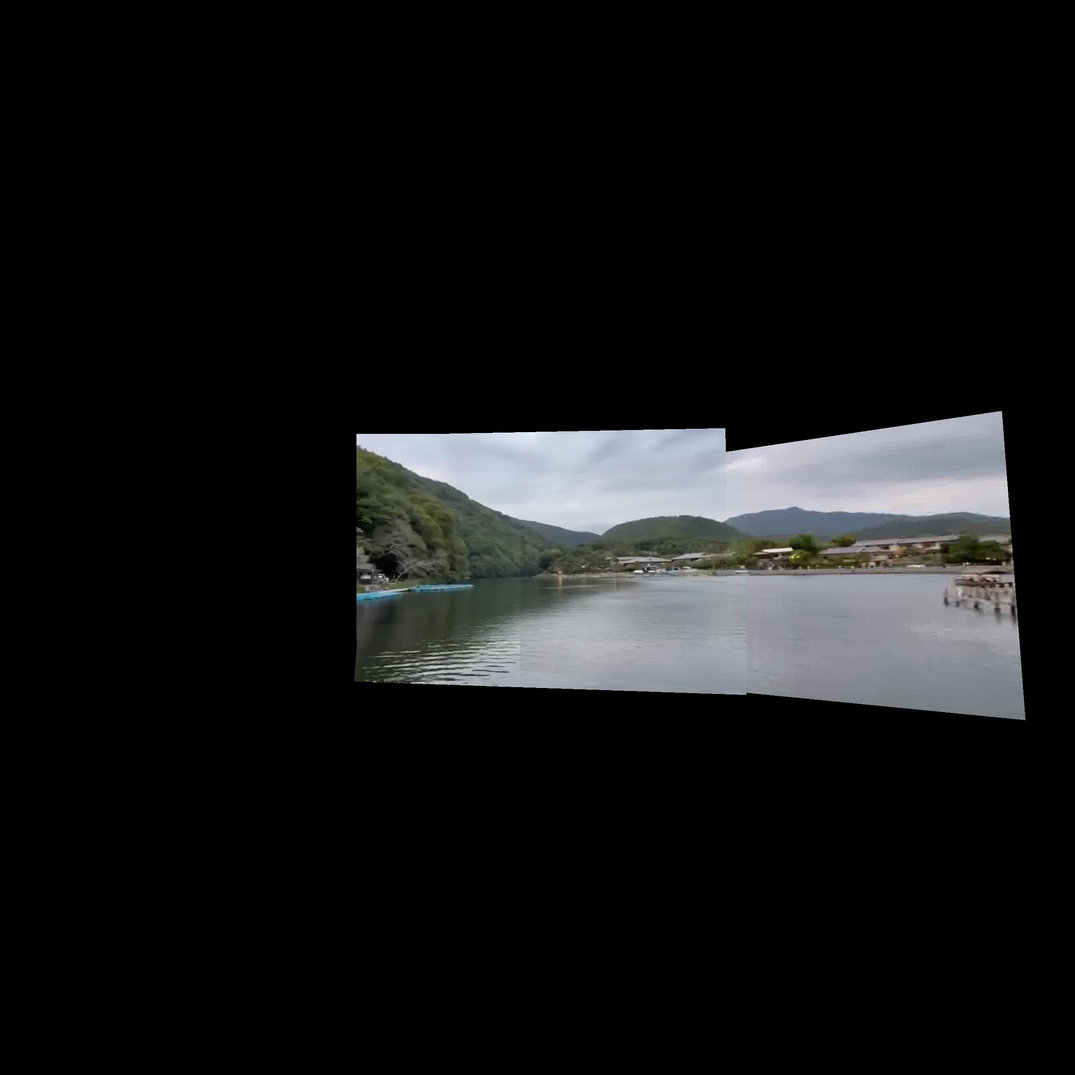} \\[-0.5mm]

& \small{$t=207$} & 
\small{$t=208$} & 
\small{$t=209$} &
\small{$t=210$} \\[2mm]

\rotatebox{90}{\parbox{2.1cm}{\centering \small GSP ~\cite{Chen:ECCV:2016:GSP}}} & 
\includegraphics[width=3.8cm, height=2.1cm,    viewport=500 800 2100 1800,   clip]{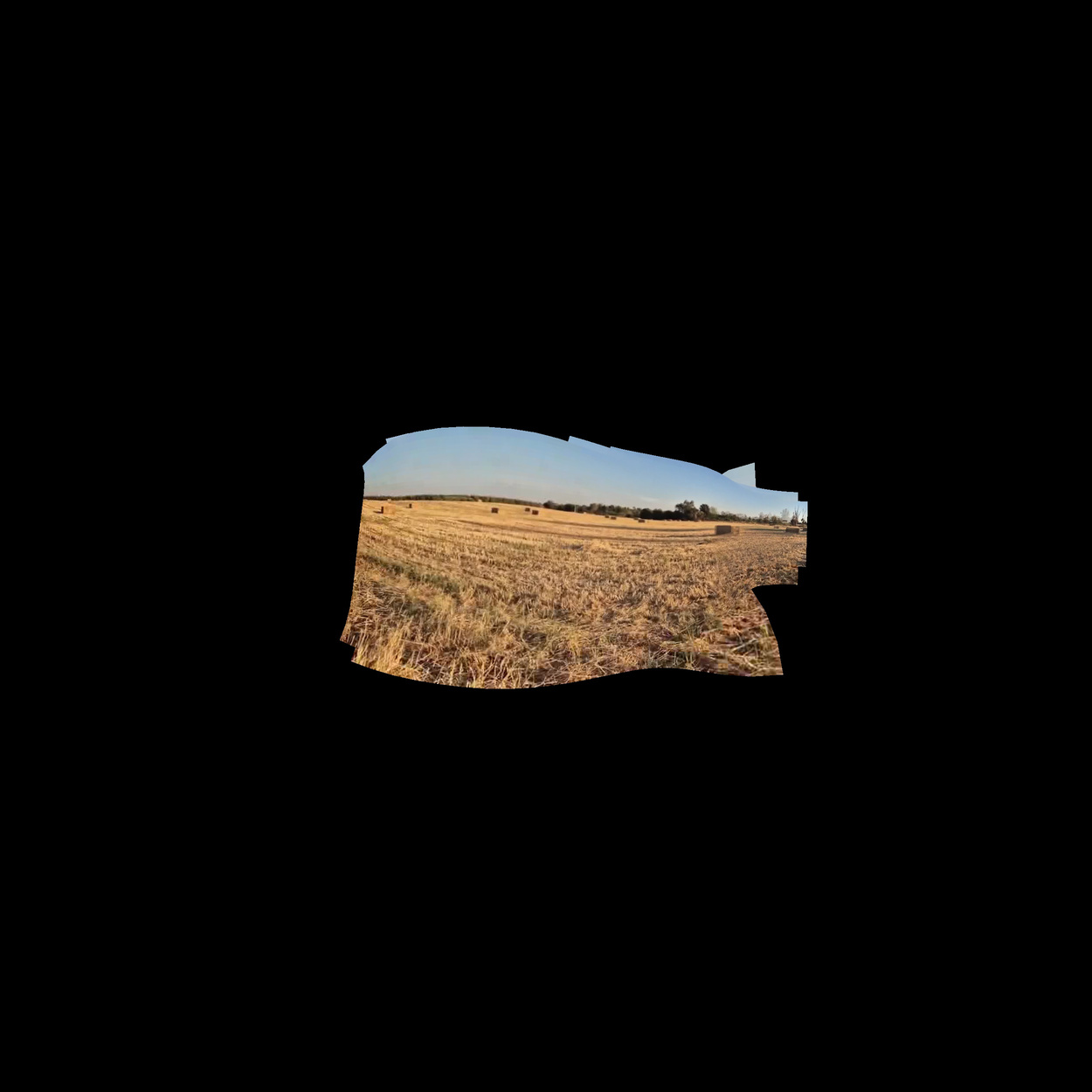} &
\includegraphics[width=3.8cm, height=2.1cm,    viewport=500 800 2100 1800,   clip]{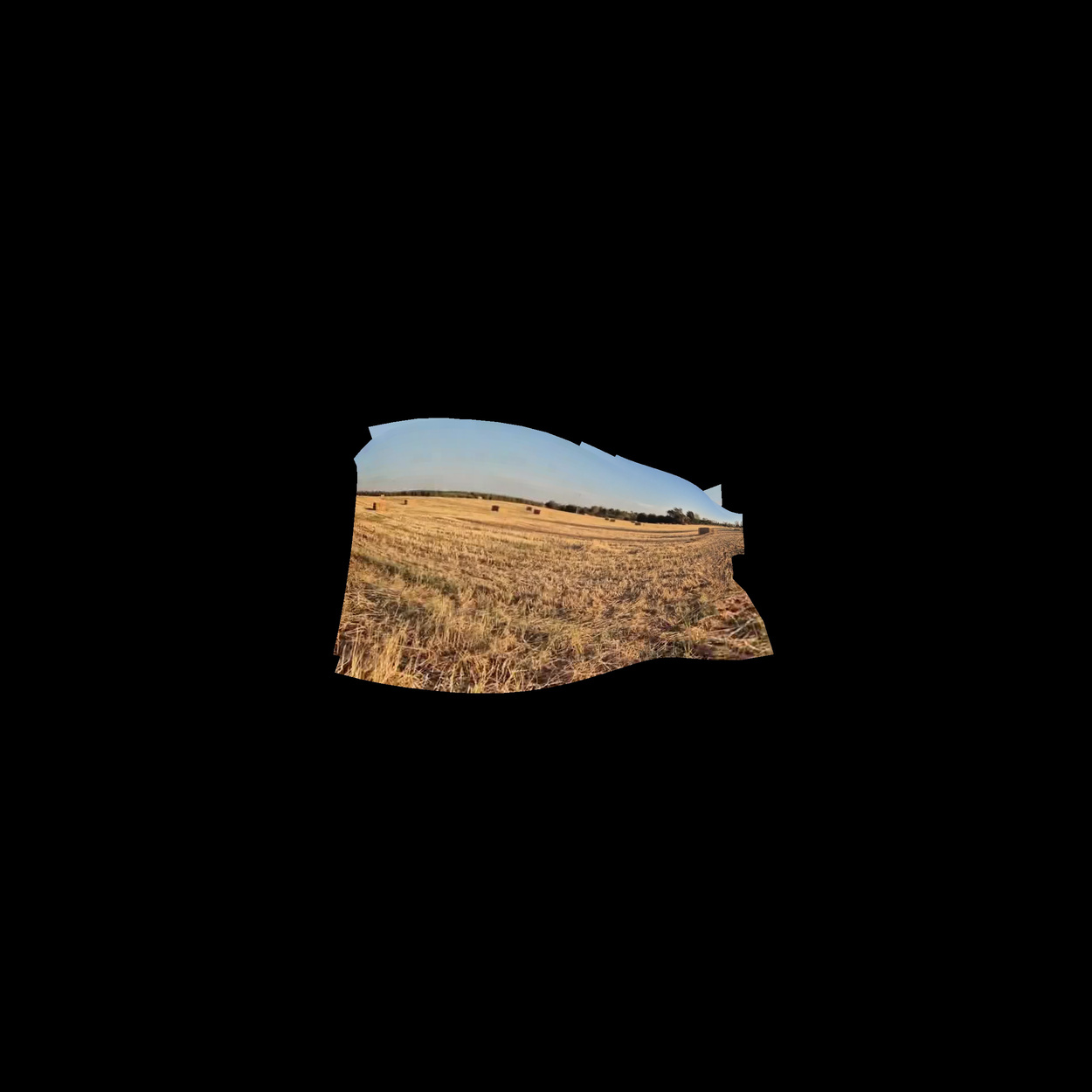} &
\includegraphics[width=3.8cm, height=2.1cm,    viewport=500 800 2100 1800,   clip]{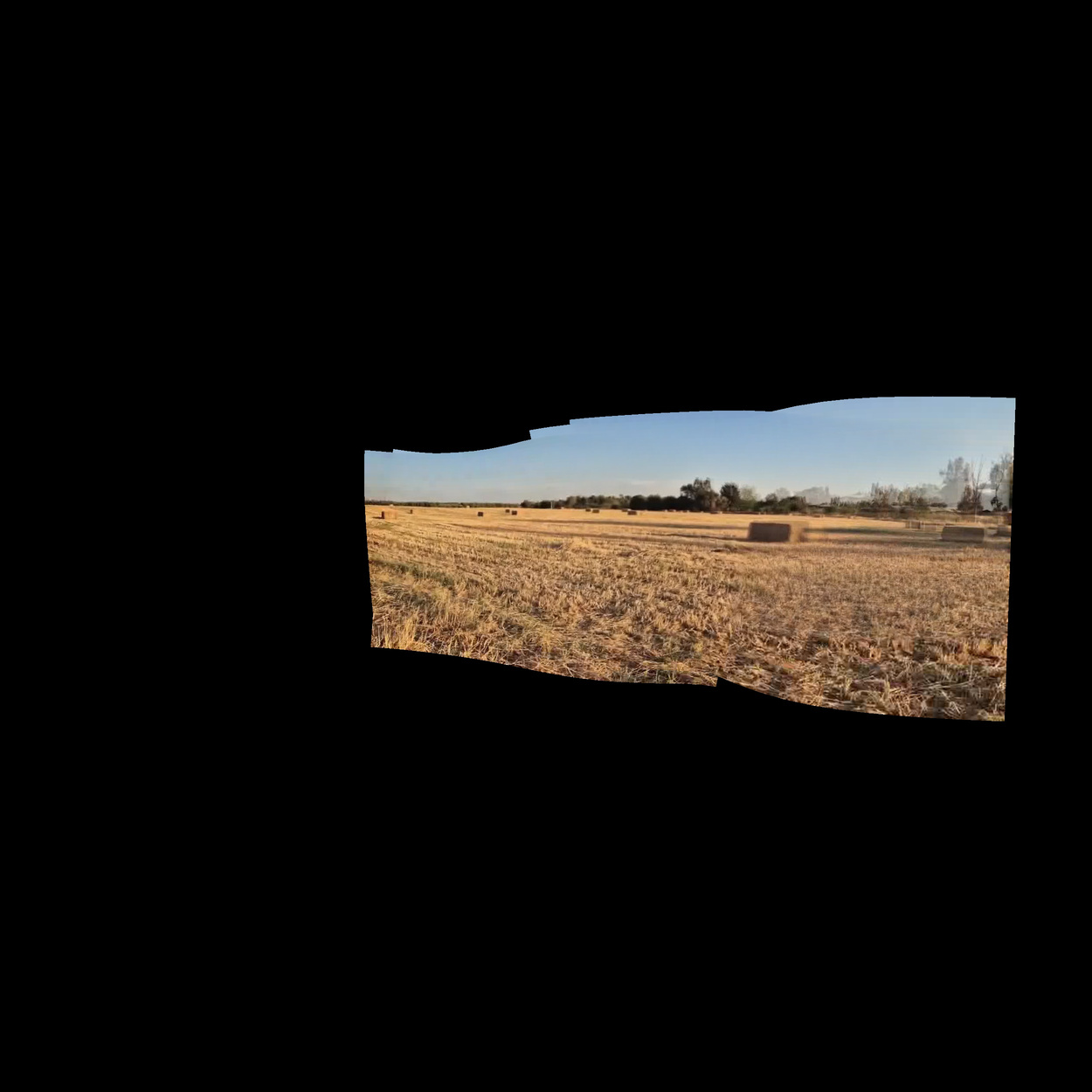}&
\includegraphics[width=3.8cm, height=2.1cm,    viewport=500 800 2100 1800,   clip]{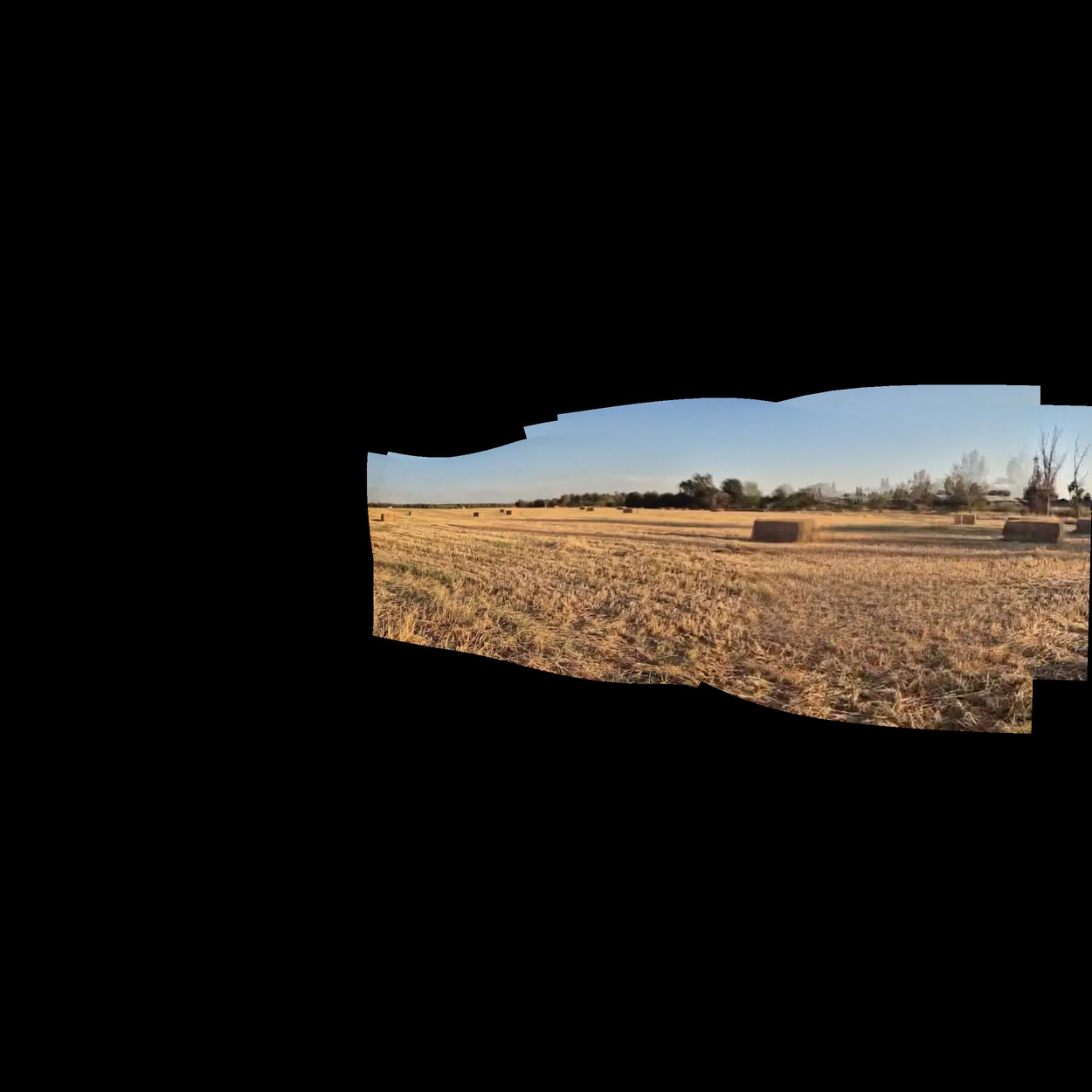} \\[0.5mm]

\rotatebox{90}{\parbox{2.1cm}{\centering \small SPW ~\cite{Liao:TIP:2019:SPW}}} & 
\includegraphics[width=3.8cm, height=2.1cm,    viewport=500 800 2100 1800,   clip]{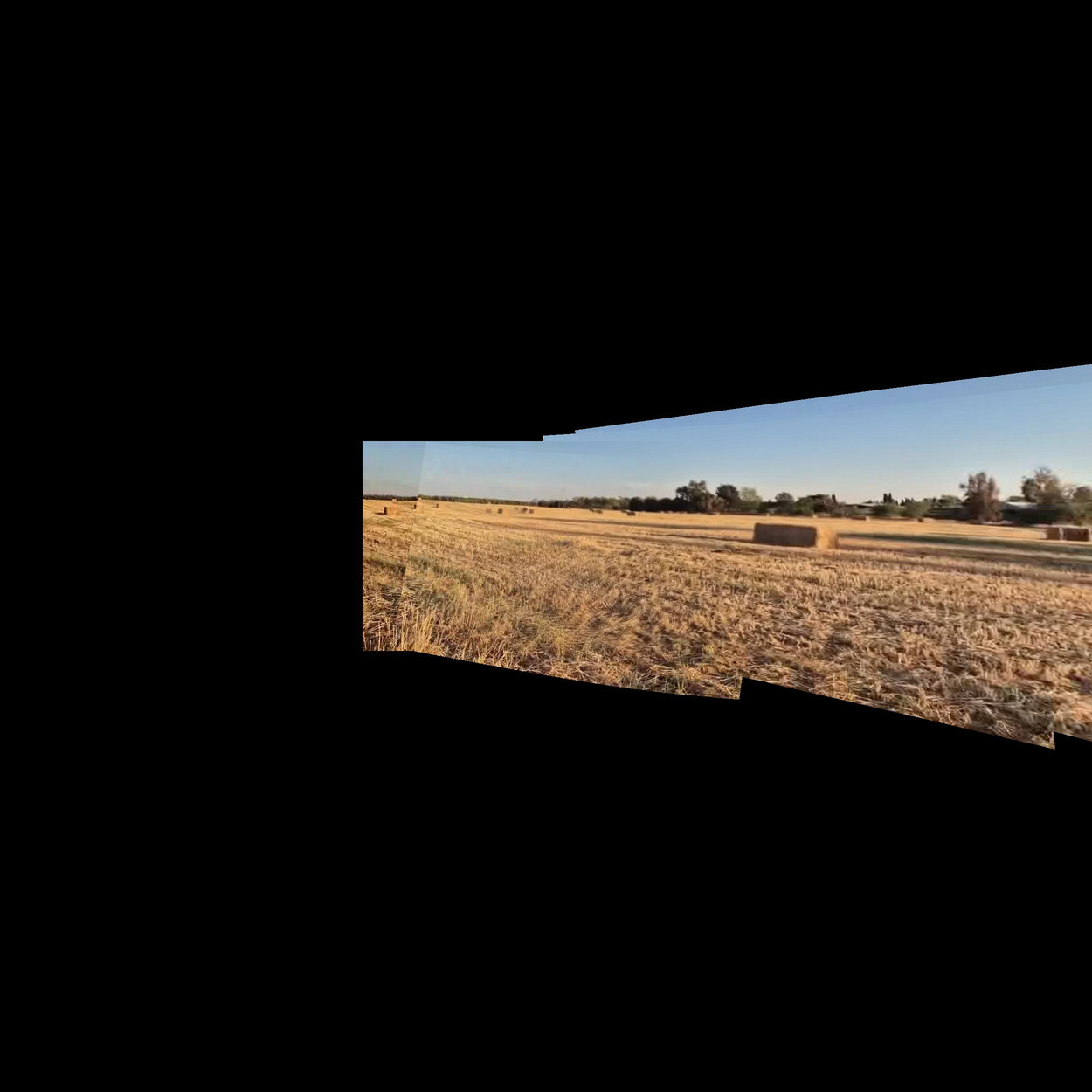} &
\includegraphics[width=3.8cm, height=2.1cm,    viewport=500 800 2100 1800,   clip]{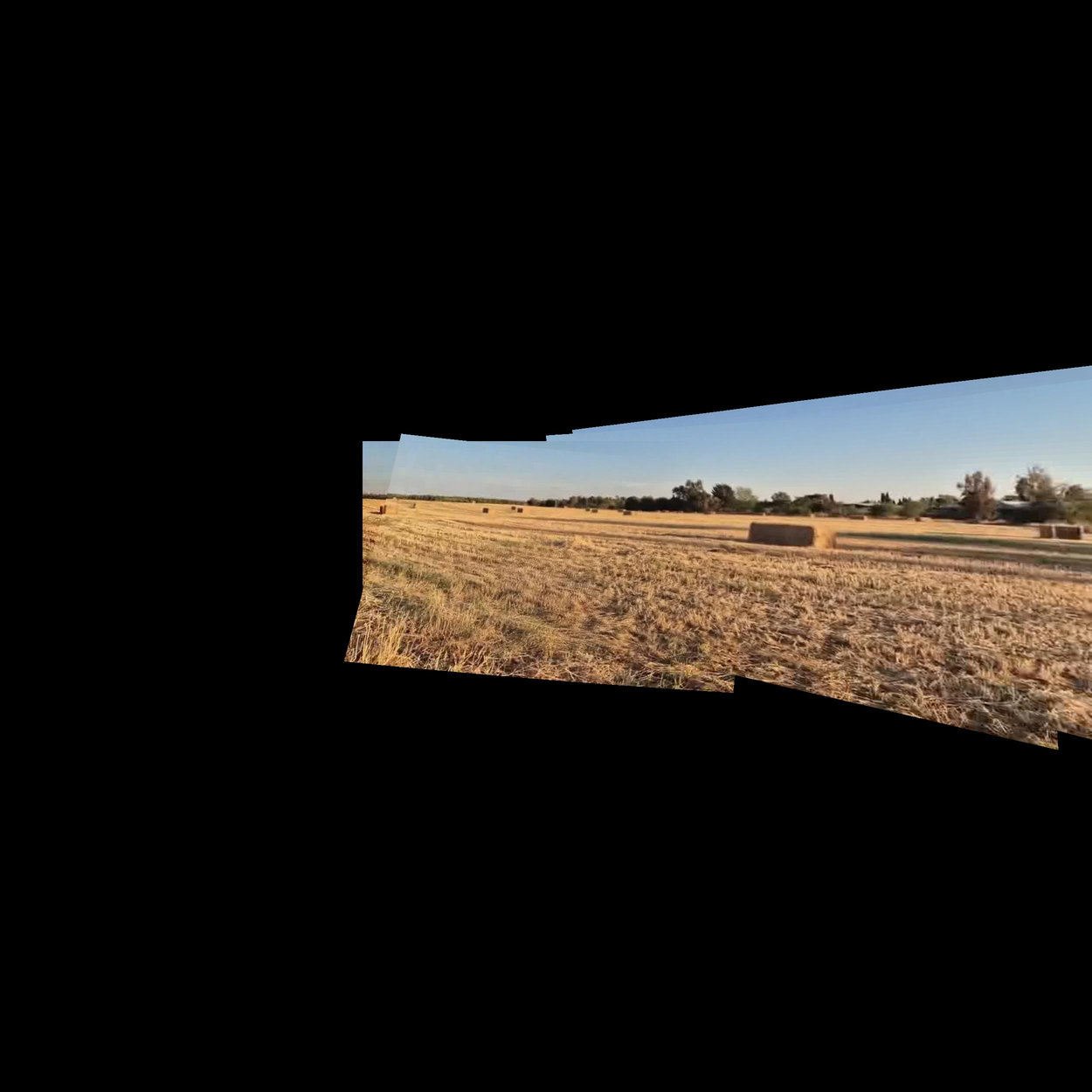} &
\includegraphics[width=3.8cm, height=2.1cm,    viewport=500 800 2100 1800,   clip]{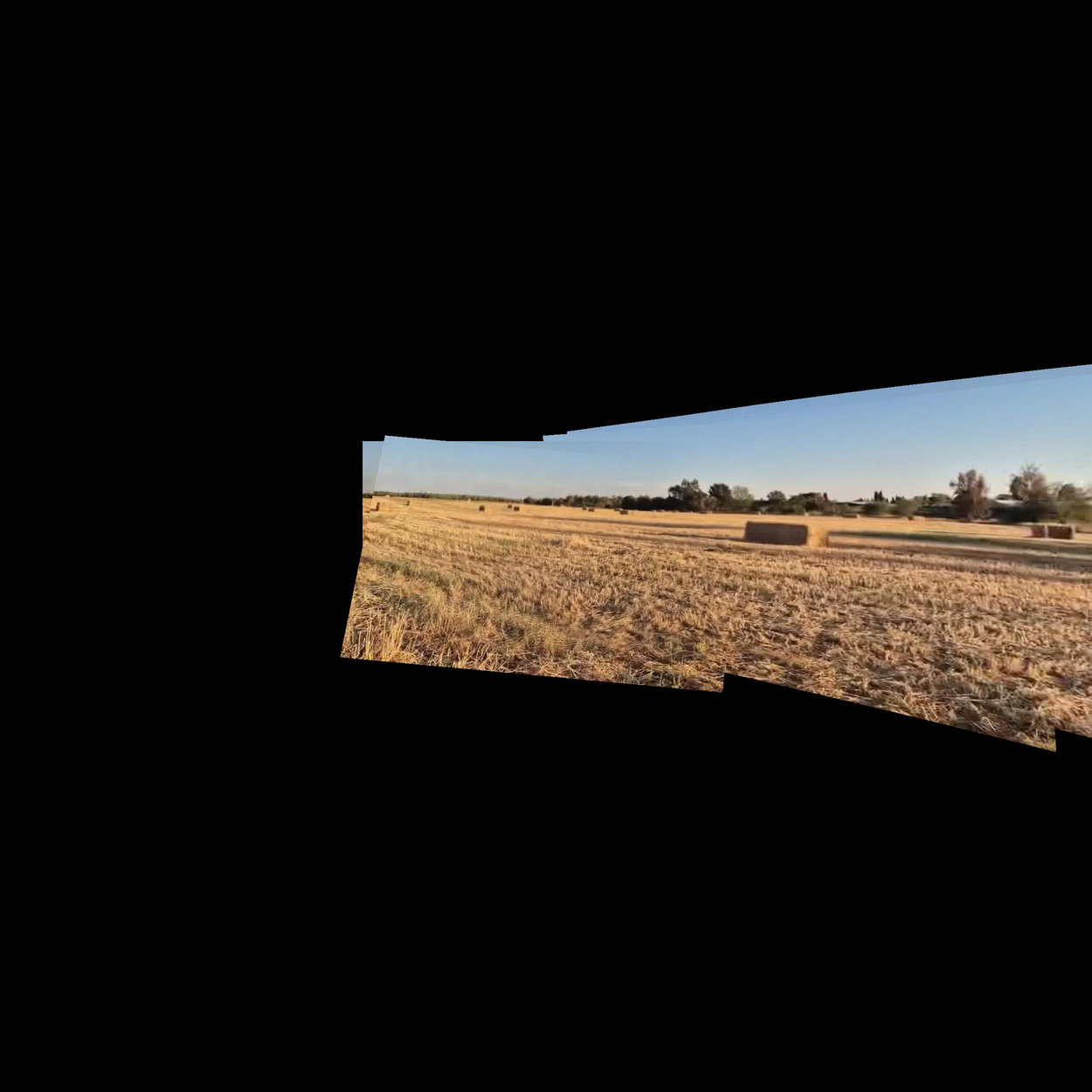}&
\includegraphics[width=3.8cm, height=2.1cm,    viewport=500 800 2100 1800,   clip]{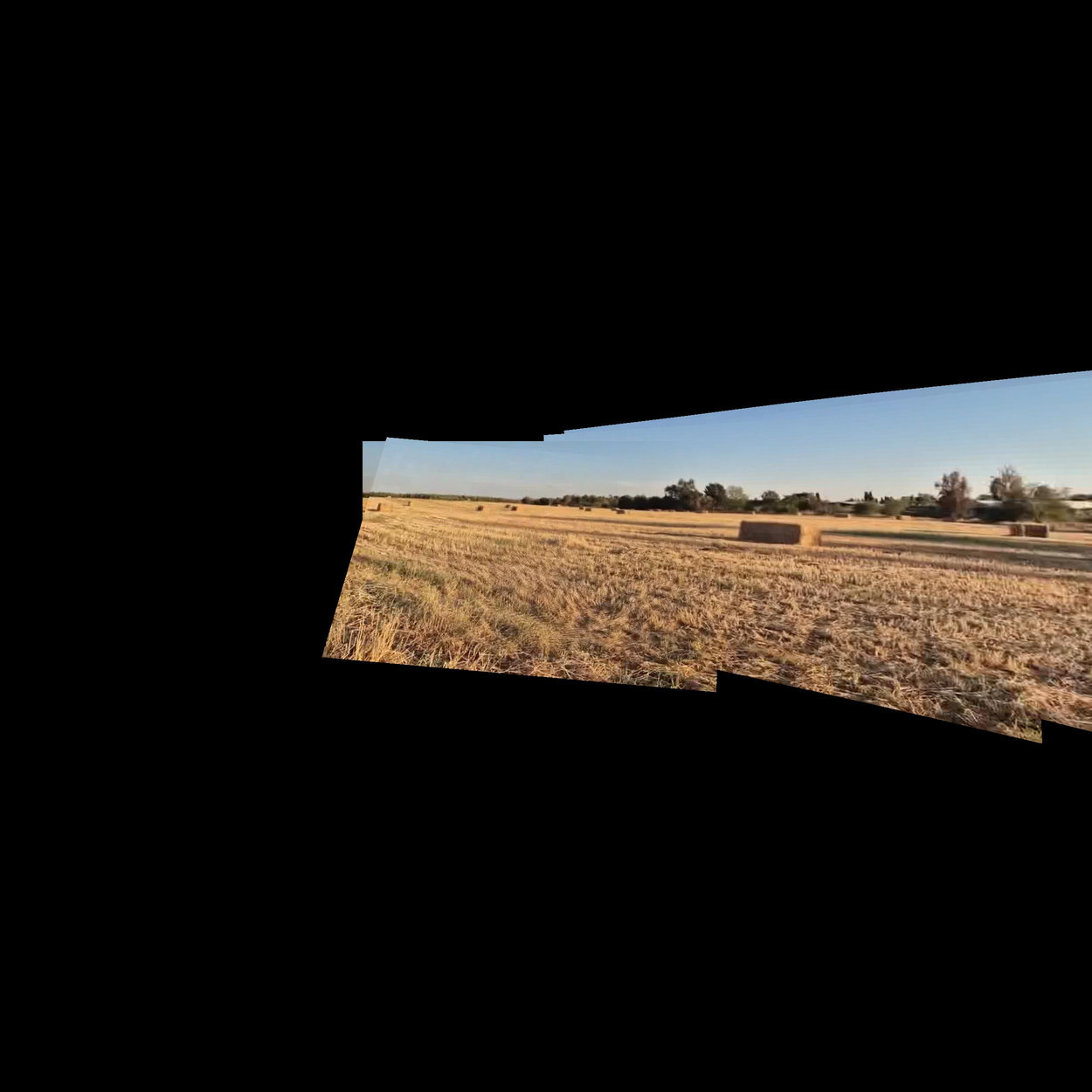} \\[0.5mm]

\rotatebox{90}{\parbox{2.1cm}{\centering \small GES-GSP ~\cite{Du:CVPR:2022:GESGSP}}} & 
\includegraphics[width=3.8cm, height=2.1cm,    viewport=500 800 2100 1800,   clip]{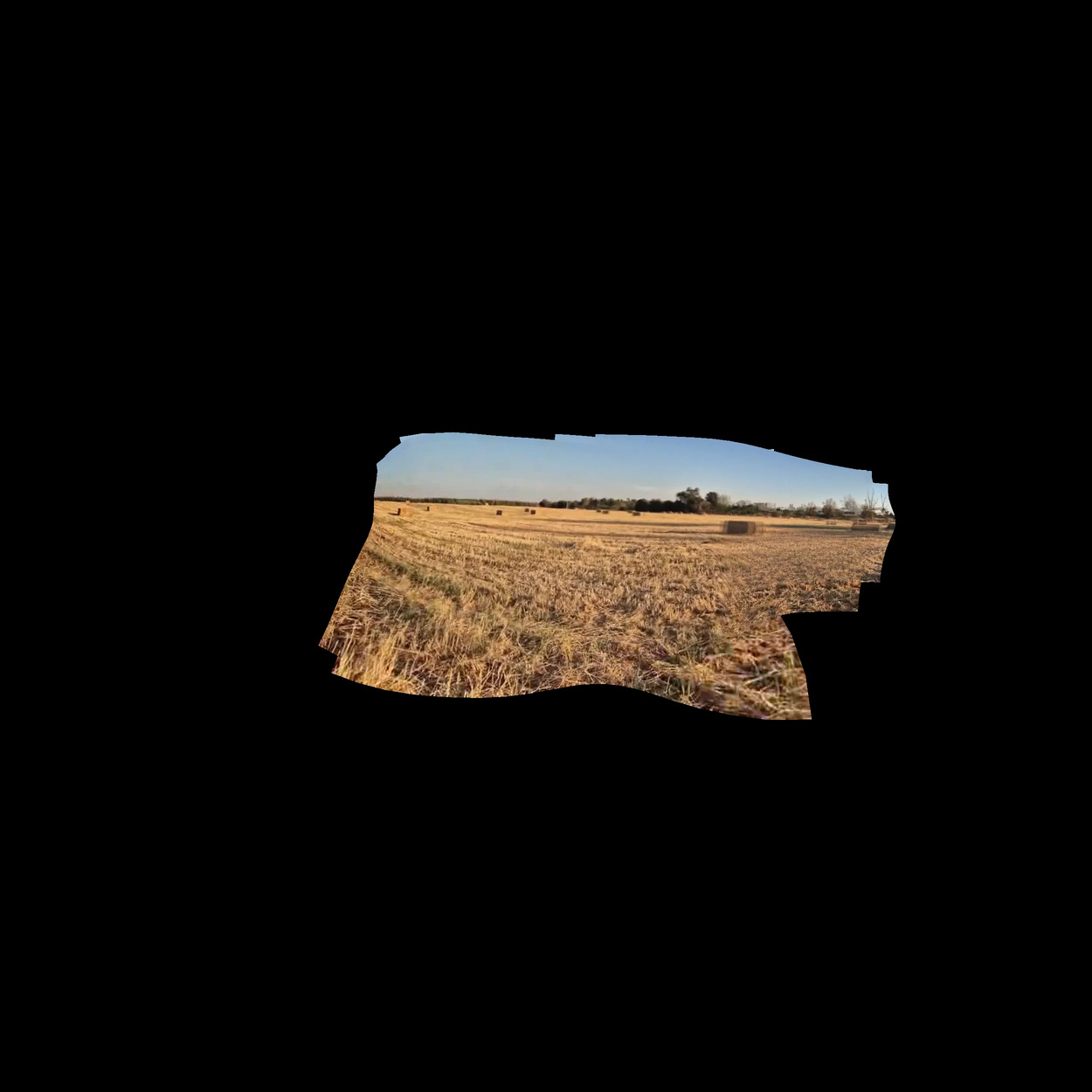} &
\includegraphics[width=3.8cm, height=2.1cm,    viewport=500 800 2100 1800,   clip]{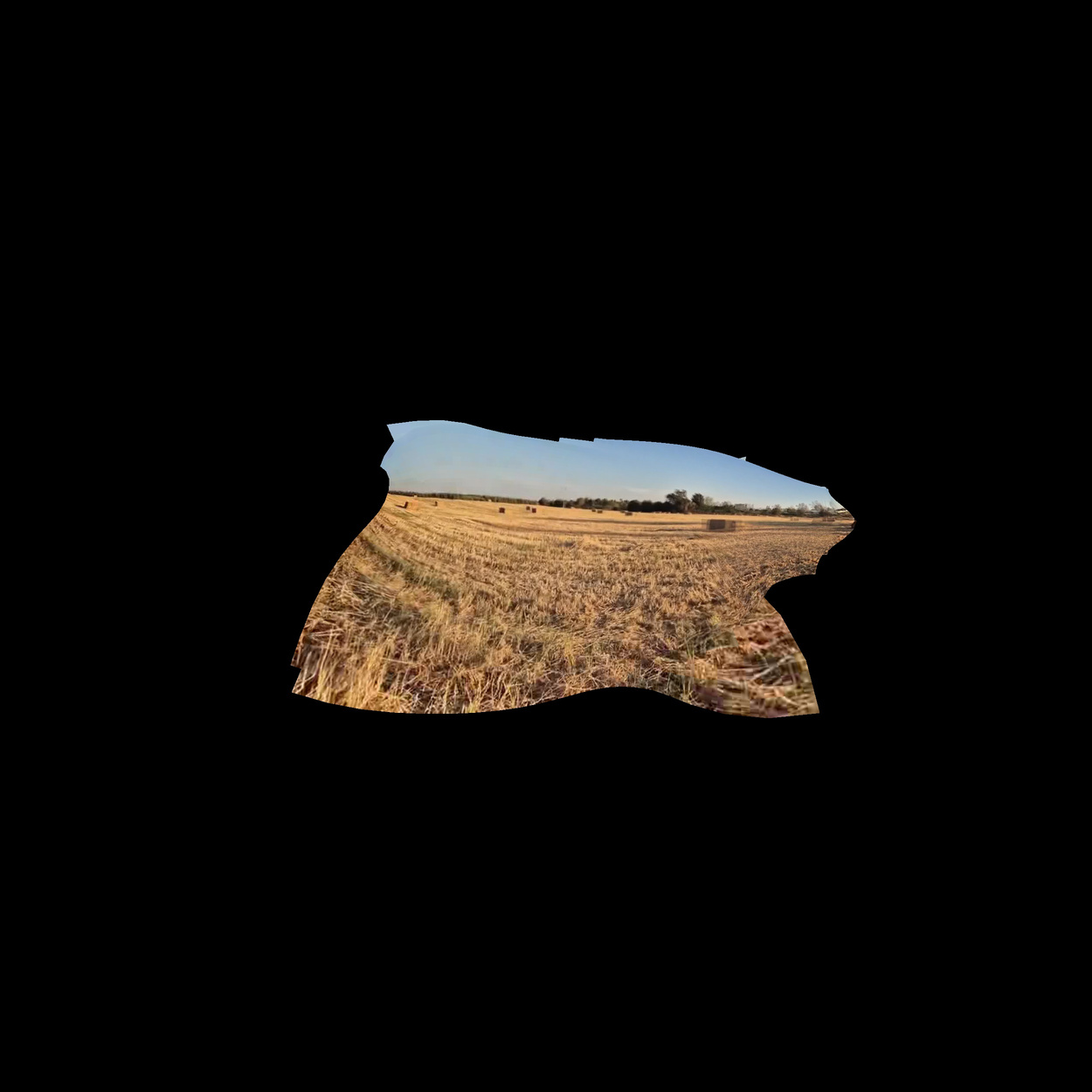} &
\includegraphics[width=3.8cm, height=2.1cm,    viewport=500 800 2100 1800,   clip]{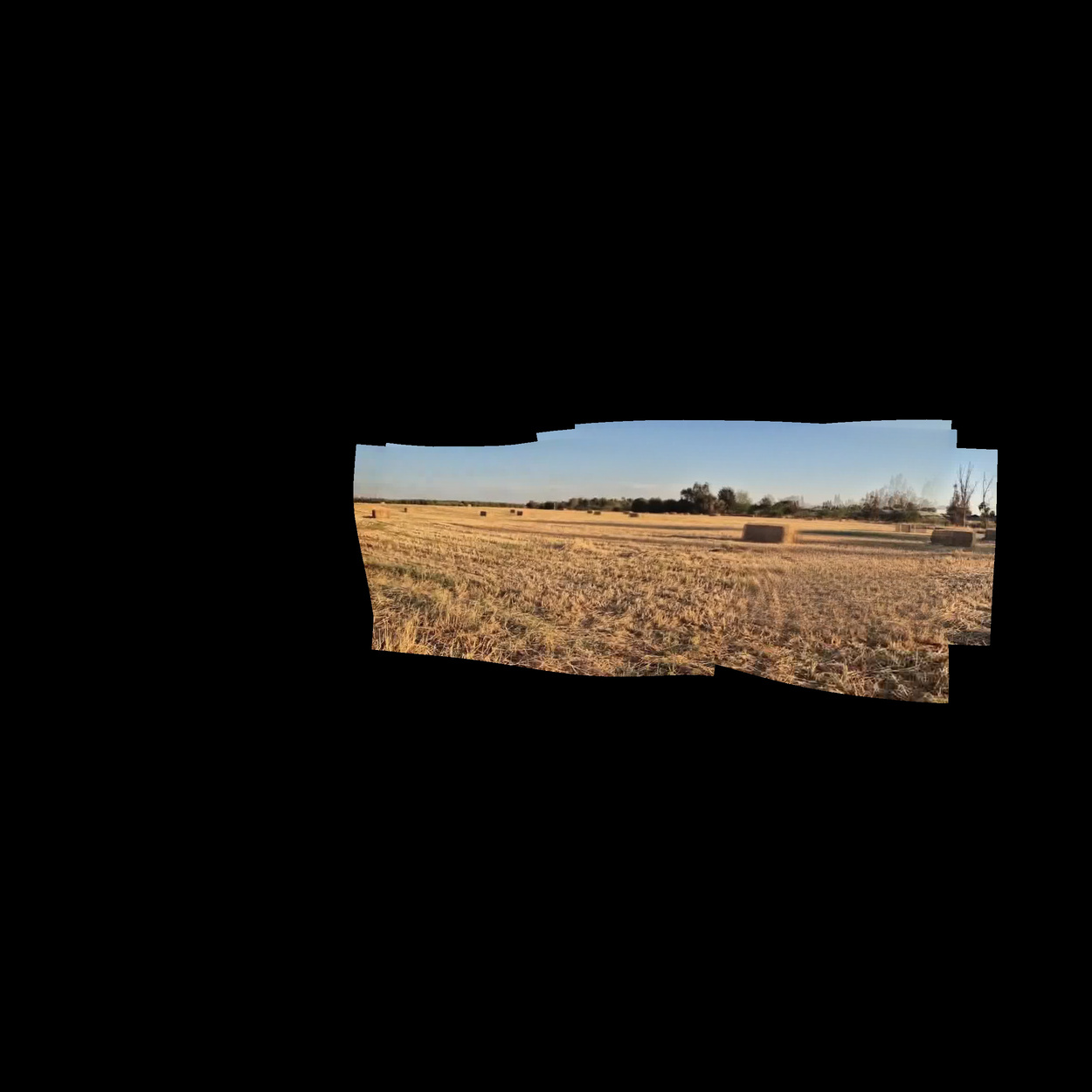}&
\includegraphics[width=3.8cm, height=2.1cm,    viewport=500 800 2100 1800,   clip]{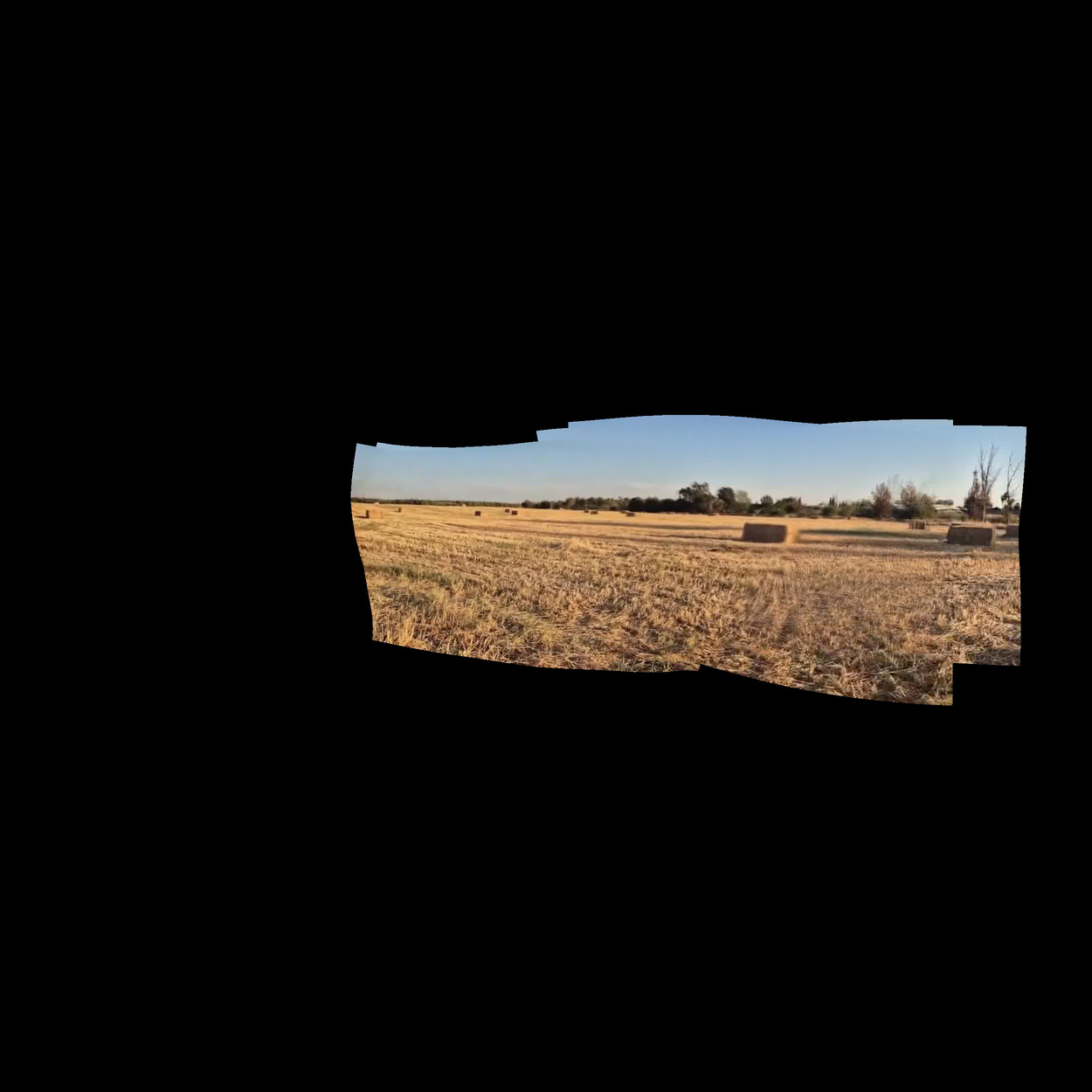} \\[0.5mm]

\rotatebox{90}{\parbox{2.1cm}{\centering \small MViS (Ours)}} & 
\includegraphics[width=3.8cm, height=2.1cm,    viewport=280 580 1692 1262,   clip]{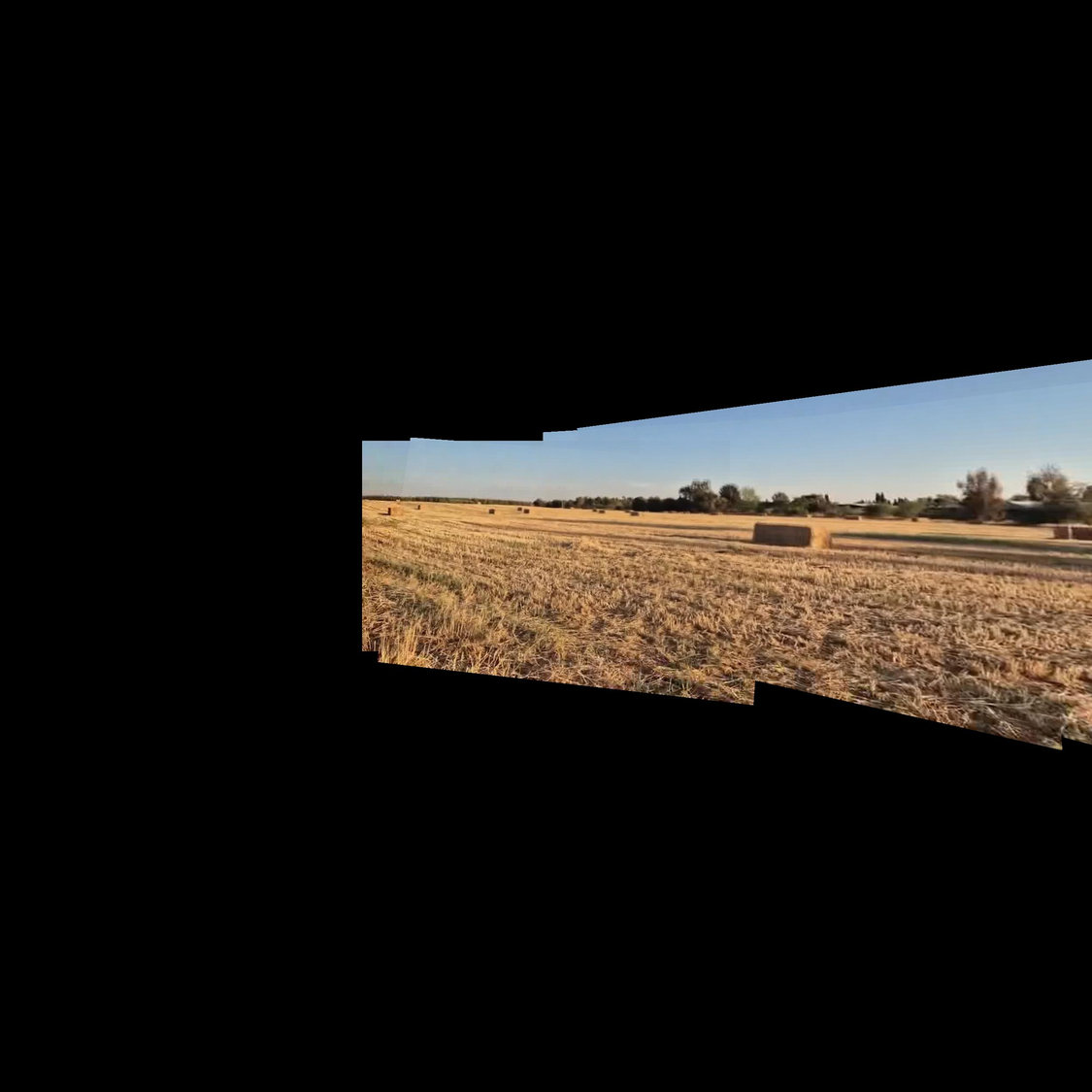} &
\includegraphics[width=3.8cm, height=2.1cm,    viewport=280 580 1692 1262,   clip]{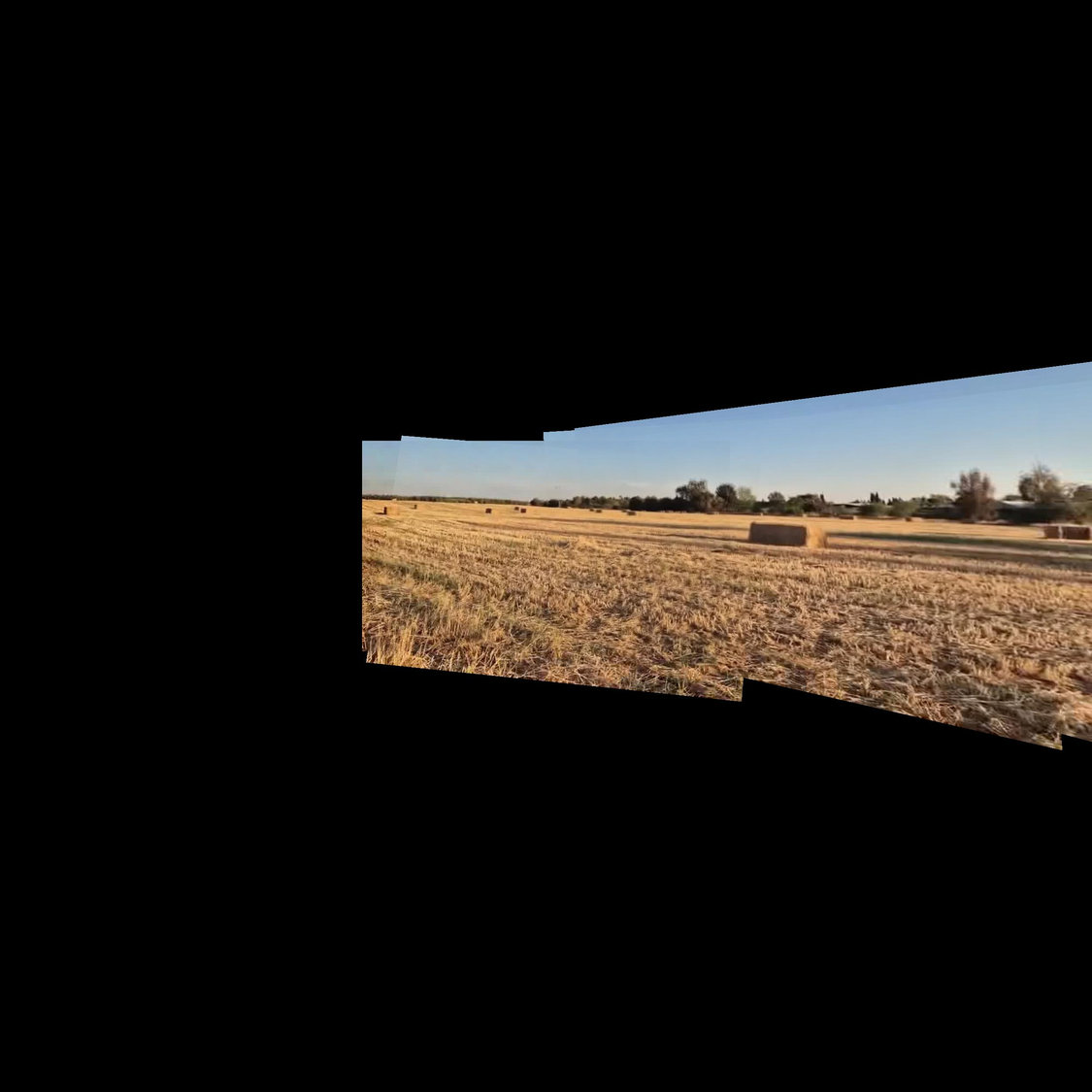} &
\includegraphics[width=3.8cm, height=2.1cm,    viewport=280 580 1692 1262,   clip]{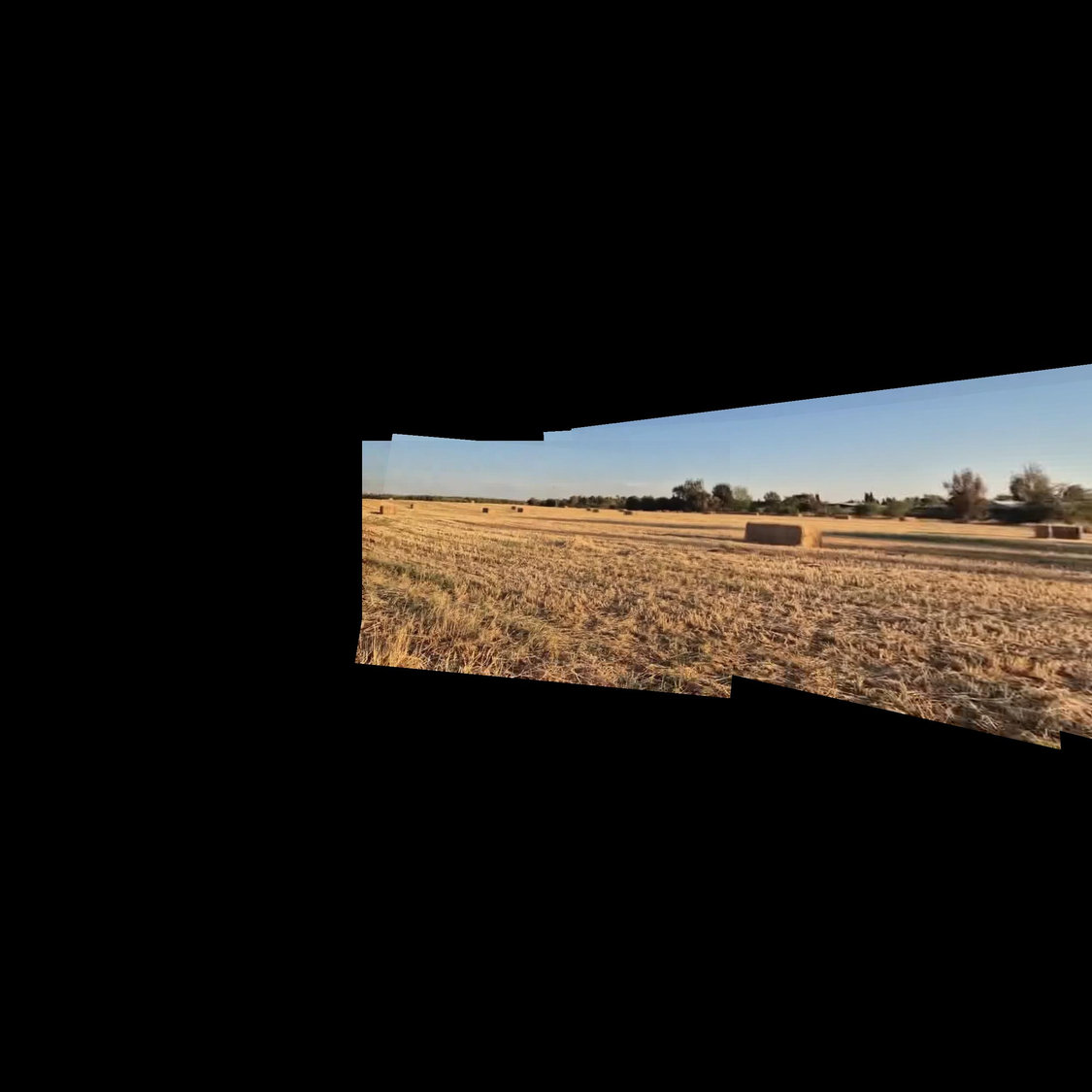}&
\includegraphics[width=3.8cm, height=2.1cm, viewport=280 580 1692 1262,   clip]{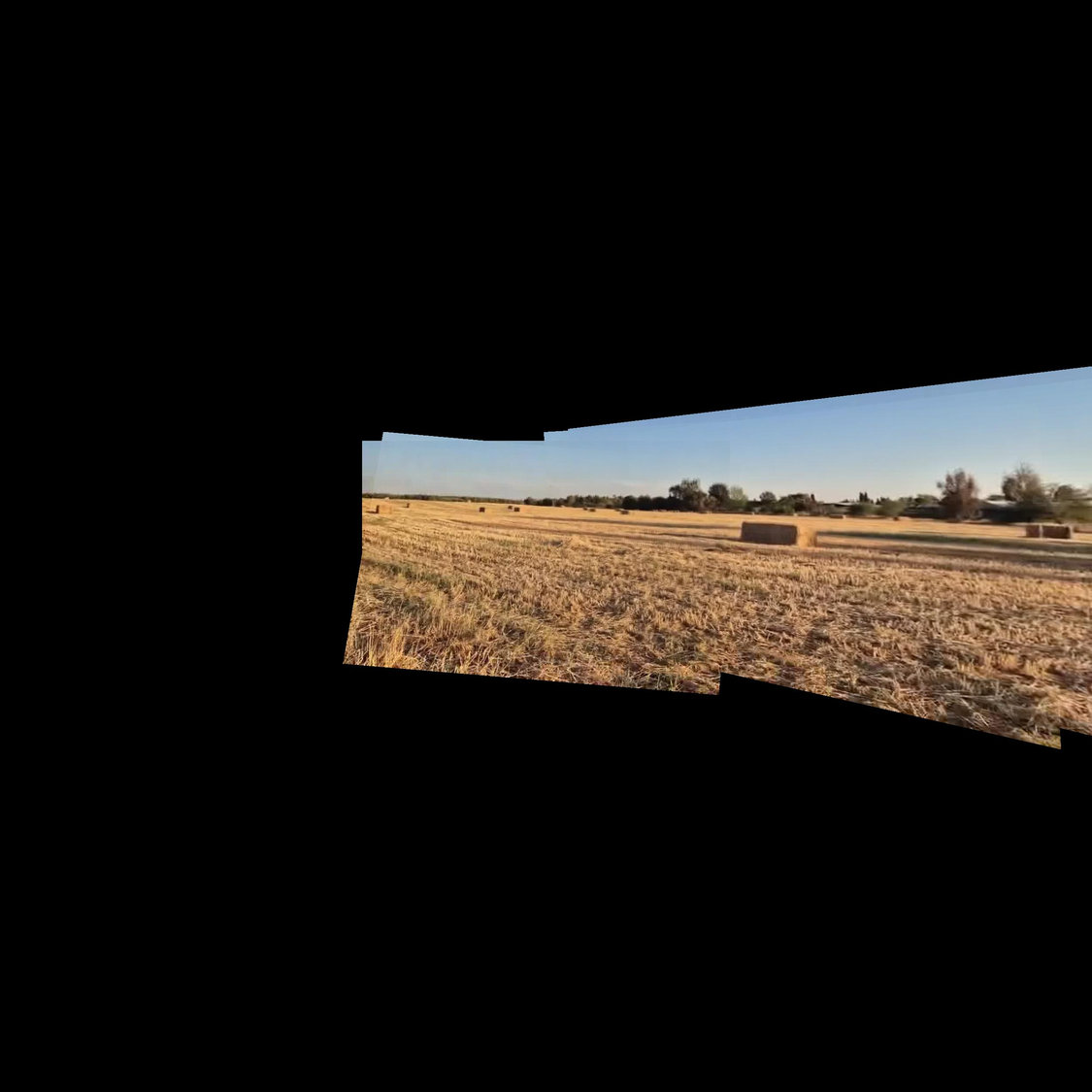} \\[-0.5mm]

& \small{$t=220$} & 
\small{$t=221$} & 
\small{$t=222$} &
\small{$t=223$} \\[3mm]

\end{tabular}
}
\caption{Visual comparison between MIS methods and MViS on scenes with $N>3$ cameras from the MMCD dataset, continued.}
  \label{fig:mis_vis2}
\end{figure}
\clearpage

\section{Parameterizing Homographies via The Special Linear Group}
\label{sec:sl}
Without special care, a deep net might predict non-invertible 
matrices. Besides the fact that this
 could hinder the optimization process~\cite{Skafte:CVPR:2018:DDTN},
 our approach requires invertible
matrices. 
Thus, we use the well-known parameterization of homographies using the Special Linear group $\mathrm{SL}(3)$ (see, \eg, \cite{mei2006homography,Arrigoni:IJCV:2020:Sync,Barel:ECCV:2024:JA}).  
Recall that a homography has 8 degrees of freedom
and every homography is identified with an equivalence class of invertible matrices, where all of the class members
 are proportional to each other and the class representative 
 can be taken to be that whose determinant is one. Let 
\begin{align}
\mathfrak{sl}(3)&= \left\{\bTheta: \bTheta=
\left[\begin{smallmatrix} \theta_1 & \theta_2 & \theta_3 \\ \theta_4 & \theta_5 & \theta_6 \\ \theta_7 & \theta_8 & -(\theta_1+\theta_5) 
\end{smallmatrix}\right]\right\}, 
%\label{Eqn:sl}
\nonumber\\
\mathrm{SL}(3)&= \left\{\bH: \bH=
\left[\begin{smallmatrix} H_1 & H_2 & H_3 \\ H_4 & H_5 & H_6 \\ H_7 & H_8 & H_9 
\end{smallmatrix}\right]
\& \det\bH=1 \right\} .
\label{Eqn:SL}
\end{align}

 $\mathrm{SL}(3)$ is the matrix Lie group of real-valued 3-by-3 matrices 
 with determinant 1, while $\mathfrak{sl}(3)$
 is the corresponding Lie algebra of zero-trace real-valued 3-by-3 matrices.
 While  $\mathfrak{sl}(3)$ is a linear space 
 and $\mathrm{SL}(3)$ is a nonlinear manifold, 
 both $\mathrm{SL}$ and  $\mathfrak{sl}$ have the same dimension, 8. 
Moreover,  $\mathrm{SL}(3)$       can be identified with the group of homographies in $\Rtwo$ 
and the matrix exponential maps  $\mathrm{sl}(3)\to\mathrm{SL}(3)$ via $\bTheta\mapsto \exp(\bTheta)$.  
Thus, we set our alignment network to predict zero-trace matrices by predicting 8-length vectors
of the form $\btheta=(\theta_1,\ldots,\theta_8)$, where the obvious bijection
between the vector $\btheta$ and the zero-trace 3-by-3 matrix $\bTheta$ is 
as in \autoref{Eqn:SL}. Then, via the matrix exponential, we obtain homographies.  
In a slight abuse of notation, we will write $\exp(\btheta)$ instead of $\exp(\bTheta)$. 
Also useful are the following two facts: 1) a zero-vector $\btheta=\bzero_{8\times 1}$ corresponds to the identity matrix (i.e., $\bH=\bI_{3\times 3}$), 
and 2) the inverse matrix of 
$\exp(\btheta)$ is, conveniently, 
$\exp(-\btheta)$.  Finally, for two different 8-length vectors, $\btheta_i$ and $\btheta_j$,  in general $\exp(\btheta_i)\exp(\btheta_j)\neq
\exp(\btheta_j)\exp(\btheta_i)$ (i.e., the homography group is not Abelian); this detail means that the order in which composition of the transformation is done (\eg, in  our loss functions) is important.

\textbf{Remark:} Recall that in general any invertible matrix defines a homography, and that if that matrix is multiplied by a nonzero scalar it still defines the same homography. In other words, in general a homography corresponds to an equivalence class of invertible matrices (that are proportional to each other). However, when using the SL parameterization above, this becomes a nonissue since, by construction, the 
$\det=1$ constraint essentially boils down to picking, for each such equivalence class, exactly one representative: the one whose determinant is 1.
\clearpage

\section{Additional Experiments}
\label{sec:exps}

\subsection{Optimization without a Neural Network}
\label{sec:optimization}
\autoref{Table:optimization} reports CMSA values (lower is better) for the \textbf{Optimization without a net} experiment described in the main paper. Instead of predicting homography parameters with the Alignment Network, we \textbf{directly optimize} the per-frame homography parameters $\{\boldsymbol{\theta}_i^t\}$ under the same losses, considering three variants:
\begin{enumerate}
    \item $\Lcal_{\mathrm{geo}}$ (w/o Net) - Geometric Loss only
    \item $\Lcal_{\mathrm{rgb}}$ (w/o Net) - Appearance Loss only
    \item w/o Net - a two stage scheme that first minimizes $\Lcal_{\mathrm{geo}}$ and then uses the resulting solution as initialization to minimize $\Lcal_{\mathrm{rgb}}$.
\end{enumerate}

To ensure proper convergence, for each variant we test 4 learning rate values and report the best result; each loss is optimized for 1000 iterations (so w/o Net runs 2000 iterations in total).

Across all $N$ values, MViS consistently achieves lower CMSA than any \newline optimization-only variant, showing that the Alignment Network contributes to temporal stabilization. As explained in the paper, we attribute this phenomenon to the inductive bias of the network towards smoother solutions: consecutive frames tend to be similar, so they lead to similar predicted homographies. A visual comparison between MViS and a direct optimization without a neural network is provided in \texttt{11\_with\_vs\_without\_net.mp4}.

\begin{table}[h]
  \caption{\textbf{Optimization without a net.} Measured CMSA on the frame's corners. In $N=2$ we used StabStitch-D dataset, and for $N>2$ the MMCD dataset.}
  \label{Table:optimization}
  \centering
  \small
  \begin{tabular}{@{}lccccc@{}}
    \toprule
    Loss &$N=2$&$N=3$& $N=4$ &$N=5$ &$N=6$\\
    \midrule
    $\Lcal_{\mathrm{}{geo}}$~(w/o~Net)& 39.9 & {\scriptsize >1k} & {\scriptsize >1k} & 485.3 & 18.0 \\
    $\Lcal_{\mathrm{rgb}}$ ~(w/o~Net) & {\scriptsize >1k} &  {\scriptsize >1k} & {\scriptsize >1k} & {\scriptsize >1k} & {\scriptsize >1k}\\
    w/o~Net & 118.7 & 35.8 & 86.4 & 11.6 & 21.3 \\
    \midrule
     MViS & \textbf{6.8} & \textbf{2.5} & \textbf{14.2} & \textbf{11.} & \textbf{2.4} \\
    
    \bottomrule
  \end{tabular}
  
\end{table}

\clearpage

\subsection{Regularization Experiments}
\label{sec:reg}
We experimented with adding explicit regularization terms to the loss formulations of both stages, weighted by a global coefficient $\alpha$ that controls the strength of the regularization.
We evaluated two terms over the predicted transformations using three different global weights $\alpha \in \{ 0.5,1,5\}$.

The first term, $\Rcal_1$=$\tfrac{1}{N}
\sum_{i=1}^{N}
\left\| \bTheta^{t+1}_{i} - \bTheta^{t}_{i} \right\|_{F}^{2}$, encourages temporal smoothness by penalizing changes in the predicted transformation parameters between consecutive frames.
The second, $\Rcal_2$=$\tfrac{1}{2N}
\sum_{k\in\{0,1\}}
\sum_{i=1}^{N}
\left\|\bTheta^{t+k}_{i} \right\|_{F}^{2}
$, penalizes the magnitude of the predicted transformations. $\bTheta$ is defined in \autoref{Eqn:SL}.
We evaluated these variants on two datasets: StabStitch-D ($N=2$ case) and the MMCD dataset ($N=3,4$ cases).

As shown in \autoref{tab:reg_exps}, explicit regularization did not provide a consistent overall improvement over MViS across the evaluated datasets, categories, and metrics.
This of course does not suggest that regularization is always inherently harmful; however, the point is 
that MViS simply does not require regularization terms to perform well. Moreover, making such regularization effective requires selecting a suitable value of $\alpha$, and choosing this value in practice requires supervision, since one must evaluate which setting performs best. Beyond the fact that the optimal $\alpha$ may vary across scenes, and potentially even across cameras and time, this tuning process introduces an additional supervised model-selection step. Avoiding this dependency is therefore a practical advantage of MViS.

\begin{table}[h]
  \centering
  \caption{Regularization experiments on StabStitch-D dataset ($N=2$) and the MMCD dataset ($N=3,4$).}

  \label{tab:reg_exps}
  {

  \footnotesize
  \begin{tabular}{@{}
      c c 
      *2{c}
      *3{c}
      *3{c} @{}
    }
    \toprule
    \multirow{2}{*}{$\Rcal$} & \multirow{2}{*}{$\alpha$}
    & \multicolumn{2}{c}{\small $N=2$}
    & \multicolumn{3}{c}{\small $N=3$} & \multicolumn{3}{c}{\small $N=4$}\\ 

    & & {\scriptsize PSNR~$\uparrow$}
      & {\scriptsize SSIM~$\uparrow$}
      & {\scriptsize GV~$\downarrow$}
      & {\scriptsize Stab.~$\uparrow$}
      & {\scriptsize CMSA~$\downarrow$}
      & {\scriptsize GV~$\downarrow$}
      & {\scriptsize Stab.~$\uparrow$}
      & {\scriptsize CMSA~$\downarrow$} \\
    \midrule

    \multirow{3}{*}{$\Rcal_1$}
      & 0.5 & \textbf{30.24} & \textbf{.904} & \textbf{3.1} & \textbf{.917} & 2.2 & \textbf{2.8} & \textbf{.925} & 35.1 \\ 
      & 1.0 & 30.17 & \textbf{.904} & \textbf{3.1} & \textbf{.917} & 2.8 & \textbf{2.8} & \textbf{.925} & 14.2 \\ 
      & 5.0 & 29.97 & .901 & \textbf{3.1} & .916 & \textbf{1.9} & \textbf{2.8} & .924 & 15.6 \\ 

    \midrule

    \multirow{3}{*}{$\Rcal_2$}
      & 0.5 & 29.02 & .890 & 7.2 & .913 & 21.7 & 6.7 & \textbf{.925} & 16.6 \\ 
      & 1.0 & 25.12 & .819 & 14.3 & .909 & 15.9 & 24.3 & .894 & \textbf{10.6} \\ 
      & 5.0 & 19.10 & .613 & 44.8 & .881 & 36.6 & 86.2 & .838 & 113.2 \\ 

    \midrule
    MViS & -- & \textbf{30.24} & \textbf{.904} & \textbf{3.1} & \textbf{.917} & 2.5 & \textbf{2.8} & \textbf{.925} & 14.2 \\ 
    \bottomrule
  \end{tabular}
  }

\end{table}
\clearpage

\subsection{Ablation on Image Matcher Choice}
\label{sec:matchers}
To evaluate the robustness of MViS to the choice of image matcher, we run our method with RoMa~\cite{Edstedt:CVPR:2024:RoMa} and LoFTR~\cite{sun:2021:loftr}. 
All other components, optimization settings, and hyperparameters are kept fixed. 
As shown in \autoref{tab:matchers}, MViS performs reliably with both matchers, indicating that our framework does not depend on a specific matcher, but only requires sufficiently reliable pairwise correspondences.

\begin{table}[h]
  \caption{ Image Matchers Comparison. 
  }
  \label{tab:matchers}
  \centering
  \scriptsize
  \renewcommand{\arraystretch}{1.05}
  \begin{tabular}{@{}
  l
  @{\hskip 3pt} c
  @{\hskip 3pt} c
  @{\hskip 6pt} c
  @{\hskip 3pt} c
  @{\hskip 3pt} c
  @{\hskip 6pt} c
  @{\hskip 3pt} c
  @{\hskip 3pt} c
  @{\hskip 6pt} c
  @{\hskip 3pt} c
  @{\hskip 3pt} c
  @{\hskip 6pt} c
  @{\hskip 3pt} c
  @{\hskip 3pt} c
@{}}
    \toprule
    \multirow{1}{*}{\footnotesize Ablation}
& \multicolumn{2}{c}{\footnotesize $N=2$}
& \multicolumn{3}{c}{\footnotesize $N=3$}
& \multicolumn{3}{c}{\footnotesize $N=4$}
& \multicolumn{3}{c}{\footnotesize $N=5$}
& \multicolumn{3}{c}{\footnotesize $N=6$} \\
     & {\scriptsize PSNR}& {\scriptsize SSIM} & {\scriptsize GV} & {\scriptsize Stab.} & {\scriptsize CMSA}& {\scriptsize GV} & {\scriptsize Stab.} & {\scriptsize CMSA}& {\scriptsize GV} & {\scriptsize Stab.} & {\scriptsize CMSA}& {\scriptsize GV} & {\scriptsize Stab.} & {\scriptsize CMSA}\\
    \midrule

    MViS (LoFTR) & 30.19 & \textbf{.905} & \textbf{3.1} & \textbf{.917} & 3.5 & \textbf{2.8} & .924 & \textbf{12.4} & \textbf{2.3} & \textbf{.874} & \textbf{8.7} & \textbf{3.0} & \textbf{.930} & 4.3\\
    MViS (RoMa) & \textbf{30.24} & .904 & \textbf{3.1} & \textbf{.917} & \textbf{2.5} & \textbf{2.8} & \textbf{.925} & 14.2 & \textbf{2.3} & \textbf{.874} & 11.0 & \textbf{3.0} & \textbf{.930} & \textbf{2.4}\\
    \bottomrule
  \end{tabular}
  
\end{table}
\clearpage

\subsection{Effect of Camera Motion}
\label{sec:camera_motion}
To study how camera motion affects MViS, we assign each MMCD scene a
scalar camera-motion score.

\textbf{Camera motion.}
For each time $t$, we compute the largest corner displacement across all cameras:
\begin{equation}
  m_t = \max_{i=1,\dots,N} \, \max_{k=1,\dots,4}
  \bigl\| \pi\bigl(\widehat{\bH}_i^{t\to t+1} \tilde{\mathbf{c}}_k\bigr) - \mathbf{c}_k \bigr\|_2 ,
\end{equation}

where $\widehat{\bH}^{t \rightarrow t+1}_{i}$ is an ORT, $\mathbf{c}_k$ is the $k$-th image corner,
$\tilde{\mathbf{c}}_k$ is its homogeneous representation, and $\pi$ converts
it to Euclidean coordinates.

The scene-level camera motion
is defined as
\begin{equation}
  \mathrm{CameraMotion} = 100 \cdot \operatorname*{median}_{t}
  \left( \frac{m_t}{\text{image diagonal}} \right).
\end{equation}

We partitioned the MMCD scenes into 3 groups by camera-motion score and report the median and IQR for each metric in \autoref{table:camera_motion_groups}. Higher camera motion is associated with lower Stab. and higher CMSA, while GV shows no monotonic trend, indicating no clear association with geometric alignment quality in these experiments.

\begin{table}[h]
\centering
\caption{Effect of camera motion on MViS}
\label{table:camera_motion_groups}
\resizebox{\linewidth}{!}{
\begin{tabular}{l c c c c c}
\toprule
Group & CameraMotion range & \# of scenes & GV $\downarrow$ & Stab. $\uparrow$ & CMSA $\downarrow$\\
\midrule
Low  & 0.17--0.46 & 17 & 2.3 [1.7--3.0] & .927 [.913--.958] & 0.1 [0.1--0.5]  \\
Mid  & 0.48--0.77 & 16 & 3.1 [2.1--3.4] & .927 [.913--.940] & 1.0 [0.3--7.1]  \\
High & 0.78--1.50 & 20 & 2.4 [1.6--3.2] & .896 [.852--.928] & 5.3 [1.1--14.2] \\
\bottomrule
\end{tabular}
}
\end{table}

\clearpage

\subsection{Robustness to ORT Corruption}
\label{sec:ort_corruption}
We perturb 10\%, 20\%, or 30\% of the temporal and multi-view ORTs using projective perturbation matrices constructed from eight Gaussian parameters ($\sigma = 0.01$ for affine/translation components and $0.0015$ for projective components), following \autoref{sec:sl}, Eq. (9).

Because each mask is derived by warping an all-ones mask with its estimated ORT, perturbing an ORT also perturbs the corresponding mask. Independently perturbing the masks would decouple them from the ORTs and would not reflect the pipeline’s actual failure mode.

Since scenes with little shared content between views are expected to be more sensitive to ORT errors, we quantify each scene’s overlap as follows.

\textbf{Overlap ratio.} 
For each directed mask $M^{t}_{i\to j}$, let
\begin{equation}
  O^{t}_{i\to j} = \frac{\text{number of white pixels in } M^{t}_{i\to j}}{\text{total number of pixels}}
\end{equation}
denote the fraction of white pixels in the mask.

We then compute the weakest
pairwise overlap at each time step, $B_t = \min_{i\neq j} O^{t}_{i\to j}$,
and define the scene-level overlap ratio as
$\mathrm{OverlapRatio} = \operatorname*{median}_{t} \, B_t $.

We group the 52 successful MMCD scenes into three categories based on their overlap ratios and report the mean for each category in \autoref{table:ort_corruption}. The remaining scene, whose minimum pairwise overlap is only 1.7\%, fails under the injected perturbations and is therefore not included in the averages.

Overall, MViS remains robust to ORT corruption.
The first-stage loss increases monotonically with the corruption level, confirming that the injected perturbations meaningfully affect the geometric optimization. Specifically, the median first-stage loss across all dataset scenes increases by 70\% when the corruption level rises from 10\% to 20\%, and by 140\% from 10\% to 30\%. In contrast, the median second-stage loss increases by only 0.5\% and 1.\%, respectively, indicating that most of the effect of the corrupted ORTs is handled during the first-stage optimization, while the subsequent appearance-refinement stage remains stable.

\begin{table}[h]
\centering
\caption{Results by overlap-ratio group and noise level.}
\label{table:ort_corruption}
\begin{tabular}{llccccc}
\toprule
Group & OverlapRatio range & Noise (\%) & GV $\downarrow$& Stab. $\uparrow$& CMSA $\downarrow$& \# of scenes \\
\midrule
\multirow{4}{*}{Low}  & \multirow{4}{*}{0.27 -- 0.52} & 0  & 2.9 & .914 & 10.7 & \multirow{4}{*}{18} \\
                      &                               & 10 & 2.9 & .913 & 7.2  & \\
                      &                               & 20 & 2.9 & .912 & 11.0 & \\
                      &                               & 30 & 3.0 & .913 & 10.1 & \\
\midrule
\multirow{4}{*}{Mid}  & \multirow{4}{*}{0.52 -- 0.66} & 0  & 2.7 & .914 & 6.6  & \multirow{4}{*}{16} \\
                      &                               & 10 & 2.7 & .912 & 6.4  & \\
                      &                               & 20 & 2.6 & .912 & 6.6  & \\
                      &                               & 30 & 2.7 & .912 & 6.3  & \\
\midrule
\multirow{4}{*}{High} & \multirow{4}{*}{0.66 -- 0.85} & 0  & 3.1 & .912 & 0.3  & \multirow{4}{*}{18} \\
                      &                               & 10 & 3.1 & .911 & 0.4  & \\
                      &                               & 20 & 3.0 & .910 & 0.5  & \\
                      &                               & 30 & 3.1 & .910 & 0.5  & \\
\bottomrule
\end{tabular}
\end{table}

\clearpage

\section{The Alignment Network Architecture}
\label{sec:arch}
Following the parameterization described in
\autoref{sec:sl}, the alignment network predicts an
$8$-dimensional Lie-algebra vector for each camera and time step.
We initialize all weights in the final fully connected layer
(FC2) to zero, so that the initial predicted transformations are the identity.

The number of parameters in the CNN module is independent of the number of cameras $N$: all BatchNorm and Conv2d layers operate on $2BN$ input frames and contain only a few thousand parameters. The dependence on $N$ appears only in the fully connected module, whose parameter count grows linearly with $N$ ($10{,}240N{+}32$ for FC1 and $264N$ for FC2).

\begin{table}[h]
\centering
\small
\caption{Alignment Network Summary.}
\begin{tabular}{llr}
\toprule
Layer (type:depth-idx) & Output Shape & \#Params \\
\midrule
Alignment Net                          & [$2B$, $8N$]              & -- \\[2pt]

\,\,In\_Norm (BatchNorm2d): 1-1        & [$2BN$, 3, 256, 256]      & 6 \\[1pt]

\,\,CNN: 1-2                  & [$2BN$, 5, 8, 8]          & -- \\
\,\,\,\,Conv2d: 2-1                    & [$2BN$, 10, 250, 250]     & 1,480 \\
\,\,\,\,BatchNorm2d: 2-2               & [$2BN$, 10, 250, 250]     & 20 \\
\,\,\,\,AdaptiveMaxPool2d: 2-3         & [$2BN$, 10, 32, 32]       & -- \\
\,\,\,\,ReLU: 2-4                      & [$2BN$, 10, 32, 32]       & -- \\
\,\,\,\,Conv2d: 2-5                    & [$2BN$, 5, 28, 28]        & 1,255 \\
\,\,\,\,BatchNorm2d: 2-6               & [$2BN$, 5, 28, 28]        & 10 \\
\,\,\,\,AdaptiveMaxPool2d: 2-7         & [$2BN$, 5, 8, 8]          & -- \\
\,\,\,\,ReLU: 2-8                      & [$2BN$, 5, 8, 8]          & -- \\[4pt]

\,\,Flatten (reshape): 1-3             & [$2B$, $320N$]            & -- \\

\,\,FC1 (Linear): 1-4                  & [$2B$, 32]                & $10,240N + 32$ \\
\,\,GELU: 1-5                          & [$2B$, 32]                & -- \\
\,\,FC2 (Linear): 1-6                  & [$2B$, $8N$]              & $264N$ \\
\bottomrule
\end{tabular}
\end{table}

\clearpage

\section{Further Details on the Multi-Camera Benchmark and MMCD Dataset}
\label{sec:details_ben}

\subsection{The Geometric Variance Metric (GV)}
\label{sec:geovar}

To the best 
of our knowledge there is no existing suitable metric for multi-view video stitching evaluation
when $N>2$.
For example,
  PSNR and SSIM,
  while sometimes used in the 
  case of $N=2$, require a reference frame and a target frame,
  which are less
  relevant
  in the more general case. 

To be able to measure the alignment quality of our stitched videos over time, we annotated keypoints on the videos. We stress
that these annotations
were used only for evaluation
purposes. 

Concretely, we annotated matching keypoints on the first frame of each video and then used CoTracker3 ~\cite{Karaev:ICCV:2025:cotracker3} to predict the trajectories of these keypoints throughout the videos. Let $p_m^{t,i}$ denote the $m$-th keypoint in $F_i^t$ (\ie, the original frame at time $t$ in video $i$) as predicted by CoTracker3.

In addition, for every predicted point in every time $t$, the CoTracker3 model returns a visibility flag $b_m^{t,i}$ which indicates whether the 
$m$-th query point  is visible in frame $t$ of video $i$ or not.
Formally, $b_{m}^{\,t,i}$ is 1 if the point is visible in video $i$
at time $t$ and 0 otherwise.

Let $n_p$ be the number of query points.
We consider keypoint $m$ at time $t$ as valid only if it is visible in all \( N \) videos:
\begin{align}
M_{m,t} \;=\; \prod_{i=1}^{N} b_{m}^{\,t,i}\, . 
\end{align}
Consequently,  the number of valid keypoints in frame $t$ is 
\begin{align}
M_t \;=\; \sum_{m=1}^{n_p} M_{m,t}
\end{align}
and the number of frames that contribute to the final score is
\begin{align}
\widehat{v_l} \;=\; \bigl|\{\,t \mid M_t>0\,\}\bigr|\,.
\end{align}
 
The per-keypoint geometric variance is calculated across all input videos' frames $t$: 
\begin{align}
\operatorname{var}_{m,t}[c] \;=\; \frac{1}{N} \sum_{i=1}^{N} \left( (T^{t,i}\cdot p_{m}^{\,t,i})[c] - \frac{1}{N} \sum_{i=1}^{N}  (T^{t,i}\cdot p_{m}^{\,t,i})[c] \right)^2,  c\in\{ 1,2\}
\end{align}
where $T^{t,i}$ is the transformation that warps frame $i$ of time $t$ to the panoramic space and the index $c$ denotes the coordinate axis.
The overall geometric variance score is
\begin{align}
S_{\text{GV}} \;=\; 
\frac{1000}{\mathrm{diag}} \cdot \sqrt{ \frac{1}{\widehat{v_l}} \sum_{t:\,M_t>0} \frac{1}{M_t} \sum_{m=1}^{n_p} M_{m,t} \left( \overline{\operatorname{var}}_{m,t} \right)
}
\end{align}
where $\mathrm{diag} = \sqrt{H^2 + W^2}$ is the
length of the diagonal of the original frames shape
(where 
$HxW$ stands for the original resolution) and $\overline{\operatorname{var}}_{m,t} = \operatorname{var}_{m,t}[1] + \operatorname{var}_{m,t}[2]$. 
The multiplicative factor $\frac{1000}{\mathrm{diag}}$
is used to account for the fact that videos from different scenes, 
even within the same dataset, may have different resolutions.
As 1000 is about the average length of the diagonal of the frames,
we divide the score by the length of the diagonal 
and multiply by 1000 so each video will influence the score in a similar way, regardless of the resolution. 

Naturally, the lower the geometric variance is,
the better the result. 

Note: The geometric variance metric needs no distance threshold and evaluates all views together in a shared space. We believe it is a more informative alternative to PCK (Percentage of Correct Keypoints) for any task that aligns multiple images into a common frame. 
\clearpage

\subsection{The Multi-Camera Benchmark Evaluation Protocol}
\label{sec:evaluation_protocol}
For MIS methods, we run each method independently on every frame. When a method yields panoramas in a drifting coordinate system, we post-align the per-frame outputs to a fixed reference to correct this drift.

If a method fails to produce a panorama for any frame in a scene, we mark the entire scene as collapsed. Quantitative metrics are averaged over non-collapsed scenes only, and we additionally report the number of successfully processed scenes. Full per-scene, per-metric, and per-method results are provided in \autoref{tab:gv_mis_table}-~\ref{tab:cmsa_exts}.

GSP, GES-GSP, and OBJ-GSP require an adjacency graph over overlapping image pairs and the selection of a reference image. We construct this graph independently at each time step
$t$, and, for improved temporal stability, we consistently choose the first camera as the reference image. 

In addition, for these methods, we anchor the resulting panoramas to a shared coordinate system by estimating a similarity transformation that maps the optimized vertices of the reference mesh to its initial vertices, and then applying this same transformation to all mesh vertices of all views at time $t$. This is conceptually similar to the panorama construction used in MViS.

\clearpage

\subsection{Implementation details of the Extended MIS Variants}
\label{sec:extended_mis}
For evaluation in the video setting, we extended GSP, GES-GSP, and OBJ-GSP in two ways, as their original formulations optimize each frame independently.

\textbf{Temporal Regularization (TR).} We augment the original optimization with a temporal smoothness term that regularizes each mesh vertex in the current frame toward its optimized position in the previous frame, thereby reducing frame-to-frame jitter. Concretely, these temporal constraints are appended as additional equations to the sparse linear system, using the optimized vertices from the previous frame as references.

\textbf{Post-hoc Smoothing (PS).} 
We also consider a post-processing variant in which the original method is first applied independently to each frame, producing a per-frame set of optimized mesh vertex positions. We then treat each vertex coordinate as a 1D signal over time and apply a Gaussian filter along the temporal axis. Finally, all frames are re-rendered using the smoothed vertex trajectories, yielding a temporally smoother panoramic video.

We experiment with $\sigma\in \{0.5,1.0,1.5\}$, where $\sigma$ is the standard deviation of the Gaussian kernel in PS
and  $w\in\{0.5,1.0,1.5\}$ is the weight of the temporal regularization term in TR. The results are summarized in \autoref{tab:weights}. The main paper reports results for the setting 
$w=1.0$ and 
$\sigma=1.0$. Across all evaluated hyperparameter settings of the MIS extensions, MViS attains the best Stab. and CMSA scores.
A visual comparison between OBJ{-}GSP, OBJ{-}GSP{+}TR, OBJ{-}GSP{+}PS, and MViS is provided in \texttt{7\_mis\_variants\_vs\_mvis.mp4}.

\begin{table}[h]
\caption{
Additional hyperparameter experiments for the MIS extensions on the MMCD dataset (the Multi-Camera Benchmark). $N$ denotes the \# of videos
per scene.  
The values shown, 
 GV (\textdownarrow),
 Stab. (\textuparrow) and
 CMSA (\textdownarrow), are averaged over the scenes (on which the respective method succeeded without collapsing)
with the same $N$. 
}
\renewcommand{\arraystretch}{1.03}
\label{tab:weights}
\centering
\footnotesize
\renewcommand{\arraystretch}{1.2}
\resizebox{\linewidth}{!}{
\begin{tabular}{w{c}{0.6cm}lcccc@{\hskip 6pt}cccc@{\hskip 6pt}cccc@{\hskip 6pt}cccc}
\toprule
&\multirow{2}{*}{\normalsize Method}
& \multicolumn{4}{c}{\normalsize $N=3$}
& \multicolumn{4}{c}{\normalsize $N=4$}
& \multicolumn{4}{c}{\normalsize $N=5$}
& \multicolumn{4}{c}{\normalsize $N=6$} \\
& & {\scriptsize GV} &{\scriptsize Stab.} & {\scriptsize CMSA}&
{\scriptsize Succ.}  & {\scriptsize GV} &{\scriptsize Stab.} & {\scriptsize CMSA}
& {\scriptsize Succ.}  & {\scriptsize GV} &{\scriptsize Stab.} & {\scriptsize CMSA}
& {\scriptsize Succ.}  & {\scriptsize GV} &{\scriptsize Stab.} & {\scriptsize CMSA}  &{\scriptsize Succ.}  \\

\midrule
\multirow{3}{*}{\rotatebox[origin=c]{90}{\footnotesize $w=0.5$}}
& {GSP\texttt{+}TR} & 2.9 & .821 & 381 & \scriptsize 19/22 & 2.5 & .872 & 165 & \scriptsize 9/10 & 2.4 & .785 & 297 & \scriptsize \textbf{11/11} & 3.0 & .839 & 204 & \scriptsize 9/10 \\
& {GES-GSP\texttt{+}TR} & 3.0 & .797 & 393 & \scriptsize 19/22 & 2.6 & .857 & 170 & \scriptsize 9/10 & 2.9 & .768 & 318 & \scriptsize \textbf{11/11} & 3.3 & .824 & 204 & \scriptsize 9/10 \\
& {OBJ-GSP\texttt{+}TR} & 3.0 & .803 & 415 & \scriptsize 19/22 & 2.6 & .842 & 244 & \scriptsize 9/10 & 2.7 & .773 & 339 & \scriptsize \textbf{11/11} & 3.1 & .833 & 213 & \scriptsize 9/10 \\
\midrule
\multirow{3}{*}{\rotatebox[origin=c]{90}{\footnotesize $w=1.0$}}
& {GSP\texttt{+}TR} & 3.1 & .863 & 62.9 & \scriptsize 19/22 & 2.8 & .898 & 52.2 & \scriptsize 9/10 & 3.2 & .817 & 93.6 & \scriptsize \textbf{11/11} & 3.5 & .879 & 38.1 & \scriptsize 9/10 \\
& {GES-GSP\texttt{+}TR} & 3.3 & .849 & 69.4 & \scriptsize 19/22 & 3.1 & .884 & 57.2 & \scriptsize 9/10 & 3.8 & .800 & 100 & \scriptsize \textbf{11/11} & 3.9 & .866 & 39.5 & \scriptsize 9/10 \\
& {OBJ-GSP\texttt{+}TR} & 3.2 & .853 & 75.3 & \scriptsize 19/22 & 3.0 & .879 & 71.6 & \scriptsize 9/10 & 3.5 & .808 & 106 & \scriptsize \textbf{11/11} & 3.7 & .872 & 42.0 & \scriptsize 9/10 \\
\midrule
\multirow{3}{*}{\rotatebox[origin=c]{90}{\footnotesize $w=1.5$}}
& {GSP\texttt{+}TR} & 3.8 & .882 & 21.0 & \scriptsize 19/22 & 3.8 & .909 & 22.4 & \scriptsize 9/10 & 5.1 & .826 & 39.4 & \scriptsize \textbf{11/11} & 4.9 & .901 & 13.6 & \scriptsize 9/10 \\
& {GES-GSP\texttt{+}TR} & 4.0 & .872 & 23.7 & \scriptsize 19/22 & 4.1 & .897 & 24.9 & \scriptsize 9/10 & 5.6 & .810 & 43.0 & \scriptsize \textbf{11/11} & 5.5 & .888 & 14.1 & \scriptsize 9/10 \\
& {OBJ-GSP\texttt{+}TR} & 4.0 & .876 & 26.7 & \scriptsize 19/22 & 4.0 & .893 & 30.6 & \scriptsize 9/10 & 5.4 & .817 & 46.7 & \scriptsize \textbf{11/11} & 5.3 & .893 & 16.1 & \scriptsize 9/10 \\
\midrule
\multirow{3}{*}{\rotatebox[origin=c]{90}{\footnotesize $\sigma=0.5$}}
& {GSP\texttt{+}PS} & 2.8 & .709 & {\scriptsize >1k} & \scriptsize 19/22 & \textbf{2.4} & .722 & {\scriptsize >1k} & \scriptsize 9/10 & \textbf{2.1} & .686 & {\scriptsize >1k} & \scriptsize \textbf{11/11} & 2.9 & .746 & {\scriptsize >1k} & \scriptsize 9/10 \\
& {GES-GSP\texttt{+}PS} & 2.8 & .685 & {\scriptsize >1k} & \scriptsize 19/22 & 2.5 & .741 & {\scriptsize >1k} & \scriptsize 9/10 & 2.4 & .683 & {\scriptsize >1k} & \scriptsize 10/11 & 3.1 & .754 & {\scriptsize >1k} & \scriptsize 9/10 \\
& {OBJ-GSP\texttt{+}PS} & 2.8 & .691 & {\scriptsize >1k} & \scriptsize 19/22 & \textbf{2.4} & .709 & {\scriptsize >1k} & \scriptsize 9/10 & \textbf{2.1} & .741 & {\scriptsize >1k} & \scriptsize 8/11 & 3.0 & .761 & {\scriptsize >1k} & \scriptsize 6/10 \\
\midrule
\multirow{3}{*}{\rotatebox[origin=c]{90}{\footnotesize $\sigma=1.0$}}
& {GSP\texttt{+}PS} & \textbf{2.7} & .816 & 216 & \scriptsize 19/22 & \textbf{2.4} & .841 & 251 & \scriptsize 9/10 & \textbf{2.1} & .783 & 577 & \scriptsize \textbf{11/11} & \textbf{2.8} & .867 & 174 & \scriptsize 9/10 \\
& {GES-GSP\texttt{+}PS} & 2.8 & .802 & 206 & \scriptsize 19/22 & 2.5 & .853 & 154 & \scriptsize 9/10 & 2.3 & .796 & 644 & \scriptsize 10/11 & 3.0 & .887 & 90.7 & \scriptsize 9/10 \\
& {OBJ-GSP\texttt{+}PS} & 2.8 & .814 & 177 & \scriptsize 19/22 & \textbf{2.4} & .836 & 196 & \scriptsize 9/10 & \textbf{2.1} & .822 & 73.6 & \scriptsize 8/11 & 2.9 & .884 & 101 & \scriptsize 6/10 \\
\midrule
\multirow{3}{*}{\rotatebox[origin=c]{90}{\footnotesize $\sigma=1.5$}}
& {GSP\texttt{+}PS} & \textbf{2.7} & .851 & 43.3 & \scriptsize 19/22 & \textbf{2.4} & .880 & 51.7 & \scriptsize 9/10 & \textbf{2.1} & .811 & 89.3 & \scriptsize \textbf{11/11} & \textbf{2.8} & .898 & 35.7 & \scriptsize 9/10 \\
& {GES-GSP\texttt{+}PS} & 2.8 & .843 & 40.6 & \scriptsize 19/22 & 2.5 & .887 & 30.0 & \scriptsize 9/10 & 2.4 & .827 & 96.2 & \scriptsize 10/11 & 3.0 & .915 & 18.3 & \scriptsize 9/10 \\
& {OBJ-GSP\texttt{+}PS} & 2.8 & .855 & 34.5 & \scriptsize 19/22 & \textbf{2.4} & .879 & 37.0 & \scriptsize 9/10 & 2.2 & .844 & 14.8 & \scriptsize 8/11 & 2.9 & .907 & 18.3 & \scriptsize 6/10 \\
\midrule
& \textbf{MViS (Ours)} & 3.1 & \textbf{.917} & \textbf{2.5} & \scriptsize \textbf{22/22} & 2.8 & \textbf{.925} & \textbf{14.2} & \scriptsize \textbf{10/10} & 2.3 & \textbf{.874} & \textbf{11.0} & \scriptsize \textbf{11/11} & 3.0 & \textbf{.930} & \textbf{2.4} & \scriptsize \textbf{10/10} \\
\bottomrule
\end{tabular}
}
\end{table}

\clearpage

\subsection{Additional MMCD Statistics}

\label{sec:mmcd_stats}
\begin{figure}[h]
\centering
  \begin{tikzpicture}
\begin{axis}[
    width=0.75\textwidth,
    height=0.5\textwidth,
    ybar,
    bar width=4.5pt,
    xlabel={Video Duration (seconds)},
    ylabel={Percentage of videos in category (\%)},
    xlabel style={font=\scriptsize},
    ylabel style={font=\scriptsize},
    xtick=data,
    xticklabels={5, 6, 7, 8, 9, 10},
    legend style={
        at={(0.98,0.98)},
        anchor=north east,
        legend columns=4,
        font=\scriptsize,
        draw=gray!40,
        fill=white,
        fill opacity=0.95,
        column sep=0.08cm
    },
    ymin=0,
    ymax=50,                
    ymajorgrids,
    grid style={gray!20},
    tick label style={font=\scriptsize},
    ytick={0,10,20,30,40,50},
]
% N=3 data
\addplot[fill=blue!60!cyan!40, draw=blue!70!cyan!50] coordinates {
    (5, 4.55) (6, 13.64) (7, 13.64) (8, 13.64) (9, 27.27) (10, 27.27)
};

% N=4 data
\addplot[fill=green!70!yellow!30!gray!50, draw=green!80!yellow!40!gray!60] coordinates {
    (5, 30) (6, 30) (7, 10) (8, 20) (9, 0) (10, 10)
};

% N=5 data
\addplot[fill=yellow!60!orange!30, draw=yellow!70!orange!40] coordinates {
    (5, 9.09) (6, 27.27) (7, 36.36) (8, 18.18) (9, 9.09) (10, 0)
};

% N=6 data 
\addplot[fill={rgb,200:red,247; green,79; blue,79}, opacity=0.8,
         draw={rgb,200:red,247; green,79; blue,79}] coordinates {
    (5, 30) (6, 30) (7, 30) (8, 0) (9, 10) (10, 0)
};

\legend{$N=3$, $N=4$, $N=5$, $N=6$}
\end{axis}
\end{tikzpicture}  
\caption{MMCD video duration statistics.}
\label{fig:dataset_fig}
\end{figure}
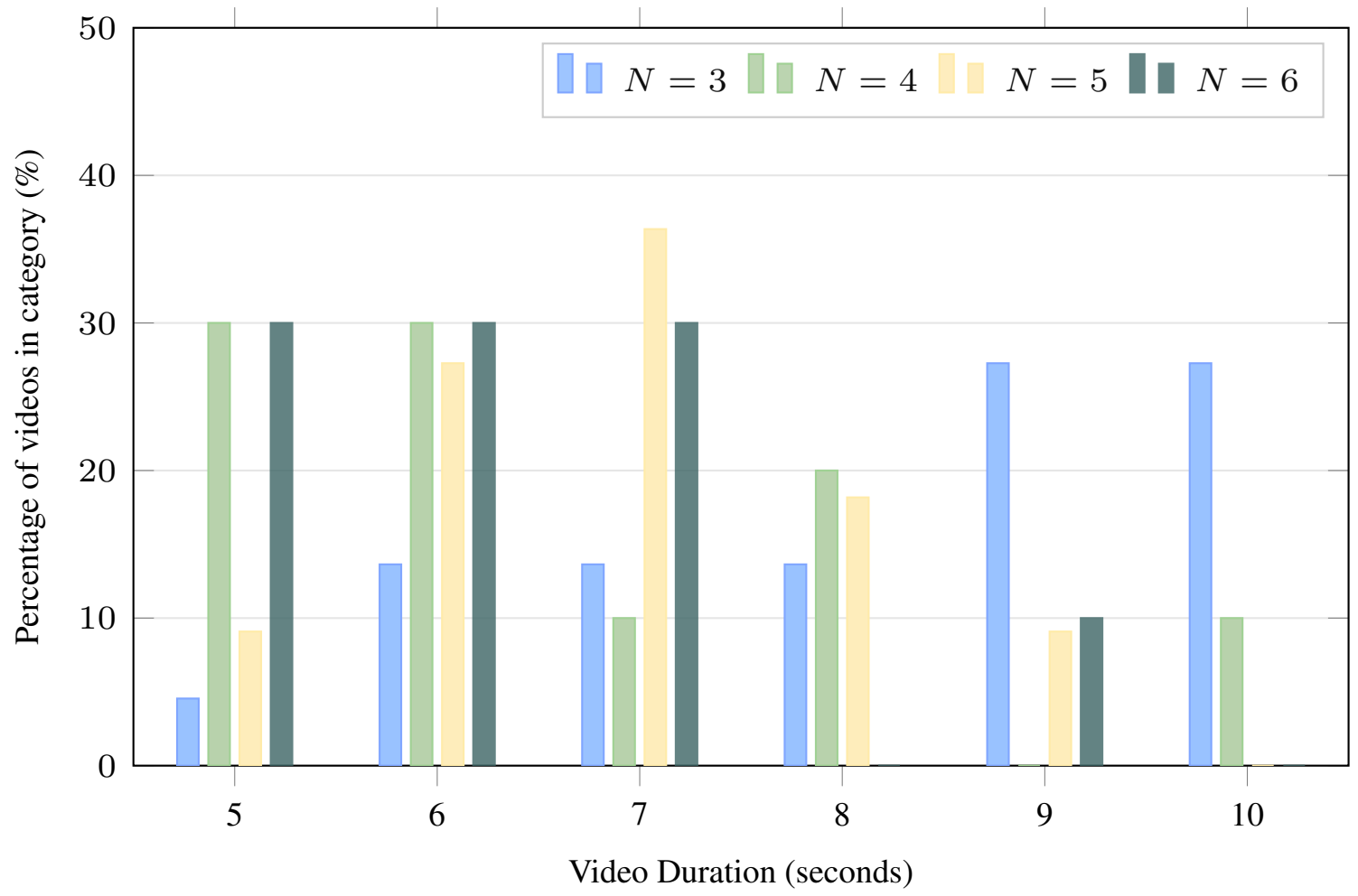

\FigureRef{fig:dataset_fig} summarizes the video lengths statistics in MMCD. \autoref{tab:overlap_ratio} shows the distribution of the 53 MMCD scenes across Overlap ratio intervals. Overlap ratio metric is defined in \autoref{sec:ort_corruption}. 

\colorlet{ovfill}{green!70!yellow!30!gray!50}
\colorlet{ovdraw}{green!80!yellow!40!gray!60}
\providecommand{\ovcol}[1]{%
  \ifnum#1>0
    \raisebox{-0.3\baselineskip}{{\setlength{\fboxsep}{0pt}\setlength{\fboxrule}{0.4pt}%
      \fcolorbox{ovdraw}{ovfill}{\rule{0pt}{\dimexpr 1.1mm*#1\relax}\hspace{4.5mm}}}}%
  \fi}

\begin{table}[ht]
  \centering
  \caption{Distribution of MMCD scenes across Overlap ratio intervals.}
  \label{tab:overlap_ratio}
  \footnotesize
  \setlength{\tabcolsep}{2.5pt}
  \begin{tabular}{l *{10}{c} c}
    \toprule
    \rule{0pt}{16.5mm}
    & \ovcol{0} & \ovcol{0} & \ovcol{3} & \ovcol{4} & \ovcol{11}
    & \ovcol{12} & \ovcol{8} & \ovcol{13} & \ovcol{2} & \ovcol{0} & \\
    \cline{2-11}
    \addlinespace[2pt]
    Overlap ratio & 0.0--0.1 & 0.1--0.2 & 0.2--0.3 & 0.3--0.4 & 0.4--0.5
    & 0.5--0.6 & 0.6--0.7 & 0.7--0.8 & 0.8--0.9 & 0.9--1.0 & Total \\
    \midrule
    \# of scenes & 0 & 0 & 3 & 4 & 11 & 12 & 8 & 13 & 2 & 0 & 53 \\
    Percentage   & 0.0\% & 0.0\% & 5.7\% & 7.5\% & 20.8\% & 22.6\% & 15.1\% & 24.5\% & 3.8\% & 0.0\% & 100\% \\
    \bottomrule
  \end{tabular}
\end{table}

\clearpage

\subsection{Full Multi-Camera Benchmark Results}
\label{sec:benchmark}
We provide here the complete per-scene benchmark results. In the main paper, the reported averages were computed only over the scenes for which each method successfully generated a panorama. In the tables below, \xmark { }denotes the method collapsed on the scene, while N/A indicates the method does not support that number of cameras.

\begin{table}[h]
\centering
\caption{Per-scene GV($\downarrow$), the Multi-Camera Benchmark.}
\label{tab:gv_mis_table}
\scriptsize
\renewcommand{\arraystretch}{0.95}
\begin{tabularx}{\linewidth}{c l *{7}{Y}}
\toprule
& Data  
& AutoStitch 
& APAP 
& GSP 
& SPW 
& GES{-}GSP 
& OBJ{-}GSP
& MViS(Ours) \\
\midrule
\multirow{22}{*}{\rotatebox{90}{$N=3$}}
& scene\_$1$ & \xmark & 1.7 & 1.7 & 1.8 & 1.7 & 1.7 & 2.0 \\
& scene\_$2$ & 2.9 & 2.0 & 1.9 & 2.2 & 1.9 & 1.9 & 3.0 \\
& scene\_$4$ & 11.4 & 7.0 & 7.1 & 7.5 & 7.1 & 7.1 & 7.8 \\
& scene\_$6$ & 1.6 & 1.5 & 1.5 & 1.5 & 1.5 & 1.5 & 1.5 \\
& scene\_$7$ & \xmark & 12.1 & 10.6 & 10.5 & 11.1 & 11.2 & 13.0 \\
& scene\_$8$ & 3.3 & 2.8 & 2.8 & 2.8 & 2.8 & 2.8 & 2.8 \\
& scene\_$9$ & \xmark & \xmark & \xmark & 3.1 & \xmark & \xmark & 3.1 \\
& scene\_$10$ & 3.5 & 3.1 & 2.4 & 2.9 & 2.4 & 2.3 & 2.3 \\
& scene\_$12$ & 1.4 & 1.4 & 1.8 & 1.6 & 1.8 & 1.8 & 1.6 \\
& scene\_$13$ & 3.0 & 2.6 & 4.2 & 3.0 & 4.2 & 4.2 & 3.3 \\
& scene\_$15$ & 3.9 & 2.4 & 2.3 & 2.8 & 2.3 & 2.4 & 3.0 \\
& scene\_$18$ & 3.5 & 2.4 & 2.4 & 2.6 & 2.4 & 2.4 & 2.7 \\
& scene\_$19$ & 1.2 & 1.2 & 1.1 & 1.2 & 1.2 & 1.2 & 1.2 \\
& scene\_$20$ & 2.8 & 1.7 & 1.5 & 2.0 & 1.5 & 1.5 & 2.0 \\
& scene\_$21$ & 1.6 & 1.6 & 1.6 & 1.7 & 1.6 & 1.6 & 1.7 \\
& scene\_$29$ & \xmark & \xmark & \xmark & 28.9 & \xmark & \xmark & 3.6 \\
& scene\_$39$ & \xmark & 3.8 & 2.9 & 13.1 & 2.9 & 3.0 & 3.1 \\
& scene\_$67$ & \xmark & 1.6 & 1.6 & 1.4 & 1.9 & 2.0 & 1.6 \\
& scene\_$69$ & \xmark & 2.2 & 2.1 & 2.1 & 2.1 & 2.1 & 2.9 \\
& scene\_$70$ & 1.8 & 1.9 & 1.9 & 2.2 & 2.1 & 1.9 & 1.9 \\
& scene\_$106$ & \xmark & \xmark & \xmark & 8.0 & \xmark & \xmark & 2.1 \\
& scene\_$110$ & 4.2 & 2.8 & 1.7 & 2.2 & 2.0 & 1.8 & 1.8 \\
\midrule
\multirow{10}{*}{\rotatebox{90}{$N=4$}}
& scene\_$25$ & \xmark & 4.8 & 4.4 & 4.2 & 4.4 & 4.4 & 3.9 \\
& scene\_$30$ & \xmark & 3.2 & 3.0 & 26.5 & 3.0 & 3.0 & 3.1 \\
& scene\_$31$ & 3.9 & 4.2 & 3.6 & 4.0 & 3.6 & 3.6 & 3.4 \\
& scene\_$35$ & 3.5 & 3.5 & 3.5 & 3.5 & 3.5 & 3.5 & 3.5 \\
& scene\_$47$ & 2.1 & 2.7 & 2.1 & 2.4 & 2.1 & 2.1 & 2.2 \\
& scene\_$54$ & 1.6 & 1.8 & 1.4 & 4.8 & 1.5 & 1.5 & 1.6 \\
& scene\_$64$ & 1.7 & 2.5 & 1.7 & 5.5 & 1.8 & 1.8 & 1.7 \\
& scene\_$49$ & 1.4 & 1.3 & 1.2 & 1.5 & 1.3 & 1.3 & 1.4 \\
& scene\_$92$ & \xmark & 20.5 & \xmark & 30.7 & \xmark & \xmark & 6.3 \\
& scene\_$81$ & \xmark & \xmark & 1.2 & 20.7 & 1.7 & 1.3 & 1.2 \\
\midrule
\multirow{11}{*}{\rotatebox{90}{$N=5$}}
& scene\_$33$ & 2.5 & 2.9 & 2.6 & 2.7 & 2.6 & 2.5 & 2.6 \\
& scene\_$46$ & 2.7 & 3.0 & 1.7 & 2.3 & 3.2 & \xmark & 1.6 \\
& scene\_$51$ & 5.4 & 2.8 & 2.6 & 3.5 & 2.6 & 2.6 & 3.1 \\
& scene\_$62$ & 4.8 & 2.1 & 1.9 & 2.2 & 2.1 & 2.1 & 2.0 \\
& scene\_$63$ & 4.0 & 2.0 & 1.9 & 2.0 & 1.9 & 1.9 & 2.1 \\
& scene\_$65$ & 4.1 & 2.9 & 3.3 & 3.2 & 3.3 & 3.3 & 3.2 \\
& scene\_$66$ & \xmark & 2.4 & 2.3 & 2.4 & 2.3 & 2.3 & 2.4 \\
& scene\_$72$ & 1.8 & 2.1 & 2.0 & 2.5 & 2.2 & \xmark & 2.3 \\
& scene\_$76$ & \xmark & 3.2 & 2.8 & 3.5 & 2.9 & \xmark & 3.5 \\
& scene\_$77$ & 1.2 & 1.2 & 1.1 & 1.2 & 1.3 & 1.2 & 1.1 \\
& scene\_$98$ & 1.4 & 1.2 & 1.0 & \xmark & \xmark & 1.3 & 1.3 \\
\midrule
\multirow{10}{*}{\rotatebox{90}{$N=6$}}
& scene\_$27$ & 4.7 & 3.6 & 3.4 & 3.4 & 3.4 & 3.3 & 3.6 \\
& scene\_$50$ & \xmark & \xmark & 2.9 & 51.6 & 3.3 & \xmark & 3.4 \\
& scene\_$52$ & 2.4 & 2.2 & 2.0 & 2.3 & 2.2 & 2.1 & 2.4 \\
& scene\_$53$ & \xmark & 4.7 & 4.6 & 4.7 & 4.6 & 4.7 & 4.8 \\
& scene\_$55$ & \xmark & 11.1 & 3.7 & 26.4 & 3.9 & 3.8 & 4.0 \\
& scene\_$57$ & 5.5 & 2.4 & 2.2 & 2.4 & 2.6 & \xmark & 2.4 \\
& scene\_$58$ & 2.0 & 2.1 & 2.5 & 3.5 & 2.5 & 2.4 & 2.5 \\
& scene\_$59$ & 2.1 & 2.1 & 1.9 & 2.2 & 1.9 & 2.1 & 2.1 \\
& scene\_$60$ & \xmark & 3.8 & 3.6 & 4.0 & 4.0 & \xmark & 3.8 \\
& scene\_$101$ & \xmark & \xmark & \xmark & 47.2 & \xmark & \xmark & 1.4 \\
\bottomrule
\end{tabularx}
\end{table}

\begin{table}[t]
\centering
\caption{Per-scene Stability score ($\uparrow$), proposed by \cite{Nie:TIP:2017:dynamic}, the Multi-Camera Benchmark.}
\label{tab:stab_mis_table}
\scriptsize
\renewcommand{\arraystretch}{1.1}
\begin{tabularx}{\linewidth}{c l *{7}{Y}}
\toprule
& Data  
& AutoStitch 
& APAP 
& GSP 
& SPW 
& GES{-}GSP 
& OBJ{-}GSP
& MViS(Ours) \\
\midrule
\multirow{22}{*}{\rotatebox{90}{$N=3$}}
& scene\_$1$ & \xmark & .889 & .736 & .889 & .717 & .730 & .910 \\
& scene\_$2$ & .870 & .901 & .855 & .889 & .704 & .643 & .911 \\
& scene\_$4$ & .641 & .898 & .726 & .857 & .725 & .615 & .926 \\
& scene\_$6$ & .937 & .926 & .775 & .928 & .772 & .712 & .932 \\
& scene\_$7$ & \xmark & .709 & .165 & .646 & .211 & .214 & .938 \\
& scene\_$8$ & .906 & .946 & .690 & .944 & .731 & .635 & .964 \\
& scene\_$9$ & \xmark & \xmark & \xmark & .944 & \xmark & \xmark & .974 \\
& scene\_$10$ & .801 & .772 & .671 & .797 & .665 & .676 & .958 \\
& scene\_$12$ & .747 & .708 & .710 & .648 & .771 & .742 & .756 \\
& scene\_$13$ & .905 & .823 & .520 & .791 & .624 & .661 & .910 \\
& scene\_$15$ & .853 & .905 & .741 & .911 & .684 & .710 & .948 \\
& scene\_$18$ & .850 & .877 & .756 & .790 & .514 & .605 & .921 \\
& scene\_$19$ & .888 & .887 & .830 & .871 & .703 & .704 & .910 \\
& scene\_$20$ & .845 & .878 & .717 & .833 & .646 & .651 & .913 \\
& scene\_$21$ & .959 & .951 & .794 & .949 & .775 & .825 & .961 \\
& scene\_$29$ & \xmark & \xmark & \xmark & .571 & \xmark & \xmark & .944 \\
& scene\_$39$ & \xmark & .812 & .677 & .616 & .676 & .750 & .934 \\
& scene\_$67$ & \xmark & .917 & .573 & .908 & .571 & .570 & .923 \\
& scene\_$69$ & \xmark & .856 & .706 & .855 & .732 & .708 & .870 \\
& scene\_$70$ & .838 & .789 & .260 & .689 & .317 & .328 & .850 \\
& scene\_$106$ & \xmark & \xmark & \xmark & .728 & \xmark & \xmark & .966 \\
& scene\_$110$ & .812 & .711 & .401 & .708 & .202 & .325 & .852 \\
\midrule
\multirow{10}{*}{\rotatebox{90}{$N=4$}}
& scene\_$25$ & \xmark & .881 & .587 & .880 & .597 & .504 & .945 \\
& scene\_$30$ & \xmark & .845 & .688 & .349 & .683 & .597 & .913 \\
& scene\_$31$ & .838 & .720 & .670 & .708 & .669 & .623 & .921 \\
& scene\_$35$ & .910 & .907 & .831 & .909 & .824 & .733 & .915 \\
& scene\_$47$ & .907 & .843 & .718 & .791 & .680 & .616 & .927 \\
& scene\_$54$ & .933 & .893 & .716 & .822 & .741 & .786 & .950 \\
& scene\_$64$ & .912 & .811 & .331 & .702 & .418 & .393 & .921 \\
& scene\_$49$ & .903 & .864 & .806 & .754 & .809 & .884 & .974 \\
& scene\_$92$ & \xmark & .771 & \xmark & .517 & \xmark & \xmark & .921 \\
& scene\_$81$ & \xmark & \xmark & .496 & .729 & .590 & .572 & .861 \\
\midrule
\multirow{11}{*}{\rotatebox{90}{$N=5$}}
& scene\_$33$ & .901 & .896 & .648 & .909 & .554 & .656 & .927 \\
& scene\_$46$ & .736 & .742 & .327 & .774 & .393 & \xmark & .823 \\
& scene\_$51$ & .872 & .920 & .697 & .865 & .644 & .588 & .939 \\
& scene\_$62$ & .903 & .857 & .531 & .788 & .565 & .659 & .927 \\
& scene\_$63$ & .701 & .769 & .686 & .744 & .716 & .761 & .790 \\
& scene\_$65$ & .906 & .892 & .751 & .896 & .722 & .821 & .905 \\
& scene\_$66$ & \xmark & .849 & .739 & .821 & .752 & .745 & .853 \\
& scene\_$72$ & .833 & .820 & .610 & .801 & .620 & \xmark & .842 \\
& scene\_$76$ & \xmark & .908 & .528 & .878 & .500 & \xmark & .952 \\
& scene\_$77$ & .886 & .885 & .689 & .877 & .670 & .653 & .887 \\
& scene\_$98$ & .770 & .764 & .667 & \xmark & \xmark & .584 & .772 \\
\midrule
\multirow{10}{*}{\rotatebox{90}{$N=6$}}
& scene\_$27$ & .961 & .978 & .866 & .977 & .843 & .884 & .979 \\
& scene\_$50$ & \xmark & \xmark & .753 & .440 & .846 & \xmark & .977 \\
& scene\_$52$ & .886 & .881 & .837 & .869 & .802 & .781 & .892 \\
& scene\_$53$ & \xmark & .810 & .695 & .772 & .543 & .599 & .917 \\
& scene\_$55$ & \xmark & .643 & .726 & .327 & .709 & .741 & .908 \\
& scene\_$57$ & .971 & .972 & .743 & .913 & .772 & \xmark & .985 \\
& scene\_$58$ & .877 & .833 & .426 & .799 & .450 & .504 & .882 \\
& scene\_$59$ & .923 & .909 & .513 & .836 & .523 & .580 & .935 \\
& scene\_$60$ & \xmark & .972 & .489 & .882 & .541 & \xmark & .998 \\
& scene\_$101$ & \xmark & \xmark & \xmark & .632 & \xmark & \xmark & .825 \\
\bottomrule
\end{tabularx}

\end{table}

\begin{table}[t]
\centering
\caption{Per-scene CMSA ($\downarrow$), the Multi-Camera Benchmark.}
\label{tab:cmsa_mis_tab}
\scriptsize
\renewcommand{\arraystretch}{1.1}
\begin{tabularx}{\linewidth}{c l *{7}{Y}}
\toprule
& Data  
& AutoStitch 
& APAP 
& GSP 
& SPW 
& GES{-}GSP 
& OBJ{-}GSP
& MViS(Ours) \\
\midrule
\multirow{22}{*}{\rotatebox{90}{$N=3$}}
& scene\_$1$ & \xmark & 109 & {\scriptsize >1k} & 148 & {\scriptsize >1k} & {\scriptsize >1k} & 0.1 \\
& scene\_$2$ & 425 & 109 & 531 & {\scriptsize >1k} & 507 & {\scriptsize >1k} & 0.2 \\
& scene\_$4$ & {\scriptsize >1k} & 177 & {\scriptsize >1k} & {\scriptsize >1k} & {\scriptsize >1k} & {\scriptsize >1k} & 0.1 \\
& scene\_$6$ & 3.8 & 97.8 & {\scriptsize >1k} & 147 & {\scriptsize >1k} & {\scriptsize >1k} & 0.1 \\
& scene\_$7$ & \xmark & 728 & {\scriptsize >1k} & {\scriptsize >1k} & {\scriptsize >1k} & {\scriptsize >1k} & 0.1 \\
& scene\_$8$ & 13.2 & 13.7 & {\scriptsize >1k} & {\scriptsize >1k} & {\scriptsize >1k} & {\scriptsize >1k} & 0.0 \\
& scene\_$9$ & \xmark & \xmark & \xmark & {\scriptsize >1k} & \xmark & \xmark & 1.3 \\
& scene\_$10$ & {\scriptsize >1k} & {\scriptsize >1k} & {\scriptsize >1k} & 426 & 756 & 751 & 0.5 \\
& scene\_$12$ & 1.4 & 102 & {\scriptsize >1k} & 360 & 822 & 954 & 0.4 \\
& scene\_$13$ & 2.8 & 113 & {\scriptsize >1k} & 204 & {\scriptsize >1k} & 755 & 0.5 \\
& scene\_$15$ & 151 & 112 & 580 & {\scriptsize >1k} & 561 & 303 & 0.1 \\
& scene\_$18$ & 142 & 58.4 & 288 & 358 & {\scriptsize >1k} & 492 & 0.1 \\
& scene\_$19$ & 14.6 & 13.5 & 307 & 561 & 287 & {\scriptsize >1k} & 0.1 \\
& scene\_$20$ & 21.4 & 339 & 318 & 146 & 320 & 495 & 0.1 \\
& scene\_$21$ & 1.4 & 8.9 & 845 & 23.7 & 846 & 506 & 0.3 \\
& scene\_$29$ & \xmark & \xmark & \xmark & {\scriptsize >1k} & \xmark & \xmark & 15.5 \\
& scene\_$39$ & \xmark & {\scriptsize >1k} & {\scriptsize >1k} & {\scriptsize >1k} & {\scriptsize >1k} & 586 & 20.0 \\
& scene\_$67$ & \xmark & {\scriptsize >1k} & {\scriptsize >1k} & {\scriptsize >1k} & {\scriptsize >1k} & {\scriptsize >1k} & 2.6 \\
& scene\_$69$ & \xmark & 565 & {\scriptsize >1k} & 281 & {\scriptsize >1k} & {\scriptsize >1k} & 7.2 \\
& scene\_$70$ & 70.0 & {\scriptsize >1k} & {\scriptsize >1k} & {\scriptsize >1k} & {\scriptsize >1k} & {\scriptsize >1k} & 3.8 \\
& scene\_$106$ & \xmark & \xmark & \xmark & {\scriptsize >1k} & \xmark & \xmark & 1.4 \\
& scene\_$110$ & 31.8 & {\scriptsize >1k} & {\scriptsize >1k} & 571 & {\scriptsize >1k} & {\scriptsize >1k} & 0.1 \\
\midrule
\multirow{10}{*}{\rotatebox{90}{$N=4$}}
& scene\_$25$ & \xmark & {\scriptsize >1k} & {\scriptsize >1k} & {\scriptsize >1k} & {\scriptsize >1k} & {\scriptsize >1k} & 13.6 \\
& scene\_$30$ & \xmark & {\scriptsize >1k} & {\scriptsize >1k} & {\scriptsize >1k} & {\scriptsize >1k} & {\scriptsize >1k} & 5.4 \\
& scene\_$31$ & {\scriptsize >1k} & {\scriptsize >1k} & {\scriptsize >1k} & {\scriptsize >1k} & {\scriptsize >1k} & {\scriptsize >1k} & 38.7 \\
& scene\_$35$ & 17.5 & 48.0 & 267 & 45.9 & 277 & {\scriptsize >1k} & 4.4 \\
& scene\_$47$ & 41.6 & {\scriptsize >1k} & 758 & {\scriptsize >1k} & 913 & {\scriptsize >1k} & 1.1 \\
& scene\_$54$ & 71.8 & 417 & {\scriptsize >1k} & {\scriptsize >1k} & {\scriptsize >1k} & 865 & 12.2 \\
& scene\_$64$ & 80.5 & {\scriptsize >1k} & {\scriptsize >1k} & {\scriptsize >1k} & {\scriptsize >1k} & {\scriptsize >1k} & 9.9 \\
& scene\_$49$ & 17.1 & 307 & {\scriptsize >1k} & 564 & {\scriptsize >1k} & 750 & 0.2 \\
& scene\_$92$ & \xmark & {\scriptsize >1k} & \xmark & {\scriptsize >1k} & \xmark & \xmark & 18.0 \\
& scene\_$81$ & \xmark & \xmark & {\scriptsize >1k} & {\scriptsize >1k} & {\scriptsize >1k} & {\scriptsize >1k} & 38.0 \\
\midrule
\multirow{11}{*}{\rotatebox{90}{$N=5$}}
& scene\_$33$ & 80.2 & 110 & {\scriptsize >1k} & 107 & {\scriptsize >1k} & {\scriptsize >1k} & 8.2 \\
& scene\_$46$ & {\scriptsize >1k} & {\scriptsize >1k} & {\scriptsize >1k} & {\scriptsize >1k} & {\scriptsize >1k} & \xmark & 60.9 \\
& scene\_$51$ & 545 & 571 & {\scriptsize >1k} & {\scriptsize >1k} & {\scriptsize >1k} & {\scriptsize >1k} & 0.3 \\
& scene\_$62$ & 147 & 676 & {\scriptsize >1k} & {\scriptsize >1k} & {\scriptsize >1k} & {\scriptsize >1k} & 2.5 \\
& scene\_$63$ & 135 & 279 & {\scriptsize >1k} & 207 & 982 & 739 & 1.0 \\
& scene\_$65$ & 40.1 & 109 & {\scriptsize >1k} & 107 & {\scriptsize >1k} & {\scriptsize >1k} & 0.3 \\
& scene\_$66$ & \xmark & 590 & 751 & {\scriptsize >1k} & 602 & 562 & 0.5 \\
& scene\_$72$ & 20.9 & 537 & {\scriptsize >1k} & {\scriptsize >1k} & {\scriptsize >1k} & \xmark & 6.8 \\
& scene\_$76$ & \xmark & 735 & {\scriptsize >1k} & {\scriptsize >1k} & {\scriptsize >1k} & \xmark & 13.8 \\
& scene\_$77$ & 33.7 & 224 & {\scriptsize >1k} & 853 & {\scriptsize >1k} & {\scriptsize >1k} & 26.4 \\
& scene\_$98$ & 7.5 & 107 & 909 & \xmark & \xmark & {\scriptsize >1k} & 0.6 \\
\midrule
\multirow{10}{*}{\rotatebox{90}{$N=6$}}
& scene\_$27$ & 56.0 & 10.2 & {\scriptsize >1k} & 77.8 & {\scriptsize >1k} & {\scriptsize >1k} & 0.2 \\
& scene\_$50$ & \xmark & \xmark & {\scriptsize >1k} & {\scriptsize >1k} & {\scriptsize >1k} & \xmark & 1.4 \\
& scene\_$52$ & 26.1 & 69.1 & 255 & 295 & 202 & {\scriptsize >1k} & 0.1 \\
& scene\_$53$ & \xmark & 266 & 542 & {\scriptsize >1k} & {\scriptsize >1k} & 576 & 0.4 \\
& scene\_$55$ & \xmark & {\scriptsize >1k} & {\scriptsize >1k} & {\scriptsize >1k} & {\scriptsize >1k} & {\scriptsize >1k} & 0.7 \\
& scene\_$57$ & 362 & 285 & {\scriptsize >1k} & {\scriptsize >1k} & {\scriptsize >1k} & \xmark & 0.5 \\
& scene\_$58$ & 15.5 & {\scriptsize >1k} & {\scriptsize >1k} & {\scriptsize >1k} & {\scriptsize >1k} & {\scriptsize >1k} & 3.7 \\
& scene\_$59$ & 93.9 & 448 & {\scriptsize >1k} & 664 & {\scriptsize >1k} & {\scriptsize >1k} & 0.8 \\
& scene\_$60$ & \xmark & {\scriptsize >1k} & {\scriptsize >1k} & {\scriptsize >1k} & {\scriptsize >1k} & \xmark & 15.4 \\
& scene\_$101$ & \xmark & \xmark & \xmark & {\scriptsize >1k} & \xmark & \xmark & 1.2 \\
\bottomrule
\end{tabularx}

\end{table}

\begin{table}[t]
\centering
\caption{Per-scene GV($\downarrow$) comparison on the Multi-Camera Benchmark, \textbf{Continued.}}
\label{tab:gv_exts}
\scriptsize
\begin{tabularx}{\columnwidth}{c l *{8}{Y}}
\toprule
& \multirow{2}{*}{\scriptsize Data}
& \multicolumn{3}{c}{\scriptsize MIS{+}TR}
& \multicolumn{3}{c}{\scriptsize MIS{+}PS}
& & \\

\cmidrule(lr){3-5} \cmidrule(lr){6-8}

&
& {\scriptsize GSP}
& {\scriptsize GES{-}GSP}
& {\scriptsize OBJ{-}GSP}
& {\scriptsize GSP}
& {\scriptsize GES{-}GSP}
& {\scriptsize OBJ{-}GSP}
& \raisebox{0pt}[0pt][0pt]{\shortstack{\scriptsize Stab\\Stitch++}}
& \raisebox{0pt}[0pt][0pt]{\shortstack{\scriptsize MViS\\(Ours)}}\\
\midrule
\multirow{22}{*}{\rotatebox{90}{$N=3$}}
& scene\_$1$ & 2.0 & 2.0 & 2.0 & 1.7 & 1.7 & 1.7 & 6.7 & 2.0 \\
& scene\_$2$ & 2.3 & 2.3 & 2.3 & 1.9 & 1.9 & 1.9 & 5.1 & 3.0 \\
& scene\_$4$ & 6.7 & 6.8 & 6.8 & 7.1 & 7.1 & 7.1 & 9.1 & 7.8 \\
& scene\_$6$ & 1.7 & 1.7 & 1.7 & 1.5 & 1.5 & 1.5 & 5.1 & 1.5 \\
& scene\_$7$ & 10.2 & 10.7 & 10.9 & 10.3 & 10.9 & 11.0 & 12.7 & 13.0 \\
& scene\_$8$ & 3.0 & 3.0 & 3.1 & 2.8 & 2.8 & 2.8 & 4.6 & 2.8 \\
& scene\_$9$ & \xmark & \xmark & \xmark & \xmark & \xmark & \xmark & 9.8 & 3.1 \\
& scene\_$10$ & 2.2 & 2.1 & 2.1 & 2.0 & 2.0 & 2.0 & 3.1 & 2.3 \\
& scene\_$12$ & 1.9 & 1.9 & 1.9 & 1.7 & 1.8 & 1.7 & 3.3 & 1.6 \\
& scene\_$13$ & 4.3 & 4.3 & 4.3 & 4.1 & 4.0 & 4.0 & 4.8 & 3.3 \\
& scene\_$15$ & 2.3 & 2.3 & 2.6 & 2.3 & 2.3 & 2.4 & 3.8 & 3.0 \\
& scene\_$18$ & 2.4 & 2.6 & 2.4 & 2.3 & 2.4 & 2.4 & 4.1 & 2.7 \\
& scene\_$19$ & 1.6 & 1.9 & 2.0 & 1.1 & 1.1 & 1.1 & 15.9 & 1.2 \\
& scene\_$20$ & 1.6 & 1.7 & 1.6 & 1.4 & 1.4 & 1.4 & 5.6 & 2.0 \\
& scene\_$21$ & 1.9 & 1.9 & 1.9 & 1.6 & 1.6 & 1.6 & 2.2 & 1.7 \\
& scene\_$29$ & \xmark & \xmark & \xmark & \xmark & \xmark & \xmark & 118 & 3.6 \\
& scene\_$39$ & 3.1 & 3.1 & 3.2 & 3.0 & 3.0 & 3.0 & 32.7 & 3.1 \\
& scene\_$67$ & 4.5 & 5.1 & 4.9 & 1.5 & 1.7 & 1.7 & 134 & 1.6 \\
& scene\_$69$ & 3.2 & 3.3 & 3.2 & 2.1 & 2.1 & 2.1 & 36.0 & 2.9 \\
& scene\_$70$ & 2.6 & 3.0 & 2.6 & 1.8 & 1.9 & 1.8 & 25.5 & 1.9 \\
& scene\_$106$ & \xmark & \xmark & \xmark & \xmark & \xmark & \xmark & 166 & 2.1 \\
& scene\_$110$ & 1.7 & 2.3 & 1.8 & 1.6 & 1.7 & 1.6 & 4.4 & 1.8 \\
\midrule
\multirow{10}{*}{\rotatebox{90}{$N=4$}}
& scene\_$25$ & 4.5 & 4.8 & 4.8 & 4.2 & 4.2 & 4.2 & N/A & 3.9 \\
& scene\_$30$ & 3.5 & 3.5 & 3.7 & 3.0 & 3.0 & 3.0 & N/A & 3.1 \\
& scene\_$31$ & 4.1 & 4.1 & 4.0 & 3.5 & 3.5 & 3.5 & N/A & 3.4 \\
& scene\_$35$ & 3.6 & 3.6 & 3.6 & 3.5 & 3.5 & 3.5 & N/A & 3.5 \\
& scene\_$47$ & 2.2 & 2.2 & 2.2 & 2.1 & 2.1 & 2.0 & N/A & 2.2 \\
& scene\_$54$ & 1.7 & 1.7 & 1.8 & 1.5 & 1.5 & 1.5 & N/A & 1.6 \\
& scene\_$64$ & 2.2 & 2.2 & 2.2 & 1.6 & 1.7 & 1.7 & N/A & 1.7 \\
& scene\_$49$ & 1.5 & 1.7 & 1.8 & 1.1 & 1.3 & 1.3 & N/A & 1.4 \\
& scene\_$92$ & \xmark & \xmark & \xmark & \xmark & \xmark & \xmark & N/A & 6.3 \\
& scene\_$81$ & 2.0 & 3.8 & 2.7 & 1.1 & 1.6 & 1.2 & N/A & 1.2 \\
\midrule
\multirow{11}{*}{\rotatebox{90}{$N=5$}}
& scene\_$33$ & 3.2 & 3.2 & 3.2 & 2.7 & 2.6 & 2.6 & N/A & 2.6 \\
& scene\_$46$ & 4.4 & 8.6 & 5.8 & 1.5 & 2.4 & \xmark & N/A & 1.6 \\
& scene\_$51$ & 2.9 & 2.9 & 3.0 & 2.5 & 2.5 & 2.5 & N/A & 3.1 \\
& scene\_$62$ & 2.3 & 2.8 & 2.6 & 1.8 & 2.0 & 2.0 & N/A & 2.0 \\
& scene\_$63$ & 2.2 & 2.2 & 2.1 & 1.9 & 1.9 & 1.9 & N/A & 2.1 \\
& scene\_$65$ & 4.1 & 4.0 & 4.2 & 3.3 & 3.3 & 3.4 & N/A & 3.2 \\
& scene\_$66$ & 3.1 & 3.1 & 3.2 & 2.3 & 2.3 & 2.3 & N/A & 2.4 \\
& scene\_$72$ & 3.9 & 4.5 & 4.4 & 1.9 & 2.0 & \xmark & N/A & 2.3 \\
& scene\_$76$ & 3.7 & 4.2 & 4.0 & 2.8 & 2.8 & \xmark & N/A & 3.5 \\
& scene\_$77$ & 2.3 & 2.6 & 2.8 & 1.2 & 1.3 & 1.3 & N/A & 1.1 \\
& scene\_$98$ & 3.4 & 3.9 & 3.9 & 1.0 & \xmark & 1.1 & N/A & 1.3 \\
\midrule
\multirow{10}{*}{\rotatebox{90}{$N=6$}}
& scene\_$27$ & 4.2 & 4.1 & 4.2 & 3.4 & 3.4 & 3.4 & N/A & 3.6 \\
& scene\_$50$ & 3.9 & 4.6 & 4.0 & 2.8 & 3.2 & \xmark & N/A & 3.4 \\
& scene\_$52$ & 2.3 & 2.4 & 2.3 & 2.0 & 2.1 & 2.1 & N/A & 2.4 \\
& scene\_$53$ & 4.5 & 4.5 & 4.5 & 4.4 & 4.4 & 4.5 & N/A & 4.8 \\
& scene\_$55$ & 4.6 & 4.6 & 4.6 & 3.5 & 3.7 & 3.6 & N/A & 4.0 \\
& scene\_$57$ & 3.3 & 4.3 & 3.9 & 2.2 & 2.5 & \xmark & N/A & 2.4 \\
& scene\_$58$ & 3.0 & 3.0 & 3.0 & 1.9 & 1.9 & 1.9 & N/A & 2.5 \\
& scene\_$59$ & 2.1 & 2.3 & 2.5 & 1.8 & 1.9 & 2.0 & N/A & 2.1 \\
& scene\_$60$ & 3.9 & 5.3 & 4.6 & 3.5 & 3.9 & \xmark & N/A & 3.8 \\
& scene\_$101$ & \xmark & \xmark & \xmark & \xmark & \xmark & \xmark & N/A & 1.4 \\
\bottomrule
\end{tabularx}

\end{table}
\begin{table}[t]
\centering
\caption{Per-scene Stab.($\uparrow$) comparison on the Multi-Camera Benchmark, \textbf{Continued.}}
\label{tab:stab_exts}
\scriptsize
\begin{tabularx}{\columnwidth}{c l *{8}{Y}}

\toprule
& \multirow{2}{*}{\scriptsize Data} 
& \multicolumn{3}{c}{\scriptsize MIS{+}TR} 
& \multicolumn{3}{c}{\scriptsize MIS{+}PS} 
& & \\

\cmidrule(lr){3-5} \cmidrule(lr){6-8}

&
& {\scriptsize GSP}
& {\scriptsize GES{-}GSP}
& {\scriptsize OBJ{-}GSP}
& {\scriptsize GSP}
& {\scriptsize GES{-}GSP}
& {\scriptsize OBJ{-}GSP}
& \raisebox{0pt}[0pt][0pt]{\shortstack{\scriptsize Stab\\Stitch++}}
& \raisebox{0pt}[0pt][0pt]{\shortstack{\scriptsize MViS\\(Ours)}}\\
\midrule
\multirow{22}{*}{\rotatebox{90}{$N=3$}}
& scene\_$1$ & .920 & .919 & .916 & .852 & .851 & .873 & .870 & .910 \\
& scene\_$2$ & .904 & .889 & .873 & .903 & .878 & .840 & .792 & .911 \\
& scene\_$4$ & .931 & .921 & .925 & .876 & .899 & .853 & .854 & .926 \\
& scene\_$6$ & .954 & .953 & .951 & .904 & .911 & .875 & .858 & .932 \\
& scene\_$7$ & .638 & .621 & .621 & .435 & .494 & .501 & .881 & .938 \\
& scene\_$8$ & .914 & .894 & .892 & .839 & .879 & .827 & .873 & .964 \\
& scene\_$9$ & \xmark & \xmark & \xmark & \xmark & \xmark & \xmark & .805 & .974 \\
& scene\_$10$ & .868 & .868 & .866 & .886 & .892 & .893 & .932 & .958 \\
& scene\_$12$ & .879 & .876 & .871 & .870 & .873 & .869 & .814 & .756 \\
& scene\_$13$ & .829 & .830 & .827 & .797 & .839 & .848 & .826 & .910 \\
& scene\_$15$ & .933 & .916 & .914 & .898 & .892 & .907 & .954 & .948 \\
& scene\_$18$ & .825 & .802 & .812 & .824 & .717 & .785 & .887 & .921 \\
& scene\_$19$ & .900 & .884 & .893 & .904 & .885 & .857 & .566 & .910 \\
& scene\_$20$ & .868 & .866 & .854 & .880 & .864 & .868 & .824 & .913 \\
& scene\_$21$ & .927 & .927 & .934 & .922 & .923 & .941 & .959 & .961 \\
& scene\_$29$ & \xmark & \xmark & \xmark & \xmark & \xmark & \xmark & .661 & .944 \\
& scene\_$39$ & .921 & .921 & .902 & .880 & .880 & .896 & .713 & .934 \\
& scene\_$67$ & .829 & .831 & .838 & .703 & .702 & .698 & .673 & .923 \\
& scene\_$69$ & .843 & .845 & .842 & .815 & .821 & .819 & .593 & .870 \\
& scene\_$70$ & .691 & .649 & .669 & .550 & .607 & .615 & .705 & .850 \\
& scene\_$106$ & \xmark & \xmark & \xmark & \xmark & \xmark & \xmark & .710 & .966 \\
& scene\_$110$ & .822 & .718 & .810 & .765 & .430 & .704 & .817 & .852 \\
\midrule
\multirow{10}{*}{\rotatebox{90}{$N=4$}}
& scene\_$25$ & .886 & .852 & .844 & .771 & .763 & .779 & N/A & .945 \\
& scene\_$30$ & .953 & .952 & .915 & .923 & .923 & .827 & N/A & .913 \\
& scene\_$31$ & .833 & .812 & .794 & .815 & .817 & .813 & N/A & .921 \\
& scene\_$35$ & .919 & .918 & .912 & .890 & .889 & .846 & N/A & .915 \\
& scene\_$47$ & .966 & .954 & .941 & .898 & .864 & .864 & N/A & .927 \\
& scene\_$54$ & .936 & .929 & .935 & .890 & .901 & .908 & N/A & .950 \\
& scene\_$64$ & .819 & .800 & .814 & .715 & .815 & .782 & N/A & .921 \\
& scene\_$49$ & .961 & .957 & .949 & .938 & .952 & .960 & N/A & .974 \\
& scene\_$92$ & \xmark & \xmark & \xmark & \xmark & \xmark & \xmark & N/A & .921 \\
& scene\_$81$ & .809 & .785 & .805 & .732 & .758 & .744 & N/A & .861 \\
\midrule
\multirow{11}{*}{\rotatebox{90}{$N=5$}}
& scene\_$33$ & .933 & .923 & .912 & .841 & .779 & .884 & N/A & .927 \\
& scene\_$46$ & .757 & .739 & .752 & .604 & .645 & \xmark & N/A & .823 \\
& scene\_$51$ & .895 & .889 & .876 & .895 & .878 & .861 & N/A & .939 \\
& scene\_$62$ & .828 & .804 & .822 & .828 & .859 & .878 & N/A & .927 \\
& scene\_$63$ & .827 & .822 & .822 & .809 & .822 & .830 & N/A & .790 \\
& scene\_$65$ & .881 & .871 & .883 & .880 & .897 & .901 & N/A & .905 \\
& scene\_$66$ & .829 & .821 & .815 & .802 & .820 & .817 & N/A & .853 \\
& scene\_$72$ & .718 & .708 & .721 & .704 & .710 & \xmark & N/A & .842 \\
& scene\_$76$ & .859 & .773 & .844 & .774 & .780 & \xmark & N/A & .952 \\
& scene\_$77$ & .761 & .762 & .743 & .775 & .771 & .741 & N/A & .887 \\
& scene\_$98$ & .698 & .693 & .696 & .704 & \xmark & .667 & N/A & .772 \\
\midrule
\multirow{10}{*}{\rotatebox{90}{$N=6$}}
& scene\_$27$ & .966 & .963 & .963 & .933 & .940 & .956 & N/A & .979 \\
& scene\_$50$ & .957 & .939 & .933 & .925 & .960 & \xmark & N/A & .977 \\
& scene\_$52$ & .886 & .881 & .887 & .902 & .893 & .897 & N/A & .892 \\
& scene\_$53$ & .810 & .809 & .811 & .871 & .822 & .855 & N/A & .917 \\
& scene\_$55$ & .911 & .902 & .897 & .915 & .913 & .920 & N/A & .908 \\
& scene\_$57$ & .972 & .938 & .961 & .855 & .923 & \xmark & N/A & .985 \\
& scene\_$58$ & .814 & .809 & .811 & .828 & .850 & .871 & N/A & .882 \\
& scene\_$59$ & .757 & .749 & .754 & .780 & .801 & .804 & N/A & .935 \\
& scene\_$60$ & .839 & .806 & .826 & .798 & .883 & \xmark & N/A & .998 \\
& scene\_$101$ & \xmark & \xmark & \xmark & \xmark & \xmark & \xmark & N/A & .825 \\
\bottomrule
\end{tabularx}

\end{table}
\begin{table}[t]
\centering
\caption{Per-scene CMSA($\downarrow$) comparison on the Multi-Camera Benchmark, \textbf{Continued.}}
\label{tab:cmsa_exts}
\scriptsize
\begin{tabularx}{\columnwidth}{c l *{8}{Y}}

\toprule
& \multirow{2}{*}{\scriptsize Data}
& \multicolumn{3}{c}{\scriptsize MIS{+}TR}
& \multicolumn{3}{c}{\scriptsize MIS{+}PS}
& & \\

\cmidrule(lr){3-5} \cmidrule(lr){6-8}

&
& {\scriptsize GSP}
& {\scriptsize GES{-}GSP}
& {\scriptsize OBJ{-}GSP}
& {\scriptsize GSP}
& {\scriptsize GES{-}GSP}
& {\scriptsize OBJ{-}GSP}
& \raisebox{0pt}[0pt][0pt]{\shortstack{\scriptsize Stab\\Stitch++}}
& \raisebox{0pt}[0pt][0pt]{\shortstack{\scriptsize MViS\\(Ours)}}\\
\midrule
\multirow{22}{*}{\rotatebox{90}{$N=3$}}
& scene\_$1$ & 4.1 & 4.0 & 6.2 & 47.3 & 46.9 & 38.8 & 1.4 & 0.1 \\
& scene\_$2$ & 7.4 & 7.9 & 8.7 & 17.2 & 17.7 & 49.1 & 5.6 & 0.2 \\
& scene\_$4$ & 17.8 & 15.9 & 16.2 & 42.0 & 28.7 & 59.2 & 1.3 & 0.1 \\
& scene\_$6$ & 2.6 & 2.7 & 4.3 & 93.2 & 55.0 & 75.3 & 1.1 & 0.1 \\
& scene\_$7$ & 536 & 550 & 533 & 842 & 555 & 427 & 1.0 & 0.1 \\
& scene\_$8$ & 17.1 & 20.0 & 94.9 & 102 & 35.1 & 91.4 & 0.9 & 0.0 \\
& scene\_$9$ & \xmark & \xmark & \xmark & \xmark & \xmark & \xmark & 9.4 & 1.3 \\
& scene\_$10$ & 14.5 & 13.9 & 14.8 & 35.4 & 21.9 & 20.6 & 0.8 & 0.5 \\
& scene\_$12$ & 9.4 & 9.4 & 12.9 & 32.9 & 38.3 & 26.7 & 0.9 & 0.4 \\
& scene\_$13$ & 7.4 & 7.7 & 8.1 & 57.5 & 27.8 & 19.1 & 2.2 & 0.5 \\
& scene\_$15$ & 5.1 & 5.2 & 6.3 & 22.0 & 20.5 & 10.4 & 0.8 & 0.1 \\
& scene\_$18$ & 3.3 & 6.9 & 4.8 & 6.7 & 48.7 & 13.9 & 0.5 & 0.1 \\
& scene\_$19$ & 1.5 & 1.8 & 108 & 6.9 & 6.1 & 119 & 45.5 & 0.1 \\
& scene\_$20$ & 10.2 & 10.2 & 11.4 & 11.3 & 10.9 & 13.3 & 1.7 & 0.1 \\
& scene\_$21$ & 18.6 & 18.6 & 18.9 & 26.2 & 25.3 & 14.7 & 0.7 & 0.3 \\
& scene\_$29$ & \xmark & \xmark & \xmark & \xmark & \xmark & \xmark & 36.9 & 15.5 \\
& scene\_$39$ & 8.6 & 8.9 & 10.0 & 43.4 & 42.5 & 16.8 & 12.4 & 20.0 \\
& scene\_$67$ & 166 & 163 & 181 & {\scriptsize >1k} & {\scriptsize >1k} & {\scriptsize >1k} & 222 & 2.6 \\
& scene\_$69$ & 99.5 & 106 & 119 & 404 & 253 & 306 & 35.2 & 7.2 \\
& scene\_$70$ & 219 & 312 & 220 & 574 & 390 & 383 & 8.0 & 3.8 \\
& scene\_$106$ & \xmark & \xmark & \xmark & \xmark & \xmark & \xmark & 51.1 & 1.4 \\
& scene\_$110$ & 47.7 & 54.1 & 50.3 & 102 & 806 & 287 & 0.8 & 0.1 \\
\midrule
\multirow{10}{*}{\rotatebox{90}{$N=4$}}
& scene\_$25$ & 261 & 290 & 278 & 603 & 339 & 340 & N/A & 13.6 \\
& scene\_$30$ & 12.4 & 12.7 & 135 & 68.7 & 68.5 & 212 & N/A & 5.4 \\
& scene\_$31$ & 15.4 & 15.8 & 22.3 & 46.5 & 45.6 & 47.7 & N/A & 38.7 \\
& scene\_$35$ & 2.9 & 2.9 & 10.0 & 10.4 & 11.3 & 41.4 & N/A & 4.4 \\
& scene\_$47$ & 4.0 & 5.1 & 8.2 & 19.1 & 21.7 & 48.4 & N/A & 1.1 \\
& scene\_$54$ & 3.9 & 4.2 & 4.1 & 50.4 & 30.3 & 19.9 & N/A & 12.2 \\
& scene\_$64$ & 61.6 & 64.1 & 66.0 & 392 & 178 & 219 & N/A & 9.9 \\
& scene\_$49$ & 11.7 & 11.8 & 15.2 & 47.6 & 29.4 & 20.2 & N/A & 0.2 \\
& scene\_$92$ & \xmark & \xmark & \xmark & \xmark & \xmark & \xmark & N/A & 18.0 \\
& scene\_$81$ & 96.9 & 108 & 107 & {\scriptsize >1k} & 661 & 819 & N/A & 38.0 \\
\midrule
\multirow{11}{*}{\rotatebox{90}{$N=5$}}
& scene\_$33$ & 9.0 & 11.2 & 15.4 & 165 & 193 & 63.7 & N/A & 8.2 \\
& scene\_$46$ & 411 & 444 & 447 & {\scriptsize >1k} & {\scriptsize >1k} & \xmark & N/A & 60.9 \\
& scene\_$51$ & 31.9 & 33.4 & 35.3 & 81.1 & 71.8 & 91.4 & N/A & 0.3 \\
& scene\_$62$ & 60.9 & 62.2 & 63.9 & 222 & 128 & 119 & N/A & 2.5 \\
& scene\_$63$ & 24.0 & 26.2 & 32.0 & 39.6 & 27.1 & 20.5 & N/A & 1.0 \\
& scene\_$65$ & 23.7 & 23.8 & 23.8 & 111 & 81.1 & 63.8 & N/A & 0.3 \\
& scene\_$66$ & 4.8 & 5.3 & 6.2 & 32.3 & 23.0 & 19.1 & N/A & 0.5 \\
& scene\_$72$ & 312 & 326 & 332 & {\scriptsize >1k} & {\scriptsize >1k} & \xmark & N/A & 6.8 \\
& scene\_$76$ & 126 & 131 & 140 & 224 & 214 & \xmark & N/A & 13.8 \\
& scene\_$77$ & 17.5 & 21.3 & 20.4 & 168 & 95.9 & 86.6 & N/A & 26.4 \\
& scene\_$98$ & 8.8 & 17.2 & 47.1 & 29.9 & \xmark & 125 & N/A & 0.6 \\
\midrule
\multirow{10}{*}{\rotatebox{90}{$N=6$}}
& scene\_$27$ & 7.3 & 9.0 & 18.0 & 92.8 & 59.6 & 36.8 & N/A & 0.2 \\
& scene\_$50$ & 11.1 & 11.1 & 18.0 & 72.9 & 27.5 & \xmark & N/A & 1.4 \\
& scene\_$52$ & 1.3 & 1.9 & 2.8 & 8.3 & 6.0 & 31.7 & N/A & 0.1 \\
& scene\_$53$ & 5.3 & 5.6 & 5.2 & 18.1 & 27.7 & 15.9 & N/A & 0.4 \\
& scene\_$55$ & 24.4 & 25.8 & 30.3 & 67.6 & 66.1 & 192 & N/A & 0.7 \\
& scene\_$57$ & 12.7 & 13.3 & 14.8 & 467 & 163 & \xmark & N/A & 0.5 \\
& scene\_$58$ & 84.0 & 85.9 & 86.6 & 222 & 186 & 182 & N/A & 3.7 \\
& scene\_$59$ & 175 & 179 & 179 & 216 & 152 & 147 & N/A & 0.8 \\
& scene\_$60$ & 21.5 & 24.0 & 23.8 & 404 & 129 & \xmark & N/A & 15.4 \\
& scene\_$101$ & \xmark & \xmark & \xmark & \xmark & \xmark & \xmark & N/A & 1.2 \\
\bottomrule
\end{tabularx}
\end{table}

\begin{table}[!t]
  \centering
  \caption{Summary of notation used in the paper.}
  \label{tab:notation}
  \footnotesize
  \resizebox{\linewidth}{!}{
  \begin{tabular}{@{}ll@{}}
  
    \toprule
    Symbol & Description \\
    \midrule
    $N$ & Number of input videos (cameras). \\
    $\tau$ & Number of frames in each video. \\
    $V_i$ & $i$-th input video, $i \in \{1,\dots,N\}$. \\
    $F_i^t$ & Frame $t$ of video $V_i$. \\
    $r$ & Preprocessing resolution, frames resized to $r \times r$. \\
    $I_i^t \in \mathbb{R}^{r \times r \times 3}$ 
          & Resized and RGB-normalized version of $F_i^t$. \\
    $\boldsymbol H_i^t \in \mathrm{SL}(3)$ 
          & Predicted homography that aligns $I_i^t$ to the global panorama. \\
    $\widehat{\boldsymbol H}_{i \to j}^t$ 
          & Observed Relative Transformation (ORT) from $I_i^t$ to $I_j^t$ (same $t$, different views). \\
    $\widehat{\boldsymbol H}_i^{t \to t+1}$ 
          & ORT from $I_i^t$ to $I_i^{t+1}$ (same view, consecutive times). \\
    $\widehat{\boldsymbol H}_i^{t+1 \to t}$ 
          & ORT from $I_i^{t+1}$ to $I_i^{t}$. \\
    $M_{i \to j}^t$ 
          & Binary mask of the overlap region implied by $\widehat{H}_{i \to j}^t$. \\
    $M_i^{t \to t+1},\, M_i^{t+1 \to t}$ 
          & Temporal overlap masks implied by $\widehat{H}_i^{t \to t+1}$, $\widehat{H}_i^{t+1 \to t}$. \\
    $S_t{=}\{I_i^s \mid s \in \{t,t{+}1\},\, i\in \{1,..,N\}\}$ 
          & Sliding window at time $t$. \\
    $\boldsymbol G \in \mathbb{R}^{d^2 \times 2}$ 
          & Regular $d \times d$ grid in $[-1,1]^2$ (sampling points for geometric loss). \\
    $\boldsymbol{\theta}_i^t \in \mathbb{R}^8$ 
          & Lie-algebra parameters of $H_i^t$ (via the $\mathfrak{sl}(3)$ exponential map). \\
    $\boldsymbol H( \boldsymbol G)\in \mathbb{R}^{d^2\times2}$ & The result of applying homography $\boldsymbol H$ to $\boldsymbol G$. \\
    \bottomrule
  \end{tabular}
  }
\end{table}

\clearpage
%%%%%%%%%%%%%%%%%%%%%%%%%%%%%%%%%%%%%%%%%%%%%%%%%%%%%%%%%%%%

%\newpage
%\input{checklist.tex}

\end{document}